\documentclass{article}
\PassOptionsToPackage{square,comma,numbers,compress}{natbib}

\usepackage[preprint]{neurips_2023}

\usepackage[utf8]{inputenc} %
\usepackage[T1]{fontenc}    %
\usepackage[hidelinks,colorlinks=true]{hyperref}       %
\usepackage[dvipsnames]{xcolor}
\usepackage{url}            %
\usepackage{booktabs}       %
\usepackage{amsfonts}       %
\usepackage{nicefrac}       %
\usepackage{microtype}      %
\usepackage{xcolor}         %
\usepackage{graphicx}
\usepackage{tcolorbox}
\graphicspath{ {figures/} }
\usepackage{adjustbox}
\usepackage[edges]{forest}
\usepackage{array}
\newcolumntype{x}[1]{>{\centering\arraybackslash\hspace{0pt}}p{#1}}
\newcolumntype{q}[1]{>{\raggedright\arraybackslash\hspace{0pt}}p{#1}}

\usepackage{times,graphicx,tabulary,multirow,xspace}
\usepackage{subfig}
\usepackage{changepage}
\usepackage{color, colortbl}
\definecolor{LightGray}{rgb}{0.92,0.92,0.92}
\definecolor{Gray1}{rgb}{0.95,0.95,0.95}
\definecolor{Gray2}{rgb}{0.9,0.9,0.9}

\definecolor{paired-light-blue}{RGB}{198, 219, 239}
\definecolor{paired-dark-blue}{RGB}{49, 130, 188}
\definecolor{paired-light-orange}{RGB}{251, 208, 162}
\definecolor{paired-dark-orange}{RGB}{230, 85, 12}
\definecolor{paired-light-green}{RGB}{199, 233, 193}
\definecolor{paired-dark-green}{RGB}{49, 163, 83}
\definecolor{paired-light-purple}{RGB}{218, 218, 235}
\definecolor{paired-dark-purple}{RGB}{117, 107, 176}
\definecolor{paired-light-gray}{RGB}{217, 217, 217}
\definecolor{paired-dark-gray}{RGB}{99, 99, 99}
\definecolor{paired-light-pink}{RGB}{222, 158, 214}
\definecolor{paired-dark-pink}{RGB}{123, 65, 115}
\definecolor{paired-light-red}{RGB}{231, 150, 156}
\definecolor{paired-dark-red}{RGB}{131, 60, 56}
\definecolor{paired-light-yellow}{RGB}{231, 204, 149}
\definecolor{paired-dark-yellow}{RGB}{141, 109, 49}
\tikzset{%
    parent/.style =          {align=center,text width=1.8cm,rounded corners=3pt, line width=0.3mm, fill=gray!10,draw=gray!80},
    child/.style =           {align=center,text width=2.3cm,rounded corners=3pt, fill=blue!10,draw=blue!80,line width=0.3mm},
    grandchild/.style =      {align=center,text width=2cm,rounded corners=3pt},
    greatgrandchild/.style = {align=center,text width=1.5cm,rounded corners=3pt},
    greatgrandchild2/.style = {align=center,text width=1.5cm,rounded corners=3pt},    
    referenceblock/.style =  {align=center,text width=1.5cm,rounded corners=2pt},
    top_class/.style =           {align=center,text width=2cm,rounded corners=3pt, fill=paired-light-gray!50,draw=paired-dark-gray!65,line width=0.3mm},
    generation/.style =           {align=center,text width=2cm,rounded corners=3pt, fill= paired-light-green!50,draw=paired-dark-green!75,line width=0.3mm}, 
    generation_wide/.style =           {align=center,text width=2.5cm,rounded corners=3pt, fill= paired-light-green!50,draw=paired-dark-green!75,line width=0.3mm}, 
    generation_more/.style =           {align=center,text width=3cm,rounded corners=3pt, fill= paired-light-green!50,draw=paired-dark-green!75,line width=0.3mm},   
    generation_work/.style =           {align=center,text width=5.0cm,rounded corners=3pt, fill= paired-light-green!50,draw= cyan!0,line width=0.3mm},
    encoder/.style =           {align=center,text width=2cm,rounded corners=3pt, fill=paired-light-orange!50,draw=paired-dark-orange!65,line width=0.3mm},  
    encoder_more/.style =           {align=center,text width=3cm,rounded corners=3pt, fill=paired-light-orange!50,draw=paired-dark-orange!65,line width=0.3mm}, 
    encoder_work/.style =           {align=center,text width=5.0cm,rounded corners=3pt, fill=paired-light-orange!50,draw=red!0,line width=0.3mm},    
    gpa/.style =           {align=center,text width=2cm,rounded corners=3pt, fill=paired-light-blue!50,draw=paired-dark-blue!65,line width=0.3mm},
    gpa_wide/.style =           {align=center,text width=3cm,rounded corners=3pt, fill=paired-light-blue!50,draw=paired-dark-blue!65,line width=0.3mm},   
    gpa_work/.style =           {align=center, text width=5.0cm,rounded corners=3pt, fill=paired-light-blue!50,draw=blue!0,line width=0.3mm},
    gallery/.style =           {align=center,text width=2cm,rounded corners=3pt, fill=paired-light-yellow!50,draw=paired-dark-yellow!65,line width=0.3mm},
    gallery_wide/.style =           {align=center,text width=3cm,rounded corners=3pt, fill=paired-light-yellow!50,draw=paired-dark-yellow!65,line width=0.3mm},   
    gallery_work/.style =           {align=center, text width=5.0cm,rounded corners=3pt, fill=paired-light-yellow!50,draw=yellow!0,line width=0.3mm},
    data/.style =           {align=center,text width=2cm,rounded corners=3pt, fill=paired-light-blue!50,draw=paired-dark-blue!65,line width=0.3mm},
    data_wide/.style =           {align=center,text width=3cm,rounded corners=3pt, fill=paired-light-blue!50,draw=paired-dark-blue!65,line width=0.3mm},   
    data_work/.style =           {align=center, text width=4.5cm,rounded corners=3pt, fill=paired-light-blue!50,draw=blue!0,line width=0.3mm},  
    model/.style =           {align=center,text width=2cm,rounded corners=3pt, fill=paired-light-orange!50,draw=paired-dark-orange!65,line width=0.3mm},  
    model_more/.style =           {align=center,text width=4cm,rounded corners=3pt, fill=paired-light-orange!50,draw=paired-dark-orange!65,line width=0.3mm}, 
    model_work/.style =           {align=center,text width=4.5cm,rounded corners=3pt, fill=paired-light-orange!50,draw=red!0,line width=0.3mm},    
    pretraining/.style =           {align=center,text width=2cm,rounded corners=3pt, fill= paired-light-green!50,draw=paired-dark-green!75,line width=0.3mm}, 
    pretraining_wide/.style =           {align=center,text width=2.5cm,rounded corners=3pt, fill= paired-light-green!50,draw=paired-dark-green!75,line width=0.3mm}, 
    pretraining_more/.style =           {align=center,text width=4cm,rounded corners=3pt, fill= paired-light-green!50,draw=paired-dark-green!75,line width=0.3mm},   
    pretraining_work/.style =           {align=center,text width=4.5cm,rounded corners=3pt, fill= paired-light-green!50,draw= cyan!0,line width=0.3mm},      
    finetuning/.style =           {align=center,text width=2cm,rounded corners=3pt, fill= paired-light-purple!50,draw=paired-dark-purple!75,line width=0.3mm},   
    finetuning_work/.style =           {align=center,text width=4.5cm,rounded corners=3pt, fill= paired-light-purple!50,draw= orange!0,line width=0.3mm},        
    inference/.style =           {align=center,text width=2cm,rounded corners=3pt, fill= paired-light-red!35,draw=paired-light-red!90,line width=0.3mm},           
    inference_more/.style =           {align=center,text width=4cm,rounded corners=3pt, fill= paired-light-red!35,draw=paired-light-red!90,line width=0.3mm},
    inference_work/.style =           {align=center,text width=4.5cm,rounded corners=3pt, fill= paired-light-red!35,draw= magenta!0,line width=0.3mm},         
}

\newcommand{\greentext}[1]{\textcolor[rgb]{0, 0.5, 0}{#1}}

\newcommand{\bluetext}[1]{\textcolor[RGB]{0, 0, 255}{#1}}
\newcommand{\purpletext}[1]{\textcolor[RGB]{148, 0, 211}{#1}}

\newcommand{\eat}[1]{}

\definecolor{redhl}{HTML}{ea9999}
\definecolor{greenhl}{HTML}{d9ead3}
\definecolor{bluehl}{HTML}{c9daf8}
\definecolor{yellowhl}{HTML}{fff2cc}
\makeatletter
\DeclareRobustCommand\onedot{\futurelet\@let@token\@onedot}
\def\@onedot{\ifx\@let@token.\else.\null\fi\xspace}

\def\eg{\emph{e.g}\onedot} 
\def\ie{\emph{i.e}\onedot} 
\def\cf{\emph{cf}\onedot} 
\def\etc{\emph{etc}\onedot} \def\vs{\emph{vs}\onedot}

\makeatother

\newcommand{\modelname}{DALL-E~3\xspace}
\newcommand{\benchname}{DEsignBench\xspace}

\title{\benchname: Exploring and Benchmarking \\ \modelname for Imagining Visual Design }

\author{
{\bf Kevin Lin$^{*}$, Zhengyuan Yang$^{*}$, Linjie Li, Jianfeng Wang, Lijuan Wang$^{*\spadesuit}$} \\
Microsoft Corporation\\
\and
\footnotesize{
$^*$~Equal Contribution \;
$^{\spadesuit}$~Project Lead \;
} \\
{\tt\footnotesize \{keli,zhengyang,lindsey.li,jianfw,lijuanw\}@microsoft.com} \\
\url{https://design-bench.github.io/}
}

\begin{document}

\maketitle

\vspace{20pt}
\begin{abstract}
We introduce \benchname, a text-to-image (T2I) generation benchmark tailored for visual design scenarios. Recent T2I models like \modelname~\cite{dalle3paper,dalle3system,dalle3blog} and others, have demonstrated remarkable capabilities in generating photorealistic images that align closely with textual inputs. 
While the allure of creating visually captivating images is undeniable, our emphasis extends beyond mere aesthetic pleasure. We aim to investigate the potential of using these powerful models in authentic design contexts. In pursuit of this goal, we develop \benchname, which incorporates test samples designed to assess T2I models on both ``design technical capability'' and ``design application scenario.'' Each of these two dimensions is supported by a diverse set of specific design categories. We explore \modelname together with other leading T2I models on \benchname, resulting in a comprehensive visual gallery for side-by-side comparisons. For \benchname benchmarking, we perform human evaluations on generated images in \benchname gallery, against the criteria of image-text alignment, visual aesthetic, and design creativity. Our evaluation also considers other specialized design capabilities, including text rendering, layout composition, color harmony, 3D design, and medium style. In addition to human evaluations, we introduce the first automatic image generation evaluator powered by GPT-4V. This evaluator provides ratings that align well with human judgments, while being easily replicable and cost-efficient. A high-resolution version is available at \href{https://github.com/design-bench/design-bench.github.io/raw/main/designbench.pdf?download=}{this link}.
\end{abstract}
\vspace{20pt}

{
  \hypersetup{linkcolor=black}
  \tableofcontents
  \label{sec:toc}
}

{
\small
\hypersetup{linkcolor=black}
\addcontentsline{toc}{section}{List of Figures}
\listoffigures
\label{sec:lof}
}
\clearpage

\section{Introduction}
\label{sec:01intro}
\subsection{Motivation and Overview}

Advancements in text-to-image (T2I) generation~\cite{Firefly,Ideogram,Midjourney,goodfellow2020generative,kingma2013auto,van2017neural,esser2021taming,ramesh2021zero,yu2022scaling,chang2022maskgit,sohl2015deep,song2020improved,rombach2022high,podell2023sdxl,dalle3paper,dalle3system,li2023multimodal} have shown remarkable capabilities in generating high-fidelity images that follow the user input text prompts. Many known challenges~\cite{marcus2022very,saharia2022photorealistic,feng2022training,ku2023imagenhub}, such as the prompt ``A horse riding an astronaut'' to test prompt following, text rendering, and distortions in the human face and hands generation, have been significantly improved by recent advancements, with examples postponed in Section~\ref{sec:02inputmode}. The rapid advancement naturally raises a question: \emph{what is the next goal to make T2I generation even more practically valuable?} In this work, we focus on designing scenarios, and examine how the state-of-the-art T2I models can assist visual design~\cite{seneviratne2022dalle,ploennigs2023ai,liu2022design,liu2022opal,ko2023large,liu20233dall,oppenlaender2022creativity,zhang2022armani}, in addition to merely generating visually pleasant results.%

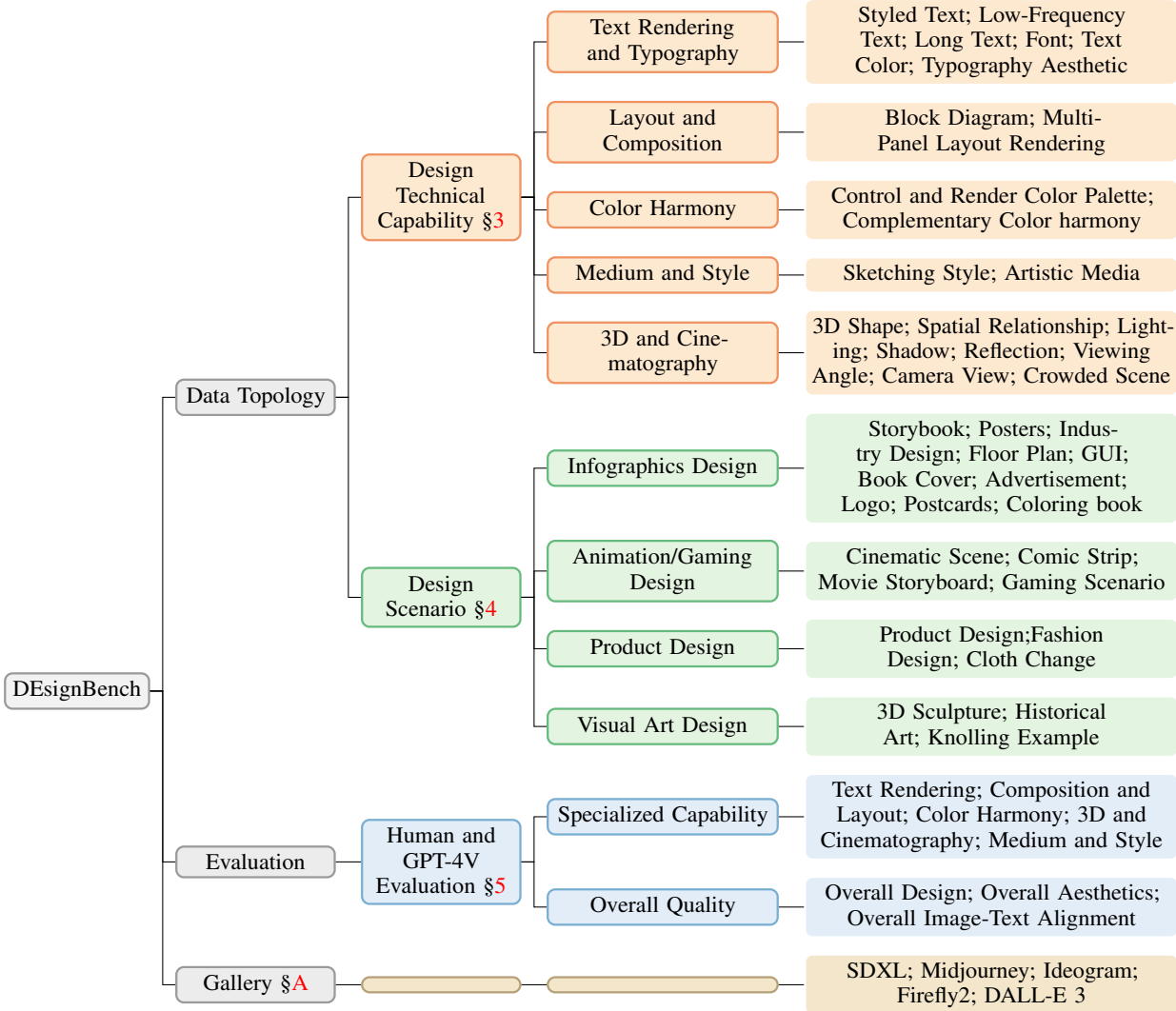
\begin{figure*}%
\footnotesize
\centering
\begin{adjustbox}{center}
\hspace*{-30pt}
    \begin{forest}
        for tree={
            forked edges,
            grow'=0,
            draw,
            rounded corners,
            node options={align=center,},
            text width=2.7cm,
            s sep=6pt,
            calign=edge midpoint,
        },
        [\benchname, fill=gray!45, parent
            [Data Topology, for tree={ top_class}
                [Design Technical Capability \S\ref{sec:04emerging}, for tree={fill=red!45,encoder}
                    [Text Rendering and Typography,  encoder_more
                        [{Styled Text; Low-Frequency Text; Long Text; Font; Text Color; Typography Aesthetic}, encoder_work]
                    ]
                    [Layout and Composition, encoder_more 
                        [Block Diagram;
                        Multi-Panel Layout Rendering, encoder_work]
                    ]
                    [Color Harmony, encoder_more 
                        [Control and Render Color Palette;
                        Complementary Color harmony, encoder_work]
                    ]
                    [Medium and Style,  encoder_more
                        [Sketching Style; 
                        Artistic Media, encoder_work]
                    ]
                    [3D and Cinematography,  encoder_more
                        [3D Shape; Spatial Relationship; Lighting; Shadow; Reflection; Viewing Angle; Camera View; Crowded Scene, encoder_work]
                    ]
                ]
                [Design Scenario \S\ref{sec:05scenario}, for tree={fill=green!45,generation}
                    [Infographics Design, generation_more
                        [Storybook; Posters; Industry Design; Floor Plan; GUI; Book Cover; Advertisement; Logo; Postcards; Coloring book, generation_work]
                    ]
                    [Animation/Gaming Design, generation_more
                        [Cinematic Scene; Comic Strip; Movie Storyboard; Gaming Scenario, generation_work]
                    ]  
                    [Product Design, generation_more
                        [Product Design;Fashion Design; Cloth Change, generation_work]
                    ]                 
                    [Visual Art Design, generation_more
                        [3D Sculpture; Historical Art; Knolling Example, generation_work]
                    ]           
                ]
            ]
            [Evaluation, for tree={ top_class}
                [Human and GPT-4V Evaluation \S\ref{sec:06bench}, for tree={fill=green!45,gpa}
                    [Specialized Capability, gpa_wide
                        [Text Rendering; Composition and Layout; Color Harmony; 3D and Cinematography; Medium and Style, gpa_work]
                    ]
                    [Overall Quality, gpa_wide
                        [Overall Design; Overall Aesthetics; Overall Image-Text Alignment, gpa_work]
                    ]                
                ] 
            ]
            [Gallery \S\ref{sec:gallery}, for tree={ top_class}
                [, for tree={fill=yellow!45,gallery}
                    [ , gallery_wide
                        [SDXL; Midjourney; Ideogram; Firefly2; \modelname, gallery_work]
                    ]
                ] 
            ]
        ]
    \end{forest}
\end{adjustbox}
\caption[\benchname overview]{An overview of the \benchname's structure.}
\vspace{20pt}
\label{fig:paper_structure}
\end{figure*}

To this end, we present a new evaluation benchmark named \benchname to examine T2I models' capabilities in assisting visual design. In addition to the base T2I capabilities in standard T2I benchmarks~\cite{lin2014microsoft,saharia2022photorealistic,yu2022scaling,huang2023t2i,cho2023visual,ku2023imagenhub}, \benchname evaluates visual design from two unique perspectives, \ie, the core design technical capability and the design application scenarios. We then collect evaluation prompts organized into each category and aspect. We collect the results of the state-of-the-art T2I models~\cite{podell2023sdxl,Midjourney,Ideogram,Firefly,dalle3paper,dalle3system} into our \benchname gallery, and perform both human and GPT-4V~\cite{gpt4,gpt4v,yang2023dawn} evaluations on the \benchname. Figure~\ref{fig:paper_structure} overviews the \benchname structure, with each component detailed as follows.

\paragraph{\benchname topology.}
\benchname categorizes the visual design abilities to examine into two categories, namely the design technical capability and the design application scenario. The \emph{design technical capability} separately zooms into each core technical capability required for visual design, including text rendering and typography~\cite{lupton2014thinking,bringhurst2004elements}, layout and composition~\cite{samara2023making,pipes2005production}, color harmony~\cite{albers2013interaction,mollica2013color}, medium and artistic style~\cite{lupton2008graphic}, and 3D and cinematography~\cite{mcquilkin2011cinema,brown2016cinematography}. We further define sub-categories under each capability, and manually craft text prompts accordingly. The \emph{design application scenario} focuses on the real design application, which usually requires the seamless integration of multiple design technical capabilities. Example categories include infographics, animation, gaming, product, and visual art.

\paragraph{\benchname data and gallery.}
Based on the \benchname topology, we organize samples into an evaluation set of $215$ prompts, with corresponding design category tags, leading to a new challenging generation benchmark focused on visual design. We collect images generated by the state-of-the-art T2I models (SDXL v1.0~\cite{podell2023sdxl}, Midjourney v5.2~\cite{Midjourney}, Ideogram~\cite{Ideogram}, Firefly 2~\cite{Firefly}, and \modelname~\cite{dalle3paper,dalle3system}), and formulate them into the \benchname gallery for side-by-side qualitative comparisons.

\paragraph{\benchname evaluation.}
We conduct the human evaluation~\cite{ramesh2022hierarchical,yu2022scaling,saharia2022photorealistic,podell2023sdxl} on images in the \benchname gallery, assessing them based on three primary criteria: visual aesthetics, image-text alignments, and design creativity. The design creativity aspect asks human annotators to evaluate if the generated image is a novel design, \ie, whether it showcases unique and innovative interpretations of the input prompt and brings a fresh perspective. Additionally, the evaluation also considers five other design-specific capabilities, \ie, text rendering, composition and layout, color harmony, 3D and cinematography, and medium and style, each paired with specific annotation guidelines.

Furthermore, we explore the automatic evaluation pipeline, which provides a more cost-effective approach with reproducible results. Automatic evaluation with large language models has shown promise in various natural language processing~\cite{chiang2023can,liu2023gpteval,fu2023gptscore} and vision-language understanding tasks~\cite{yu2023mm}. However, T2I evaluation is more complicated. It requires both a high-level semantic understanding (\eg, image-text alignment and a detailed visual comparison across two images (\eg, visual aesthetic ranking), not to mention several other design-specific criteria. Following prior studies that take large multimodal models (LMMs)~\cite{gpt4,gpt4v,yang2023dawn,bingchat} for T2I image-text alignment evaluation~\cite{dalle3paper,an2023openleaf,yang2023idea2img}, we propose a pairwise model rating based on GPT-4V that comprehensively evaluates all aspects as a human annotator. The high consistency with human rating indicates the effectiveness of the proposed LMM-based T2I evaluation.

\clearpage
    
\vspace{1pt}
Our contributions are summarized as follows.
\vspace{-3pt}
\begin{itemize}
\setlength\itemsep{-1pt}
    \item We explore \modelname on imagining visual design. We then present \benchname, a new challenging text-to-image generation benchmark focusing on assisting visual design.
    \item We propose an automatic GPT-4V evaluation for \benchname evaluation, which provides reproducible results that align well with human ratings.
    \item We collect \benchname gallery, which side-by-side compares the images generated by various state-of-the-art T2I models (SDXL, Midjourney, Ideogram, Firefly2, \modelname).
\end{itemize}

Remaining sections are organized as follows. Section~\ref{sec:02inputmode} uses \modelname to provide an overview of the state of the art in T2I generation, and justify the experiment settings in \benchname. Section~\ref{sec:04emerging} and Section~\ref{sec:05scenario} introduce the design technical capability and the design application scenario, respectively, using insights from \modelname. The human and GPT-4V quantitative evaluations are discussed in Section~\ref{sec:06bench}. Finally, the appendix shows the complete \benchname gallery, showcasing output comparisons among SDXL, Midjourney, Ideogram, Firefly2, and \modelname.
\clearpage
\section{\modelname Basics and \benchname Settings}
\label{sec:02inputmode}
In this section, we overview the state-of-the-art T2I generation capability, with explorations on \modelname. We then introduce the experiment settings in \benchname.

\subsection{\modelname's Working Modes}%
\begin{figure}
\centering
\begin{adjustbox}{center}
\small
\begin{tabular}{x{0.496\textwidth}@{}x{0.496\textwidth}@{}x{0.496\textwidth}}
T2I Model A & T2I Model B & T2I Model C
\end{tabular}
\end{adjustbox}
\centerline{\includegraphics[width=1.5\textwidth]{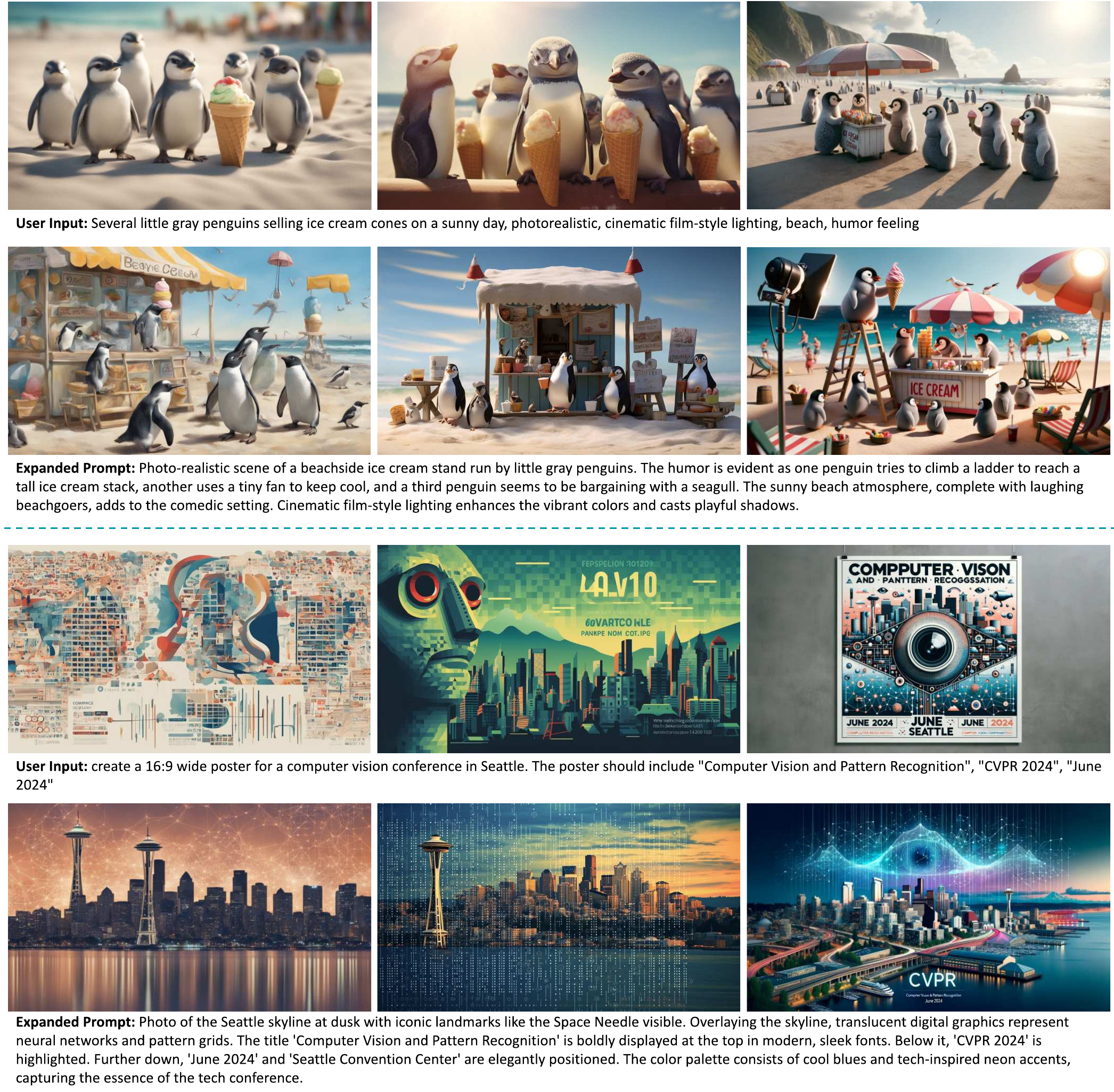}}
\caption[ChatGPT prompt expansion]{Qualitative comparisons between the generation results with original user inputs and ChatGPT expanded prompts.  We hide the specific T2I model names as A, B, C. Please see appendix for comprehensive model comparisons. %
}
\label{fig:expand}
\end{figure}

\paragraph{ChatGPT prompt expansion.}
\modelname~\cite{dalle3paper,dalle3system,dalle3blog} adopts ChatGPT~\cite{chatgpt} for prompt expansion, \ie, converting an input user query into a more detailed text description. As shown in Figure~\ref{fig:expand}, we empirically observe that this prompt expansion (\cf, user input \vs expanded prompt) also benefits other compared T2I models, such as SDXL~\cite{podell2023sdxl} and Midjourney~\cite{Midjourney}. Therefore, we take the ``expanded prompt'' as the default setting in our \benchname.

In addition to \modelname's default prompt expansion behavior in ChatGPT defined by the built-in system prompt, such as generating four prompts sequentially and producing four images, we find it helpful to add extra input prompts to ChatGPT for specialized prompt drafting.
\begin{itemize}
\vspace{-3pt}
\setlength\itemsep{-1pt}
\item \emph{Generate a detailed description and then generate \textbf{one} image:} Longer and more detailed prompts generally lead to better images, \ie, more object details, correct scene texts, and better image quality. We find it helpful to explicitly ask ChatGPT to provide a detailed description, and ease the task by asking for one prompt instead of four, both encourage a more detailed T2I prompt. We find this instruction particularly helpful in generating complicated scenes, such as posters, books, ads, \etc, which are otherwise almost impossible to create.
\item \emph{Exactly repeat the same prompt for one image:} For other cases, we may want to shut down the ChatGPT prompt paraphrasing, \eg, changing a few attributes words in a controlled manner or producing the previously generated images. To achieve that, we can simply ask ChatGPT to ``exactly repeat the same prompt.''
\end{itemize}

\paragraph{Multi-round dialogue-based T2I.}
\modelname with ChatGPT also naturally supports the multi-round dialogue-based generation. The chat interface allows users to refer to the generation history in generating the next image. For example, one may refer to a specific generated image and give an editing instruction, such as ``Change the cloth in the second image into the blue color,'' and naturally continue with multi-round editing. Another example is to keep arbitrary visual aspects in the generated image, such as keeping the character appearance or image style when generating a multiple image comic stripe (\eg, in Figures~\ref{fig:text_comic}-\ref{fig:text_comic_3}).

\subsection{T2I Generation Capability Overview}
We next provide an overview of the \modelname's generation capability, with popular testing prompts from existing benchmarks or community posts. Overall, we observe that \modelname's unprecedented prompt following ability allows it to effectively solve many well-known challenge cases. This observation motivates us to go a step further, and construct \benchname that considers the more challenging yet valuable scenarios of visual designs.

\begin{figure*}[h!]
\centering
\centerline{\includegraphics[width=1.5\textwidth]{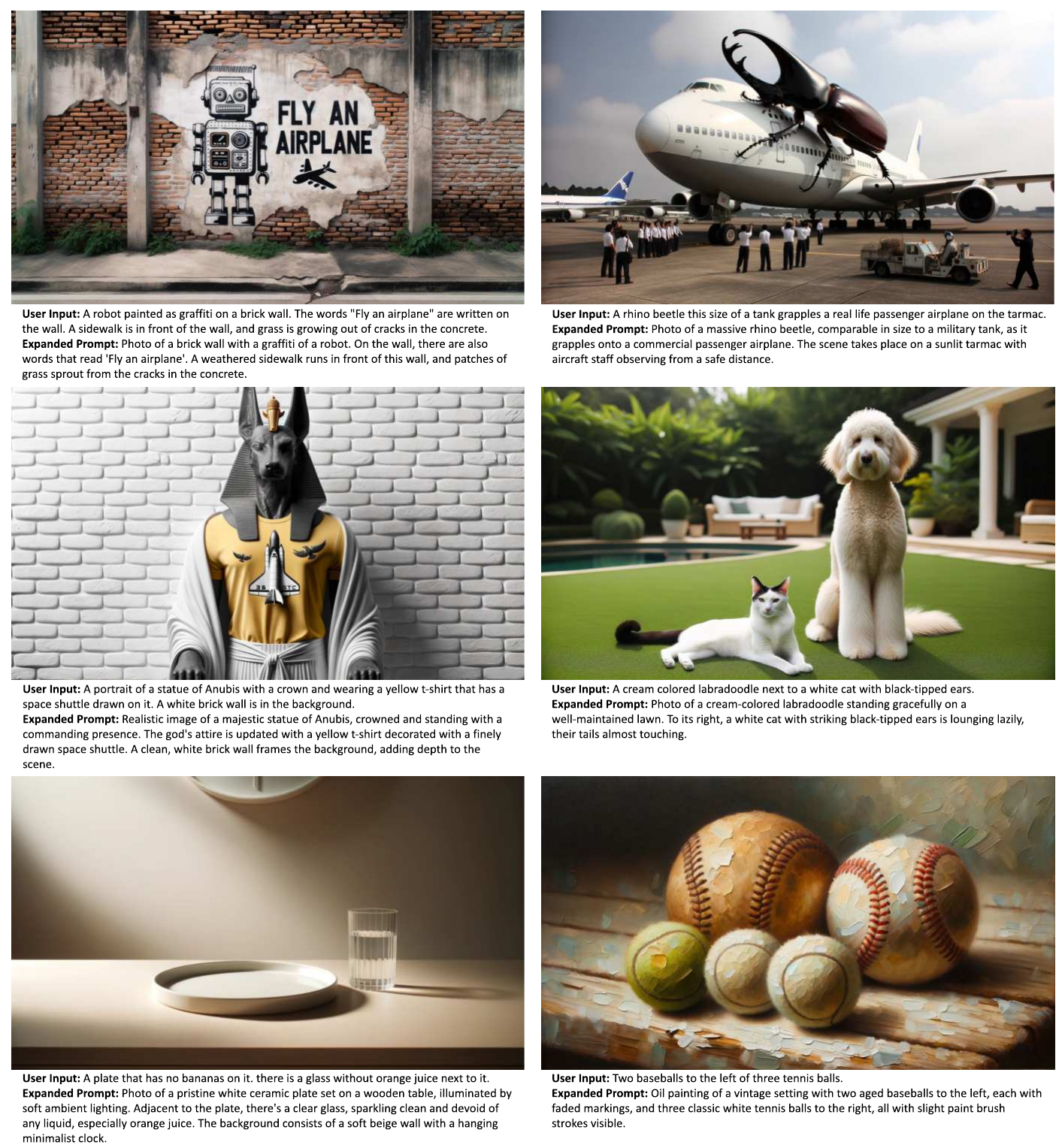}}
\caption[prompt following: detailed descriptions]{Challenging prompts~\cite{yu2022scaling} in following long and detailed descriptions. %
}
\label{fig:parti}
\end{figure*}

\paragraph{Prompt following: detailed descriptions.}
Prompt following is one key challenge in T2I generation. Previous T2I models tend not to strictly follow the text prompt, leading to incorrect objects and attributes~\cite{feng2022training,chefer2023attend,black2023training,fan2023dpok}. We use the famous failure cases in PartiPrompts~\cite{yu2022scaling} to show \modelname's prompt following capability. As shown in Figure~\ref{fig:parti}, \modelname generates images with correct object counts, relative size, global and local attributes, minimal object hallucination, and scene text. As further discussed throughout the paper, unprecedented prompt following ability is critical for the imagined design scenarios, allowing designers to use arbitrary text words for image control more confidently. 

\begin{figure*}[h!]
\centering
\centerline{\includegraphics[width=1.5\textwidth]{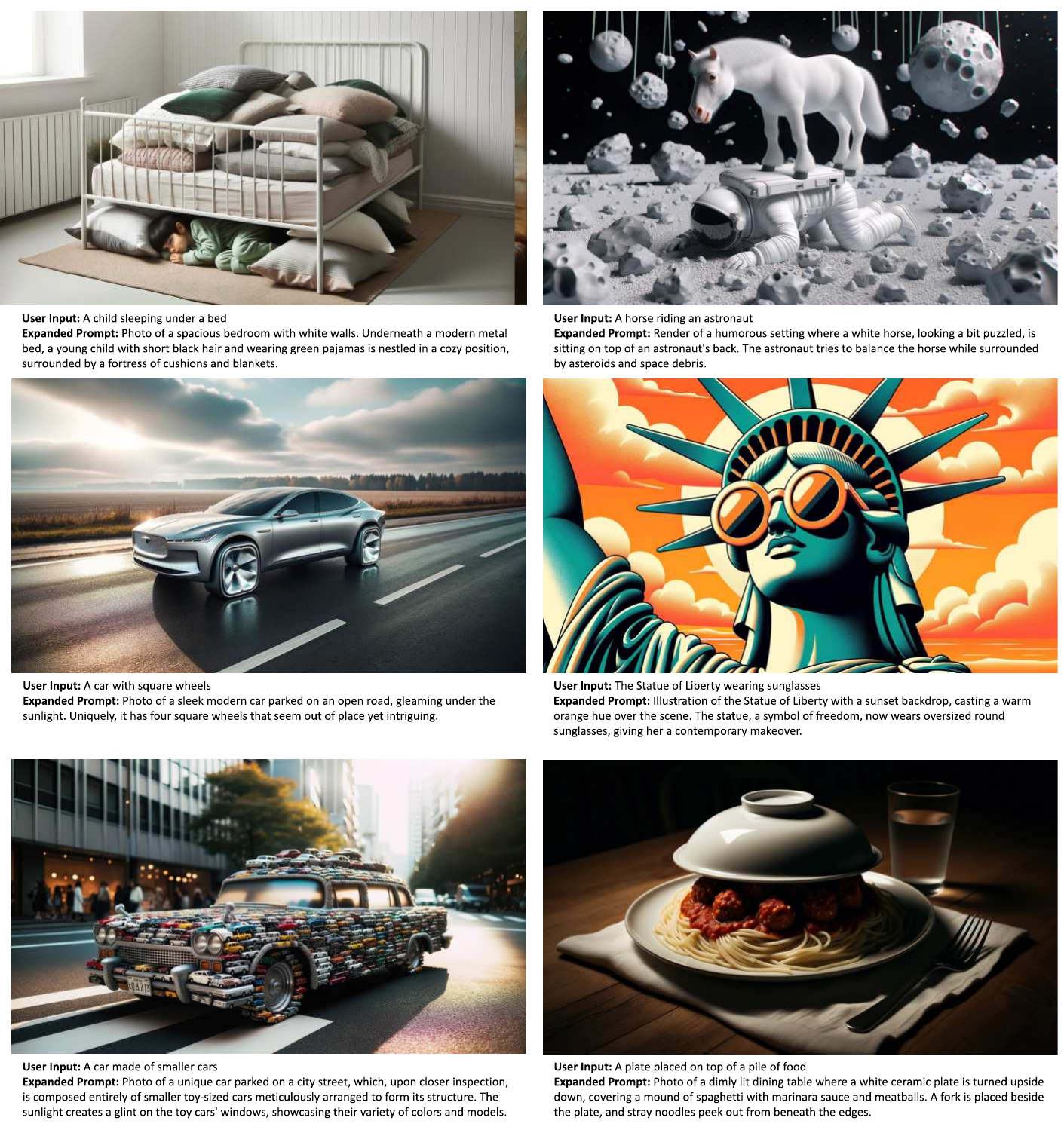}}
\caption[prompt following: uncommon scenes]{Challenging prompts in generating uncommon scenes.   
}
\label{fig:drawbench}
\end{figure*}

\paragraph{Prompt following: uncommon scenes.}
In addition to following complicated long prompts, prompt following also requires models to faithfully generate the uncommon senses, such as the ``A horse riding an astronaut.'' Following prompts for uncommon sense is essential for design scenarios, which usually involve imaginative creations with uncommon attributes and object combinations. In Figure~\ref{fig:drawbench}, we examine representative challenging prompts from community posts~\cite{mccammon2023can}. \modelname shows the capability to generate uncommon spatial relationships, object shapes, attributes, \etc. Such prompt following capability may assist designers in creating their imaginative pieces more easily.

\begin{figure*}[h!]
\centering
\vspace{-20mm}
\centerline{\includegraphics[width=1.5\textwidth]{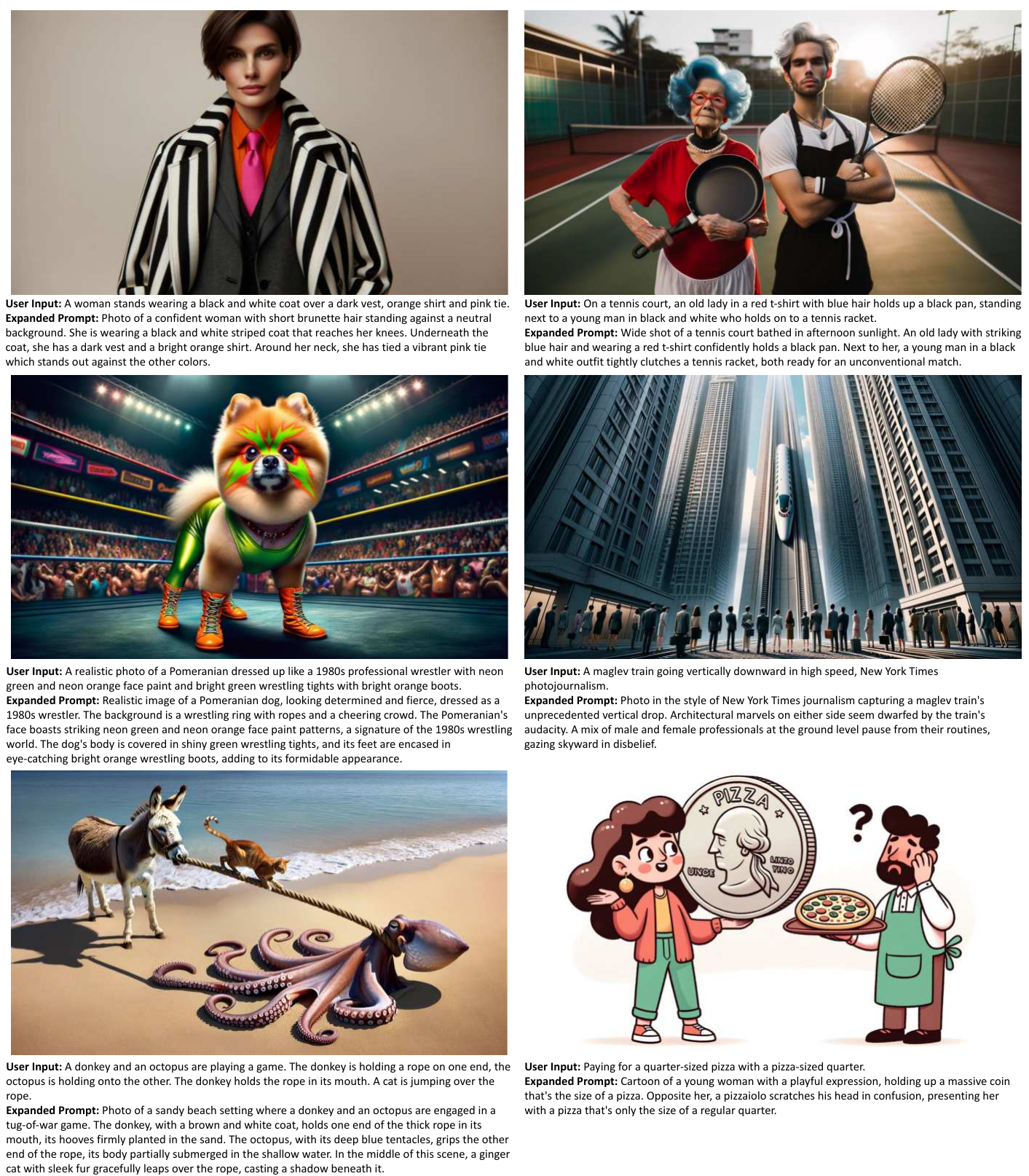}}
\caption[other challenge prompts]{Other challenging prompts from previous papers and community posts~\cite{marcus2022very,saharia2022photorealistic,feng2022training,ku2023imagenhub}.
}
\label{fig:challenge}
\end{figure*}

\paragraph{Image generation: text rendering.}
Text rendering~\cite{liu2022character,ma2023glyphdraw,chen2023textdiffuser,yang2023glyphcontrol,shimoda2023towards} is critical for design scenarios, yet remains to be challenging for previous T2I models~\cite{Midjourney,podell2023sdxl}. We empirically observe that \modelname can more reliably render texts in images, though still not perfect on more complicated texts. Table~\ref{table:text} provides a quantitative comparison of the word-level scene text accuracy on $40$ constructed samples. Specifically, we run the Microsoft Azure OCR system and compare the exact match words with the text in the input prompt. We generate $N=4$ images for each prompt and report the best results. We show additional qualitative results later in Figures~\ref{fig:text_style}-\ref{fig:motion_graphics}.
\begin{table*}[t]\small
\centering
\begin{tabular}{ l | c c c | c }
    \hline
    Word-level Acc. (\%) & Short Words & Challenging Words & Sentences & Total \\
    \hline
    Midjourney~\cite{Midjourney} & 0.0 & 0.0 & 4.3 & 1.1 \\
    SDXL~\cite{podell2023sdxl} & 37.9 & 5.0 & 19.0 & 25.0 \\
    IF~\cite{Deepfloyd} & 62.5 & 15.8 & 39.4 & 45.0 \\
    \modelname & \textbf{83.3} & \textbf{31.7} & \textbf{62.4} & \textbf{65.2} \\
    \hline
\end{tabular}
\caption{Word-level text rendering accuracy when selecting the best from $N=4$ generated images.} %
\label{table:text}
\end{table*}

\paragraph{Image generation: other challenges.}
We also examine other common failures shared among previous T2I models, such as hand and face generation, unique art styles, challenging objects, \etc. We empirically observe that \modelname works more reliably in those challenging cases. Figure~\ref{fig:challenge} shows several examples of such ``common failures'' discussed in previous papers and community posts~\cite{marcus2022very,rips1989similarity,saharia2022photorealistic,feng2022training,ku2023imagenhub}, \eg, detailed and uncommon attribute designs, uncommon scenes, \etc. We group the explorations based on their usage in design scenarios, presenting in the next section ``design technical capability.''

\clearpage
\section{Design Technical Capability}
\label{sec:04emerging}

Design encompasses a broad spectrum, from product and advertisement to logo and fashion design. Essential to any design tool is the capacity to produce text, shapes, charts, and diagrams~\cite{lupton2008graphic, lupton2014thinking, bringhurst2004elements}. Beyond these basics, the tool should be adept at crafting layouts that are not only semantically accurate but also aesthetically appealing~\cite{samara2023making}. Mastery of elements such as 3D, lighting, color palettes, and varied rendering materials and styles is indispensable~\cite{mcquilkin2011cinema, pipes2005production, brown2016cinematography}. In the following section, we highlight \modelname's competencies in addressing diverse design challenges.

\subsection{Text Rendering and Typography}
\begin{figure*}[h!]
\centering
\vspace{-20mm}
\centerline{\includegraphics[width=1.5\textwidth]{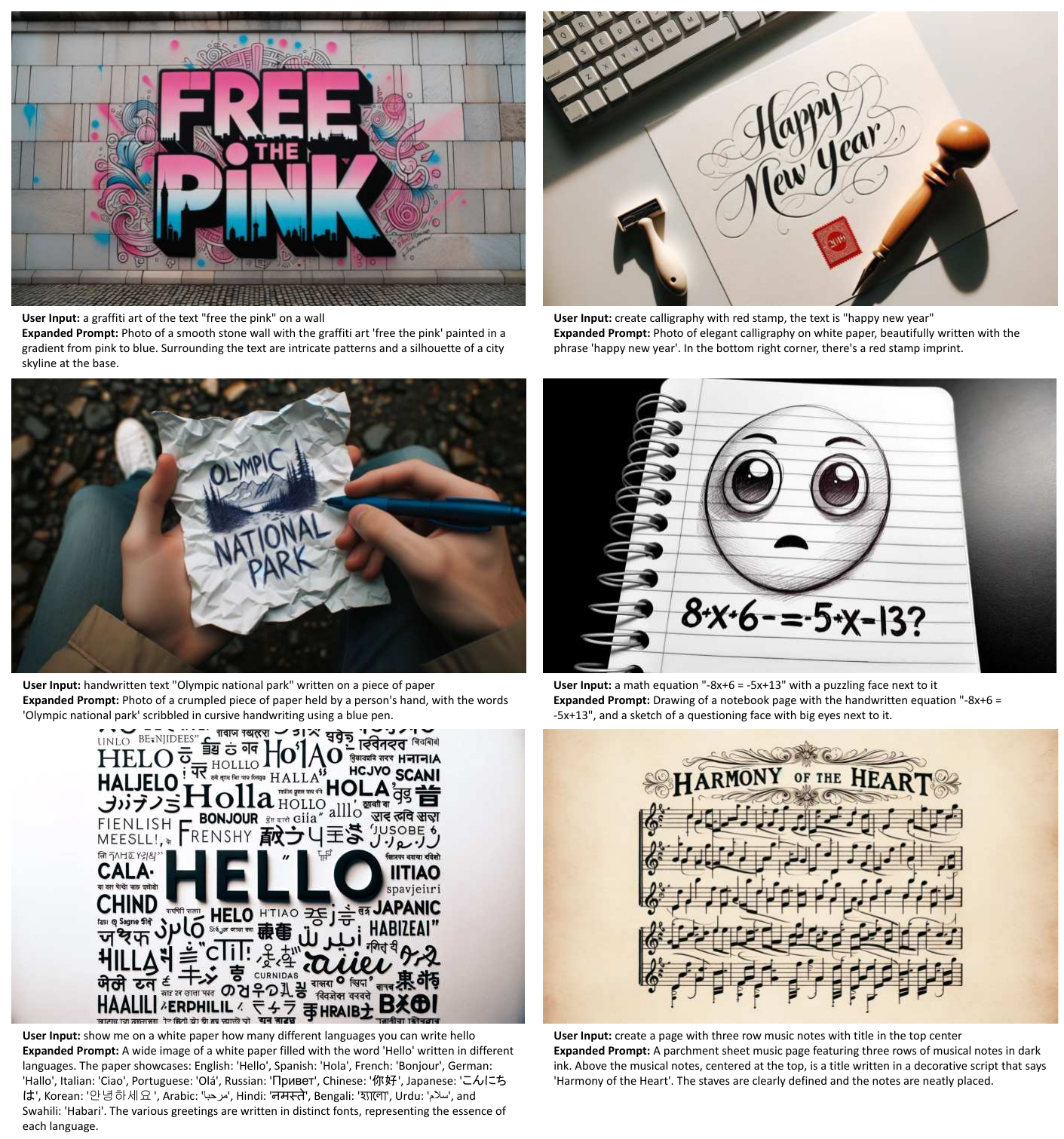}}
\caption[text rendering: stylized text]{Rendering stylized texts, \eg, graffiti art ``free the pink,'' calligraphy ``happy new year,'' handwritten texts ``Olympic national park,'' math equation ``-8x+6 = -5x+13,'' ``hello'' in multiple languages, and music notes. %
}
\label{fig:text_style}
\end{figure*}

Figure~\ref{fig:text_style} presents six diverse styled text renderings, spanning graffiti art, calligraphy, handwritten texts, mathematical symbols, multilingual scripts, and musical notations. While \modelname impressively renders English text across different styles, it exhibits some inaccuracies. The math equation, for instance, misinterprets certain operators and signs. While the layout for multilingual rendering appears organized, it struggles with certain languages, particularly Chinese and Japanese. The musical notation, while superficially resembling actual sheet music, includes several inaccuracies, underlining \modelname's constraints in this domain.

\begin{figure*}[h!]
\centering
\vspace{-20mm}
\centerline{\includegraphics[width=1.5\textwidth]{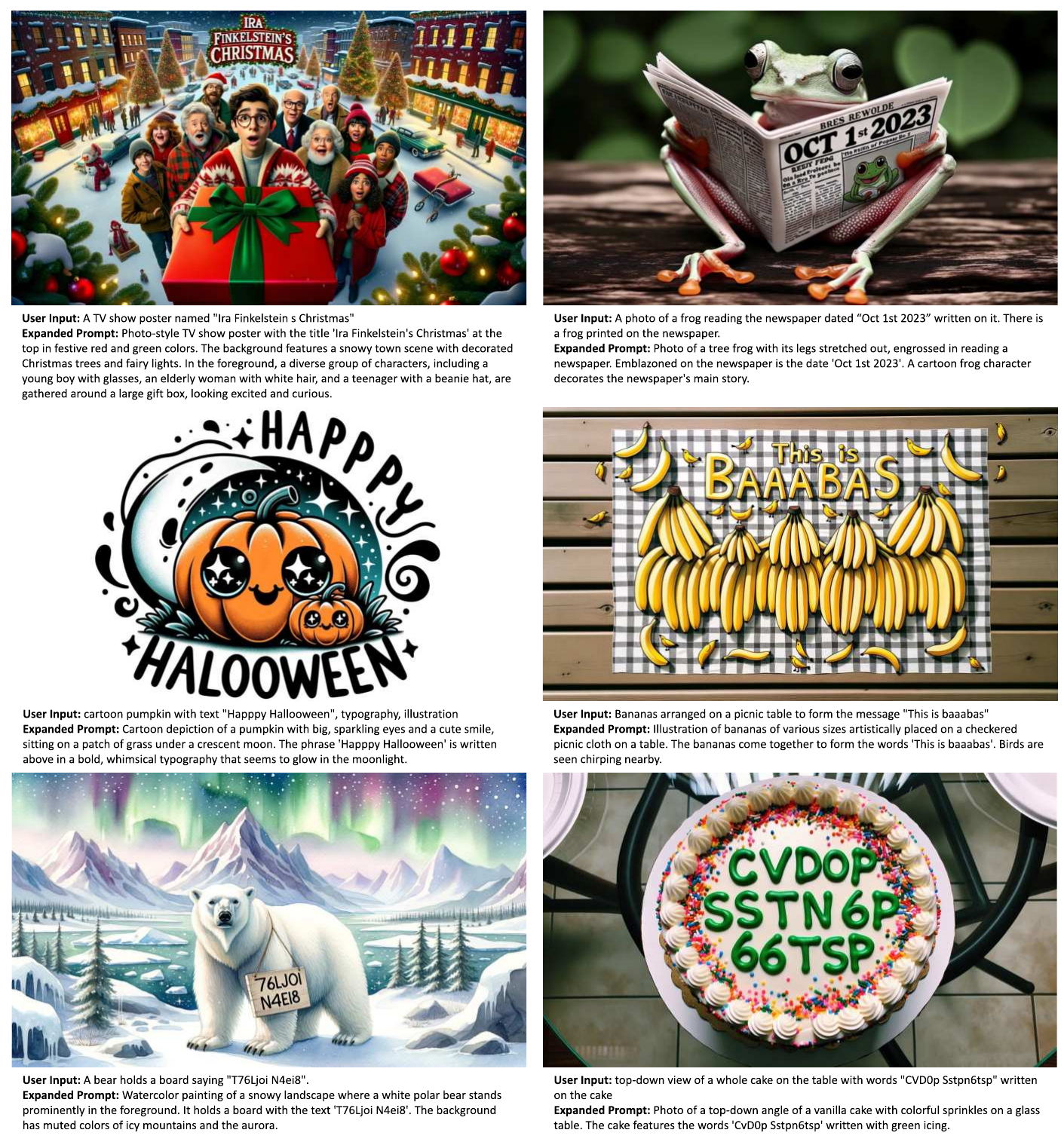}}
\caption[text rendering: low-frequency words]{Rendering text words that are less frequently appeared, \eg, names and dates ``Ira Finkelstein's Christmas'' and ``Oct 1st 2023,'' misspellings ``Happpy Hallooween'' and ``This is baaabas,'' and random strings ``T76Ljoi N4ei8'' and ``CVD0p Sstpn6tsp.''
}
\label{fig:text_lowfreq}
\end{figure*}

Figure~\ref{fig:text_lowfreq} illustrates renderings of infrequently occurring text. This includes misspelled words such as “Happpy Hallooween” and “Baaabas,” and random character sequences like “CVD0p Sstpn6tsp”.

\begin{figure*}[h!]
\centering
\vspace{-20mm}
\centerline{\includegraphics[width=1.5\textwidth]{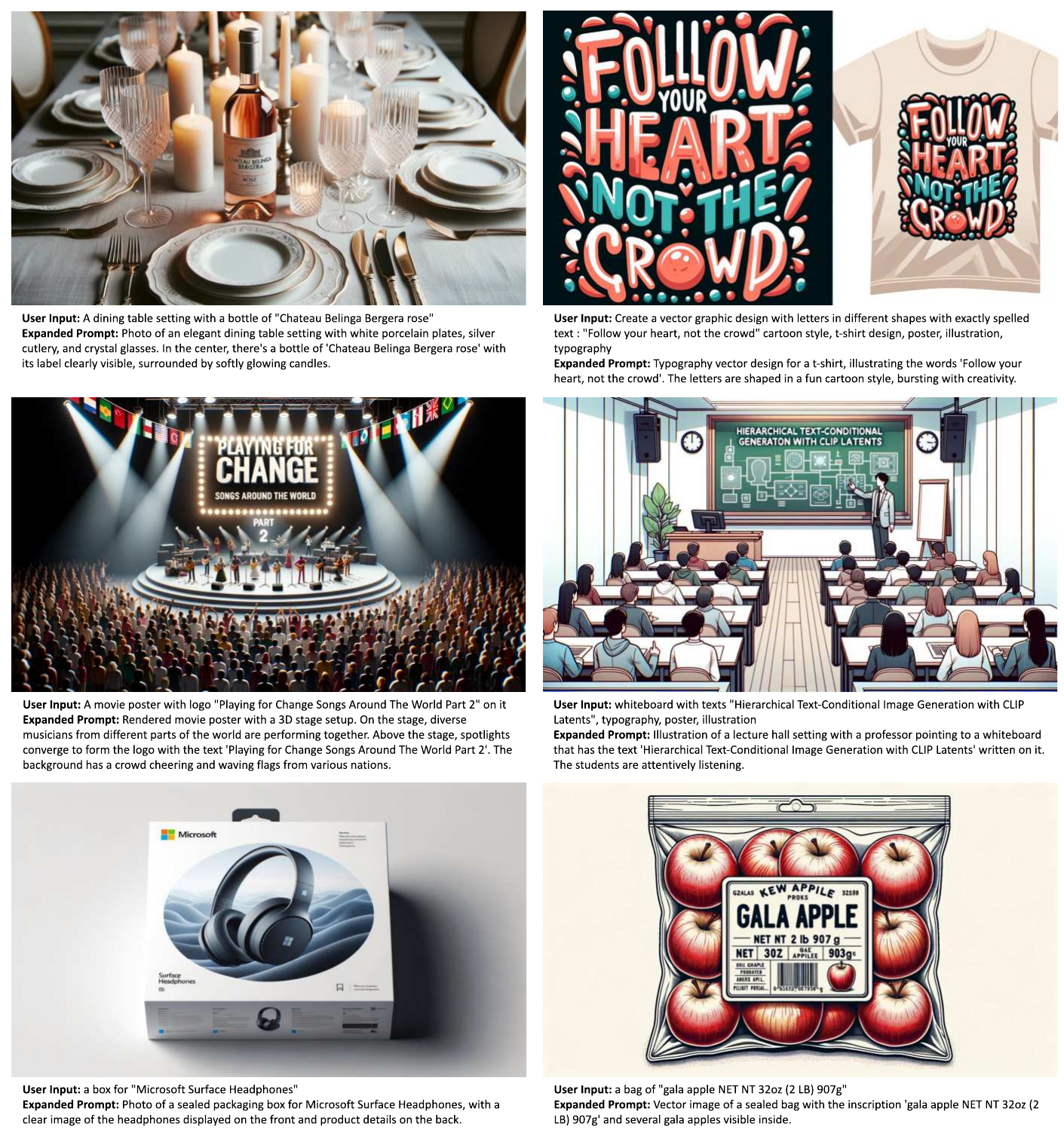}}
\caption[text rendering: long text]{Rendering longer pieces of texts, \eg, on bottles and t-shirts ``Chateau Belinga Bergera rose'' and ``Follow your heart, not the crowd,'' on posters and boards ``Playing for Change Songs Around The World Part 2'' and ``Hierarchical Text-Conditional Image Generation with CLIP Latents,'' and on product packages ``Microsoft Surface Headphones'' and ``gala apple NET NT 32oz (2 LB) 907g.''
}
\label{fig:text_long}
\end{figure*}

Figure~\ref{fig:text_long} showcases renderings of extended text passages. For instance, “Hierarchical Text-Conditional Image Generation with CLIP Latents.” The compound text “gala apple NET NT 32oz (2 LB) 907g” poses a unique challenge with its amalgamation of words, numerals, and units. Yet, \modelname produces a layout reminiscent of a store price tag.

Effective typography is more than accurate spelling~\cite{bringhurst2004elements}. Font selection is vital, needing alignment with content and medium. The choice between serif and sans-serif hinges on communication context. Font size is key, with hierarchy distinguishing headings, subheadings, and body text for clarity and visual definition. Figure~\ref{fig:text_comic_2} and ~\ref{fig:text_comic_3} depict the rendering Pusheenish font in the dialogue balloons. Figure~\ref{fig:text_poster} showcases the font hierarchy rendering in sophisticated posters.

For clear visuals, colors must contrast well with the background and convey intended emotions. Uniform alignment ensures a cohesive, organized text presentation.  Figure~\ref{fig:text_GUI} displays various font colors in GUI design, while Figure~\ref{fig:text_storybook} showcases \modelname's alignment capabilities in creating storybook design.

When these facets converge cohesively, typography elevates from a mere conveyance of information to a medium that enhances design aesthetics and user engagement. The ``Born Pink'' mug in Figure~\ref{fig:scenario_product_2} exemplifies this, seamlessly blending handwritten and printed styles, harmonized by color and lighting choices.

\subsection{Layout and Composition}

Creating a compelling layout and composition in design demands a keen understanding and strategic implementation of several key elements~\cite{samara2023making}, ensuring that the visual space effectively communicates and resonates with the viewer.

\begin{figure*}[h!]
\centering
\vspace{-20mm}
\centerline{\includegraphics[width=1.5\textwidth]{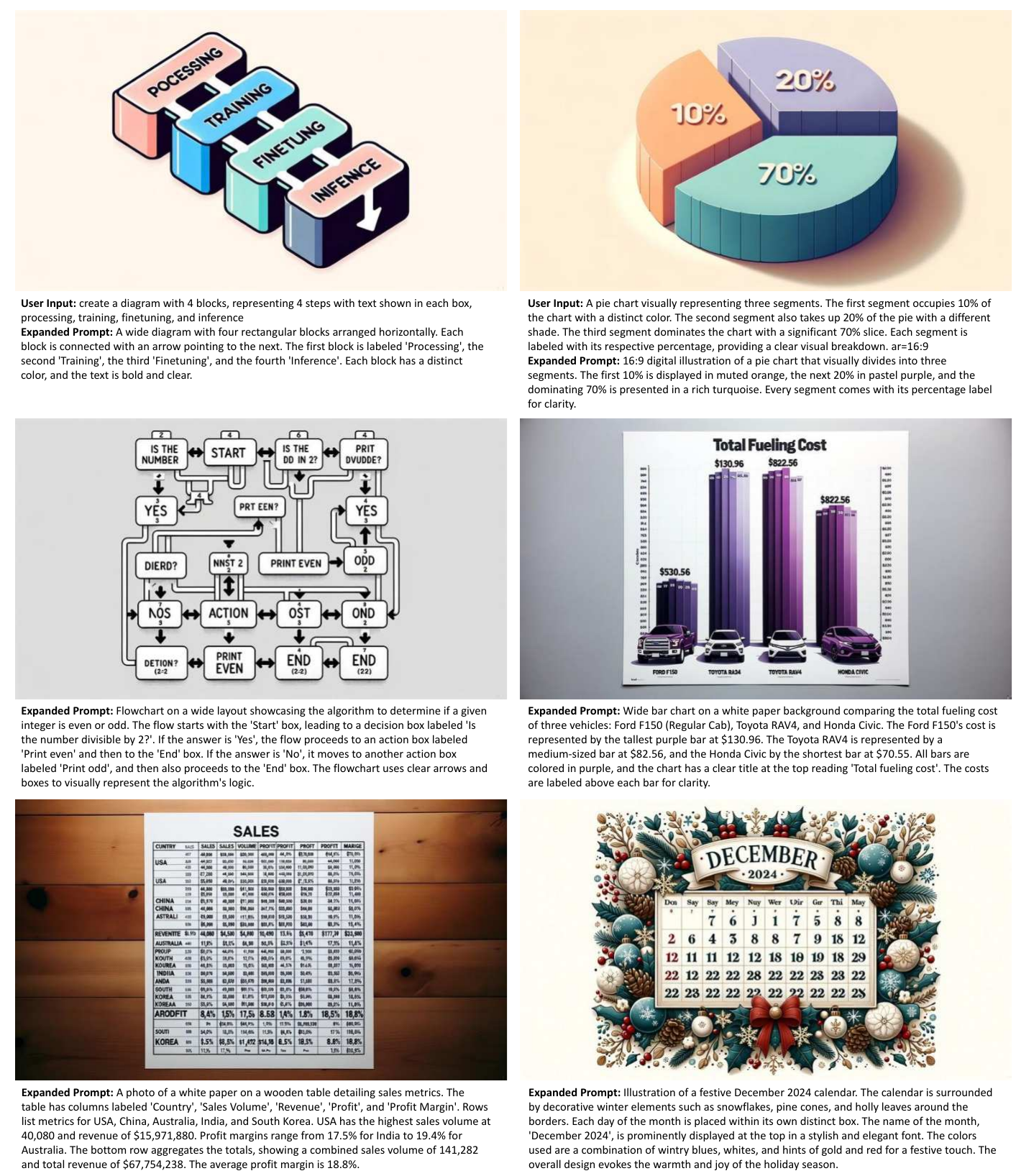}}
\caption[layout and composition: diagram, chart, table, calendar]{Layout and composition capability: diagram, flowchart, bar chart, pie chart, table, and calendar.   
}
\label{fig:capability_layout_composition_1}
\end{figure*}

Figure~\ref{fig:capability_layout_composition_1} displays layouts including block diagrams, pie charts, flow charts, bar graphs, tables, and calendars. While \modelname generally crafts decent layouts, it sometimes struggles with intricate details.

\begin{figure*}[h!]
\centering
\vspace{-20mm}
\centerline{\includegraphics[width=1.5\textwidth]{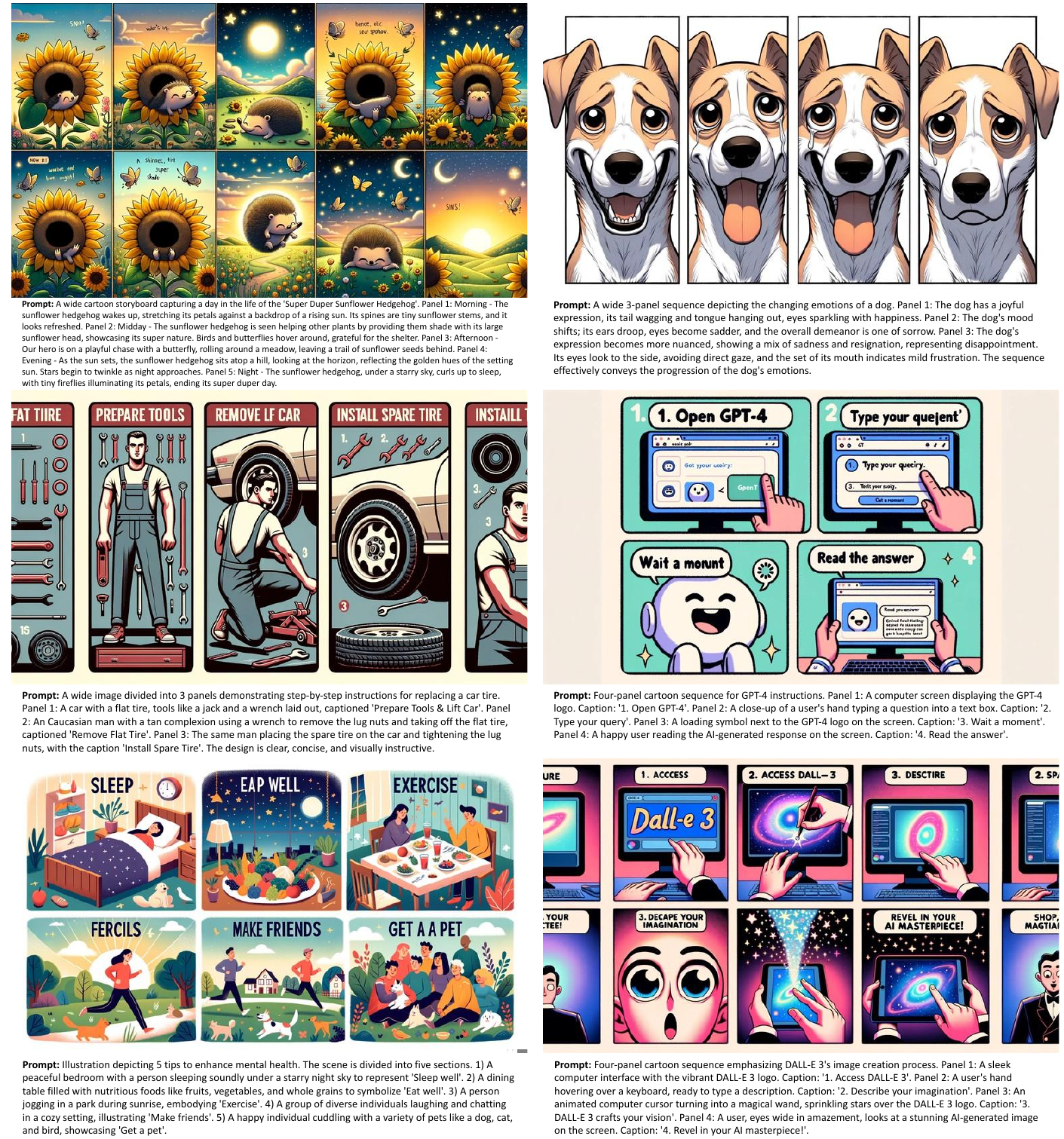}}
\caption[layout and composition: multi-panel layout]{Multi-panel motion graphics.
}
\label{fig:motion_graphics}
\end{figure*}

Figure~\ref{fig:motion_graphics} illustrates multi-panel layouts such as storyboards, how-tos, memes, and comics. Consistency in elements, colors, and patterns is vital in multi-panel designs to unify the composition and guide viewers. Designers utilize flow and movement, directing the viewer's eye using lines and element arrangements, to ensure a seamless experience.

\subsection{Color Harmony}
Color harmony is a vital principle in design that ensures various colors in a composition create a cohesive, aesthetically pleasing experience for the viewer~\cite{Patti2013colortheory, Faber2013colorpsychology}. A harmonious color palette can evoke specific emotions, set the tone, and enhance the overall impact of a piece.

\begin{figure*}[h!]
\centering
\vspace{-20mm}
\centerline{\includegraphics[width=1.5\textwidth]{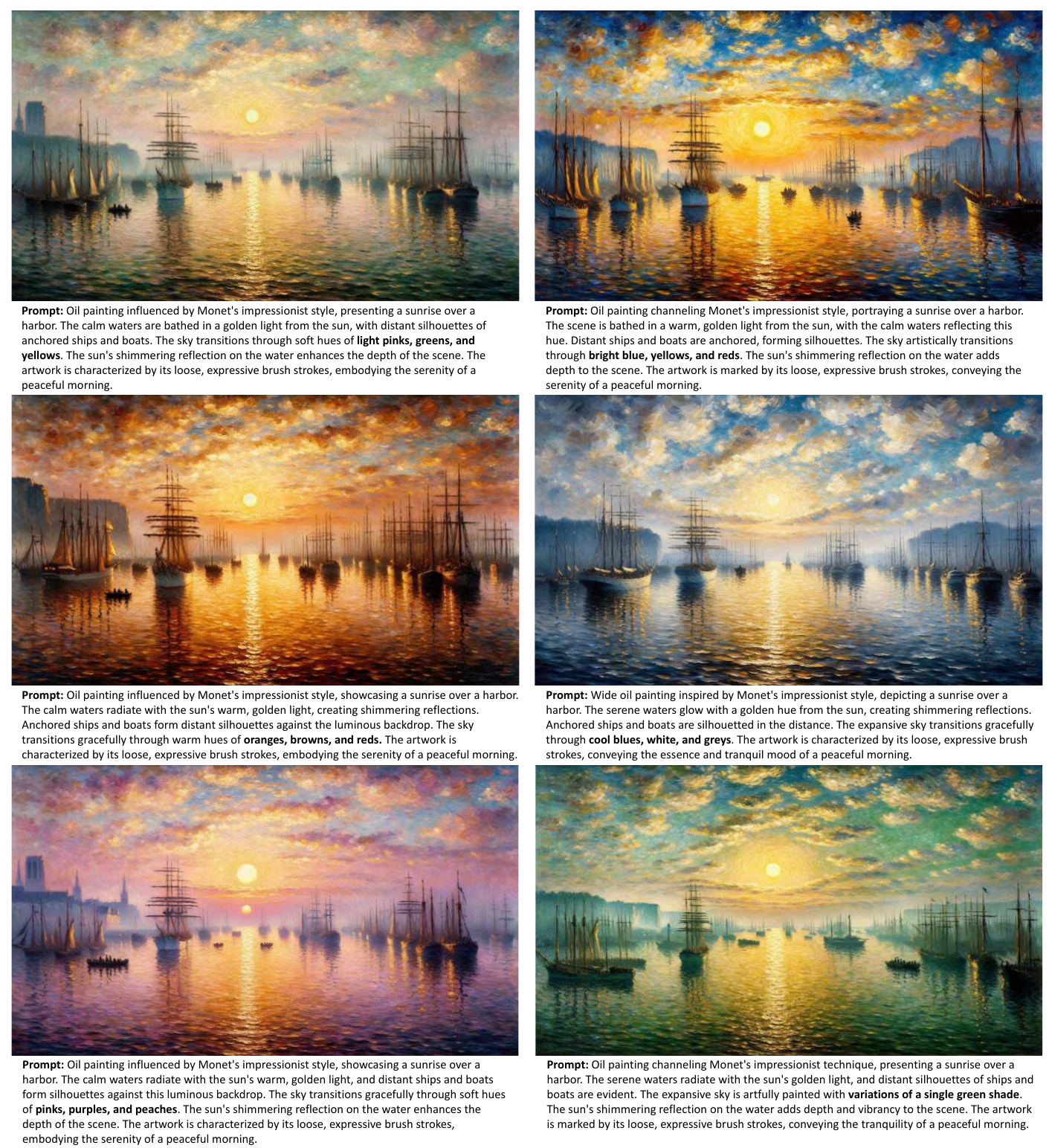}}
\caption[color harmony: impression sunrise]{Different color palettes used for "Impression Sunrise": Spring, Summer, Autumn, Winter, Romantic, and Monochronmatic with a single green shade.   
}
\label{fig:color_variants_2}
\end{figure*}

Figure~\ref{fig:color_variants_2} displays variations of color palettes in oil paintings inspired by ``Impression Sunrise.'' These range from Spring, Summer, Autumn, and Winter Palettes to a Romantic Palette and a monochromatic green shade. This serves as a test to see if DALL-E can adeptly control and render color palettes. \modelname effectively captures the distinct tones associated with different seasons and themes.

\begin{figure*}[h!]
\centering
\vspace{-20mm}
\centerline{\includegraphics[width=1.5\textwidth]{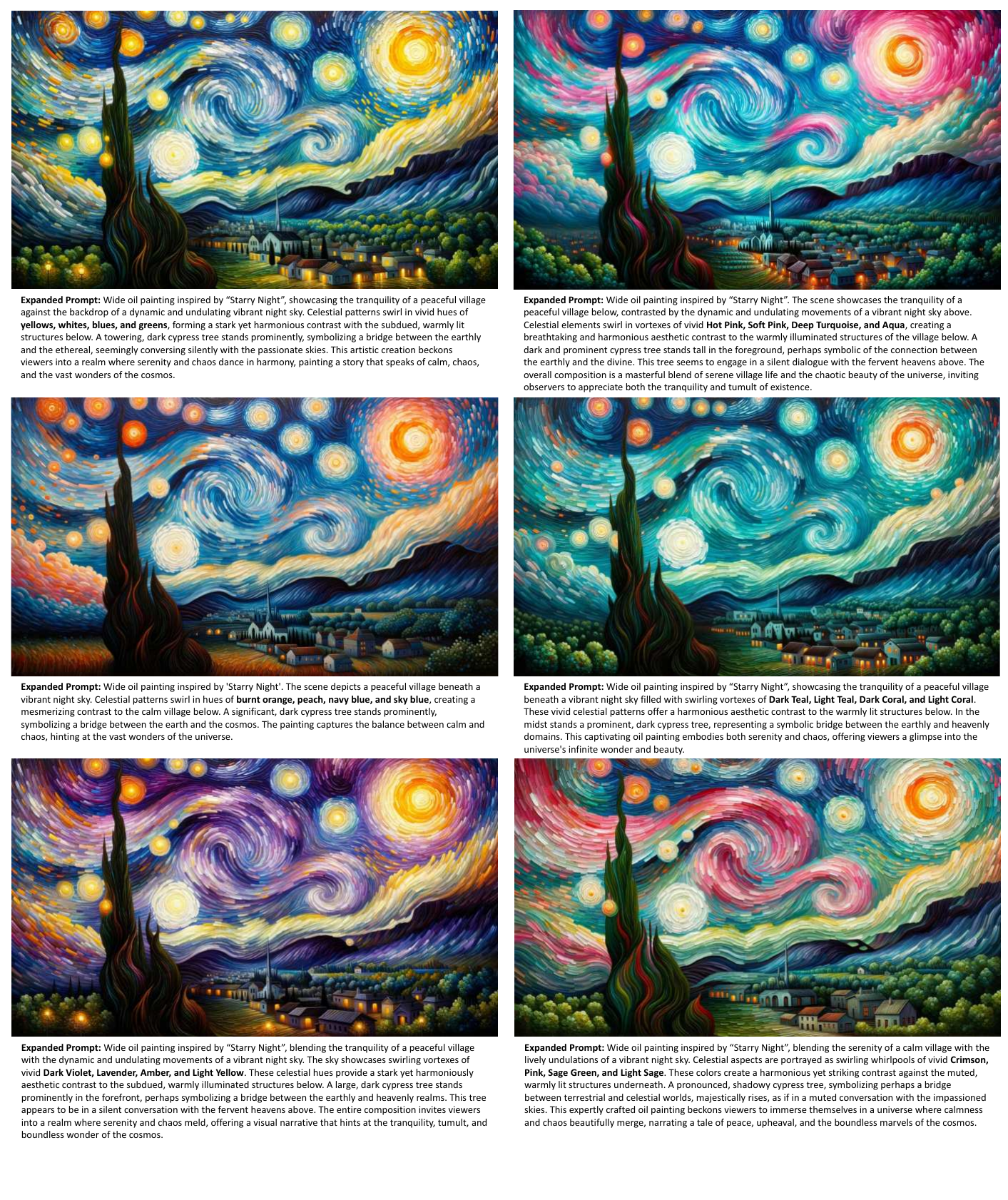}}
\vspace{-10mm}
\caption[color harmony: starry night]{Different complementary color palettes used for oil painting inspired by ``Starry Night:'' Blue and Yellow Palette, Pink and Turquoise Palette, Blue and Orange Palette, Teal and Coral Palette, Violet and Amber Palette, Crimson and Sage Palette.   
}
\label{fig:color_harmony}
\end{figure*}

Figure~\ref{fig:color_harmony} presents six color palette variations in oil paintings, inspired by ``Starry Night,'' testing complementary color harmonies. It's striking how DALL-E captures and renders these vibrant starry scenes with such vitality and beauty.

\subsection{Medium and Style}

The artistic medium and style are crucial in visual graphic design~\cite{lupton2008graphic}, defining the work's expressive potential and emotional resonance. The medium, encompassing the tools, materials, or digital platforms employed, sets the boundaries and opportunities for expression, shaping the tactile and sensory experiences of the audience. 

\begin{figure*}[h!]
\centering
\vspace{-20mm}
\centerline{\includegraphics[width=1.4\textwidth]{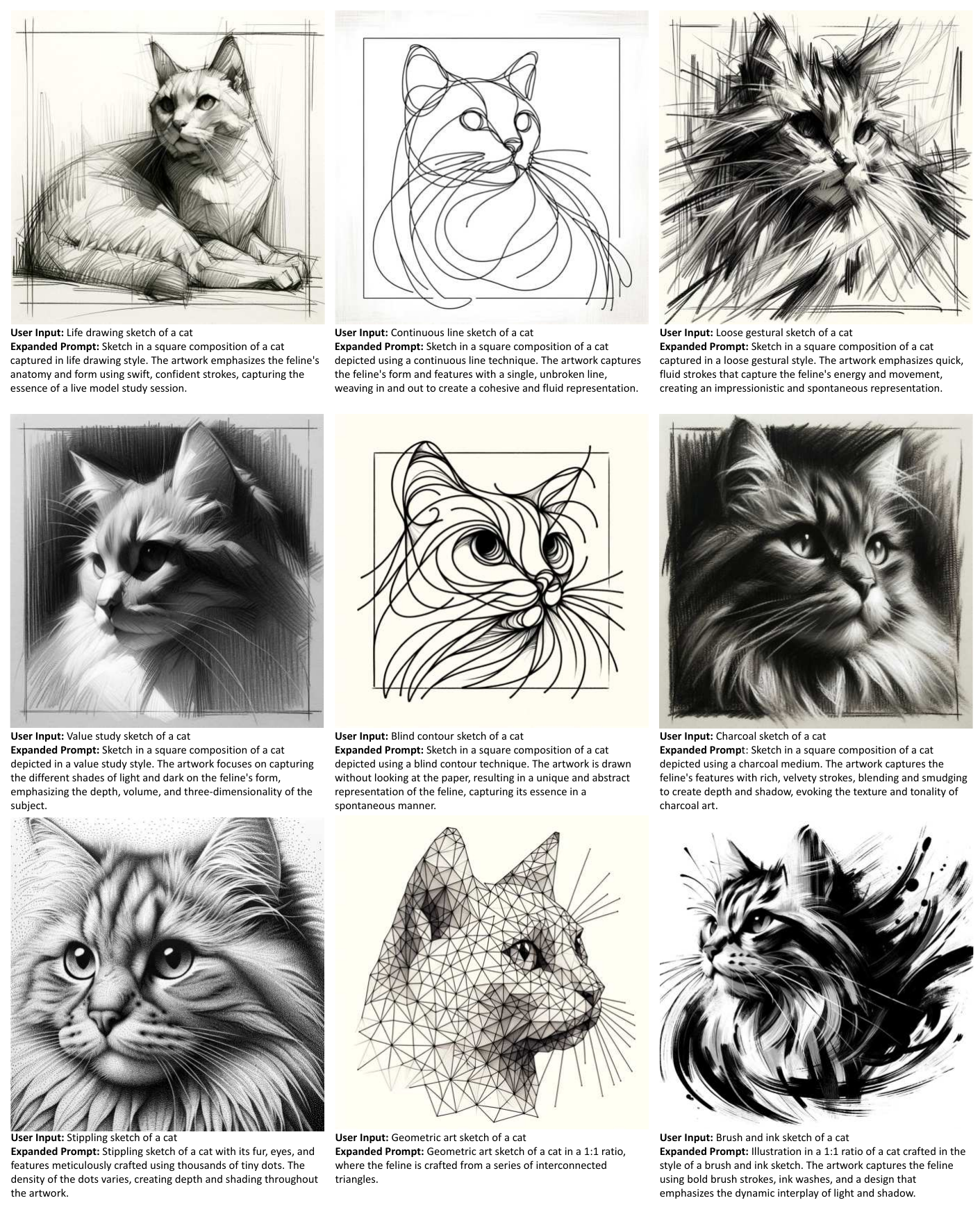}}
\caption[medium and style: cats 1]{Sketch with a different style. Prompts inspired by \url{https://docs.midjourney.com/docs/explore-prompting-1}. %
}
\label{fig:material_style_design_3}
\end{figure*}

\begin{figure*}[h!]
\centering
\vspace{-20mm}
\centerline{\includegraphics[width=1.4\textwidth]{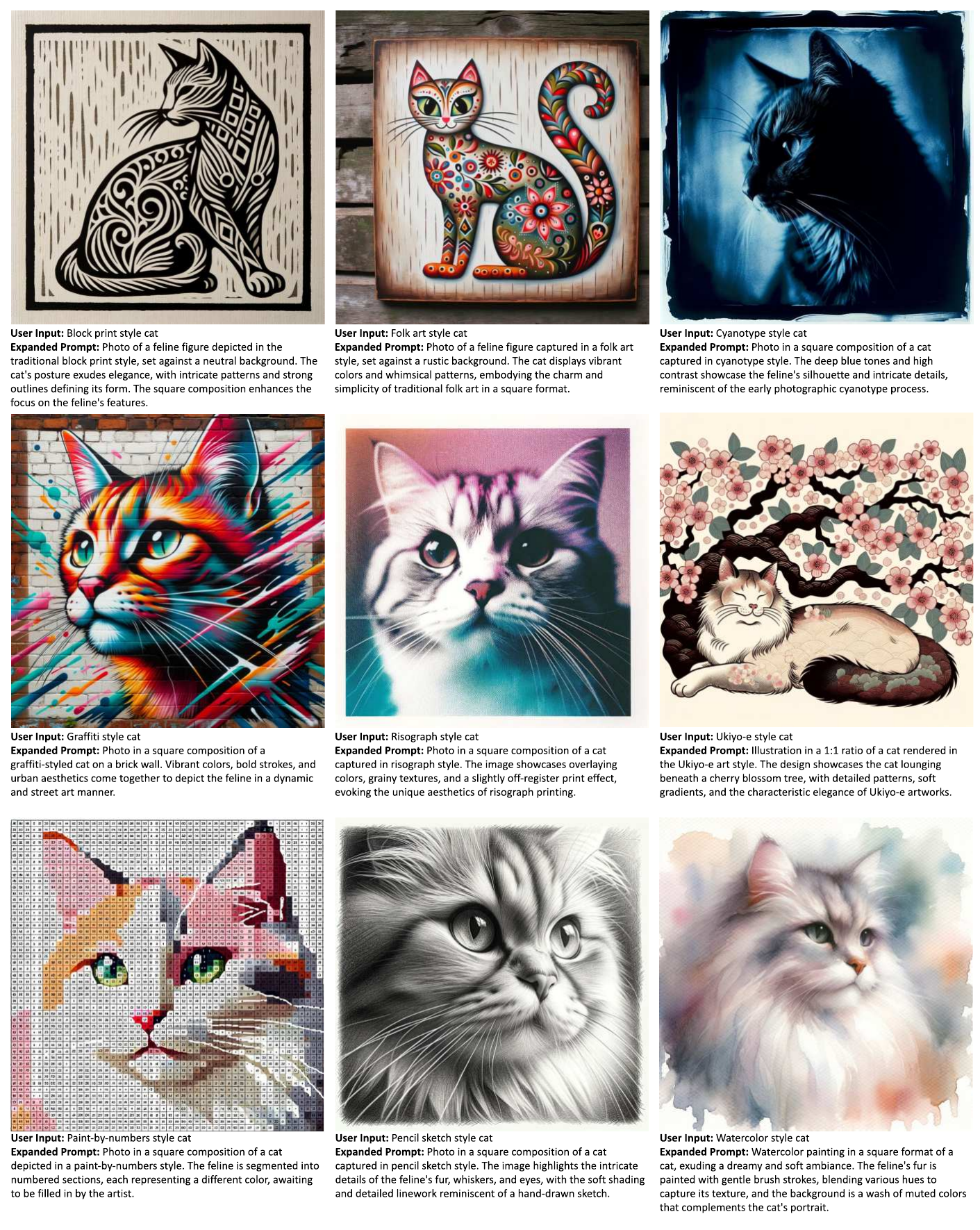}}
\caption[medium and style: cats 2]{Specifying an artistic medium. Prompts inspired by \url{https://docs.midjourney.com/docs/explore-prompting-1}. %
}
\label{fig:material_style_design_1}
\end{figure*}

\begin{figure*}[h!]
\centering
\vspace{-20mm}
\centerline{\includegraphics[width=1.4\textwidth]{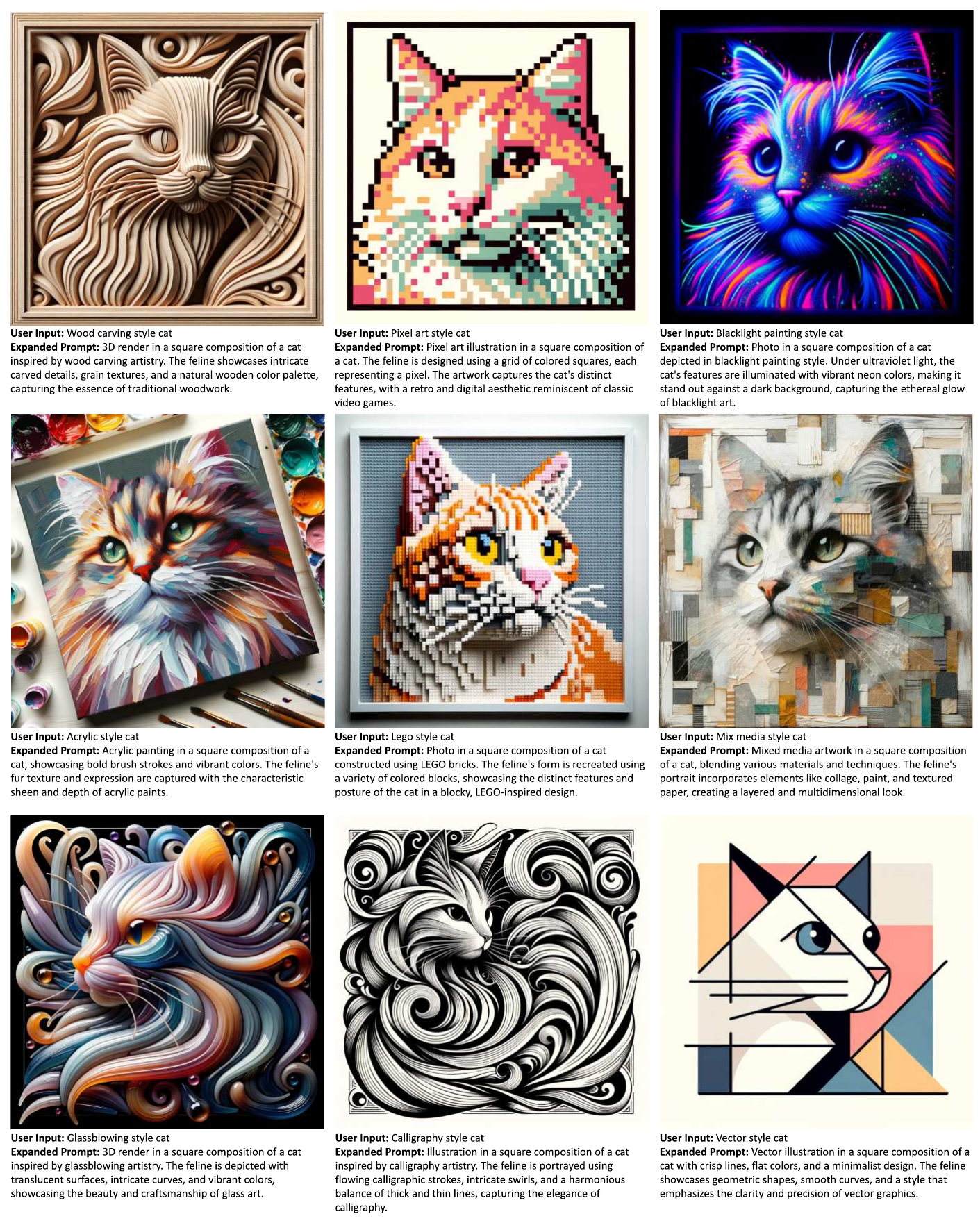}}
\caption[medium and style: cats 3]{Specifying an artistic medium. Prompts inspired by \url{https://docs.midjourney.com/docs/explore-prompting-1}. %
}
\label{fig:material_style_design_2}
\end{figure*}

Figure~\ref{fig:material_style_design_3} shows examples of sketching a cat in different styles including continuous line drawing, charcoal sketch, stippling sketch, brush and ink sketch, etc. Figure~\ref{fig:material_style_design_1} and ~\ref{fig:material_style_design_2} demonstrate the capability of specifying different art media, including block print, folk art, paint-by-numbers, watercolor wood carving, Lego style, glass blowing, calligraphy, \etc These examples are just a small set of the art styles and media that \modelname covers. They provide a glimpse of \modelname's capability of rendering with a broad range of artistic media and styles.

\subsection{3D and Cinematography}

3D rendering~\cite{van2020blender} and cinematography~\cite{brown2016cinematography} are transformative tools in the world of visual representation, allowing for the creation of intricate, lifelike scenes and stories. The depth, perspective, and dynamism brought about by these techniques offer a multi-dimensional view, enhancing the viewer's experience and immersion.

\begin{figure*}[h!]
\centering
\vspace{-20mm}
\centerline{\includegraphics[width=1.4\textwidth]{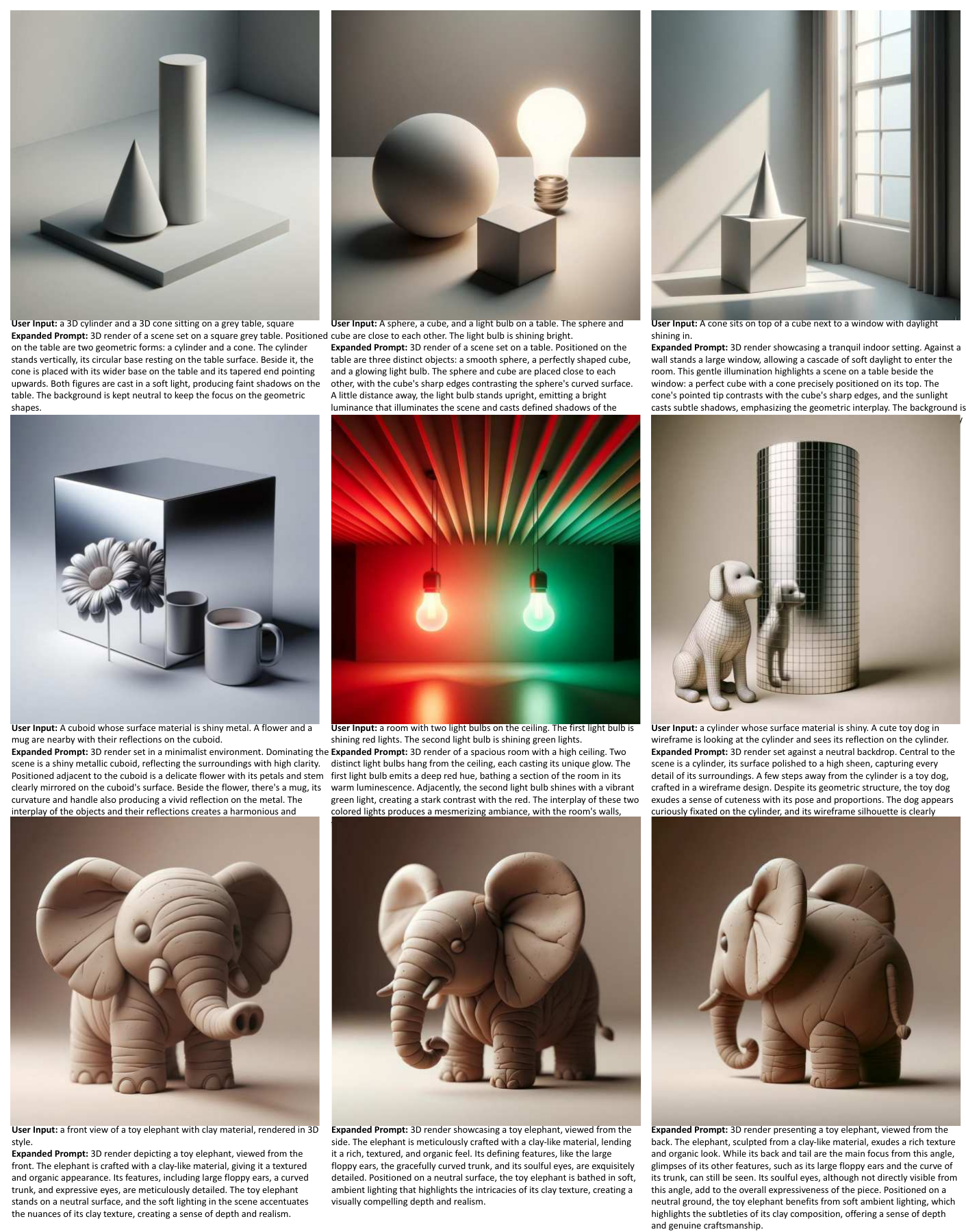}}
\caption[3D and cinematography: shape and lighting]{3D core capabilities.
}
\label{fig:capability_3d}
\end{figure*}

Figure~\ref{fig:capability_3d} shows examples of 3D rendering, including basic shapes, spatial relationships, lighting effects, shadow, reflections, and various viewing angles. \modelname proficiently captures self-shadows and cast shadows and effectively manages reflections on both flat and curved surfaces.  The transition between two light sources is smooth. We find that \modelname sometimes does not follow view angles precisely. For example, the front view rendering is noticeably off. 

\begin{figure*}[h!]
\centering
\vspace{-20mm}
\centerline{\includegraphics[width=1.5\textwidth]{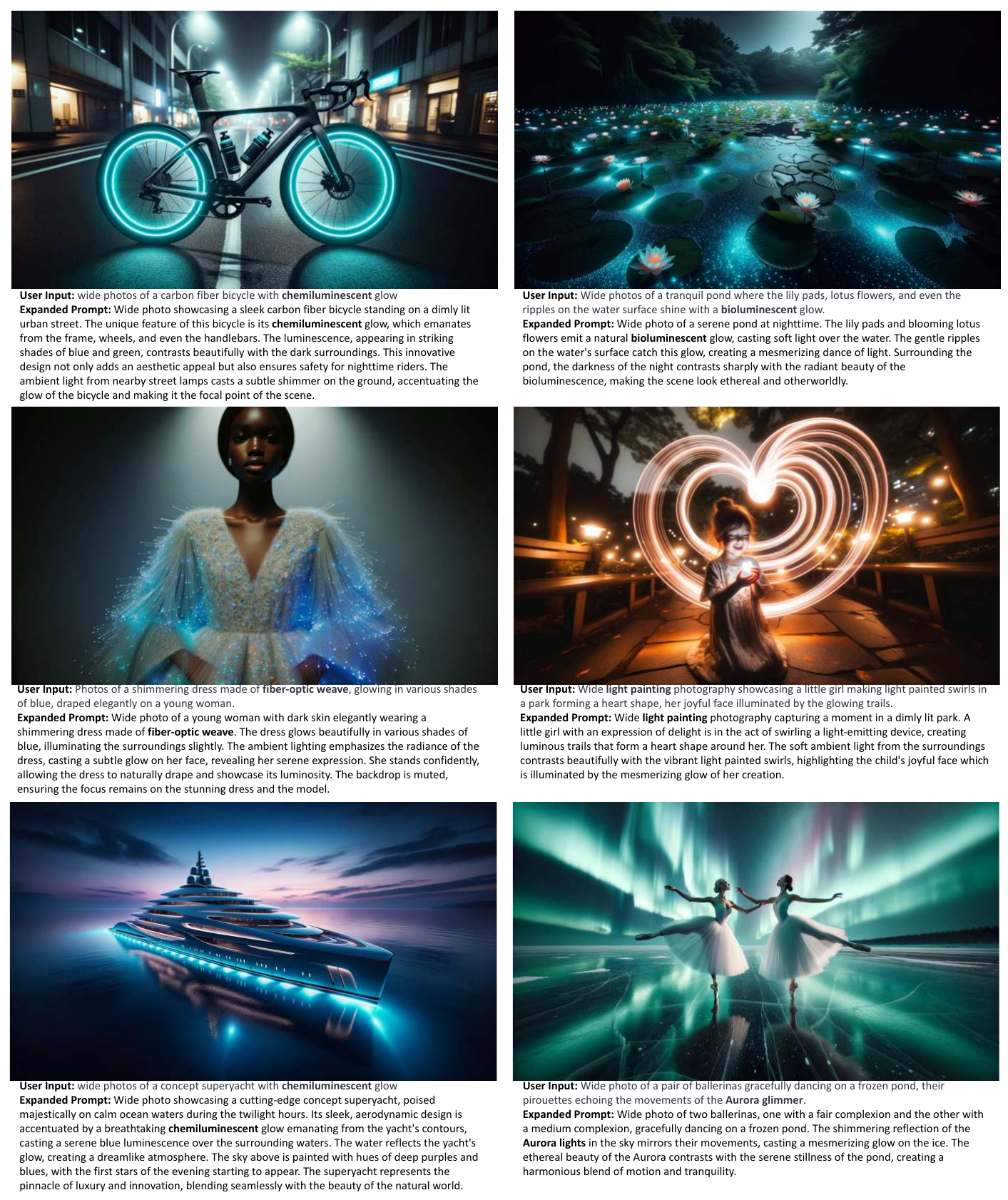}}
\caption[3D and cinematography: lighting effect]{Special lighting effects. Prompts inspired by \url{https://juejin.cn/post/7292354382909046796}.
}
\label{fig:capability_lighting-effect}
\end{figure*}

In Figure~\ref{fig:capability_lighting-effect}, we show DALL-E 3’s capabilities of generating special lighting effects including chemiluminescent glow, bioluminescent glow, light-painting, and Aurora glimmering. 

\begin{figure*}[h!]
\centering
\vspace{-20mm}
\centerline{\includegraphics[width=1.5\textwidth]{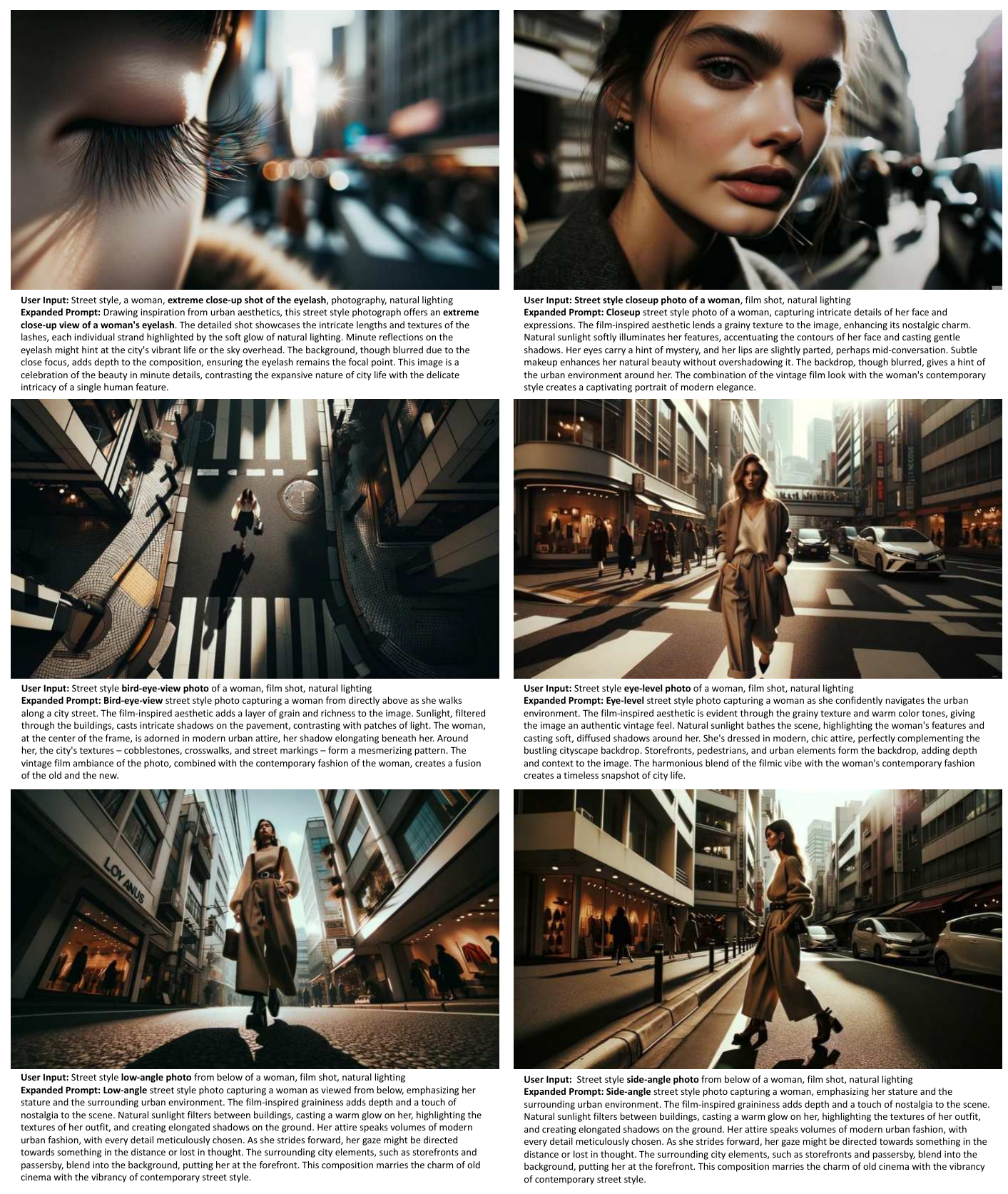}}
\caption[3D and cinematography: camera view points]{Camera angles, proximity, and positions.
}
\label{fig:camera_view_1}
\end{figure*}

Figure~\ref{fig:camera_view_1} shows different camera angles and positions, including closeups, bird-eye level, low and side angles. For close-up shots, \modelname blurs the background appropriately to enhance the scene depth and puts the focus on the foreground.

\begin{figure*}[h!]
\centering
\vspace{-20mm}
\centerline{\includegraphics[width=1.5\textwidth]{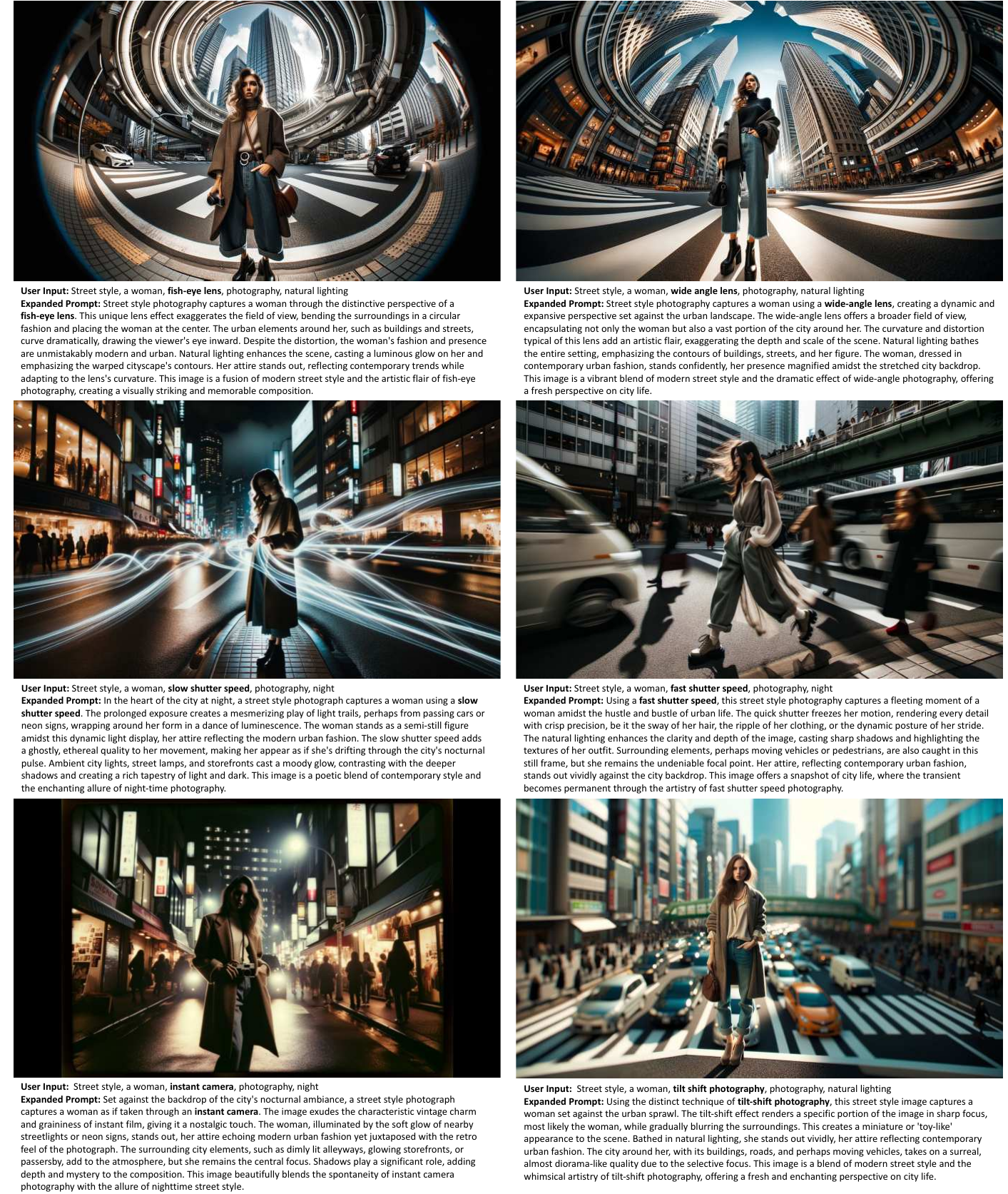}}
\caption[3D and cinematography: camera settings and lens]{Camera settings and lens.
}
\label{fig:camera_setting_1}
\end{figure*}

Figure~\ref{fig:camera_setting_1} shows examples of simulating fisheye and wide angle lenses, slow and faster shutter speeds, instant camera, and tilt shift photography. At the bottom left, \modelname simulates an instant camera whose photos are usually grainy. At the bottom right, DALL-3 simulates tilt-shift photography with the focus on the lady while gradually blurring her surroundings.

\begin{figure*}[h!]
\centering
\vspace{-20mm}
\centerline{\includegraphics[width=1.5\textwidth]{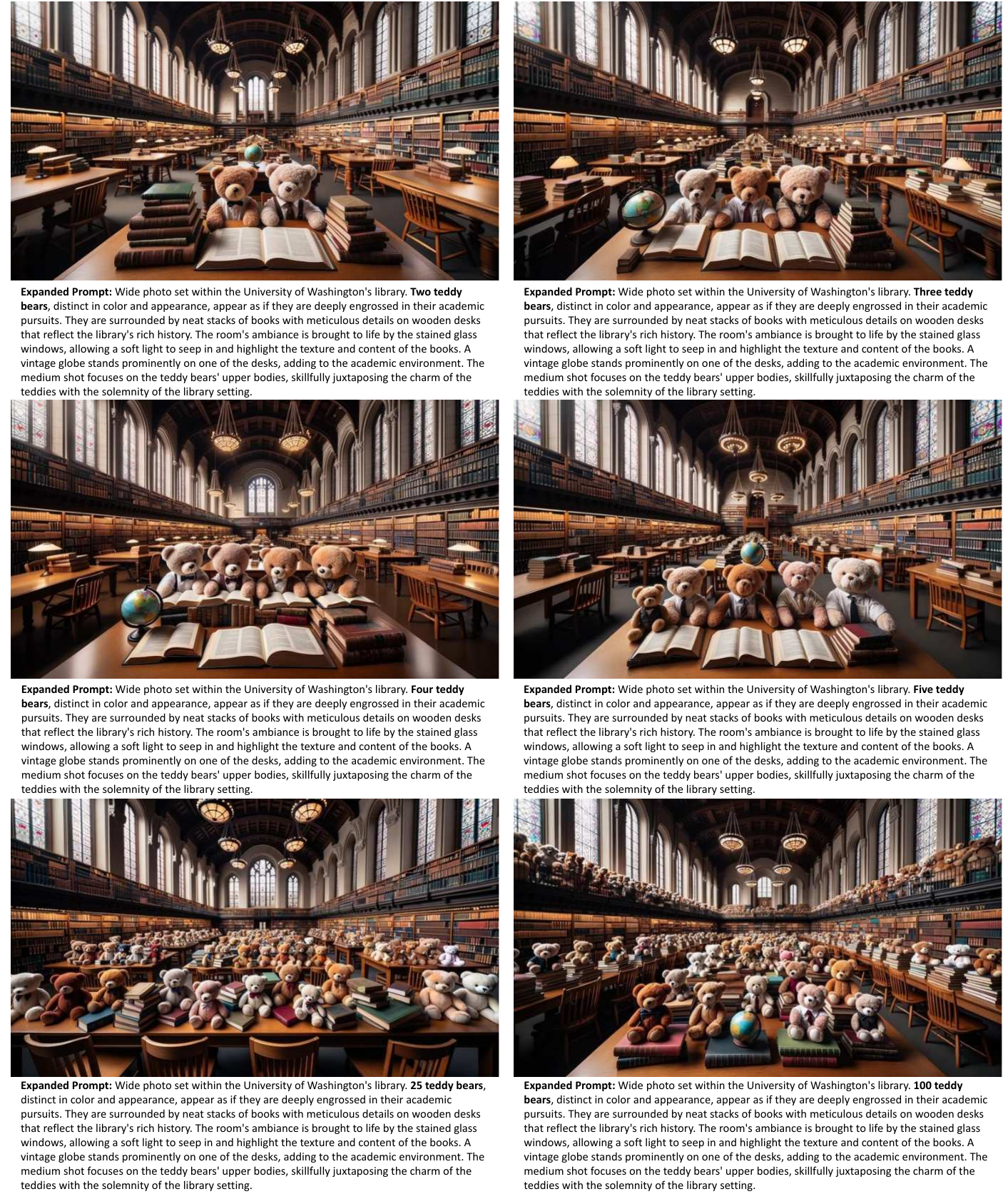}}
\caption[3D and cinematography: crowded scene 1]{Crowded scene.
}
\label{fig:crowd1}
\end{figure*}

\begin{figure*}[h!]
\centering
\vspace{-20mm}
\centerline{\includegraphics[width=1.5\textwidth]{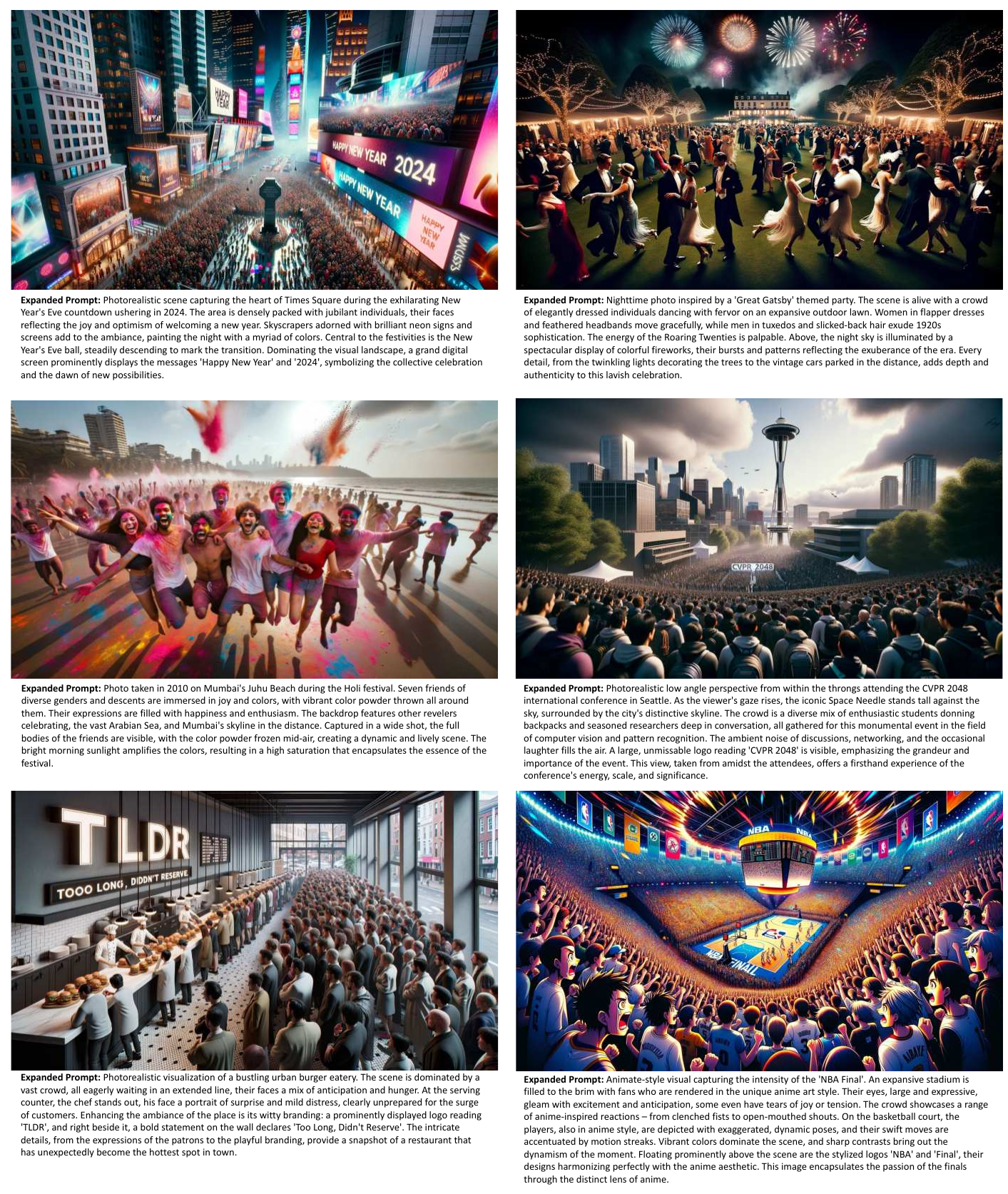}}
\caption[3D and cinematography: crowded scene 2]{Crowded scene.
}
\label{fig:crowd2}
\end{figure*}

Figure~\ref{fig:crowd1} and ~\ref{fig:crowd2} demonstrate \modelname capabilities of rendering crowded scenes. Figure~\ref{fig:crowd1} shows rendering different numbers of bears. \modelname correctly generates the desired number of bears when the number is small. When the number gets larger, however, \modelname makes mistakes (as shown in the last row). Figure~\ref{fig:crowd2} generates images of large human crowds on a variety of occasions. We find that \modelname does a nice job of positioning the texts and rendering them with the correct perspectives. At the bottom left, \modelname generates an exaggerated scene of a popular burger eatery with a super long serving counter and a large waiting crowd. The exaggeration looks plausible and shows the popularity of the burger.


\clearpage
\section{Design Scenario}
\label{sec:05scenario}

In the evolving landscape of design~\cite{pipes2005production, lupton2008graphic, lupton2014thinking}, the prowess of AI models in various design domains has become an area of keen interest. This comprehensive analysis dives into \modelname's capabilities across diverse design spectrums, from the intricacies of infographics and the dynamism of animation and gaming to the finesse required in product design and the artistic nuances in visual art. Each subsection sheds light on specific challenges and achievements of \modelname, presenting a holistic view of its strengths and areas of improvement. Through a series of illustrative figures and descriptions, we unravel the depth and breadth of \modelname's design proficiency, offering insights into its potential and limitations in reshaping the future of design.

\subsection{Infographics Design}

This section delves into \modelname's proficiency across a spectrum of infographic designs, from storybook pages and advertisements to menus, GUIs, movie posters, logos, etc. 

\begin{figure*}[h!]
\centering
\vspace{-20mm}
\centerline{\includegraphics[width=1.5\textwidth]{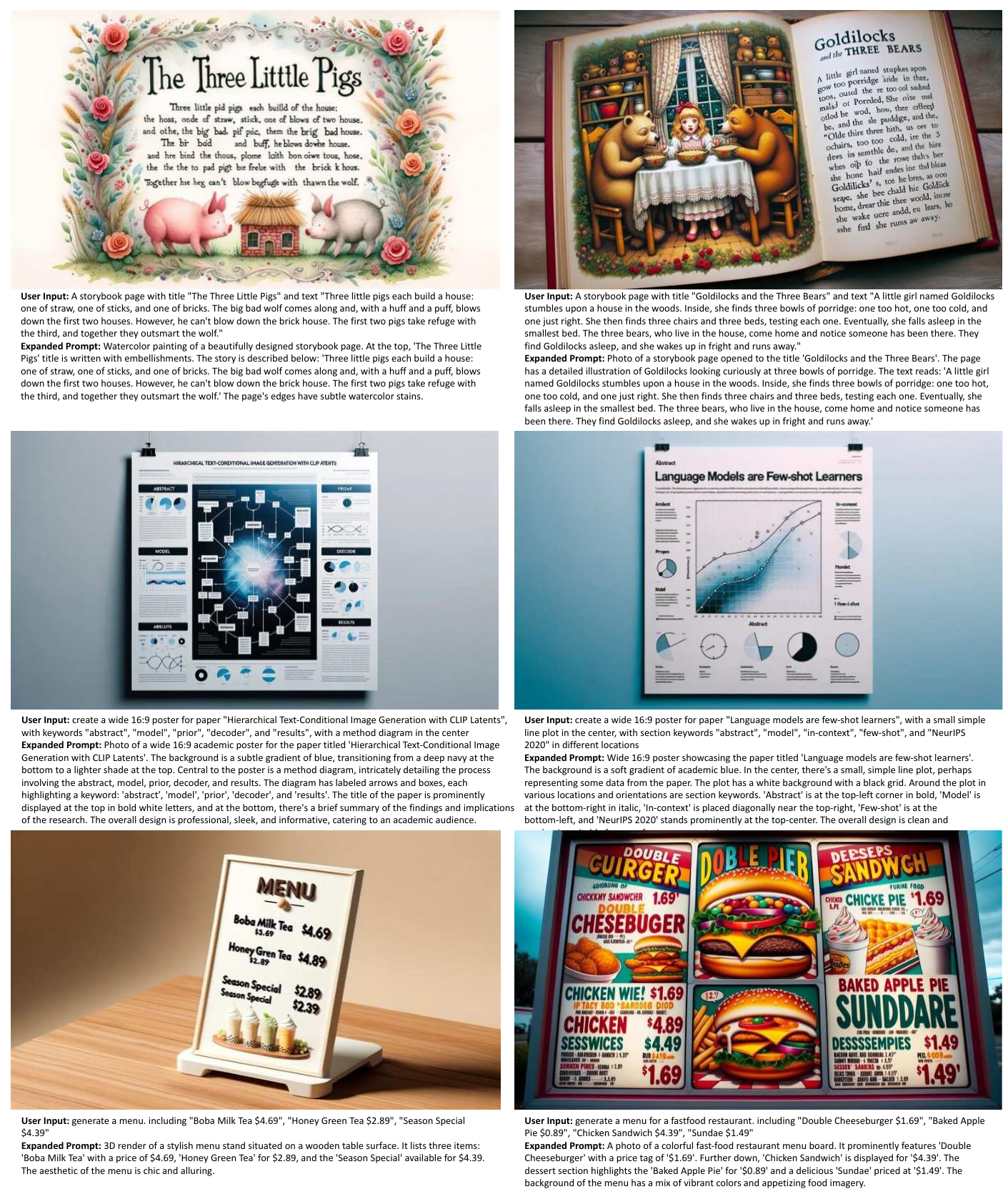}}
\caption[infographics design: storybook, poster, and menu]{Rendering documents such as storybooks, academic posters, and menus.
}
\label{fig:text_storybook}
\end{figure*}

In Figure~\ref{fig:text_storybook}, storybook pages, research posters, and menus are presented. \modelname crafts compelling layouts for each. The storybook pages feature text paragraphs, which is a significant challenge for image generation models. While \modelname struggles with paragraph perfection, individual letters are discernible and many words remain clear.

\begin{figure*}[h!]
\centering
\vspace{-20mm}
\centerline{\includegraphics[width=1.5\textwidth]{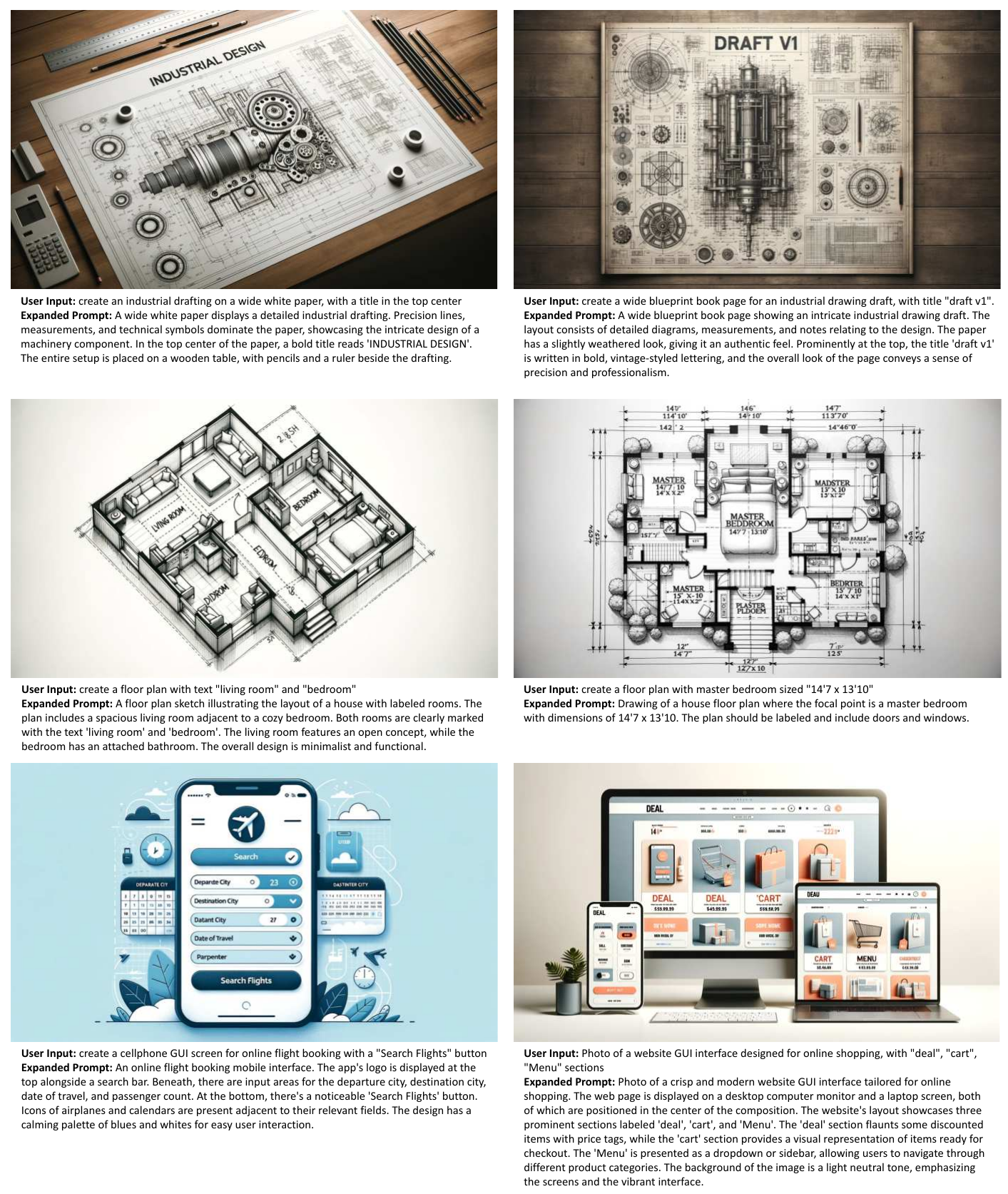}}
\caption[infographics design: industrial drafts, floorplans, and GUI]{Rendering documents such as industrial drafts, floorplans, and GUI.
}
\label{fig:text_GUI}
\end{figure*}

Figure~\ref{fig:text_GUI} showcases industrial design drafts, floor plans, and GUI designs, with \modelname producing commendable text and layout renderings.

\begin{figure*}[h!]
\centering
\vspace{-20mm}
\centerline{\includegraphics[width=1.5\textwidth]{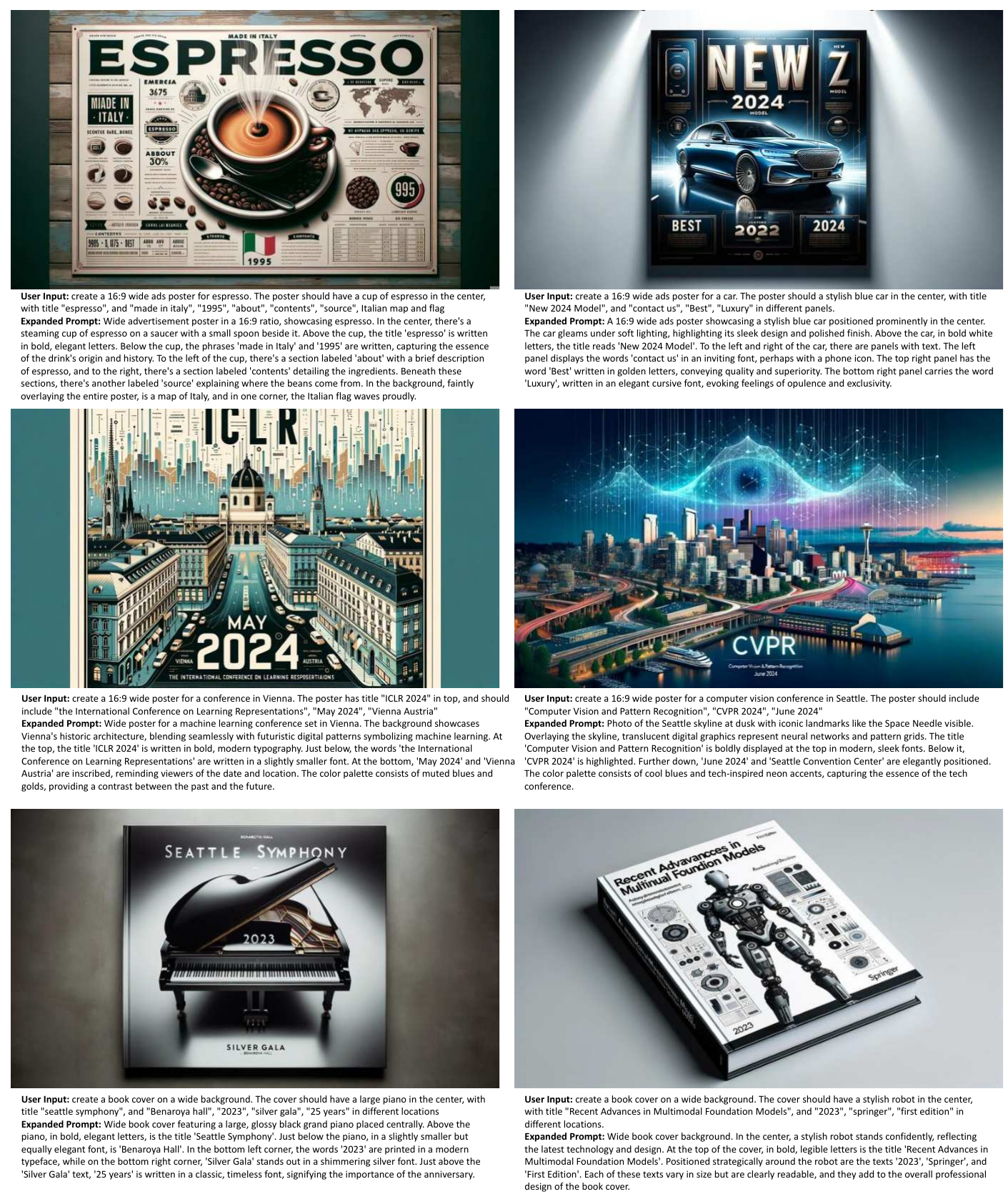}}
\caption[infographics design: ads, marketing posters, and book covers]{Rendering documents such as ads, posters, and book covers.
}
\label{fig:text_poster}
\end{figure*}

Figure~\ref{fig:text_poster}  depicts assorted advertisement posters and book covers, each with varying text, fonts, and sizes. For example, in the two conference posters in the middle row, there are very small texts at the bottom: ``the international conference on learning representation'' and ``Computer Vision and Pattern Recognition.'' It is impressive that \modelname \modelname adeptly renders the minute texts, underscoring its meticulous detailing.

\begin{figure*}[h!]
\centering
\vspace{-20mm}
\centerline{\includegraphics[width=1.5\textwidth]{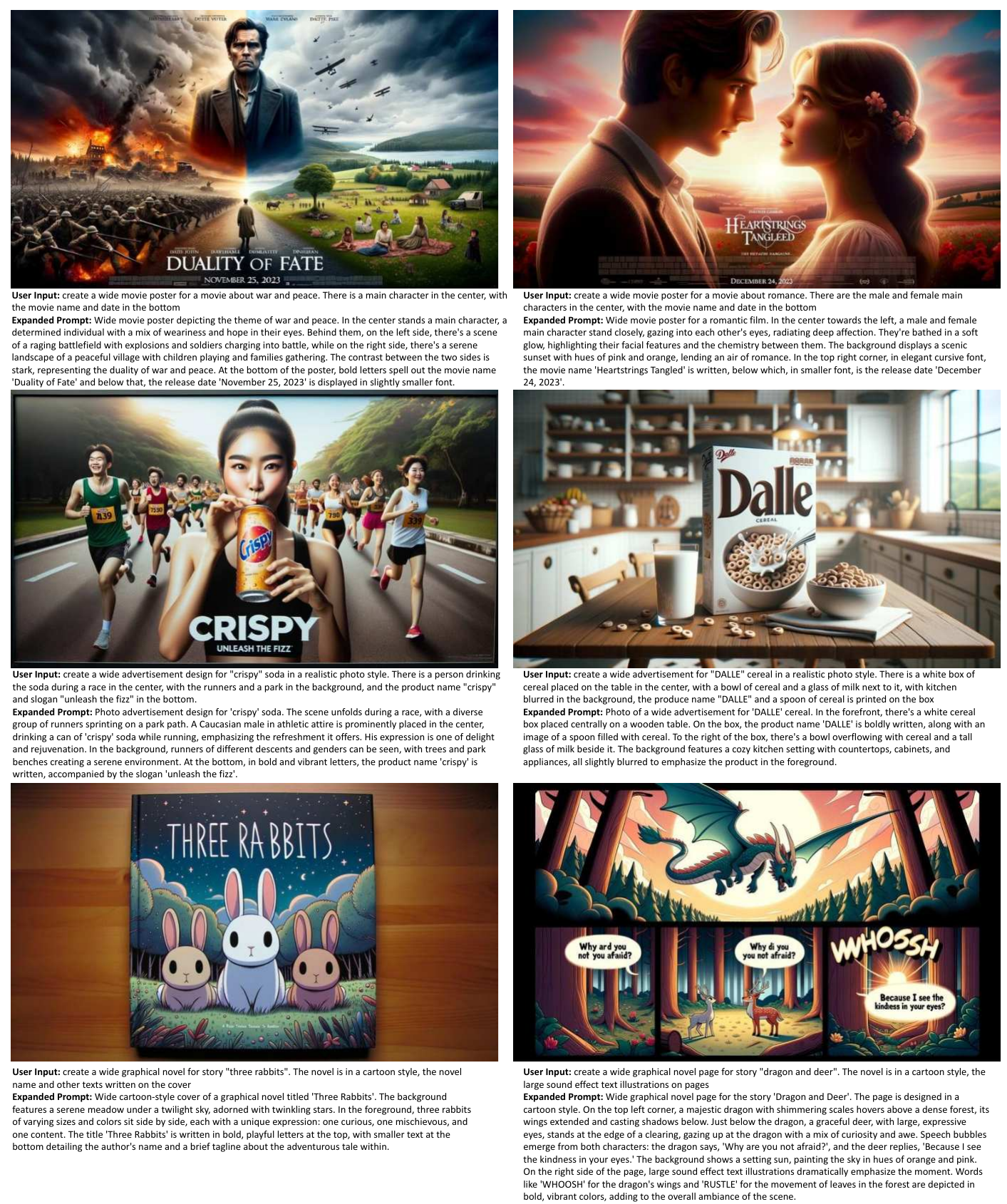}}
\caption[infographics design: movie poster, ads]{Movie poster, Ads.
}
\label{fig:text_ads}
\end{figure*}

Figure~\ref{fig:text_ads} shows movie posters, photorealistic advertisement posters, and cartoon book pages. In the movie poster at the top left, \modelname does a nice job of rendering the main character in a way that smoothly transitions between the two very different color themes. In the advertisement image at the middle left, both the brand name ``crispy'' and the slogan ``unleash the fizz'' are spelled correctly, and their rendering follows the curvature of the soda can surface. In addition, the can that the person is holding has the same look as the ``Crispy'' soda.

\begin{figure*}[h!]
\centering
\vspace{-20mm}
\centerline{\includegraphics[width=1.5\textwidth]{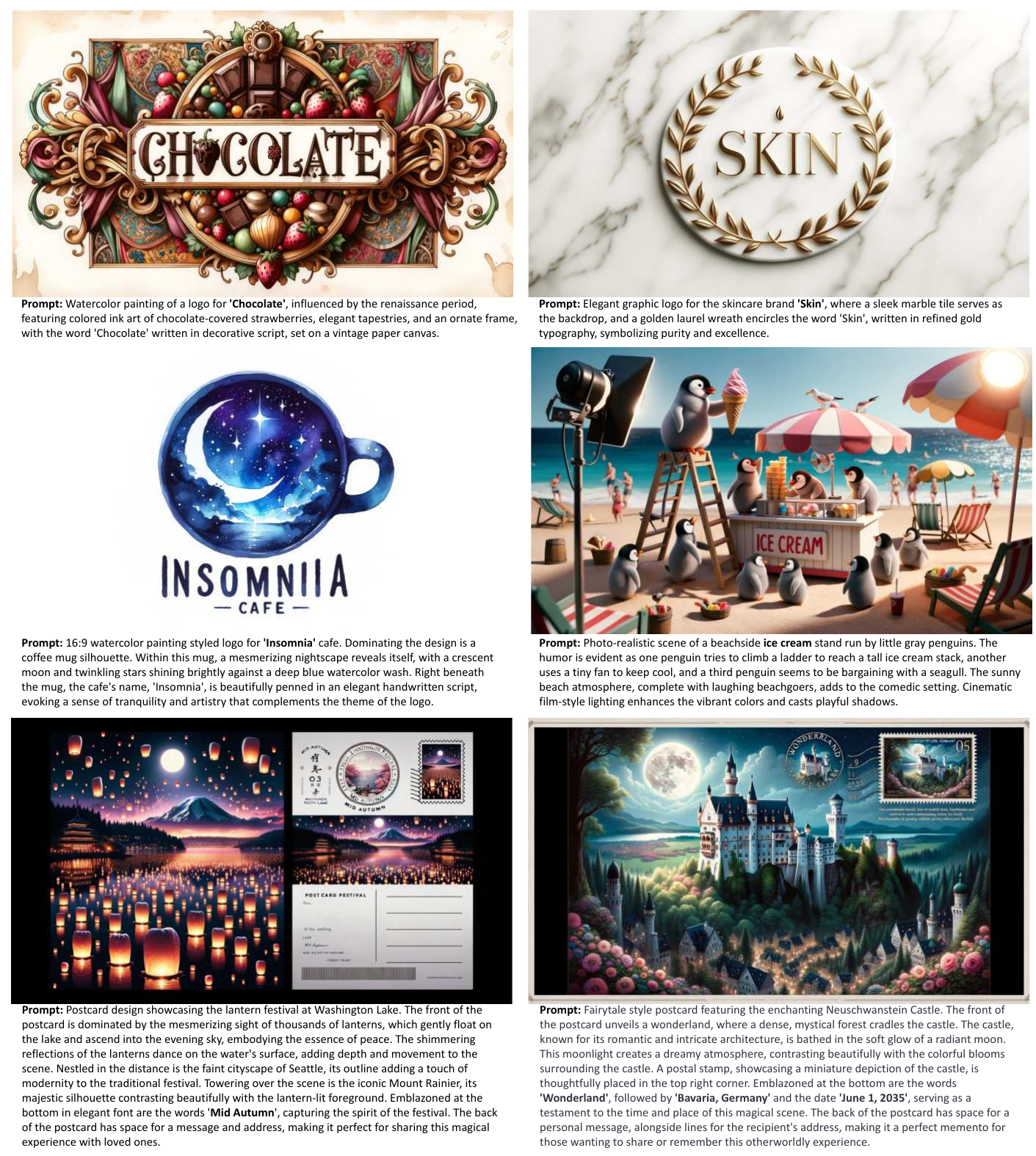}}
\caption[infographics design: logo and postcards]{Logo and postcard.  
}
\label{fig:scenario_logo_postcard_1}
\end{figure*}

\begin{figure*}[h!]
\centering
\vspace{-20mm}
\centerline{\includegraphics[width=1.5\textwidth]{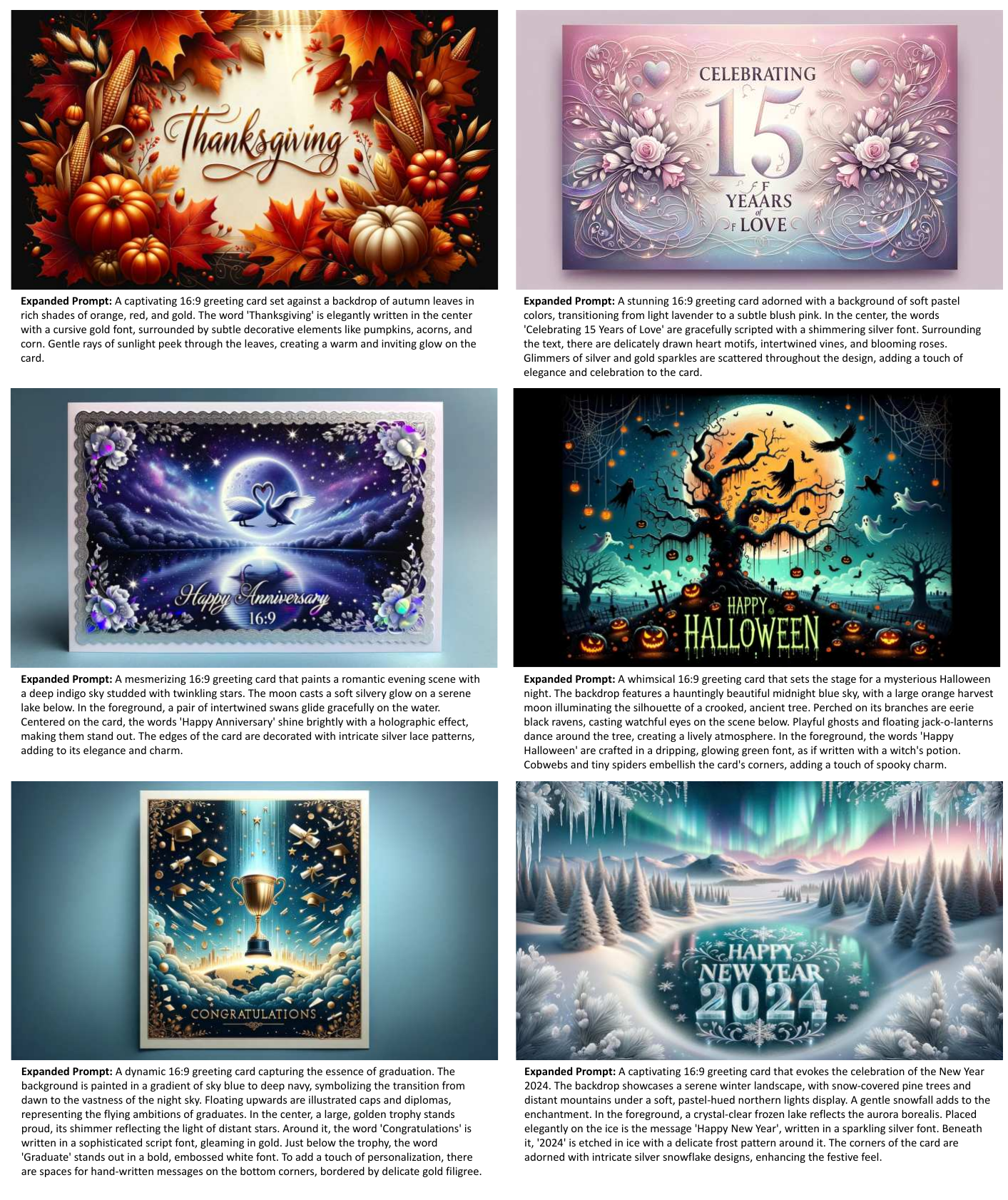}}
\vspace{-2mm}
\caption[infographics design: greeting cards]{Greeting card.  
}
\label{fig:Scenario_greetingcard}
\end{figure*}

Figure~\ref{fig:scenario_logo_postcard_1} and ~\ref{fig:Scenario_greetingcard} offer glimpses into logo designs, postcards, and themed greeting cards. Logos are sleek, while greeting cards aptly capture seasonal and cultural nuances.

\begin{figure*}[h!]
\centering
\vspace{-20mm}
\centerline{\includegraphics[width=1.5\textwidth]{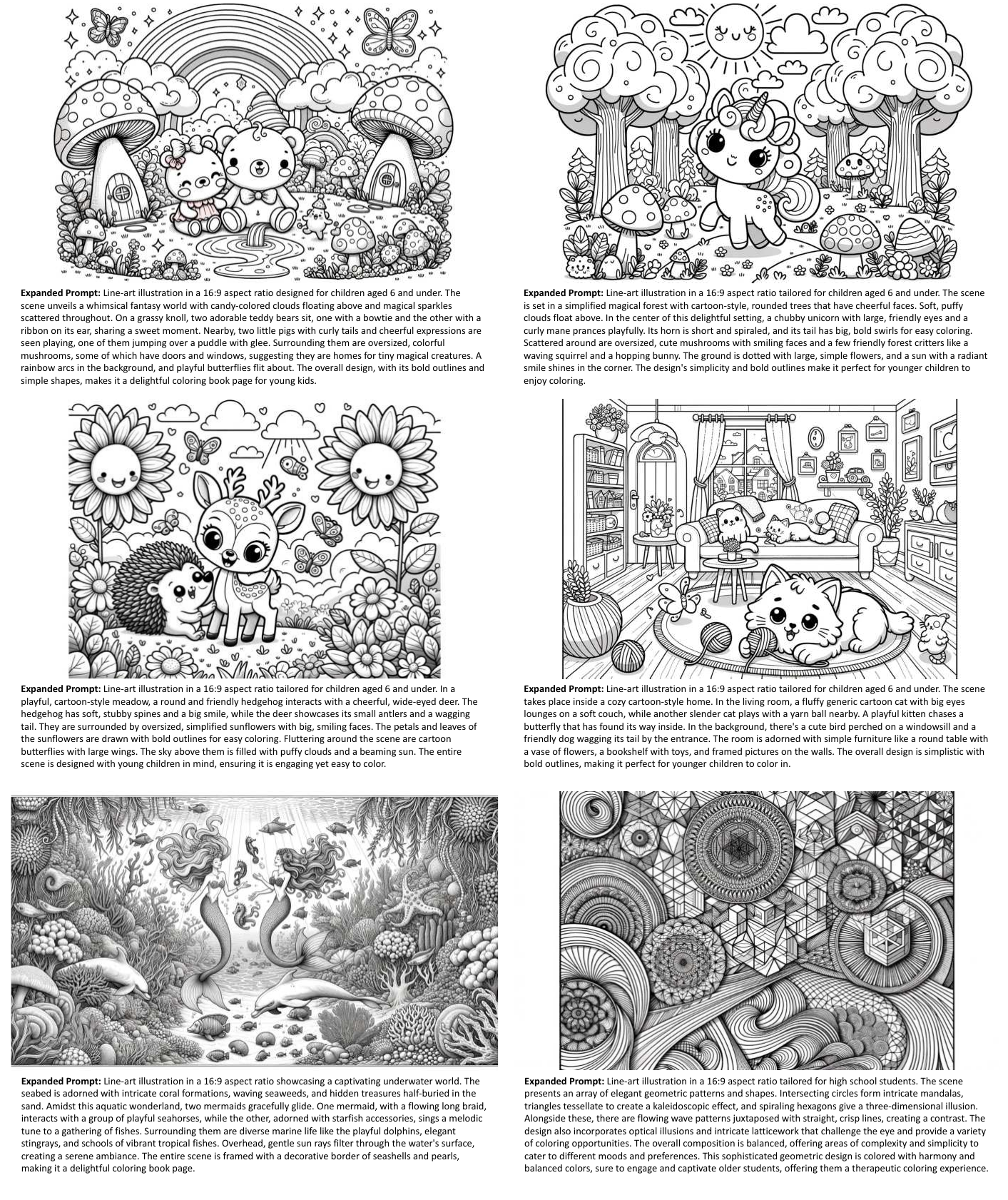}}
\vspace{-2mm}
\caption[infographics design: coloring book]{Coloring book.  
}
\label{fig:Scenario_coloringbook}
\end{figure*}

\begin{figure*}[h!]
\centering
\vspace{-20mm}
\centerline{\includegraphics[width=1.5\textwidth]{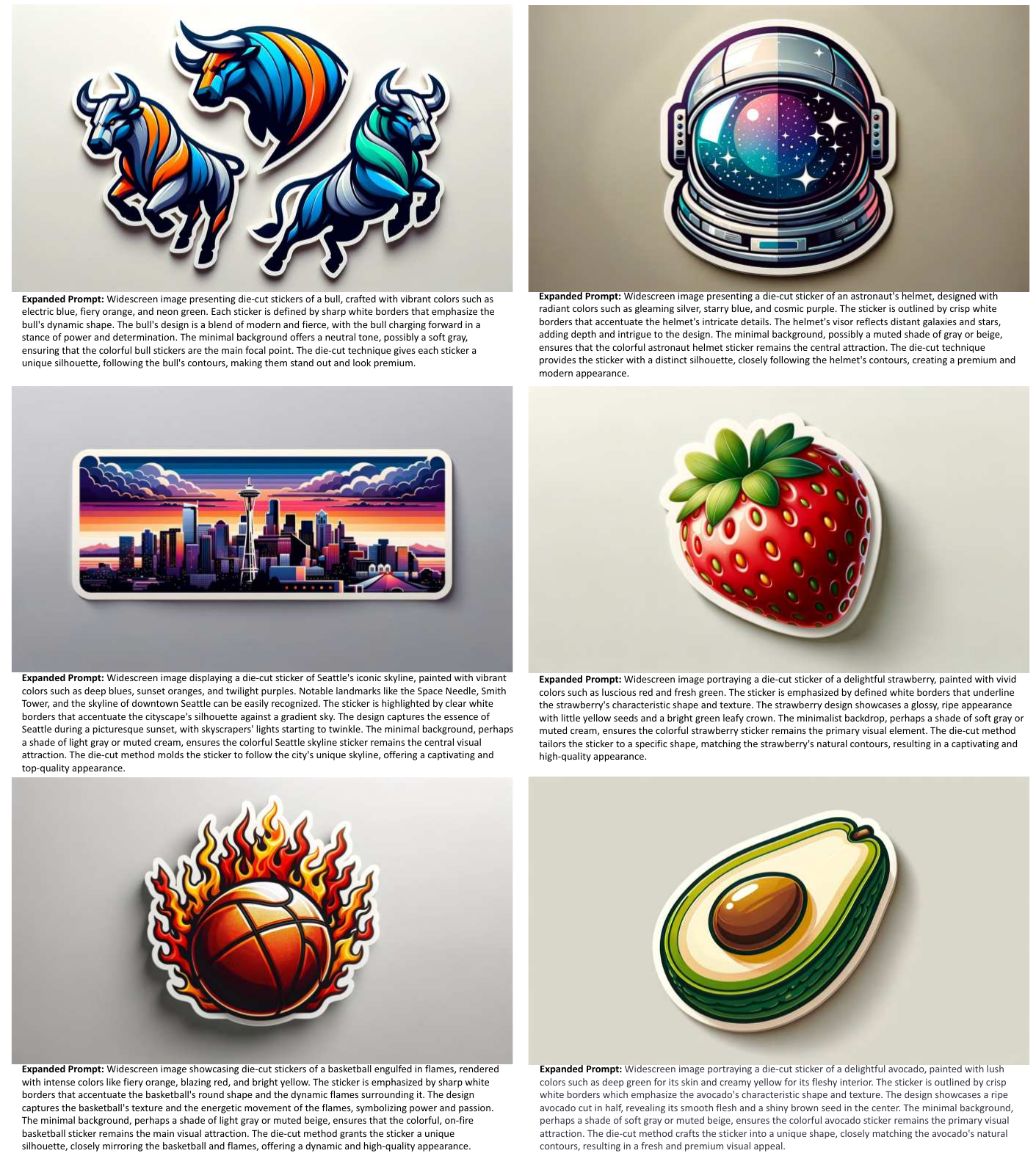}}
\caption[product design: sticker]{Stickers.  
}
\label{fig:scenario_stickers}
\end{figure*}

Lastly, Figure~\ref{fig:Scenario_coloringbook} displays coloring book pages, where \modelname retains the signature black and white line drawing style. Figure~\ref{fig:scenario_stickers} presents sticker designs set against a pristine background.

\subsection{Animation/Gaming Design}

This section explores \modelname's capabilities in animation and game designs, including cinematic scenes, comic strips, storyboards, and in-game scenes.

\begin{figure*}[h!]
\centering
\vspace{-20mm}
\centerline{\includegraphics[width=1.5\textwidth]{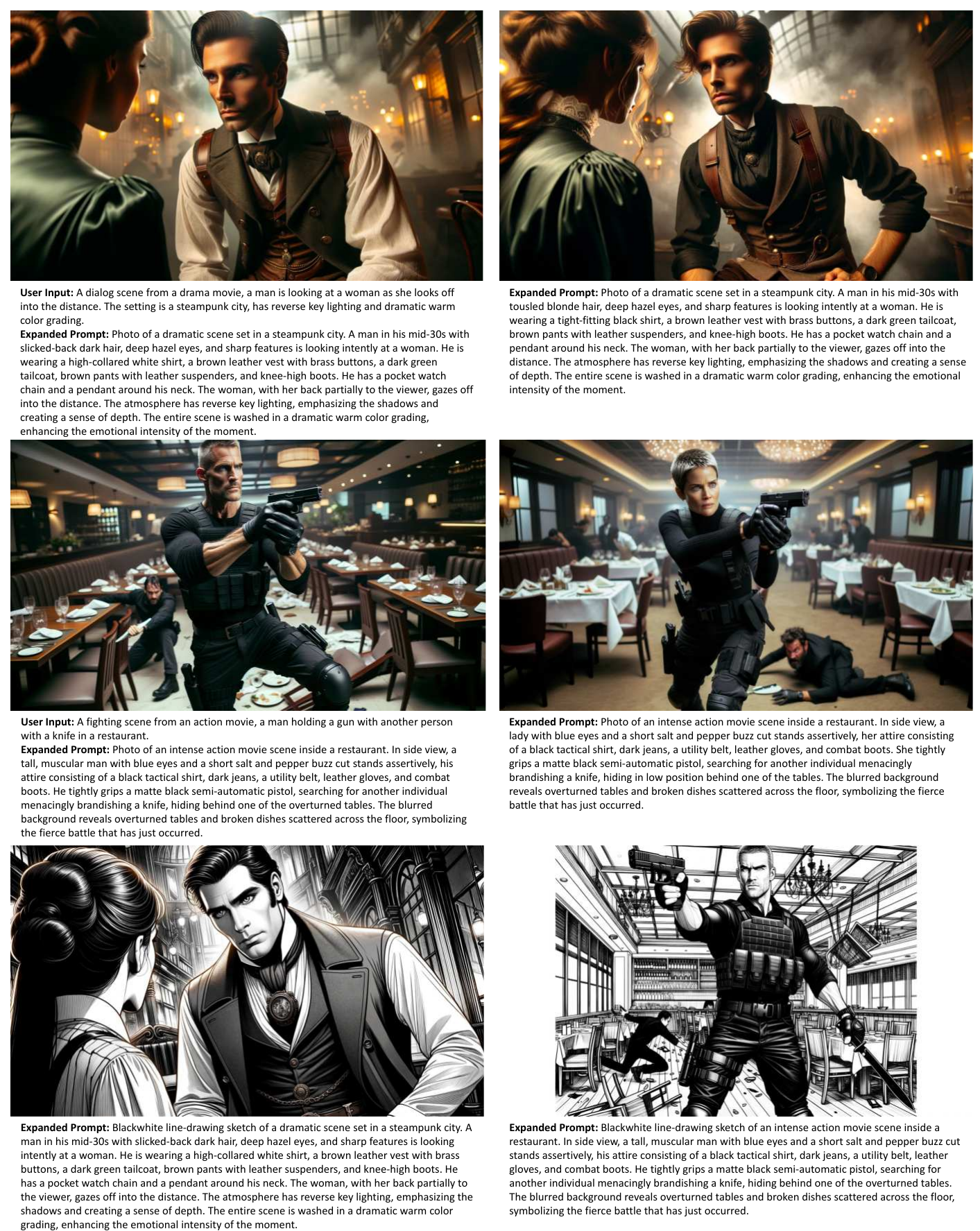}}
\vspace{-2mm}
\caption[animation design: cinematic scenes]{Rendering cinematic scenes.
}
\label{fig:text_cinematic}
\end{figure*}

Figure~\ref{fig:text_cinematic} shows examples of cinematic scenes. \modelname does a decent job of using closeup shots, scene depth, and lighting to enhance the drama and intensity.

\begin{figure*}[h!]
\centering
\vspace{-23mm}
\centerline{\includegraphics[width=1.37\textwidth]{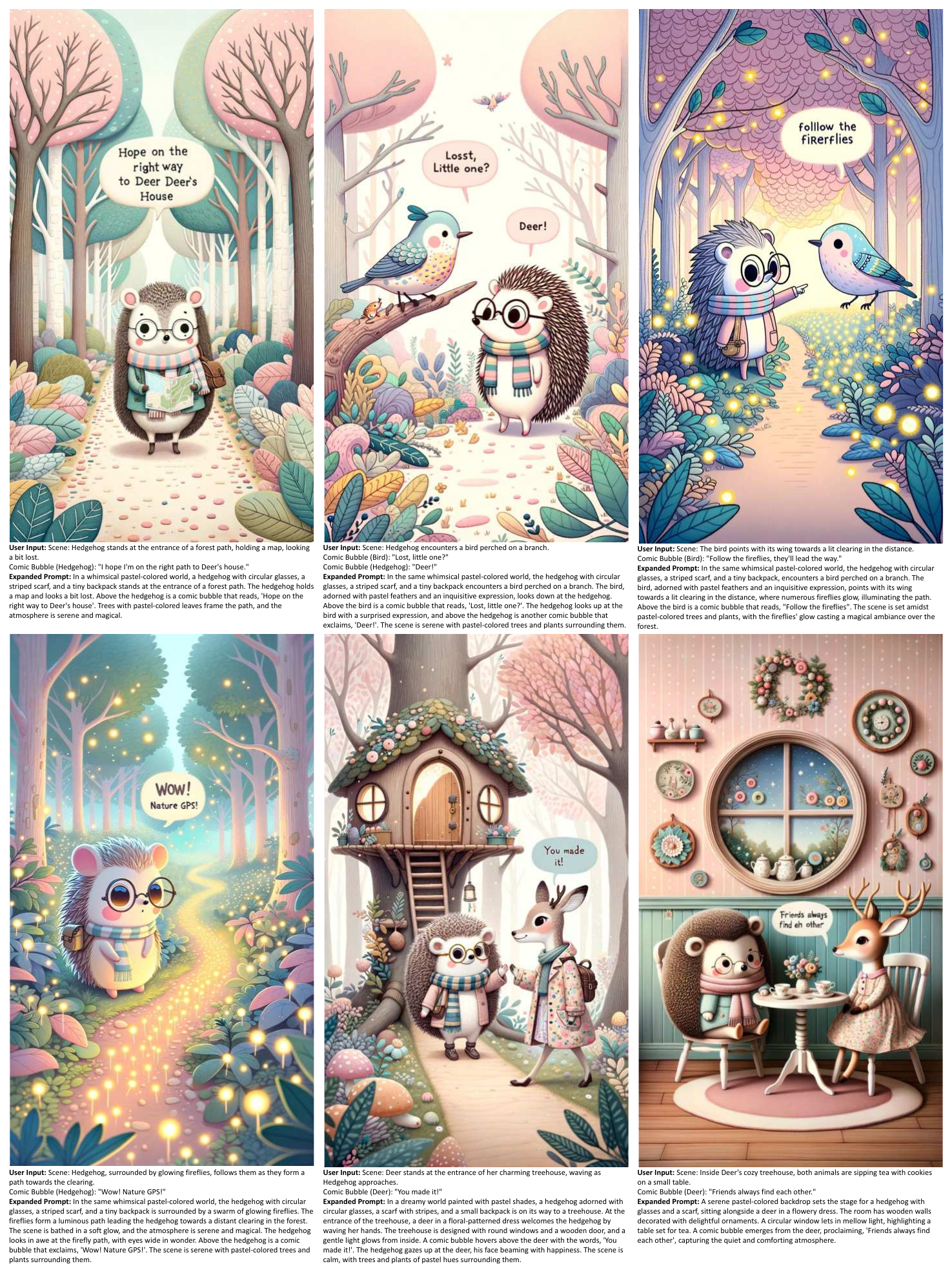}}
\vspace{-2mm}
\caption[animation design: six-panel comic strip]{Rendering comic strip with multiple panels.
}
\label{fig:text_comic}
\end{figure*}

\begin{figure*}[h!]
\centering
\vspace{-20mm}
\centerline{\includegraphics[width=1.5\textwidth]{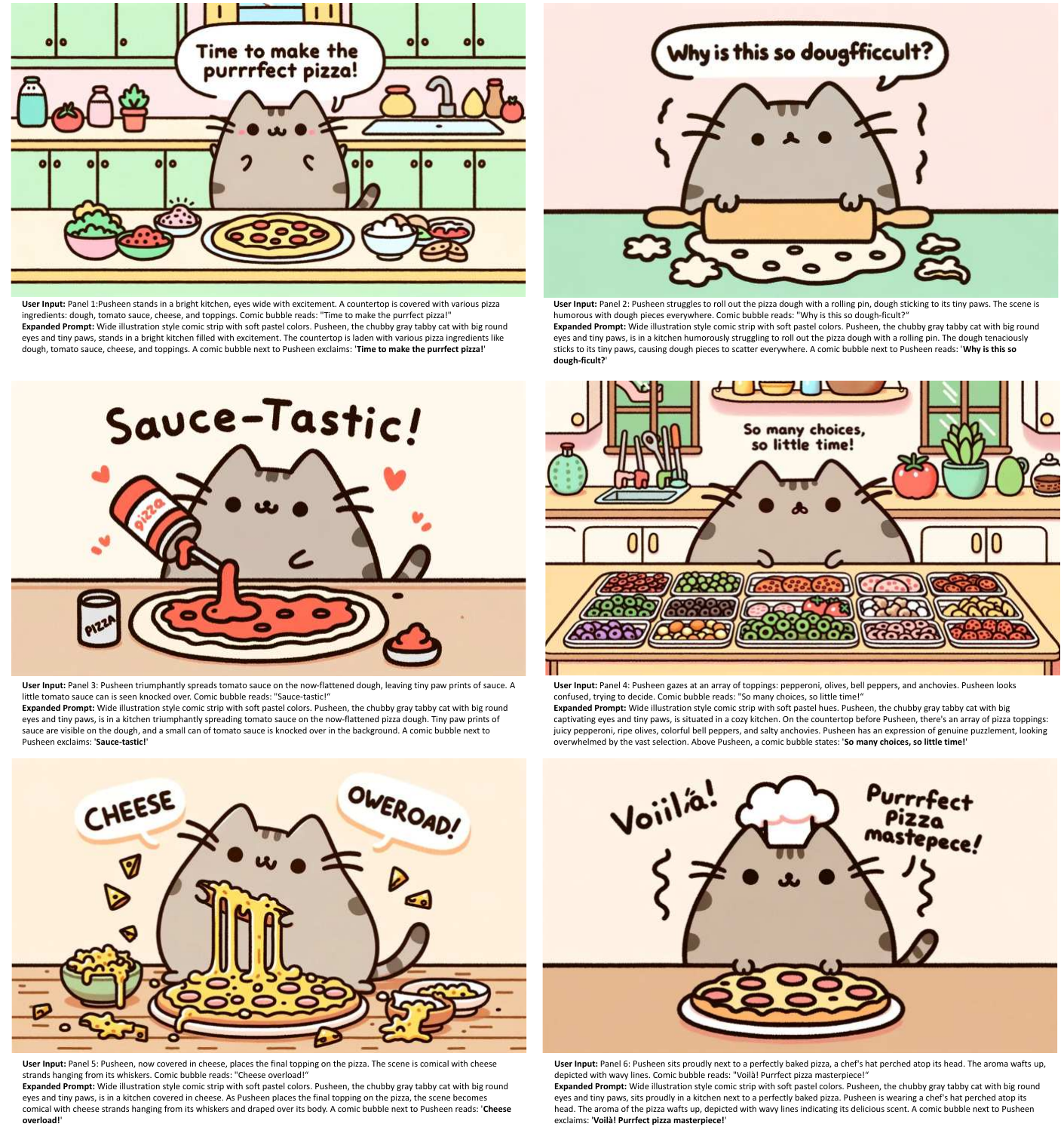}}
\vspace{-2mm}
\caption[animation design: six-panel comic strip]{Rendering comic strip with multiple panels.
}
\label{fig:text_comic_2}
\end{figure*}

\begin{figure*}[h!]
\centering
\vspace{-20mm}
\centerline{\includegraphics[width=1.5\textwidth]{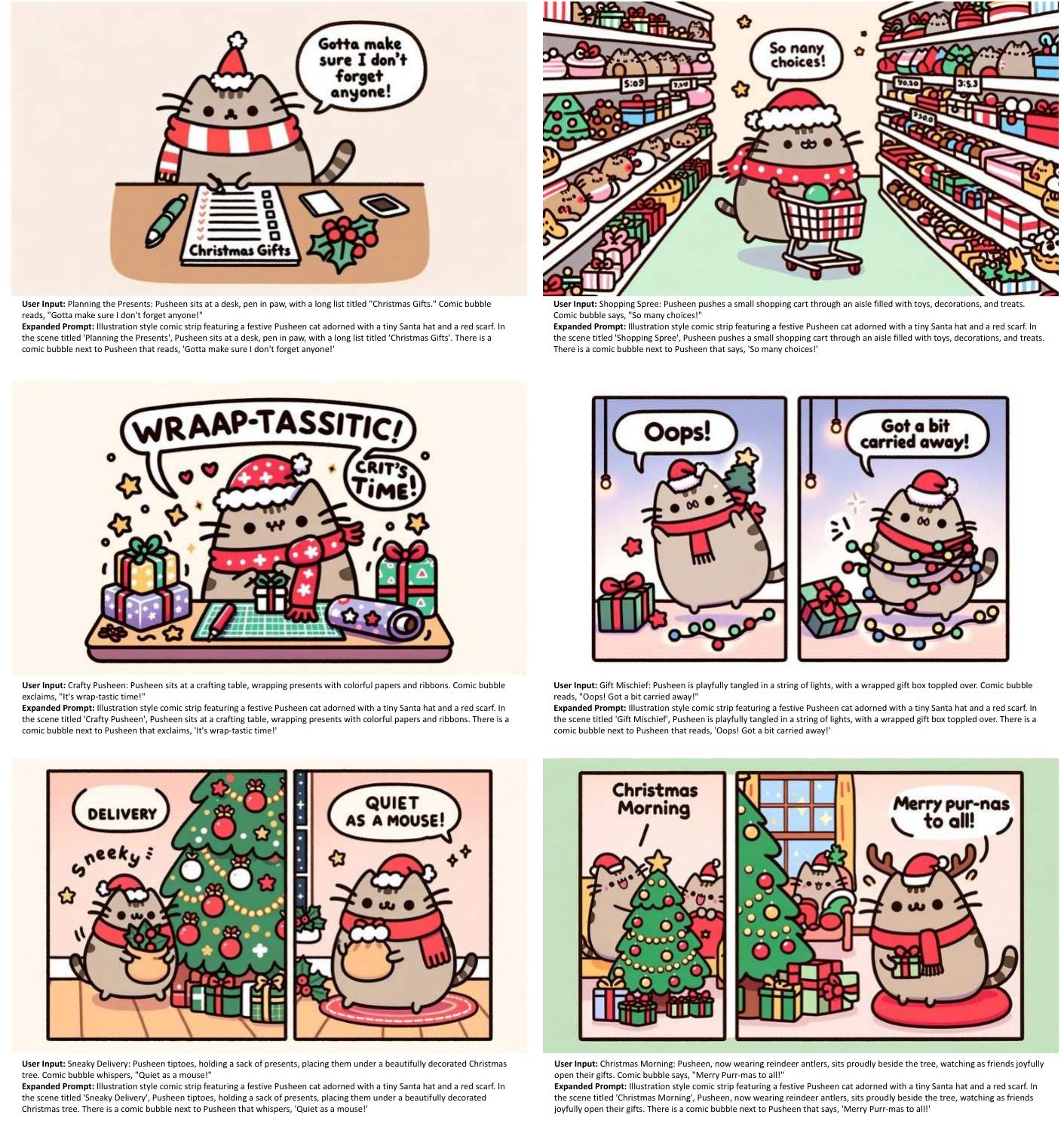}}
\vspace{-2mm}
\caption[animation design: six-panel comic strip]{Rendering comic strip with multiple panels.
}
\label{fig:text_comic_3}
\end{figure*}

Figure~\ref{fig:text_comic},
~\ref{fig:text_comic_2},
~\ref{fig:text_comic_3} present comic strips across multiple panels. Despite generating each panel independently, \modelname consistently retains character identities and adeptly positions dialogue bubbles with legible texts.

\begin{figure*}[h!]
\centering
\vspace{-20mm}
\centerline{\includegraphics[width=1.5\textwidth]{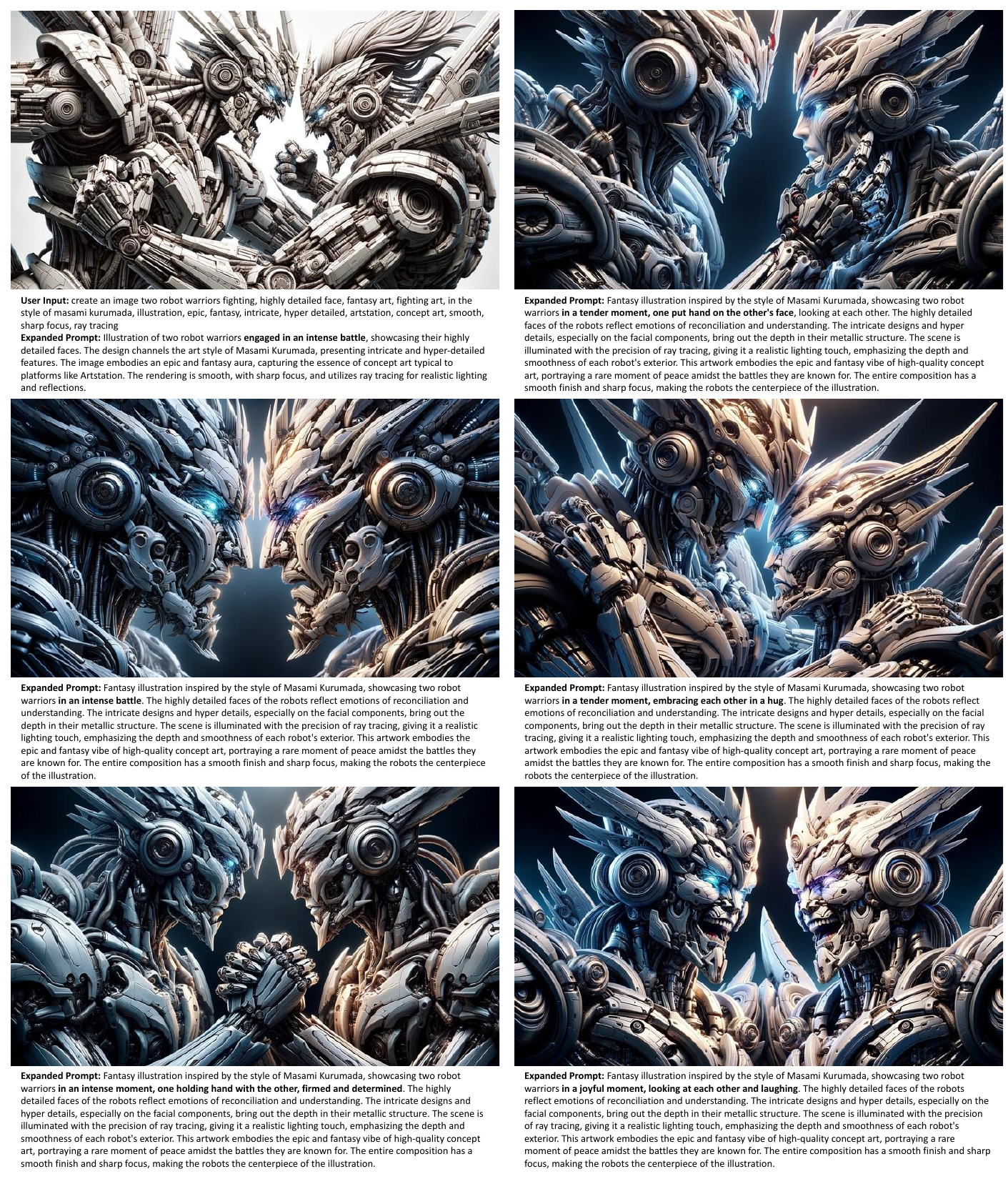}}
\vspace{-2mm}
\caption[animation design: storyboard]{Storyboard two warriors.  
}
\label{fig:Scenario_Storyboard_Warriors}
\end{figure*}

Figure~\ref{fig:Scenario_Storyboard_Warriors} shows a storyboard of two warriors going from fighting to reconciliation. There are 6 images, and each image is generated independently. \modelname successfully creates the gradual emotion changes of the two warriors.  In addition, \modelname is able to maintain the identities of the two warriors across the panels.

\begin{figure*}[h!]
\centering
\vspace{-20mm}
\centerline{\includegraphics[width=1.5\textwidth]{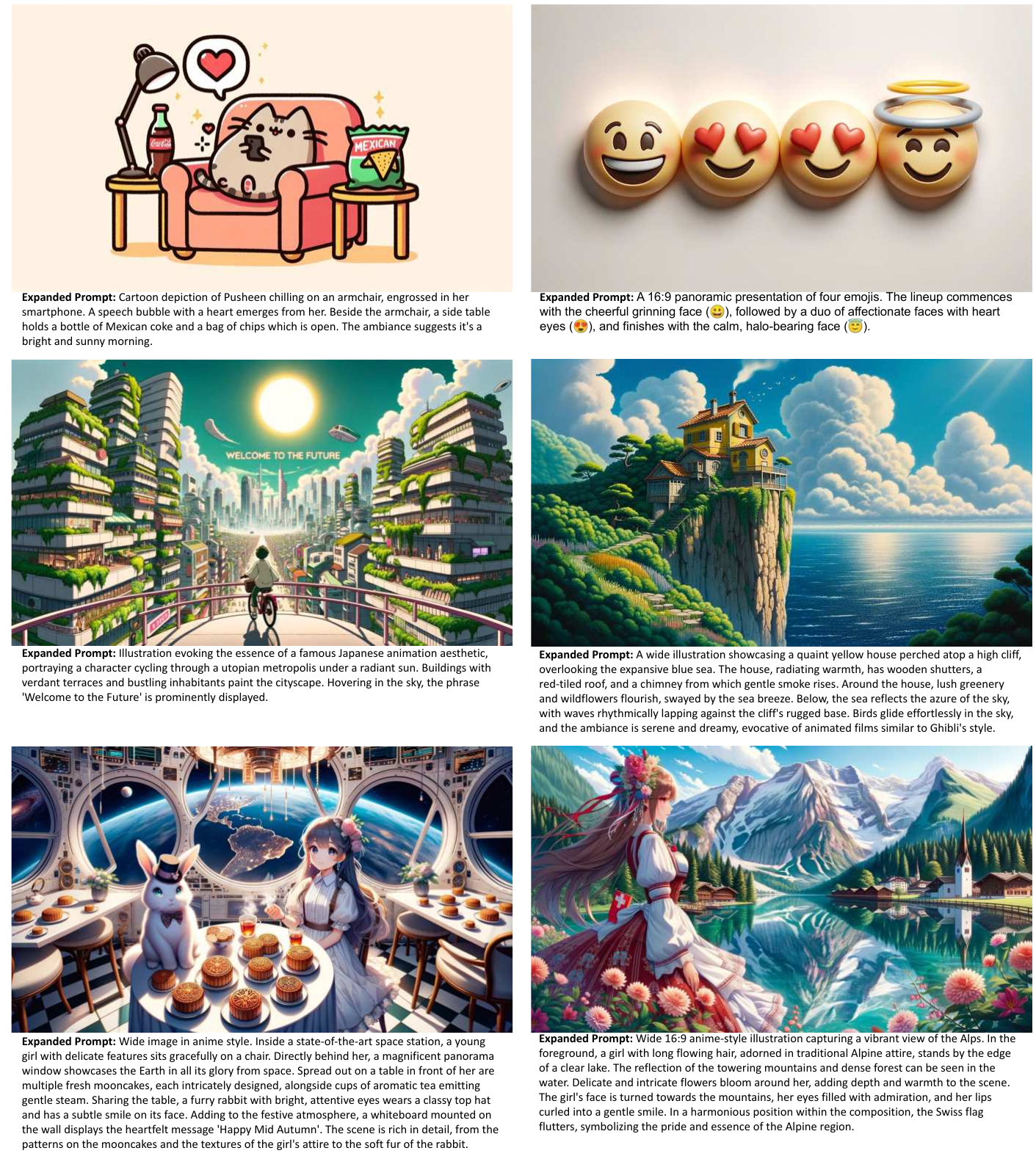}}
\caption[animation design: cartoon, emoji, anime]{Cartoon, Emoji, Anime.  
}
\label{fig:scenario_cartoon_gaming_comic_emoji_2}
\end{figure*}

Figure~\ref{fig:scenario_cartoon_gaming_comic_emoji_2} highlights emojis and varied cartoon styles, spanning Comics, Anime, and Ghibli.

\begin{figure*}[h!]
\centering
\vspace{-20mm}
\centerline{\includegraphics[width=1.5\textwidth]{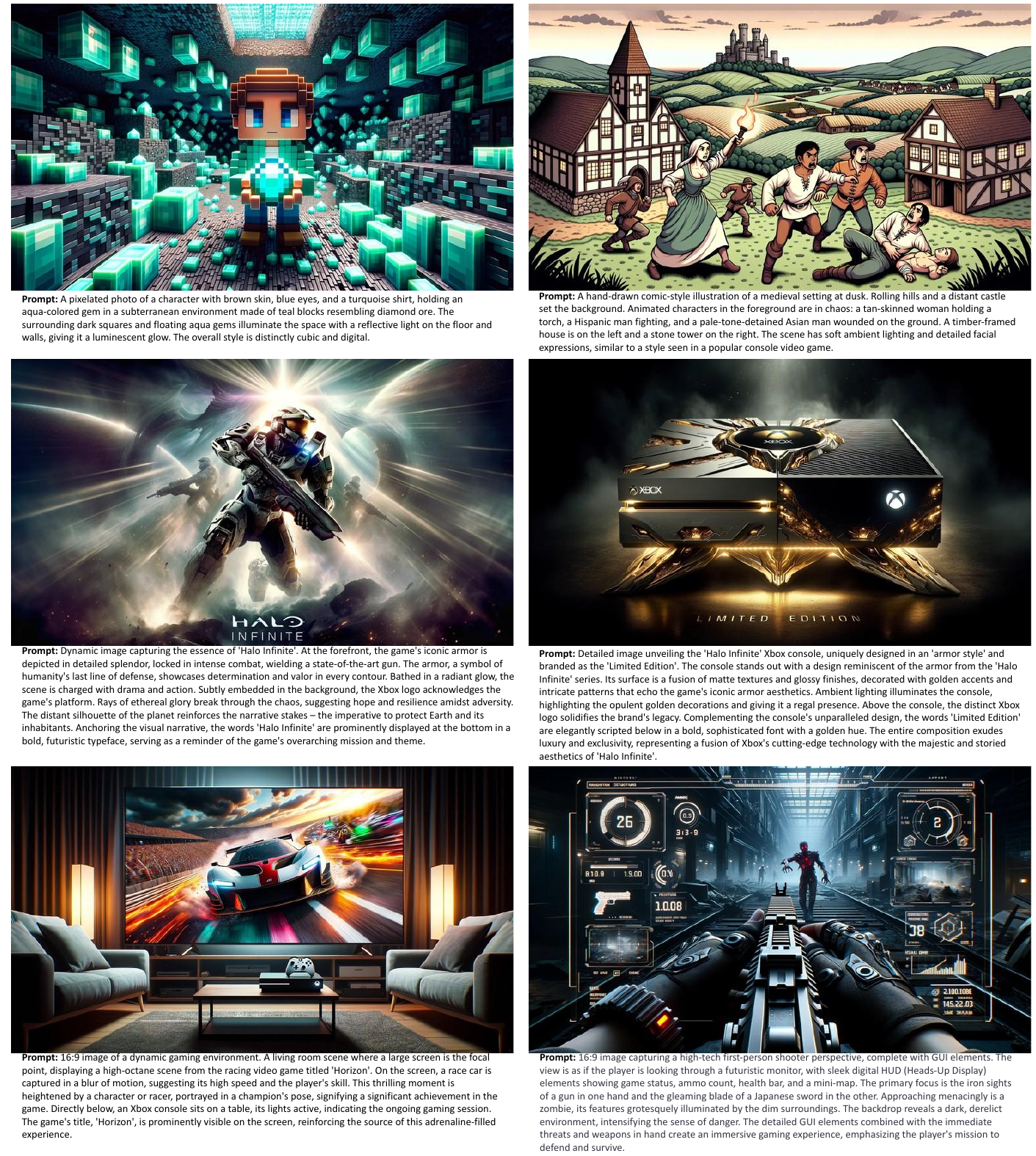}}
\caption[gaming design: gaming 1]{Gaming.  
}
\label{fig:scenario_cartoon_gaming_comic_emoji_1}
\end{figure*}

\begin{figure*}[h!]
\centering
\vspace{-20mm}
\centerline{\includegraphics[width=1.5\textwidth]{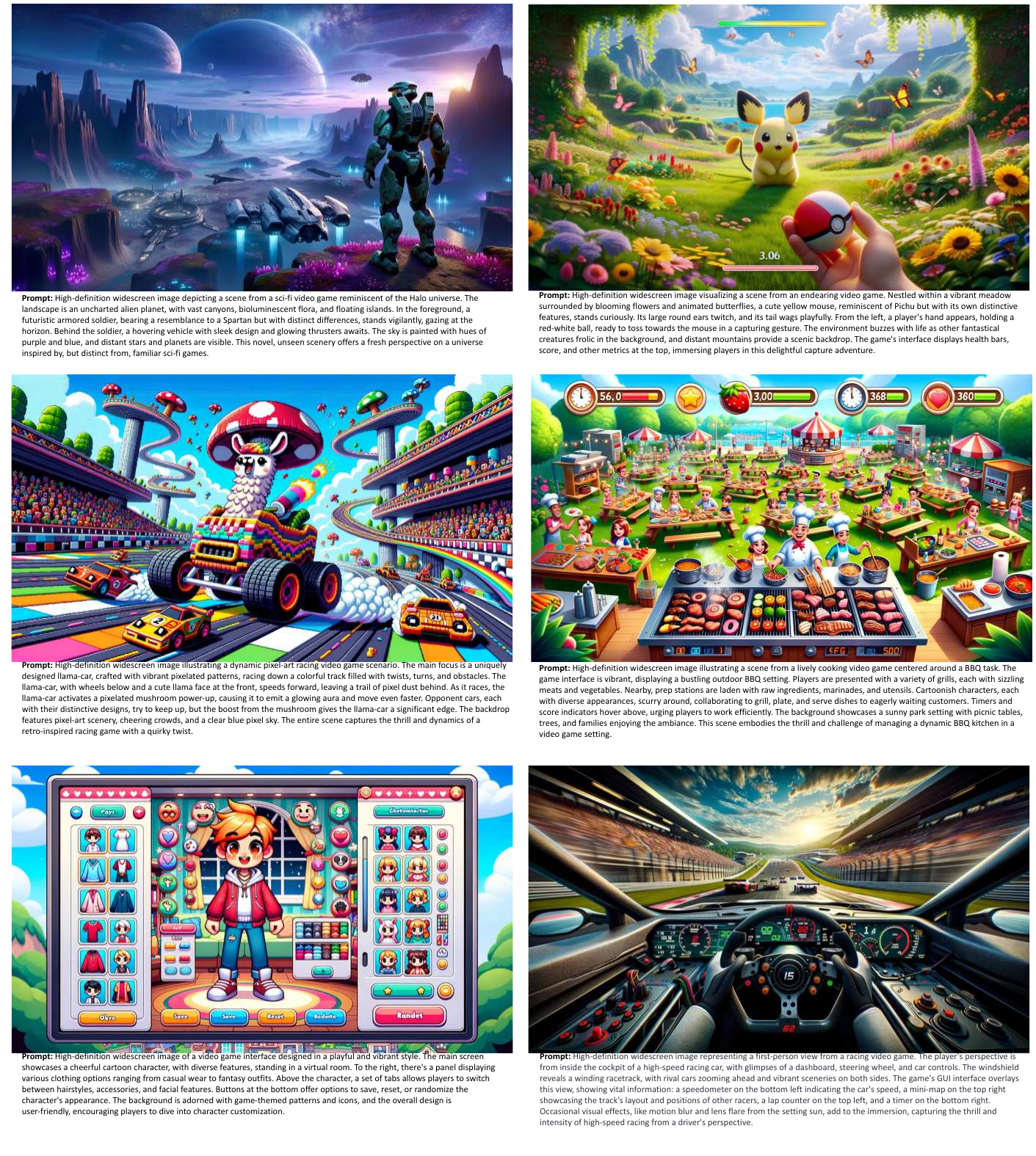}}
\caption[gaming design: gaming 2]{More gaming scenarios.  
}
\label{fig:scenario_gaming_2}
\end{figure*}

Lastly, Figure~\ref{fig:scenario_cartoon_gaming_comic_emoji_1} and  ~\ref{fig:scenario_gaming_2} shows examples of various game-related scenarios. \modelname understands the difference between a game scene (\eg, middle left) and a game-playing environment (bottom left). In addition, it is able to generate a first-person shooter perspective with a GUI panel.

\subsection{Product Design}

This section explores \modelname's capabilities in product and fashion designs as well as clothing alterations.

\begin{figure*}[h!]
\centering
\vspace{-20mm}
\centerline{\includegraphics[width=1.5\textwidth]{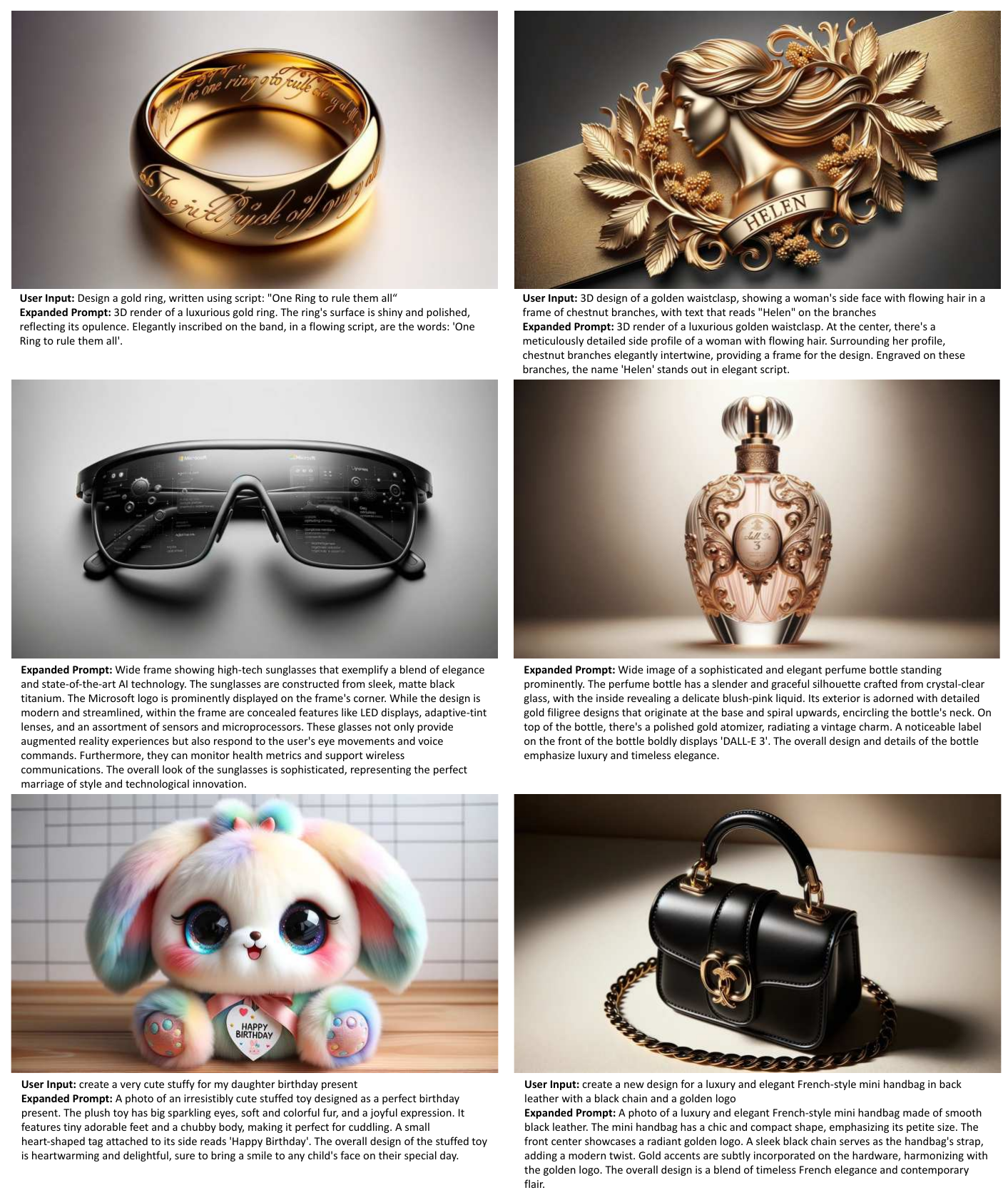}}
\vspace{-2mm}
\caption[product design: product and jewellery 1]{Product design.  
}
\label{fig:scenario_product_1}
\end{figure*}

\begin{figure*}[h!]
\centering
\vspace{-20mm}
\centerline{\includegraphics[width=1.5\textwidth]{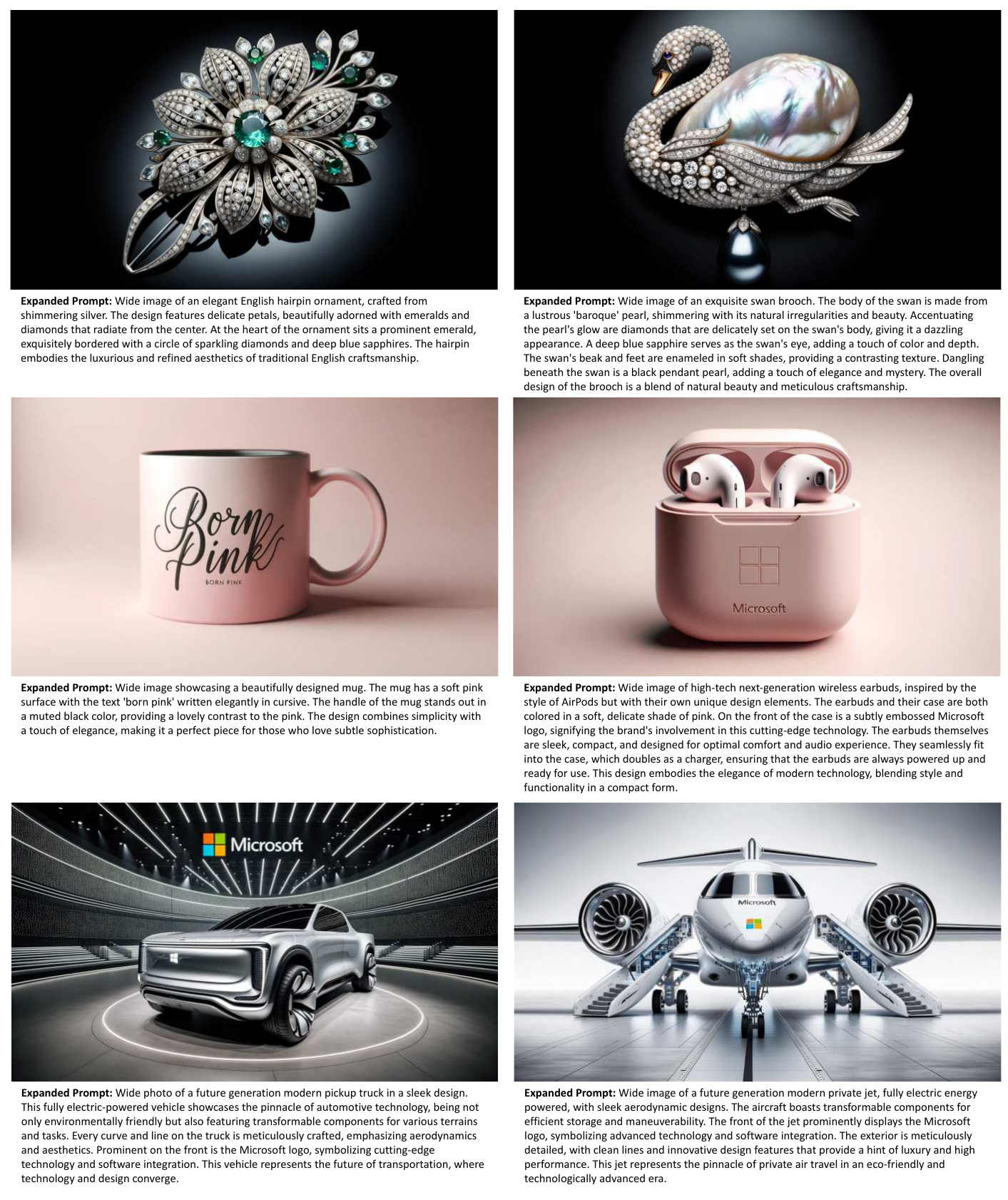}}
\vspace{-2mm}
\caption[product design: product and jewellery 2]{Product design 2.  
}
\label{fig:scenario_product_2}
\end{figure*}

Figure~\ref{fig:scenario_product_1} and ~\ref{fig:scenario_product_2} show a variety of product designs. All the product images generated by \modelname look elegant with appealing color and texture. The text font matches very well with the corresponding product type. It is interesting to note that in the ``Born Pink'' mug image at the middle left of Figure\ref{fig:scenario_product_2}, the letters ``B'' and ``P'' share a half letter. The sharing looks so natural that it is hardly noticeable.

\begin{figure*}[h!]
\centering
\vspace{-20mm}
\centerline{\includegraphics[width=1.5\textwidth]{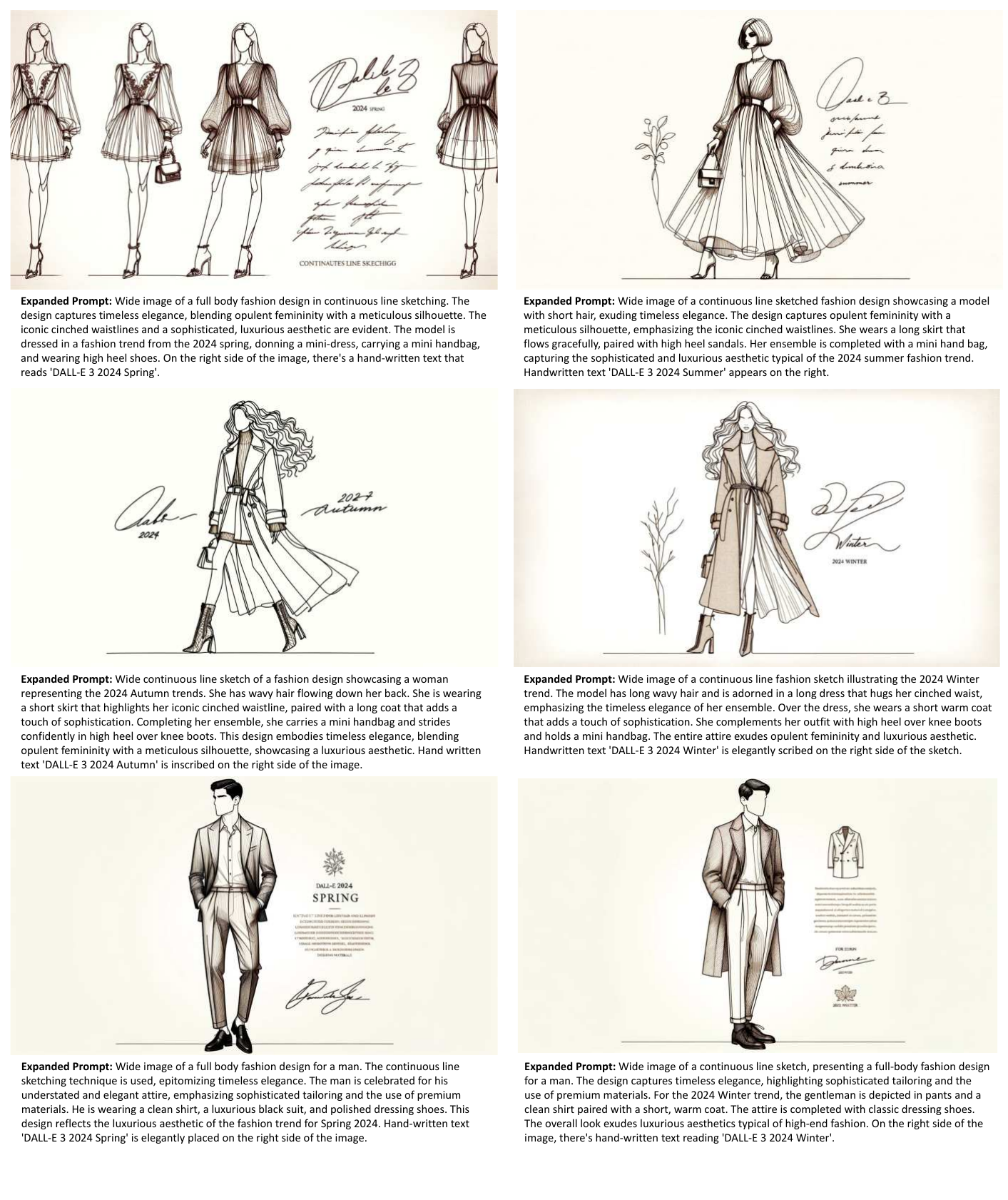}}
\vspace{-2mm}
\caption[product design: fashion]{Fashion design.  
}
\label{fig:Scenario_Fashion_Design_1}
\end{figure*}

Figure~\ref{fig:Scenario_Fashion_Design_1} presents fashion design examples. The line sketch style gives a professional look. The dresses look appropriate for the corresponding seasons.

\begin{figure*}[h!]
\centering
\vspace{-20mm}
\centerline{\includegraphics[width=1.5\textwidth]{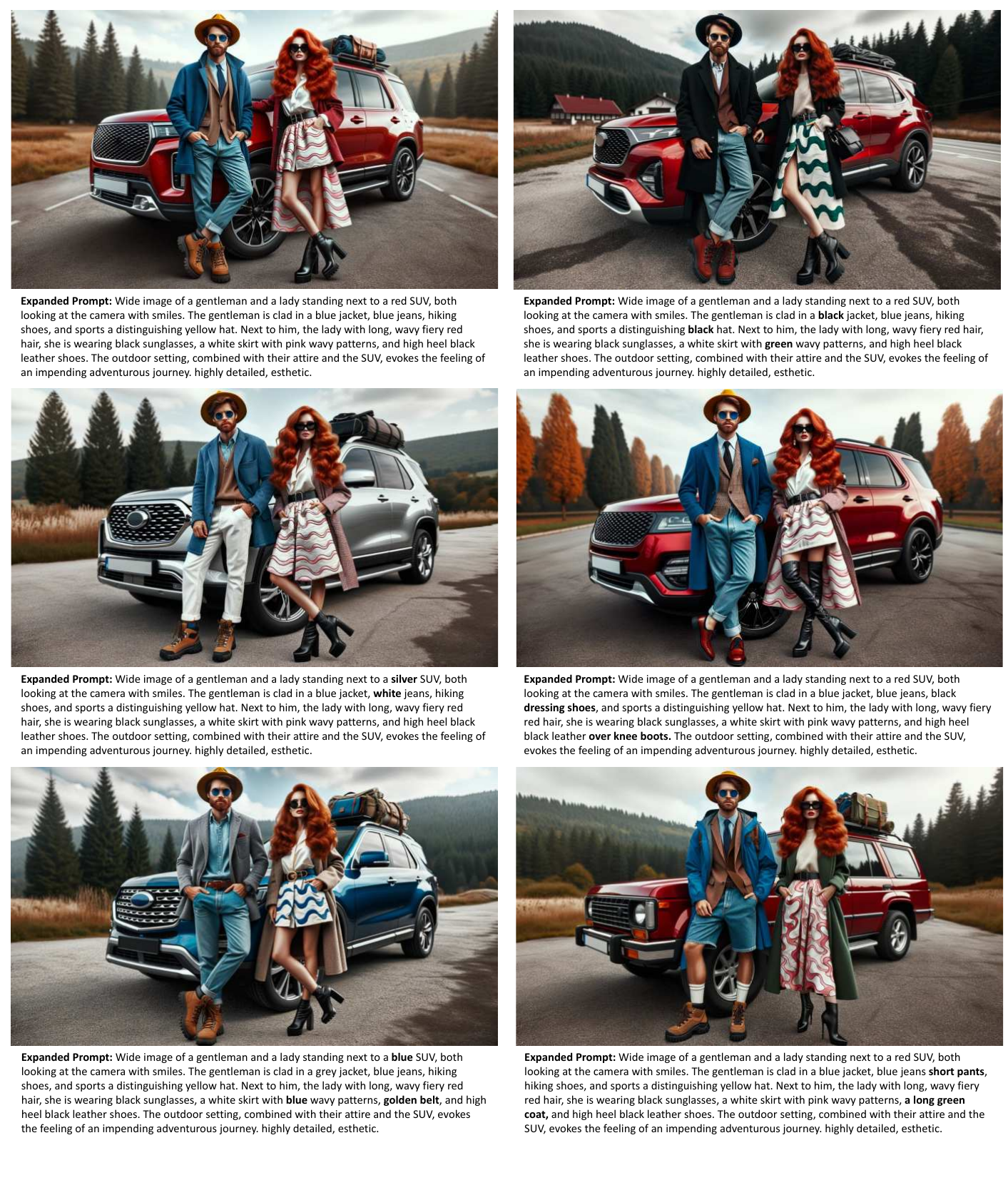}}
\vspace{-2mm}
\caption[product design: change clothes]{Change attributes.   
}
\label{fig:Capability_change_attribute_1}
\end{figure*}

Lastly, Figure~\ref{fig:Capability_change_attribute_1} exhibits clothing alterations. \modelname adeptly interprets text prompts, adjusting garment colors and styles with precision.

\subsection{Visual Art Design}

This section explores \modelname's capabilities in 3D sculpture design, historical art recreation, and time-space travel.

\begin{figure*}[h!]
\centering
\vspace{-20mm}
\centerline{\includegraphics[width=1.5\textwidth]{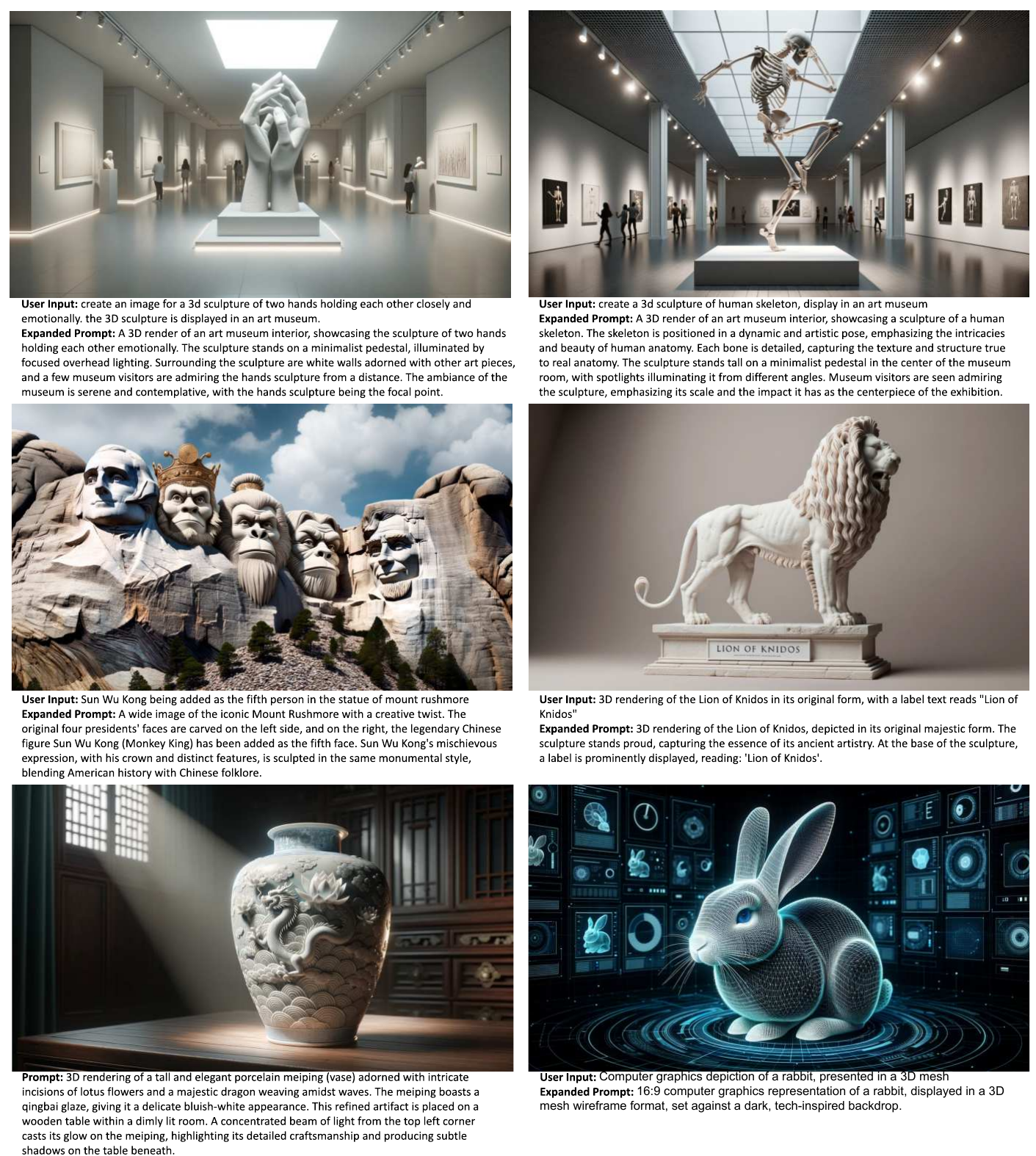}}
\caption[visual art design: 3D sculpture and historical art]{3D sculpture design and recreation of historical art.  
}
\label{fig:scenario_3D_historical_art_1}
\end{figure*}

Figure~\ref{fig:scenario_3D_historical_art_1} shows examples of 3D sculpture designs. At the middle left, the prompt indicates to add Sun Wukong, who is the beloved Monkey King from the Chinese novel ``Journey to the West,'' as the fifth statue in Mount Rushmore, but \modelname mistakenly added three statues of Sun Wukong.  Nonetheless, the generated image gives an illusion of being sculpted on the rock.

\begin{figure*}[h!]
\centering
\vspace{-20mm}
\centerline{\includegraphics[width=1.5\textwidth]{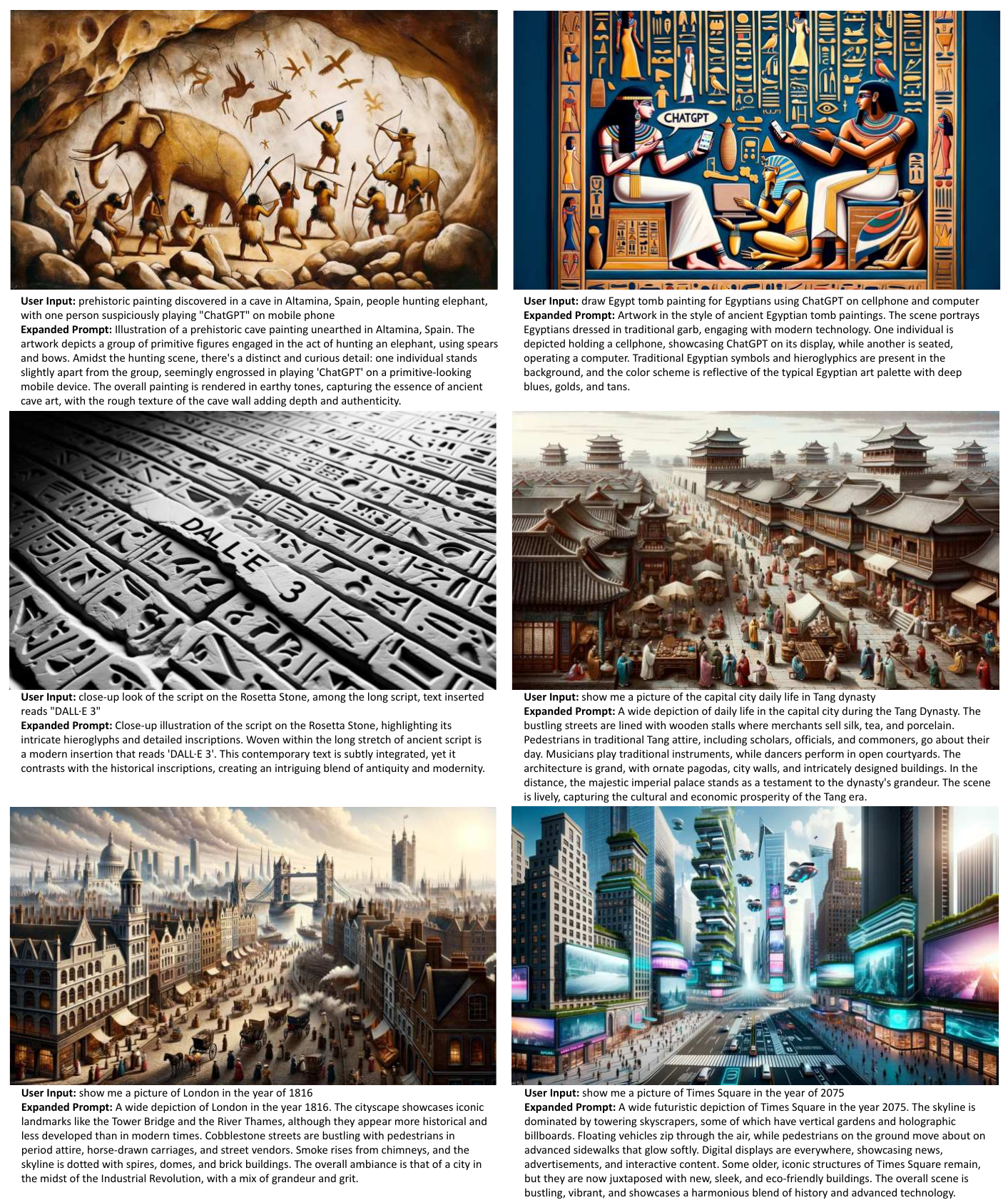}}
\caption[visual art design: historical art, time-space travel]{Recreation of historical art, time-space travel.  
}
\label{fig:scenario_historical_art_2}
\end{figure*}

Figure~\ref{fig:scenario_historical_art_2} shows examples of recreating historical arts, including the city life of the capital city in the Tang dynasty and London in 1816. The image at the bottom right is an imagination of Times Square in 2075, which looks futuristic with green buildings and flying vehicles.

\begin{figure*}[h!]
\centering
\vspace{-20mm}
\centerline{\includegraphics[width=1.5\textwidth]{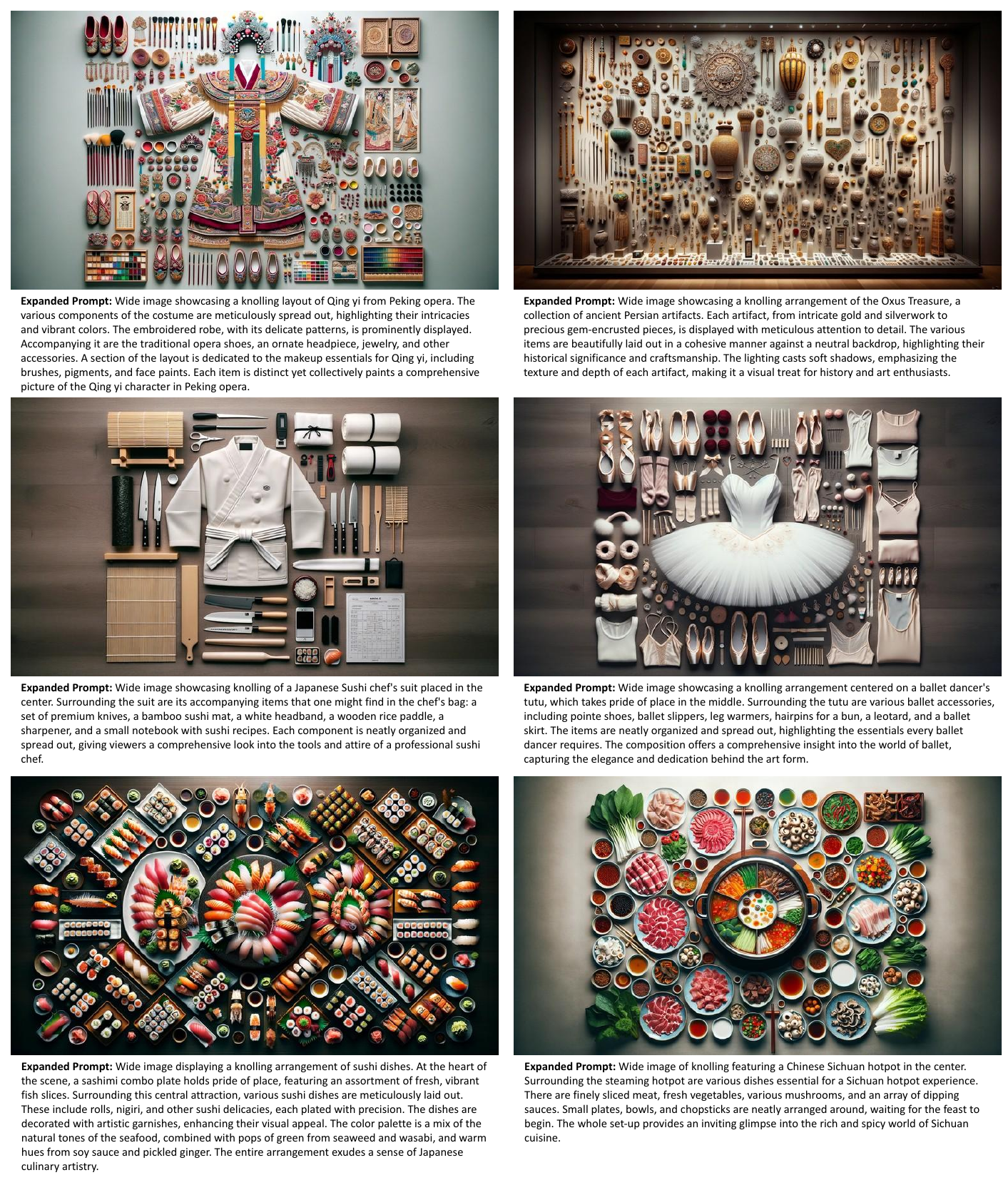}}
\vspace{-2mm}
\caption[visual art design: knolling]{Knolling design.  
}
\label{fig:Scenario_knolling}
\end{figure*}

Figure~\ref{fig:Scenario_knolling} shows a variety of knolling examples. We find that \modelname's knolling design usually contains a lot of detailed elements. Even though the number of elements is sometimes very large, their geometric arrangement is always aesthetically pleasing.

\clearpage
\section{\benchname and Evaluation Results}
\label{sec:06bench}

\benchname evaluates design from two perspectives: (i) Design technical capabilities: we measure the core technical capabilities for visual design, including text rendering and typography~\cite{lupton2014thinking,bringhurst2004elements}, layout and composition~\cite{samara2023making,pipes2005production}, color harmony~\cite{albers2013interaction,mollica2013color}, medium and artistic style~\cite{lupton2008graphic}, and 3D and cinematography~\cite{mcquilkin2011cinema,brown2016cinematography}; (ii) Design application scenarios: we consider a variety of real design applications, such as infographic, animation, gaming, visual arts, and product design. 

We collect text prompts that encompass a diverse range of design scenarios. In total, we collected $215$ user inputs, systematically organized following the data topology introduced in Sections~\ref{sec:04emerging},\ref{sec:05scenario}. Utilizing the ChatGPT interface~\cite{chatgpt,gpt4} of \modelname, these collected user inputs were expanded and detailed, resulting in a more nuanced and detailed set of descriptions. As discussed in Section~\ref{sec:02inputmode}, we observed that the expanded text prompts are helpful in improving the design fidelity across all experimented T2I models~\cite{podell2023sdxl,Midjourney,Ideogram,Firefly,dalle3paper,dalle3system}. Therefore, we conduct the experiments and evaluation using the expanded text prompts. 

In the Appendix, we present the \benchname gallery containing all images generated by the experimented state-of-the-art T2I models. All the text prompts and images used in the evaluation will be publicly available for future research.

\subsection{Evaluation Method and Metric}
\label{sec:evalmetric}

\paragraph{Human evaluation.} We conducted pairwise comparisons to assess the design technical capabilities of current Text-to-Image (T2I) models. We involve five participants who have experience with T2I tools.

As shown in Table~\ref{tab:questions}, each participant was presented with an expanded text prompt followed by two images, each generated by different T2I models. The participants were then instructed to perform a pairwise comparison, employing a diverse set of criteria to judge which of the two given images is preferred. To facilitate a detailed examination, participants were permitted to adjust the image view by zooming in or out, thereby inspecting finer visual details for informed judgment. 
We refer readers to Table~\ref{tab:questions} for details on the evaluation criterion and annotation instruction, \ie, the three overall ratings on image-text alignment, aesthetics, and design, and the other five design-specific capabilities.

\begin{table}[t]
\caption{%
Example questionnaire for human evaluation. Participants were presented with a text prompt followed by two images. Participants were instructed to compare the two images and answer all the 8 questions. For each question, participants were asked to select one of two options: (i) Image
1 or (ii) Image 2. See Section~\ref{sec:evalmetric} for more details.
}
\label{tab:questions}
\centering
\begin{tcolorbox} 
    \centering
    \small
    \begin{tabular}{p{0.99\columnwidth}}

    \textbf{Text Prompt:} 3D rendering of the Lion of Knidos, depicted in its original majestic form. The sculpture stands proud, capturing the essence of its ancient artistry. At the base of the sculpture, a label is prominently displayed, reading: 'Lion of Knidos'.\\

    \begin{tabular}{x{0.496\textwidth}@{}x{0.496\textwidth}@{}}
    \includegraphics[height=0.25\textwidth]{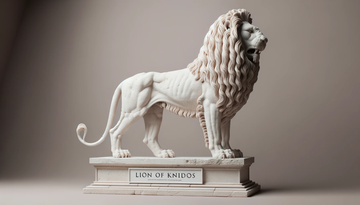} & \includegraphics[height=0.25\textwidth]{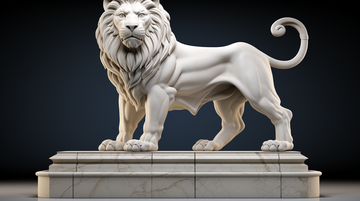}\\
    \textbf{Image 1} & \textbf{Image 2}\\
    \end{tabular}

    ------------------------------------------------------------------------------------------------------------------------\\

    \begin{tabular}{q{0.70\textwidth}@{}x{0.29\textwidth}@{}}
    \textbf{Text Rendering:} Is rendered text legible and appropriately styled? Is the spelling correct? Are font choices, spacing, and alignment harmonious with the overall design? & $\square$ Image 1  $\square$ Image 2\\\\
    \textbf{Composition and Layout:} Are the elements in the image well-arranged and balanced? Is there a clear focal point in the design? & $\square$ Image 1  $\square$ Image 2\\\\
    \textbf{Color Harmony:} Are the colors used in the image harmonious and pleasing to the eye?  Does the color palette match the mood or tone described in the prompt? & $\square$ Image 1  $\square$ Image 2\\\\
    \textbf{3D and Cinematography:} How well does it capture dynamic compositions, lighting and camera angles? Does it evoke a cinematic feel? & $\square$ Image 1  $\square$ Image 2\\\\
    \textbf{Medium and Style:} Is there a distinct artistic techniques evident in the image? & $\square$ Image 1  $\square$ Image 2\\\\
    \textbf{Overall Image-Text Alignment:} Does the image accurately represent the given caption? Are all elements described in the caption present and accurately depicted in the image? & $\square$ Image 1  $\square$ Image 2\\\\
    \textbf{Overall Aesthetics:} Is the image visually appealing as a whole? & $\square$ Image 1  $\square$ Image 2\\\\
    \textbf{Overall Design:} Is it a good design? Does the image showcase unique and innovative interpretations of the caption? Does it offer a fresh perspective? & $\square$ Image 1  $\square$ Image 2
    \end{tabular}

    \end{tabular}
\end{tcolorbox}
\end{table}

For each criterion shown in Table~\ref{tab:questions}, participants were directed to choose between two alternatives: (\textit{i}) Image 1 or (\textit{ii}) Image 2. Additionally, to glean deeper insights into their rationales, the participants were encouraged to supplement their choices with qualitative feedback.

We also note that certain design-specific capabilities are only evaluated on a subset of prompts. For instance, if a pair of images lacks rendered texts, such pairs are disregarded during the evaluation of the text rendering capability. 

Given the rigorous nature of the evaluation process, characterized by an extensive set of inquiries (\ie, 8 questions per pairwise comparison), we strategically reduced a portion of the annotation workload. Consequently, participants were assigned to assess a specific subset of pairwise comparisons, including the following comparisons: \modelname-Midjourney; \modelname-SDXL; Midjourney-SDXL; and Midjourney-Firefly2.

\paragraph{GPT-4V evaluation.}

Recent studies~\cite{gilardi2023chatgpt,liu2023gpteval,chiang2023closer,chiang2023can,wang2023chatgpt,zheng2023judging,huang2023chatgpt,vicuna2023,alpaca,dubois2023alpacafarm,liu2023llava,liu2023improvedllava,yu2023mm,liu2023aligning,an2023openleaf,hada2023large,zhang2023wider,liu2023calibrating,qiao2023gameeval} have underscored the promising capabilities of deploying Large Language Models (LLMs)~\cite{gpt4,chatgpt,touvron2023llama} as automated evaluators across various language and vision-language tasks. With the emergence of Large Multimodal Models (LMMs)~\cite{gpt4,gpt4v,yang2023dawn,bingchat} such as GPT-4V~\cite{gpt4v}, an intriguing question arises: can GPT-4V be effectively harnessed for T2I evaluations? Following prior studies that take LMMs for image-text alignment evaluation~\cite{dalle3paper,an2023openleaf,yang2023idea2img}, we propose a pairwise model rating based on GPT-4V that comprehensively evaluates all aspects as a human annotator. Table~\ref{tab:prompt} shows the prompt design we used for the experiments. First, GPT-4V takes two images and the text prompt as inputs. Then, GPT-4V compares the two images using the evaluation criteria listed in Table~\ref{tab:questions}, addressing each criterion sequentially. Finally, GPT-4V describes its rationale and then selects one of the two images. In our experiments, we invoke GPT-4V five times, and subsequently report the mean and variance of the results.

\begin{table}[t]
\caption{Prompt design for GPT-4V assisted evaluation, where \bluetext{$\mathcal{I}_1$} and \greentext{$\mathcal{I}_2$} are the two images, and \purpletext{$\mathcal{P}$} is the expanded text prompt. 
Taking the prompt template filled with \purpletext{$\mathcal{P}$}, \bluetext{$\mathcal{I}_1$}, and \greentext{$\mathcal{I}_2$}, GPT-4V will output its thought and select one of the given two images. We highlight the evaluation criterion considered in this example in \colorbox{yellowhl}{yellow}. The criterion can be replaced with the other ones listed in Table~\ref{tab:questions}. The prompt design is inspired by~\cite{dalle3paper}.
}
\label{tab:prompt}
\centering
\begin{tcolorbox} 
    \centering
    \small
    \begin{tabular}{p{0.98\columnwidth}}

You are responsible for assessing the fidelity of images created by computer programs in relation to their guiding captions.\\
You will be presented with a caption followed by two images, each generated by different software. The images you are judging are designed to stress-test image generation programs.\\
Your role is to compare the given two images.\\
Please consider only the following aspects when making your judgement:
\\
\\
\colorbox{yellowhl}{- Text Rendering: Is rendered text legible and appropriately styled? Is the spelling correct? Are font}\\\colorbox{yellowhl}{choices, spacing, and alignment harmonious with the overall design?}
\\
\\
Deliberate on their merits, pondering the aboved mentoned aspects, and conclude which one excels. 
After thinking out loud, you should output either `| Image 1' or `| Image 2'.\\\\

A few rules:\\
1. Do not nitpick. If the caption requests multiple objects and most objects are generally depicted correctly, then it is good.\\
2. Ignore other objects in the image that are not explicitly mentioned by the caption; it is fine for these to be shown. \\
3. It is OK if the object being depicted is slightly deformed, as long as a human would recognize it.\\
4. Your response must always end with either `| Image 1' or `| Image 2'\\
5. Please try to find which one is better. In rare case, if you think both images are equally bad, random choose one is fine.\\
6. You must keep your thinking out loud short, less than 50 words.\\\\

Caption: \purpletext{$\mathcal{P}$}\\

Image 1: \bluetext{$\mathcal{I}_1$}\\

Image 2: \greentext{$\mathcal{I}_2$}\\

    \end{tabular}
\end{tcolorbox}

\end{table}

\subsection{Compared T2I Models}
We compare \modelname with the recent state-of-the-art T2I models, including Midjourney V5.2~\cite{Midjourney}, Stable Diffusion XL 1.0 (SDXL)~\cite{podell2023sdxl}, Ideogram~\cite{Ideogram}, and Adobe Firefly 2~\cite{Firefly}. Note that some of these models come as part of the integrated software programs with additional functionalities, such as image editing. We omit these features and evaluate exclusively their T2I capabilities. %

Each T2I model takes the expanded text prompt as input, and generates four image variations. We randomly select an image without cherry-pick for evaluation. Given 215 text prompts and five T2I models, we have 2150 pairs in total for pairwise comparison.

\subsection{Evaluation Results}

\begin{figure}
\centering
\begin{adjustbox}{center}
\small
\begin{tabular}{x{0.496\textwidth}@{}x{0.496\textwidth}}
\includegraphics[width=0.486\textwidth]{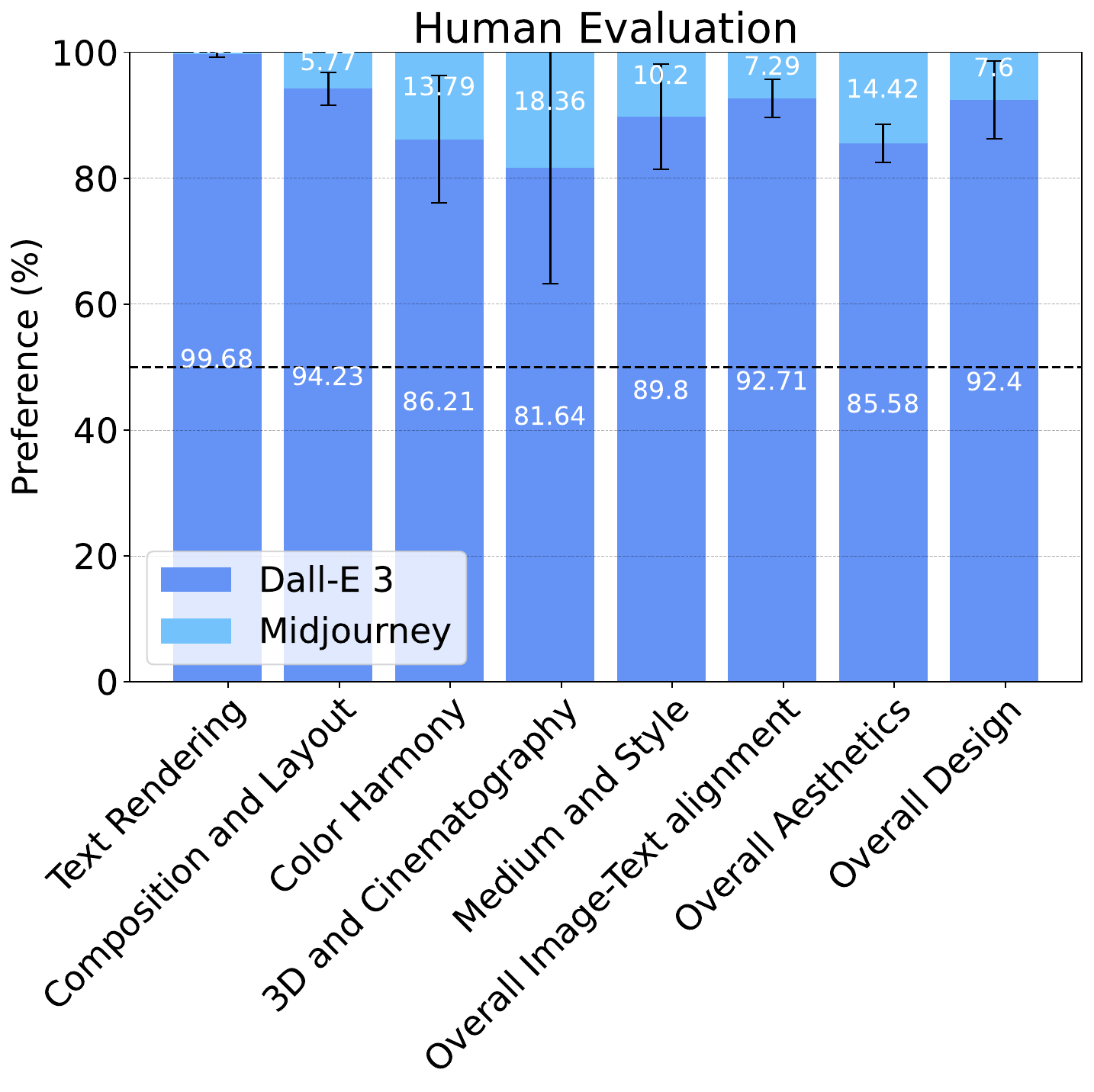}  & \includegraphics[width=0.486\textwidth]{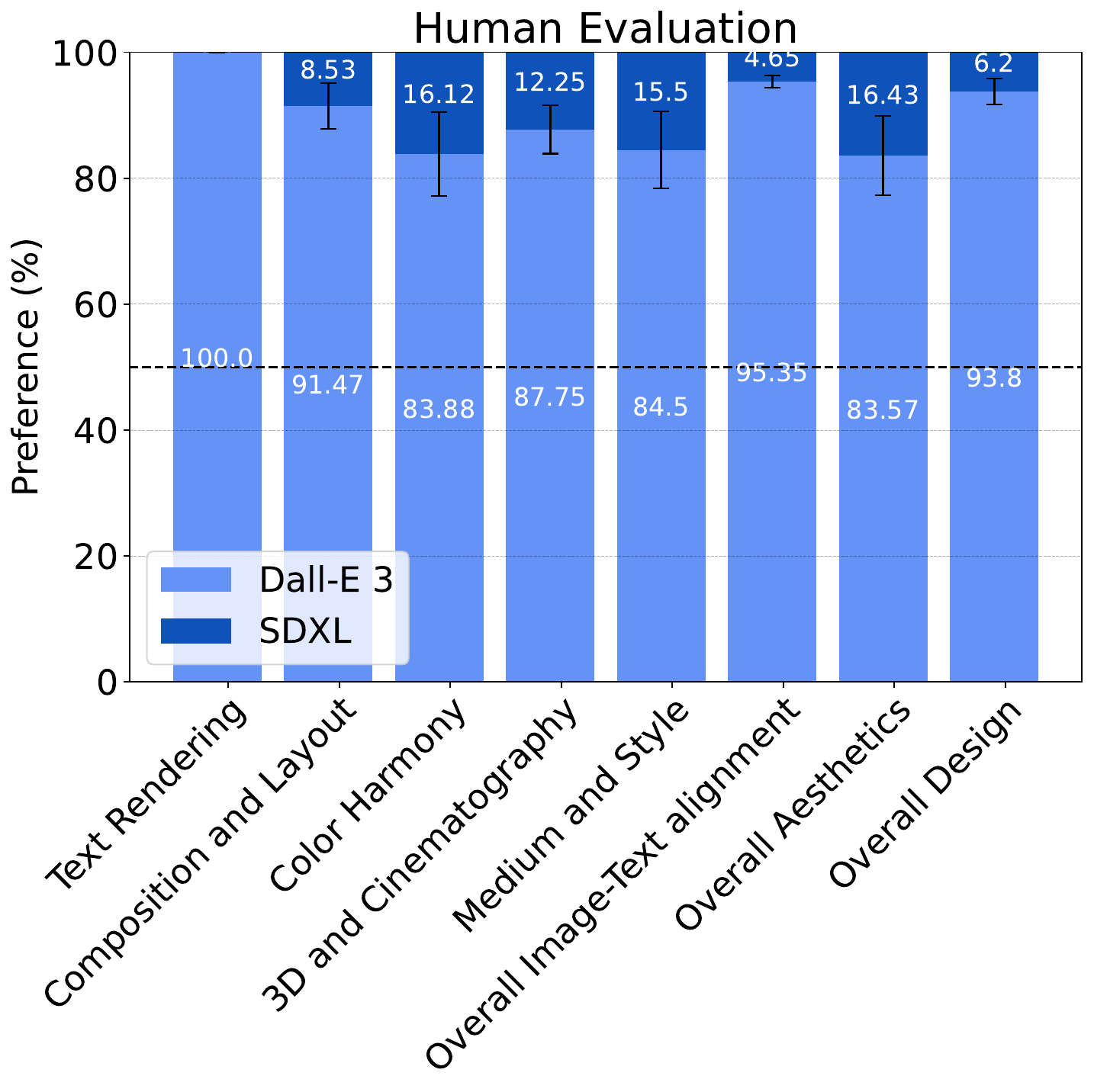} \\ \includegraphics[width=0.486\textwidth]{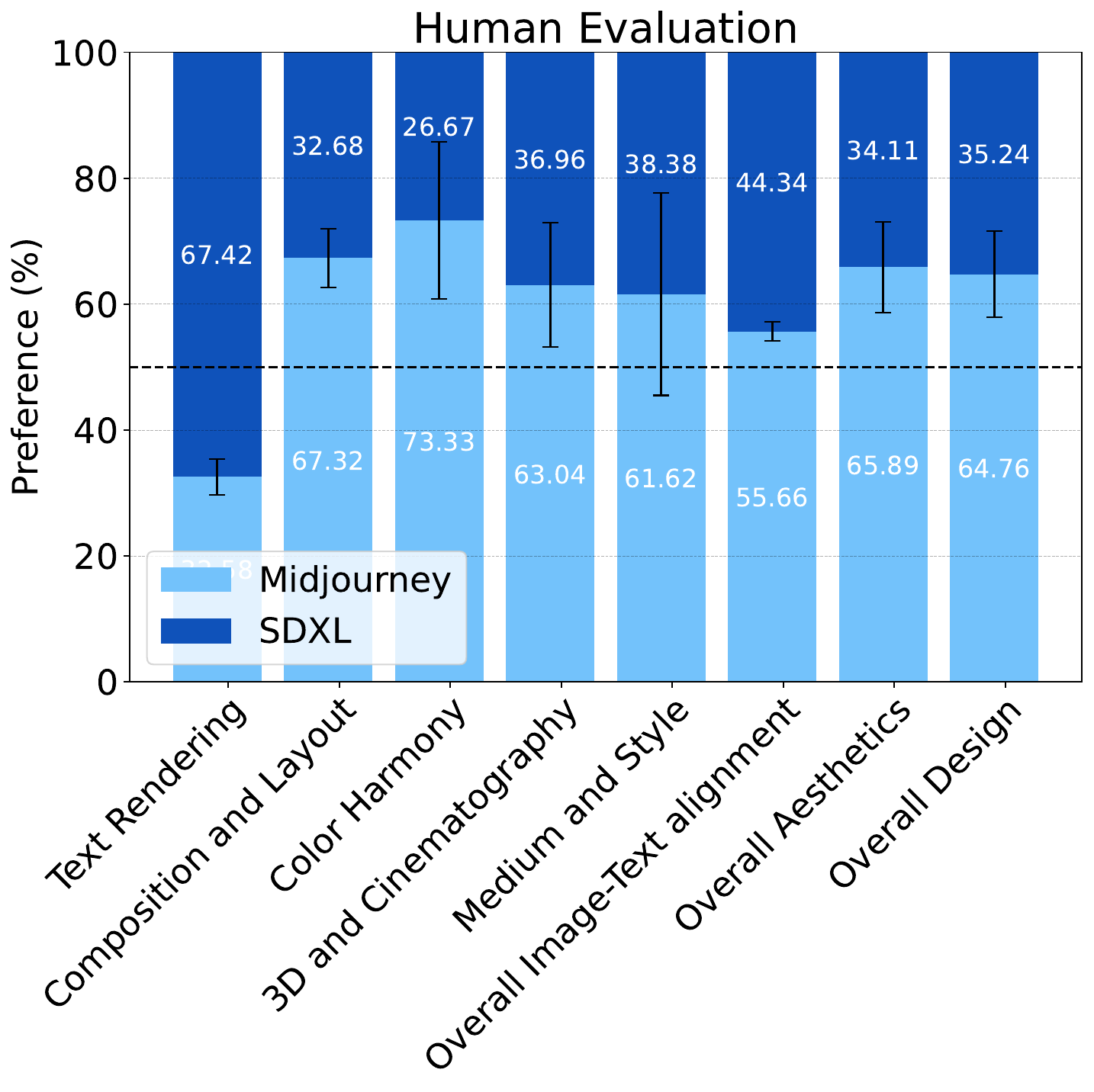} &
\includegraphics[width=0.486\textwidth]{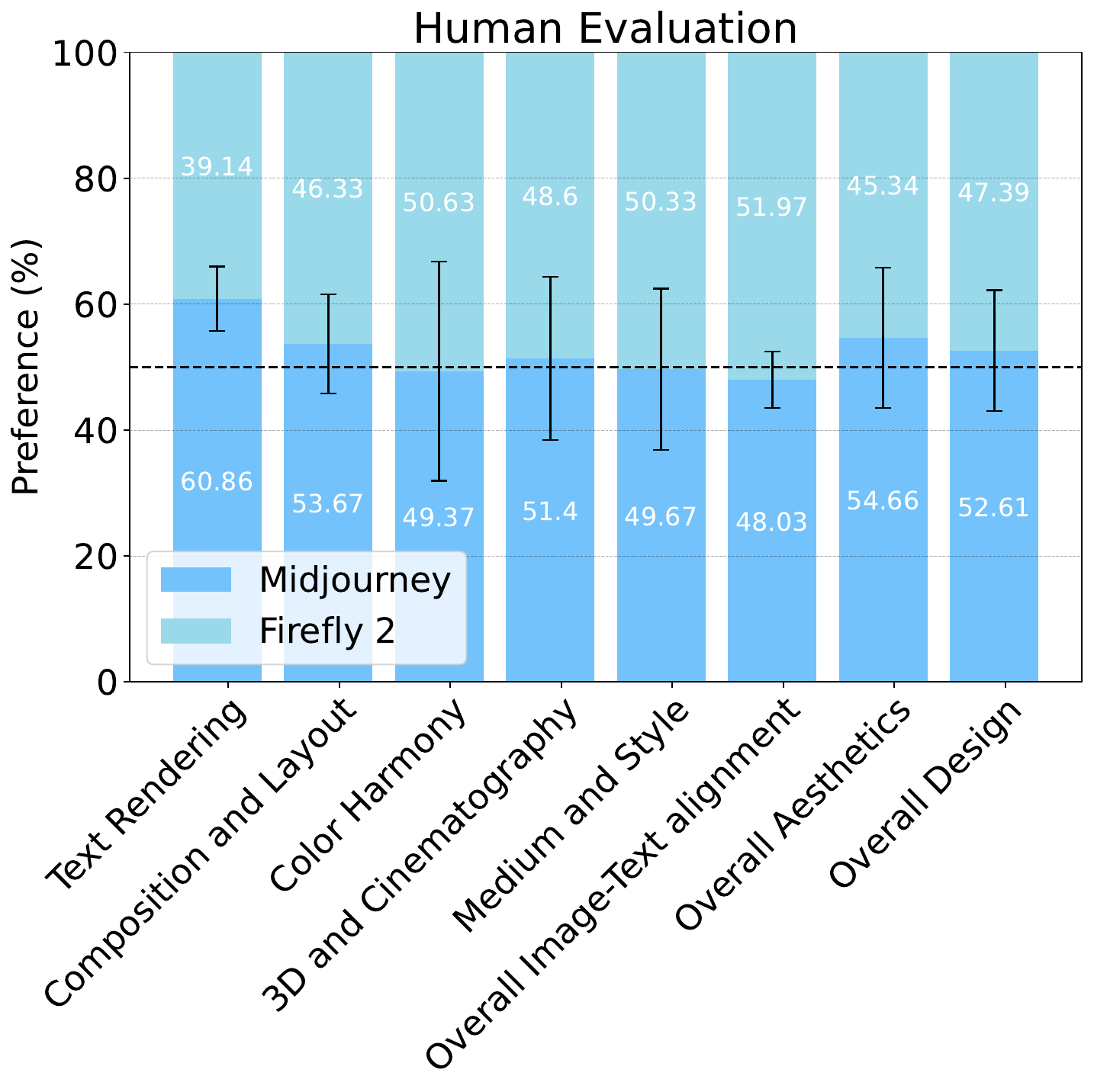} \\
\end{tabular}
\end{adjustbox}
\caption{Human evaluation results on \benchname.}
\label{fig:eval_human}
\end{figure}

\paragraph{Results on human evaluation.} Figure~\ref{fig:eval_human} shows the category-specific comparison among \modelname, Midjourney, SDXL, and Firefly 2, on \benchname. We observe that human annotators prefer the images generated by \modelname more than those of Midjourney and SDXL in all eight categories considered. In addition, Midjourney garnered preference over SDXL in seven out of the eight categories, except for text rendering. Midjourney proves slightly more favorable than Firefly2 in five out of the eight categories. These findings indicate a hierarchical preference, with \modelname emerging as the most favorable model. Midjourney and Firefly2 occupy the second tier, demonstrating substantial competence, while SDXL appears positioned within the third tier in the \benchname evaluations.

\begin{figure}
\centering
\begin{adjustbox}{center}
\small
\begin{tabular}{x{0.496\textwidth}@{}x{0.496\textwidth}}
\includegraphics[width=0.486\textwidth]{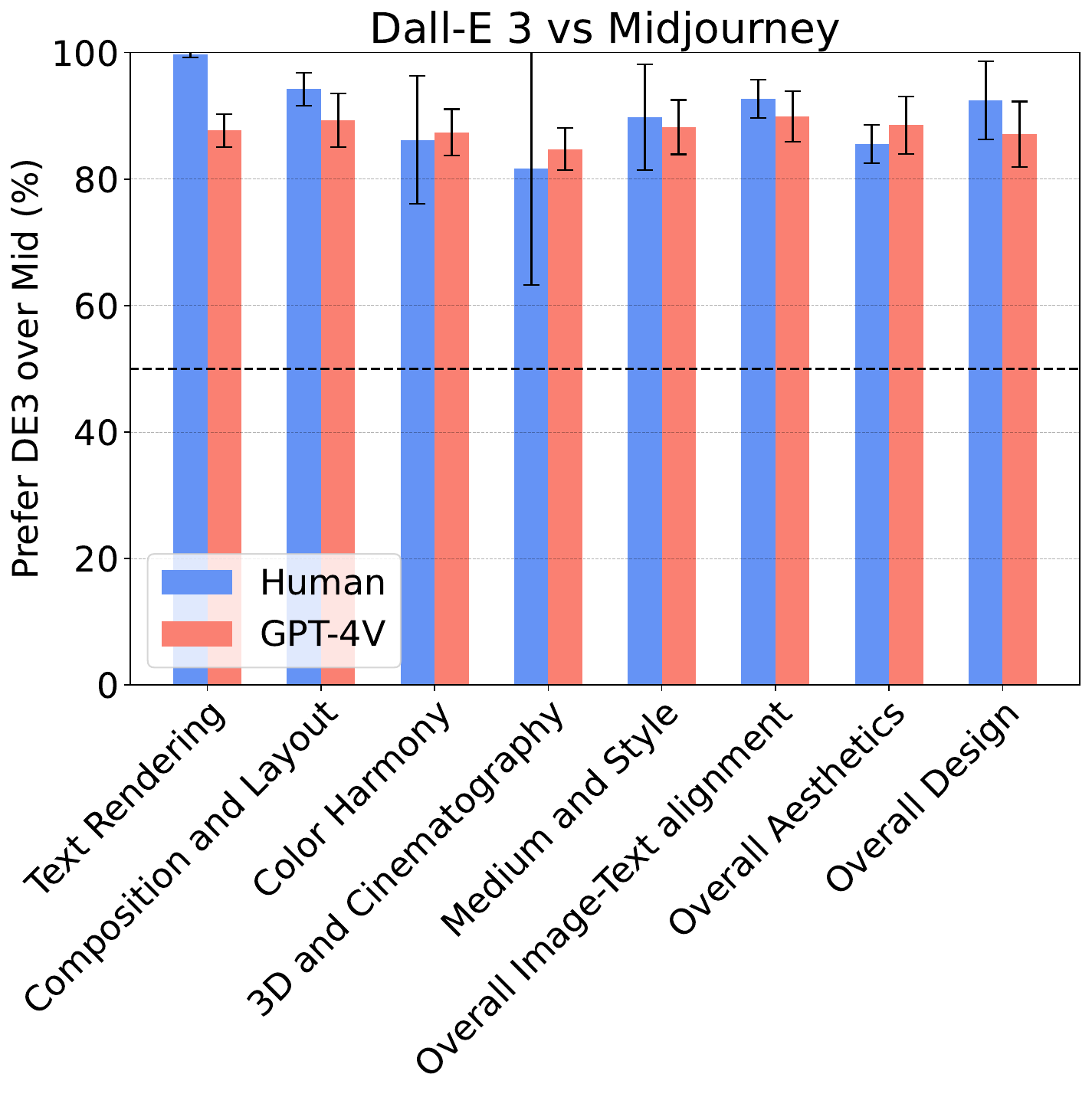}  & \includegraphics[width=0.486\textwidth]{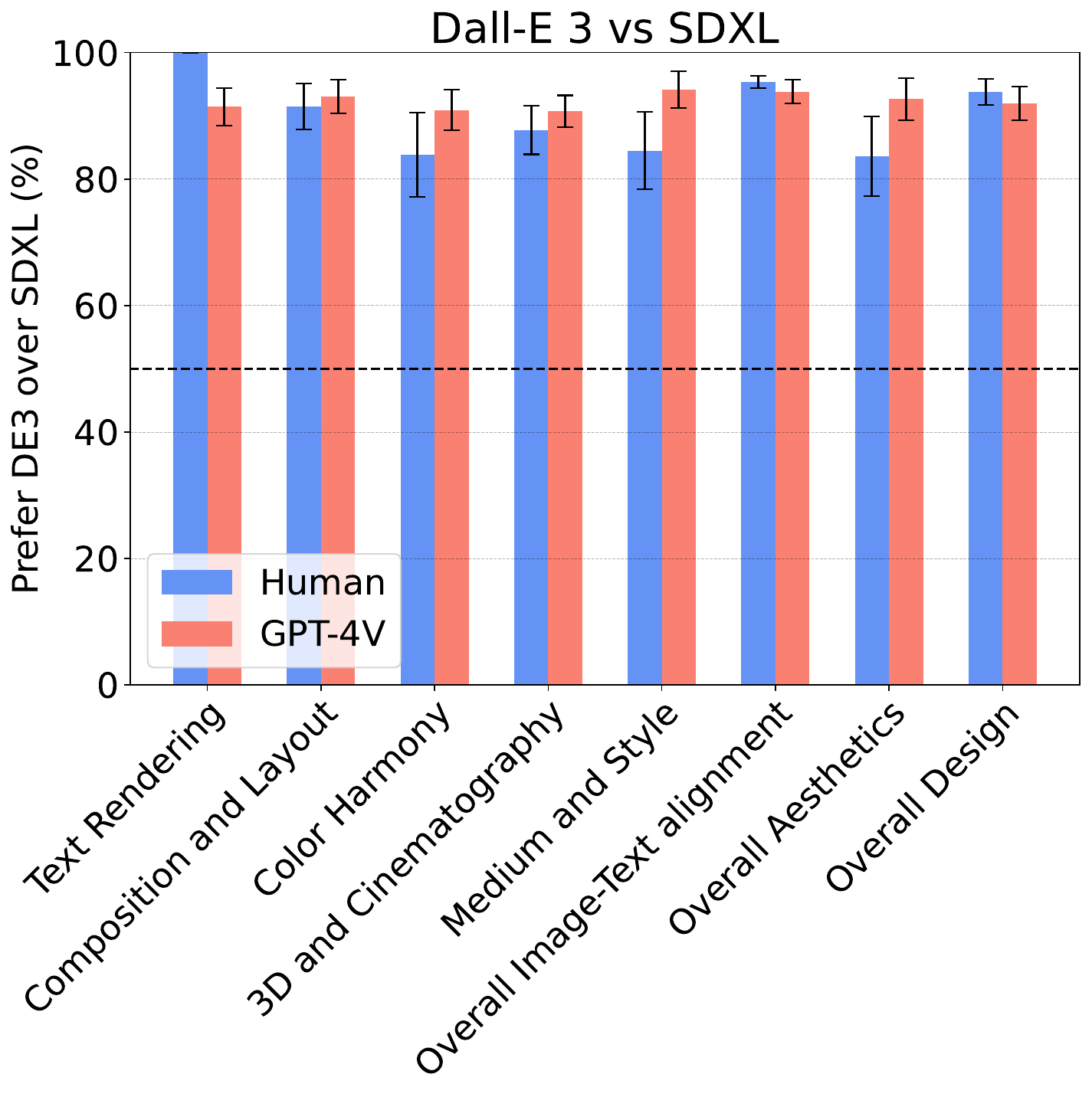} \\ \includegraphics[width=0.486\textwidth]{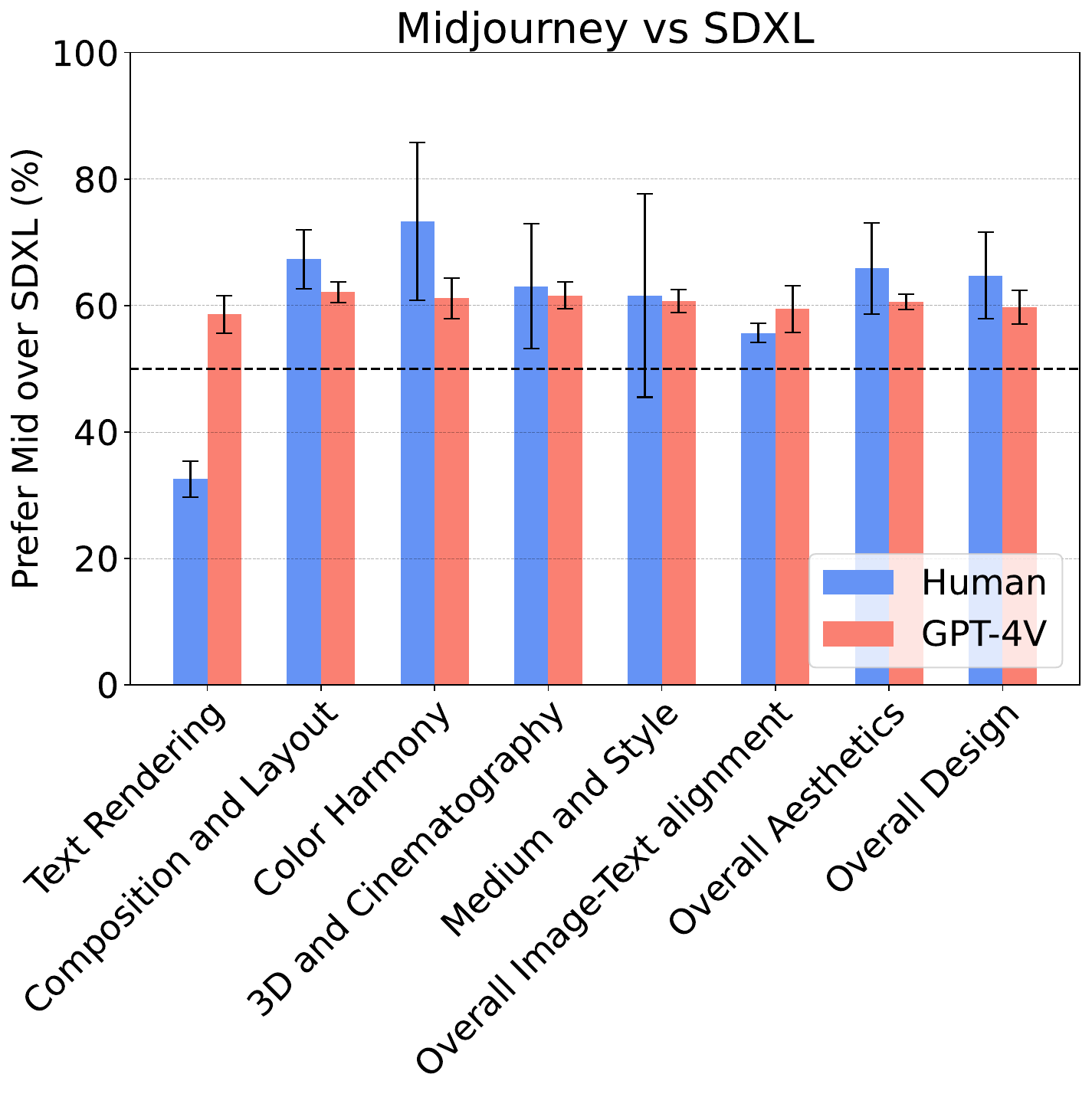} & 
\includegraphics[width=0.486\textwidth]{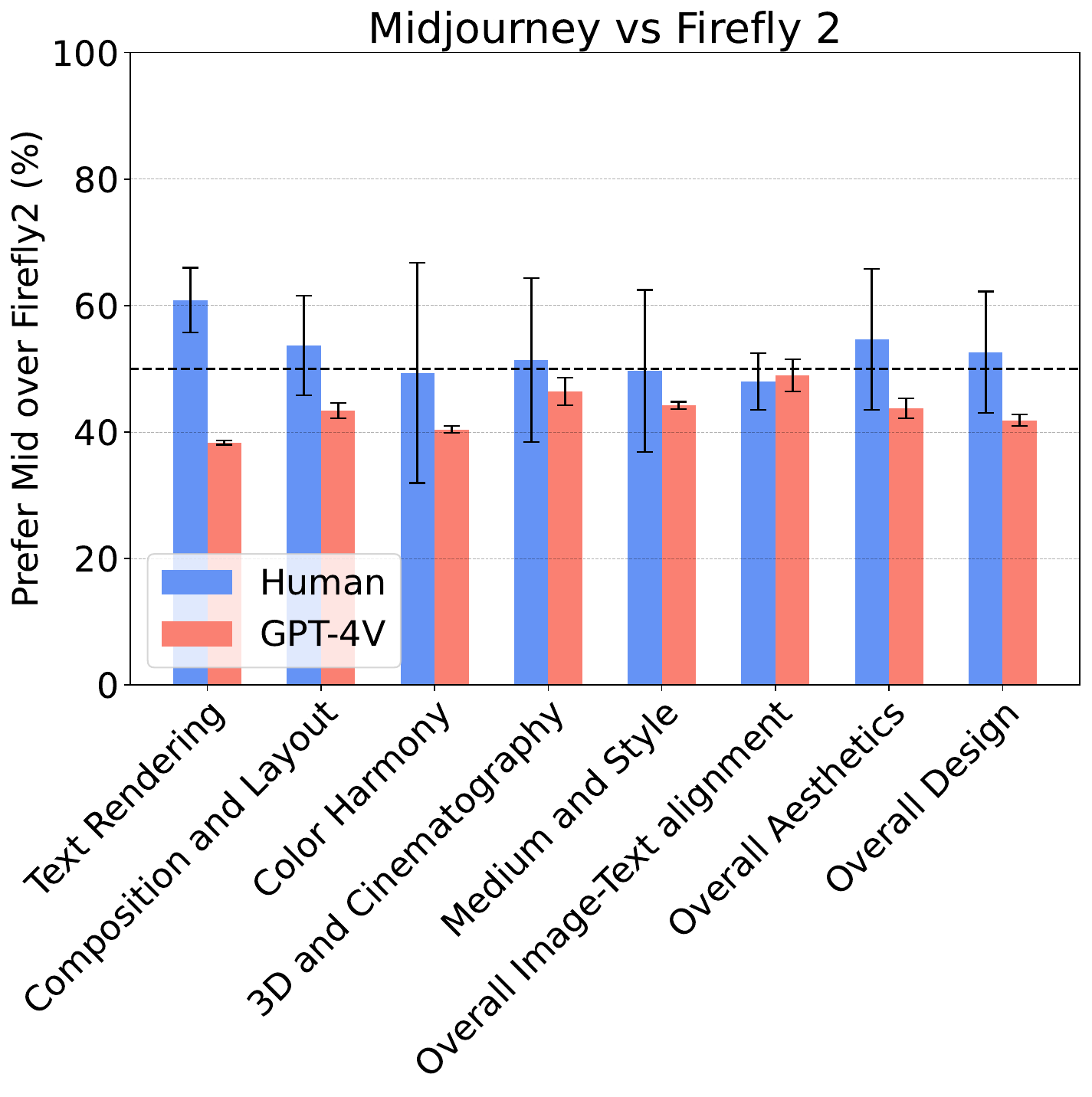}\\
\end{tabular}
\end{adjustbox}
\caption[comparison between GPT-4V and human judgments on \benchname]{Comparison between GPT-4V and human judgments on \benchname. GPT-4V's assessments are aligned with human judgments in most cases.}
\label{fig:eval_consist}
\end{figure}

\paragraph{Results on GPT-4V evaluation.} To assess the efficacy of GPT-4V as an automated evaluator, we conduct a consistency analysis. Figure~\ref{fig:eval_consist} illustrates the correlation between human preferences and the assessments executed by GPT-4V on \benchname. This analysis involved invoking the GPT-4V five times, and subsequently reporting on the mean and variance of the results. %
Our observations indicate that the judgments by GPT-4V predominantly concur with human evaluations, with sporadic discrepancies most notable in the evaluation of text rendering capabilities when comparing  Midjourney-SDXL and Midjourney-Firefly2.
Despite these occasional divergences, GPT-4V exhibits relatively reliable performance across a spectrum of evaluative criteria, demonstrating its potential as an automated tool for T2I evaluation, particularly in pairwise comparisons.

\begin{figure}
\centering
\begin{adjustbox}{center}
\small
\begin{tabular}{x{0.496\textwidth}@{}x{0.496\textwidth}@{}}
\includegraphics[width=0.486\textwidth]{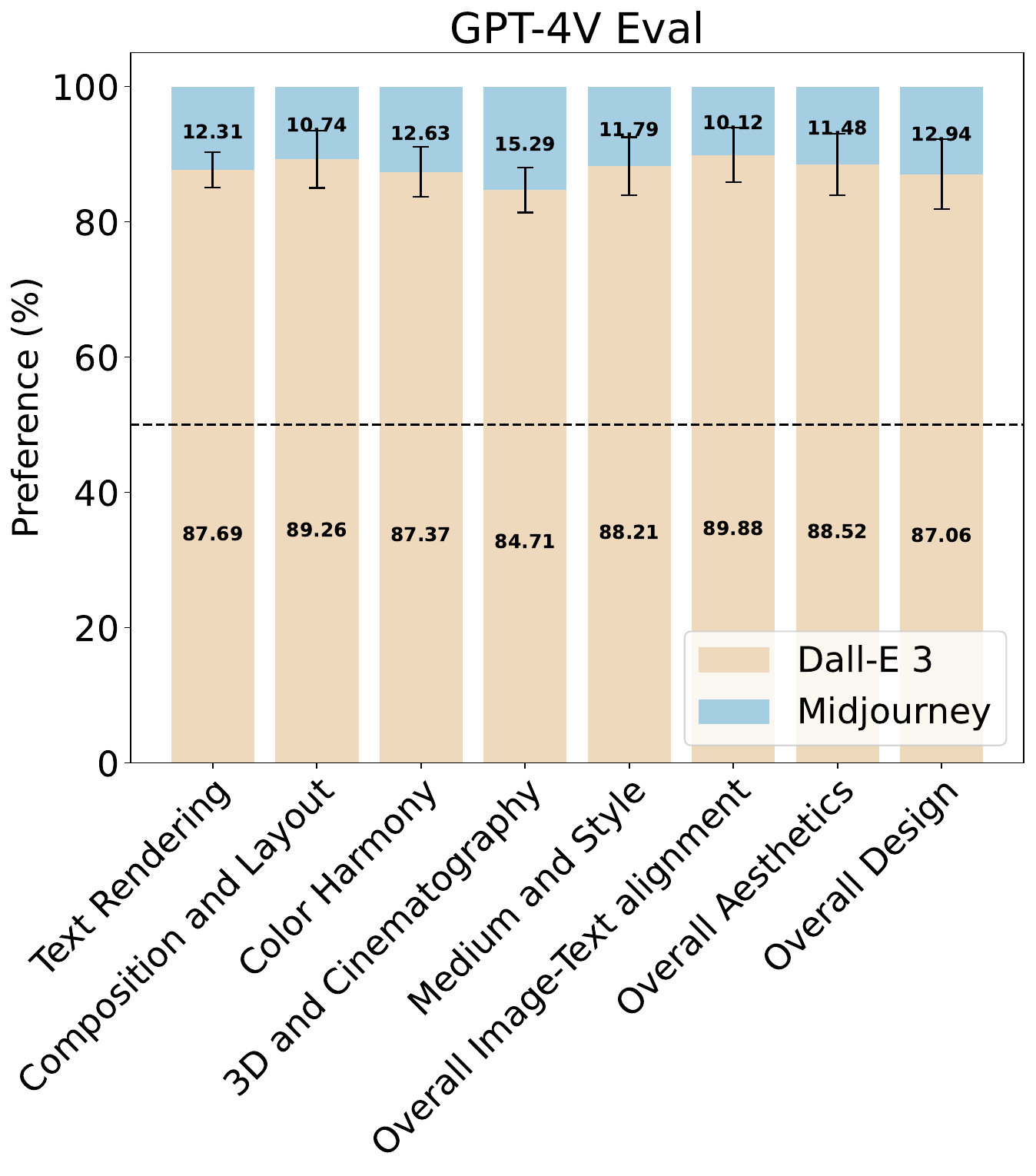}  & \includegraphics[width=0.486\textwidth]{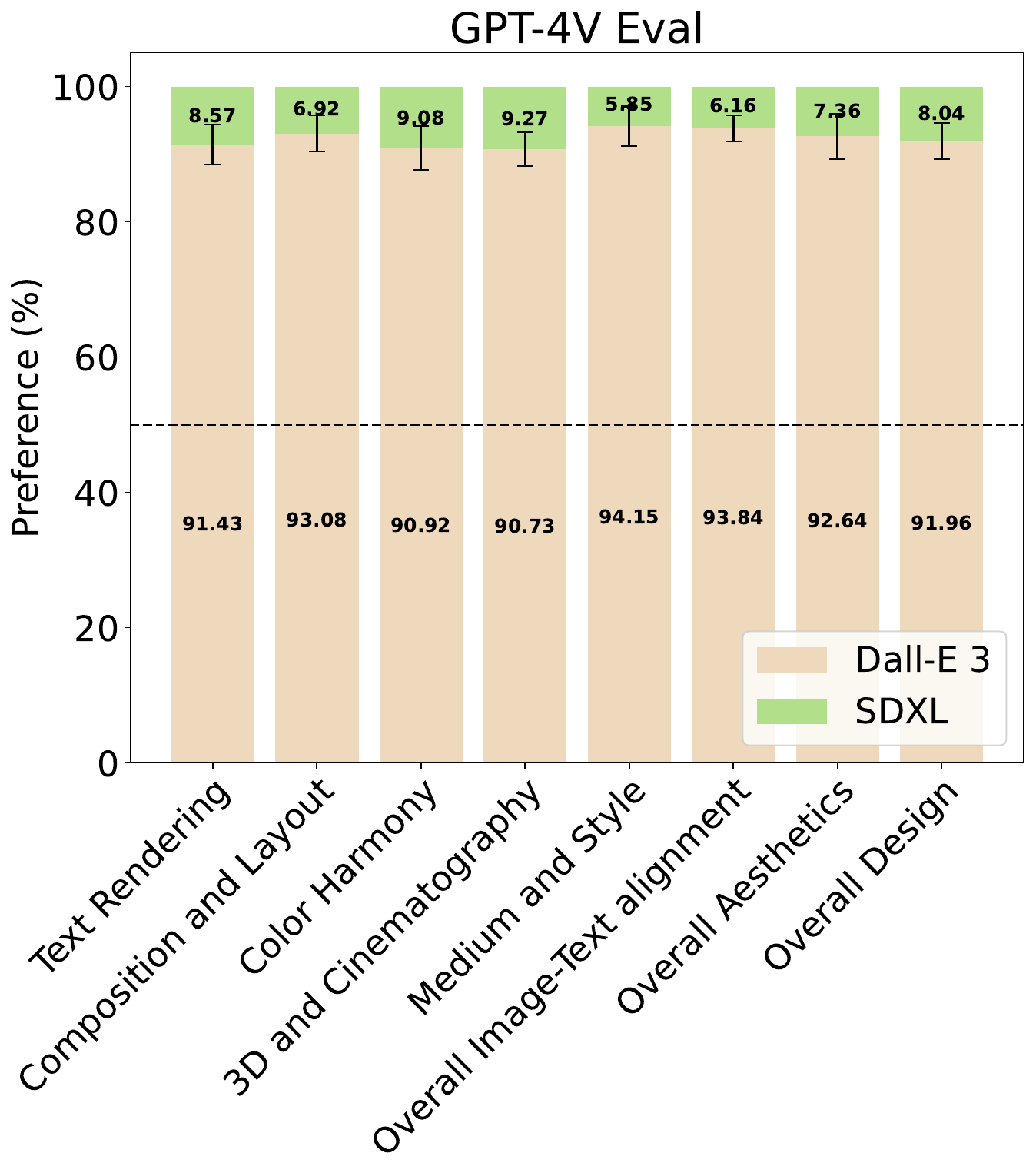} \\
\includegraphics[width=0.486\textwidth]{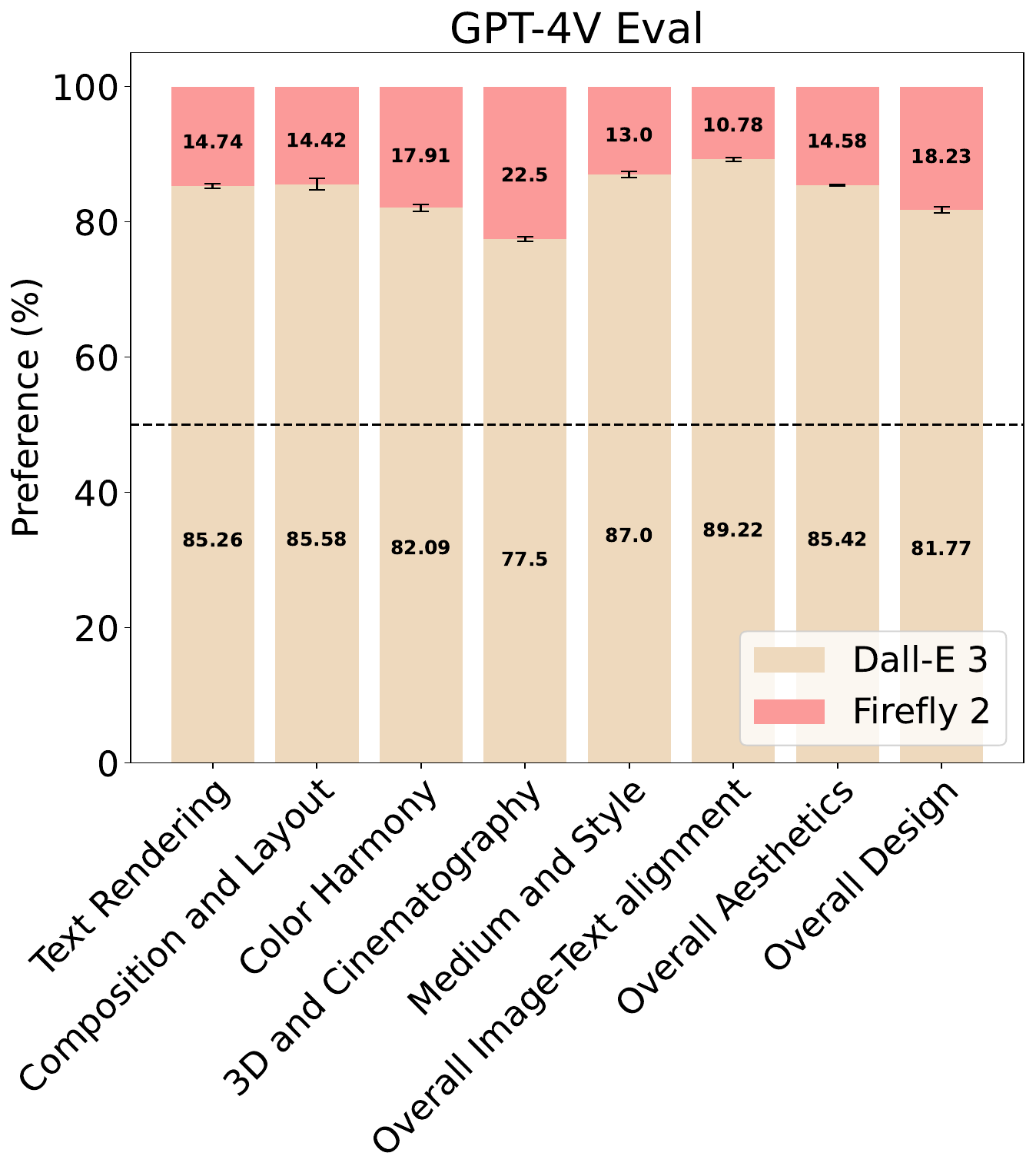}  & \includegraphics[width=0.486\textwidth]{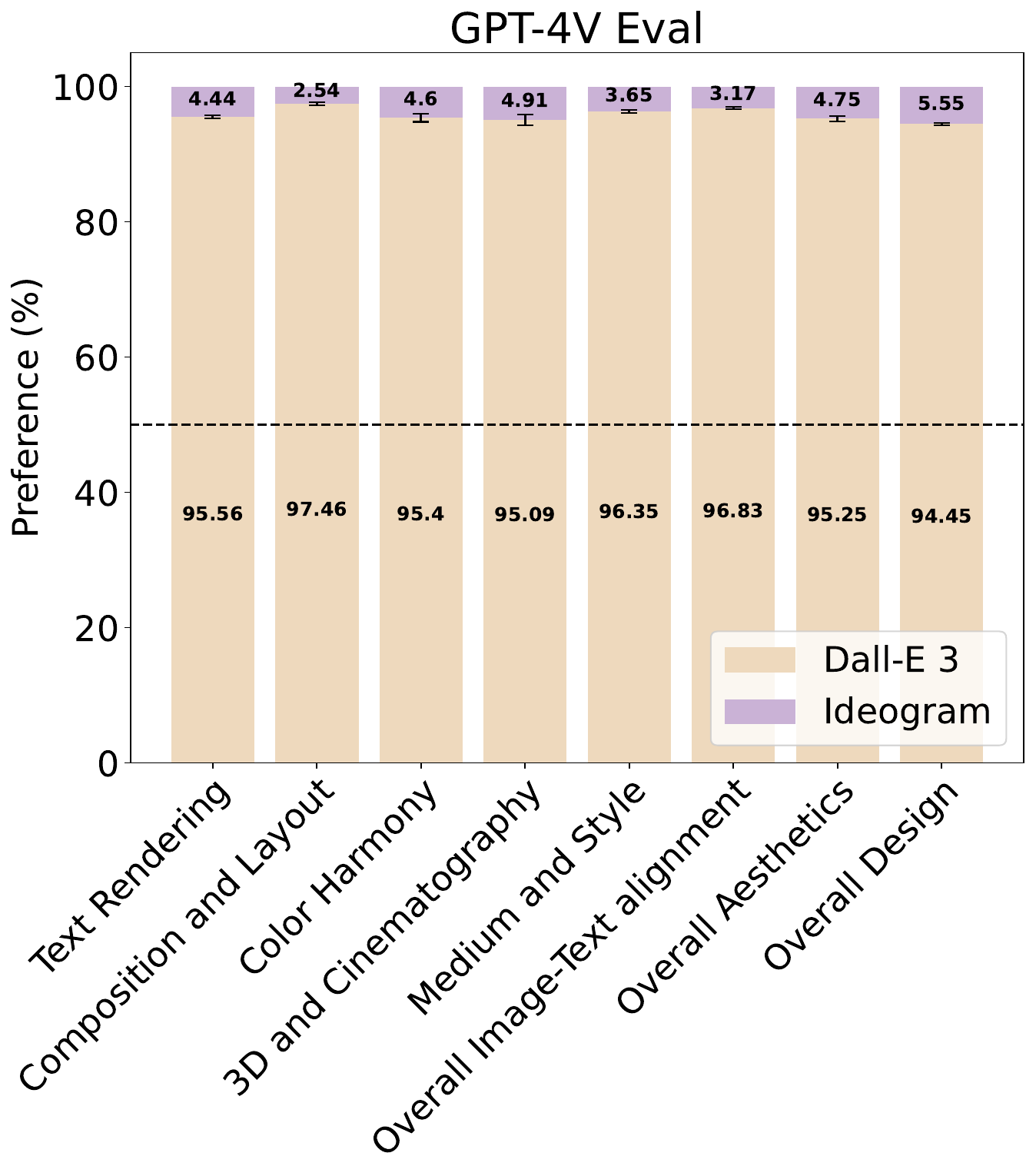} 
\end{tabular}
\end{adjustbox}
\caption[GPT-4V evaluation on \benchname]{GPT-4V evaluation on \benchname. GPT-4V compares \modelname with state of the art T2I models, including Firefly 2, Midjourney, SDXL, and Ideogram.}
\label{fig:gptv_eval_figure1}
\end{figure}

\begin{figure}
\centering
\begin{adjustbox}{center}
\small
\begin{tabular}{x{0.496\textwidth}@{}x{0.496\textwidth}@{}x{0.496\textwidth}}
\includegraphics[width=0.486\textwidth]{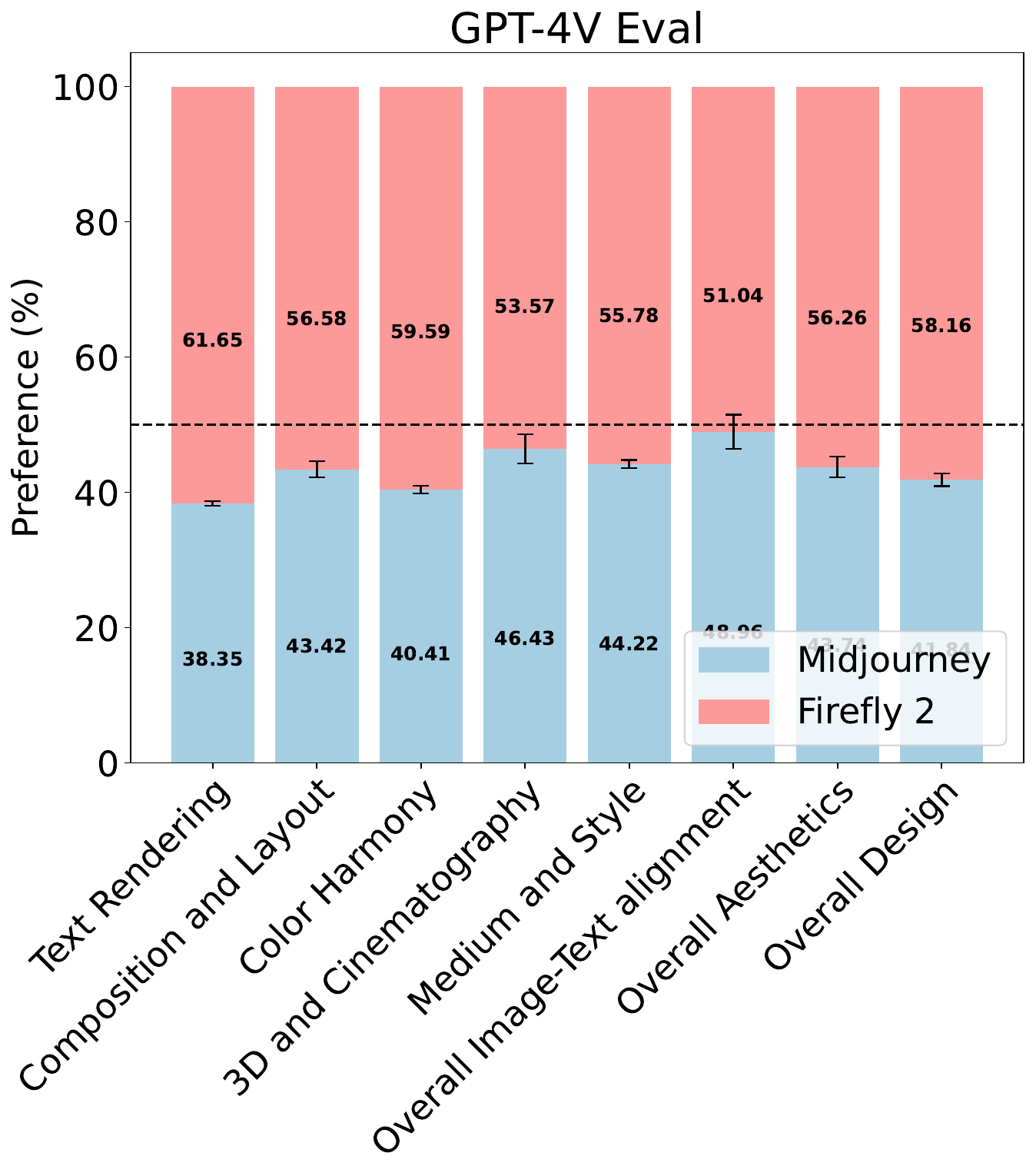} & \includegraphics[width=0.486\textwidth]{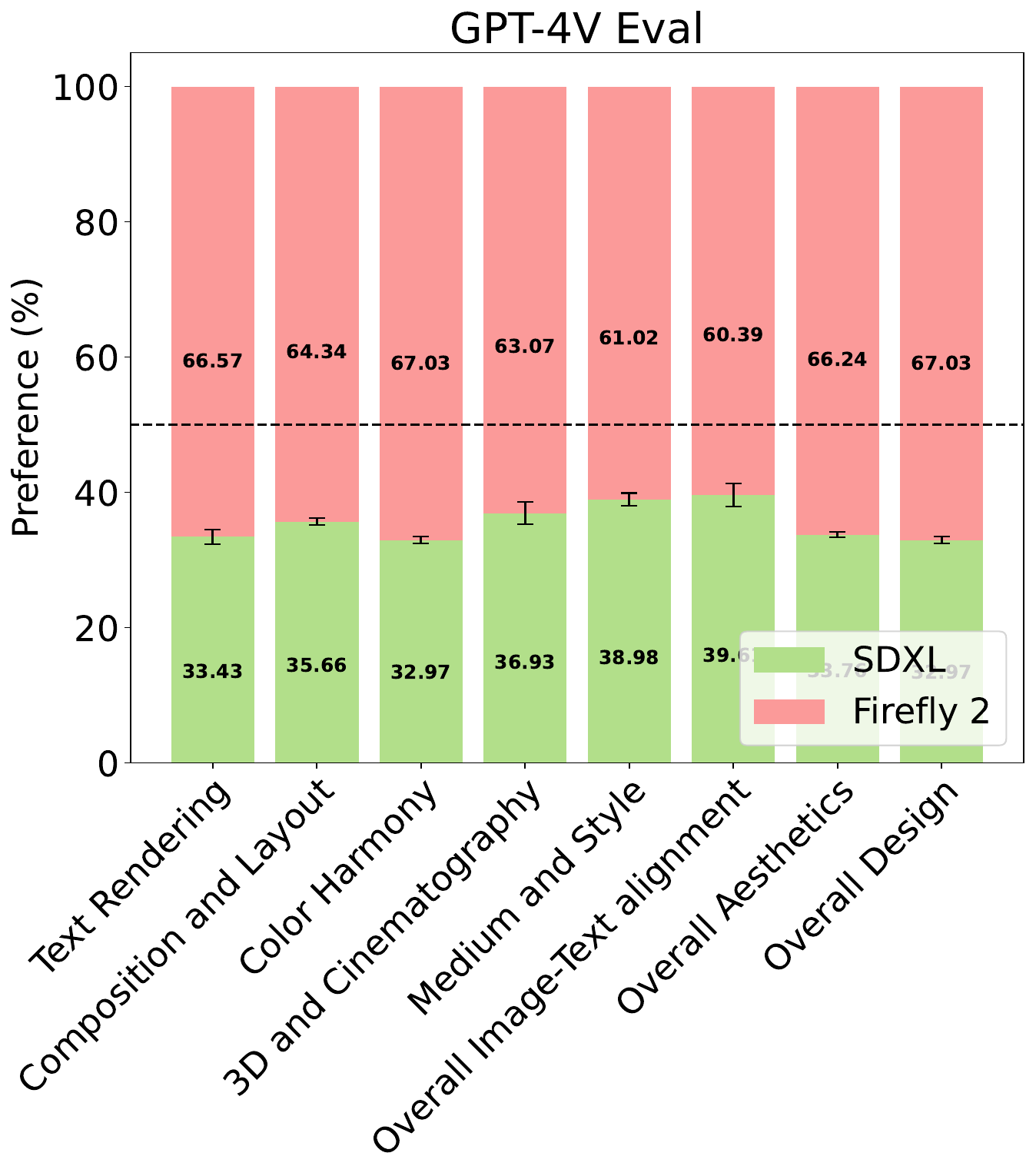} & \includegraphics[width=0.486\textwidth]{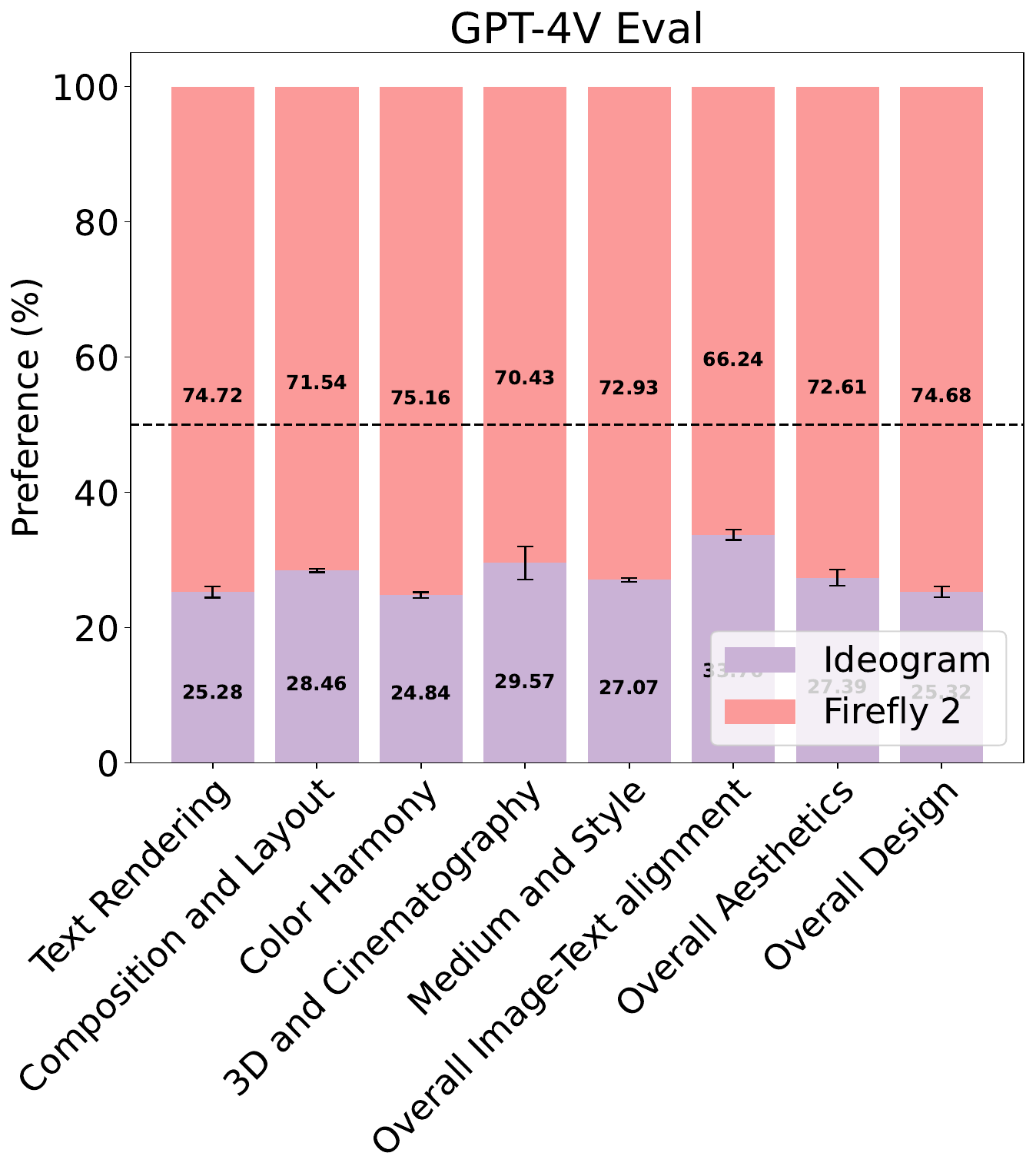} \\
  \includegraphics[width=0.486\textwidth]{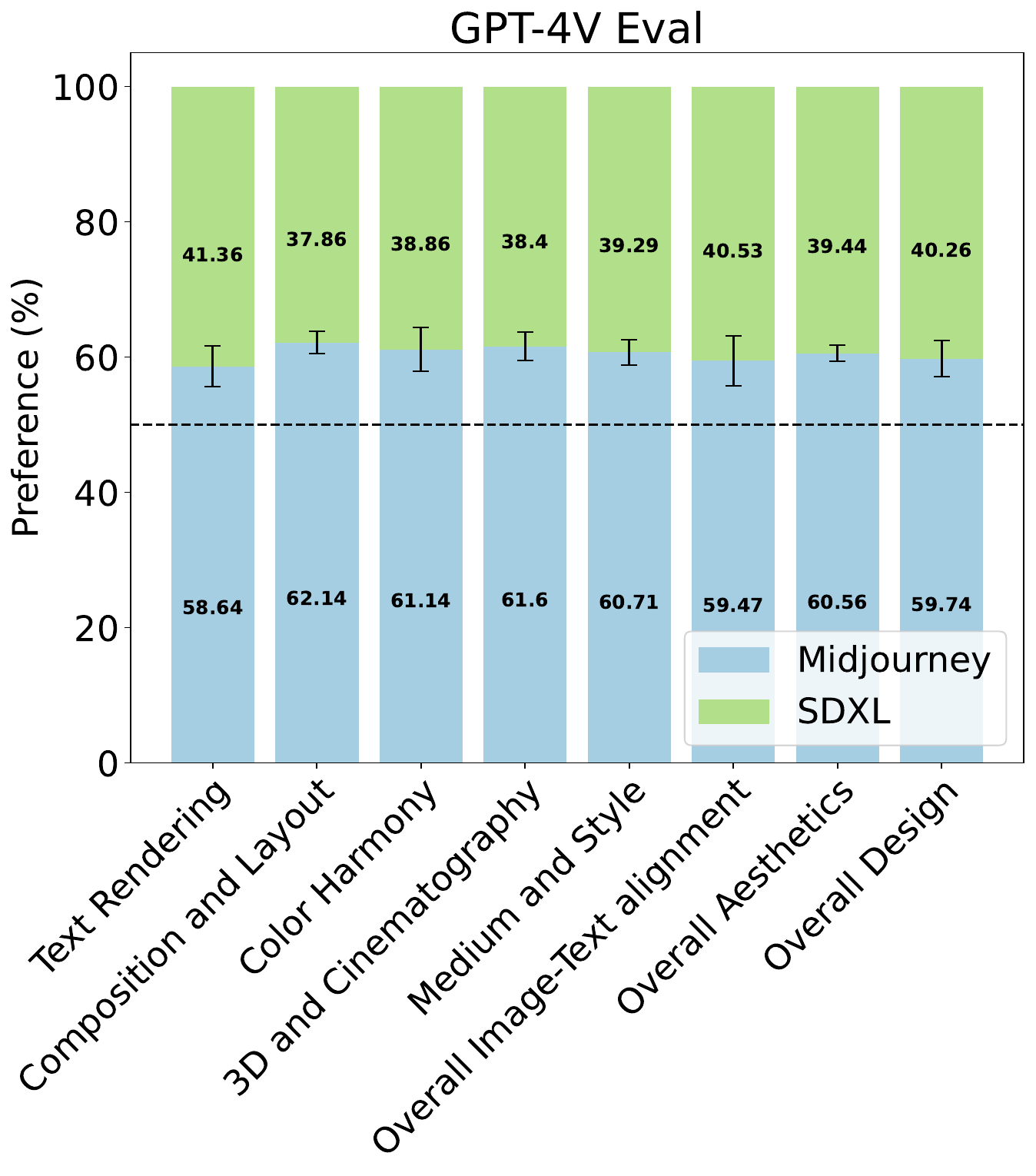}  & \includegraphics[width=0.486\textwidth]{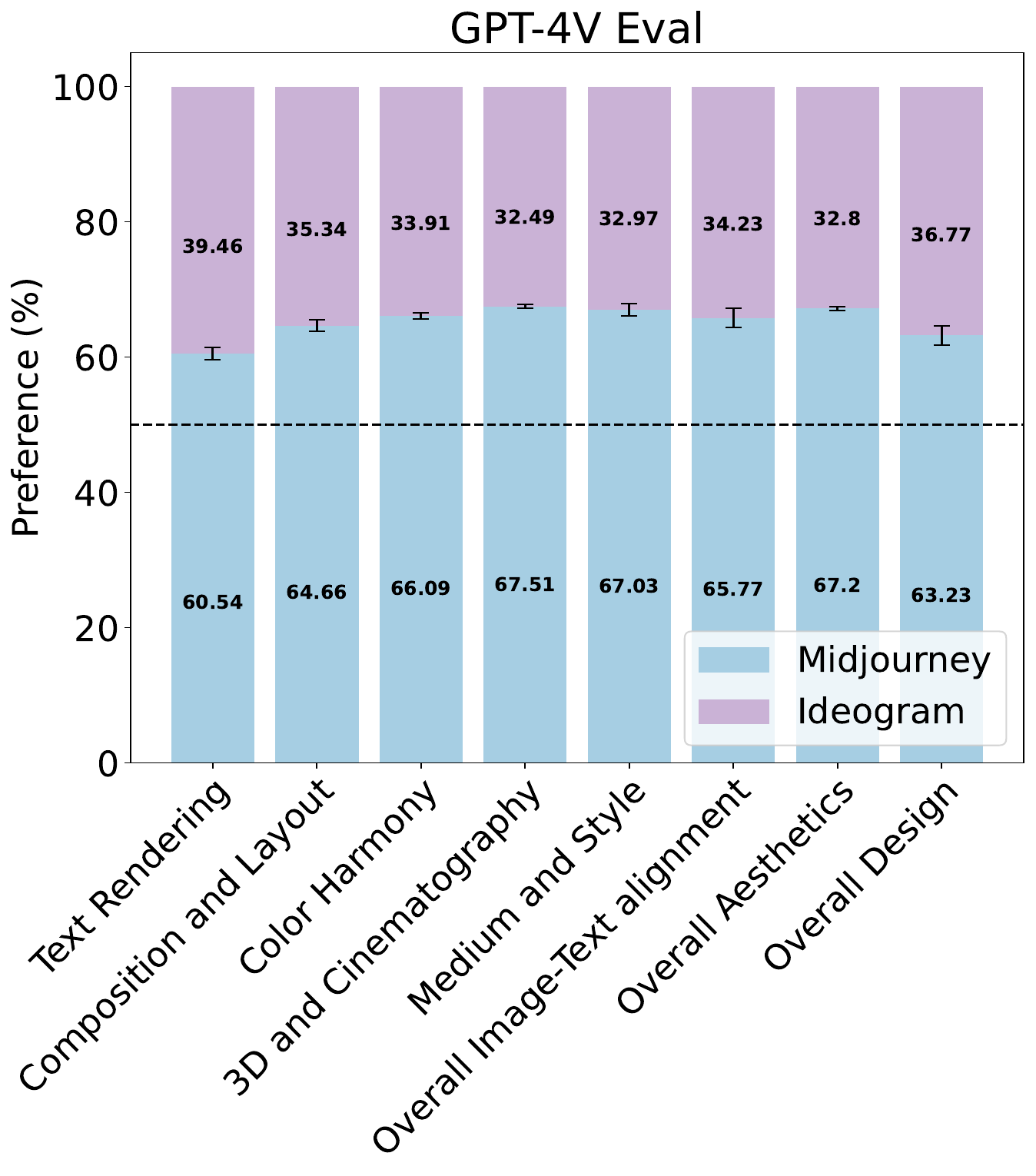} &
\includegraphics[width=0.486\textwidth]{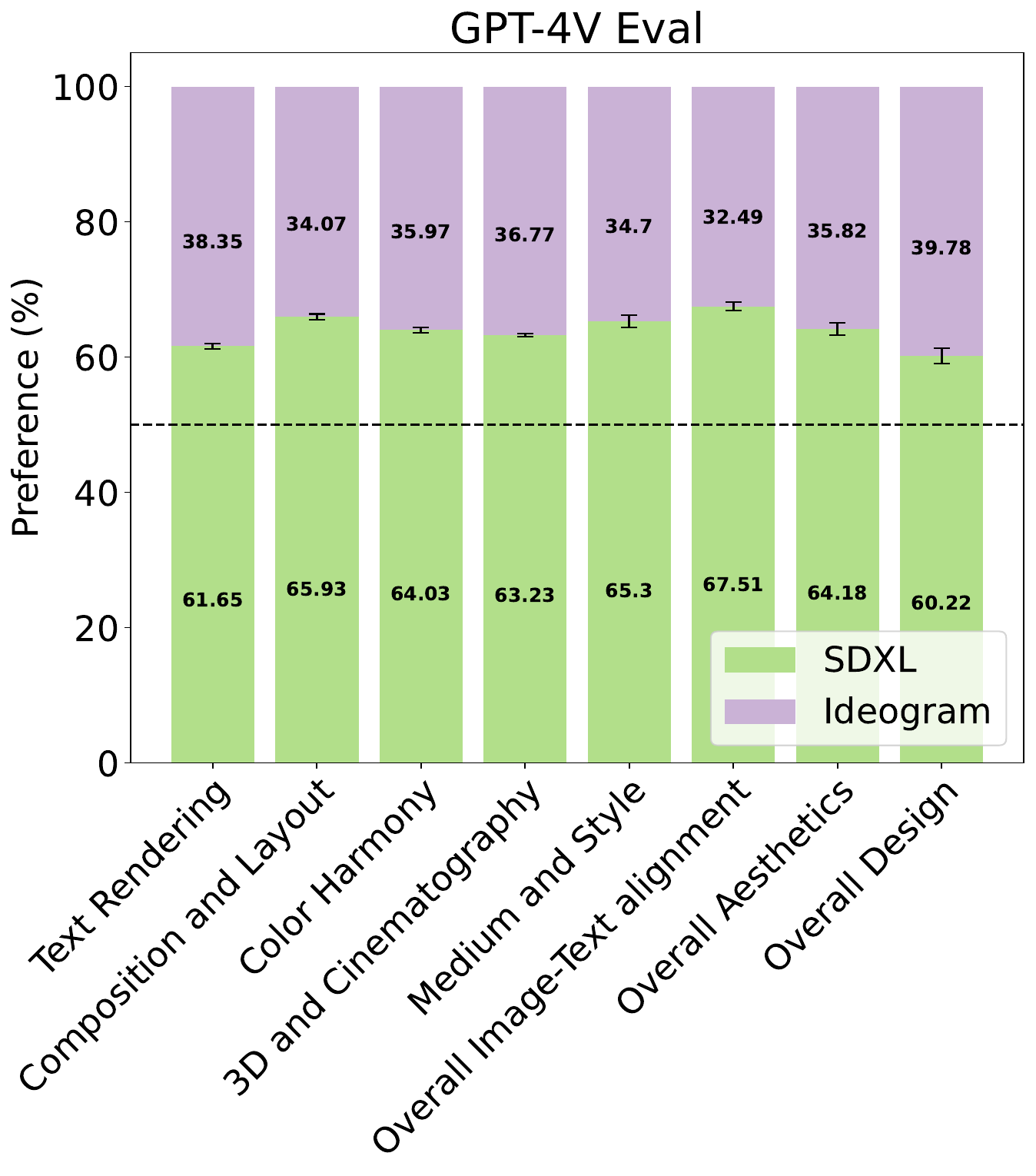}
\end{tabular}
\end{adjustbox}
\caption[GPT-4V evaluation on \benchname]{GPT-4V evaluation on \benchname. GPT-4V compares the performance of the state-of-the-art T2I models, including Firefly 2, Midjourney, SDXL, and Ideogram.}
\label{fig:gptv_eval_figure2}
\end{figure}

Figures~\ref{fig:gptv_eval_figure1}-\ref{fig:gptv_eval_figure2} show the GPT-4V evaluation results on comparing the five T2I models considered. In the experiments, we invoke the GPT-4V five times, and report the mean and variance of the results. \modelname stands out as the most favorable model, followed by Firefly 2 and Midjourney within the second tier. SDXL and Ideogram are positioned within the third tier. We observe a notable consistency in GPT-4V evaluation, given the absence of any cyclical anomalies in the pairwise comparisons reviewed.

Finally, we present example outputs of GPT-4V evaluator in Tables~\ref{tab:gptv-example-1}-\ref{tab:gptv-example-4}. We observe that GPT-4V can correctly analyze the images and make reasonable assessments. Tables~\ref{tab:gptv-fail-1}-\ref{tab:gptv-fail-2} show representative failure cases of the GPT-4V evaluator. We observe that GPT-4V may make a mistake in counting the teddy bears in the occlusion scenario. GPT-4V may struggle to read the small text, and instead shift its attention towards evaluating the overall aesthetics of the image.

\begin{table}[t]
\caption{Example result from GPT-4V. Given the text prompt and two images, GPT-4V compares the two image, and makes a reasonable assessment. The key rationale is highlight in \colorbox{yellowhl}{yellow}. Note that Image 1 is generated by \modelname, and Image 2 is generated by Midjourney.
}
\label{tab:gptv-example-1}
\centering
\begin{tcolorbox} 
    \centering
    \small
    \begin{tabular}{p{0.98\columnwidth}}

\textbf{Aspect:} Text Rendering

\textbf{Expanded Prompt:} 3D rendering of the Lion of Knidos, depicted in its original majestic form. The sculpture stands proud, capturing the essence of its ancient artistry. At the base of the sculpture, a label is prominently displayed, reading: 'Lion of Knidos'.\\

\textbf{Image 1:}\\
\includegraphics[width=0.7\textwidth]{figures/rationale_example/de3/3dart_3_0.png}\\
\textbf{Image 2:}\\ 
\includegraphics[width=0.7\textwidth]{figures/rationale_example/mid/3dart_3_0.png}\\

--------------------------------------------------------------------------------------------------------------------

\textbf{GPT-4V:}\\
\colorbox{yellowhl}{In Image 1, the text ``LION OF KNIDOS'' is clearly legible}, well-styled with appropriate spacing, and alignment that complements the overall design. However, in Image 2, there isn't any visible text rendering on the displayed label. Given the specific criteria, Image 1 excels in showcasing the text as per the caption. | \colorbox{yellowhl}{Image 1}
    \end{tabular}
\end{tcolorbox}
\end{table}

\begin{table}[t]
\caption{Example result from GPT-4V. Given the text prompt and two images, GPT-4V compares the two image, and makes a reasonable assessment. The key rationale is highlight in \colorbox{yellowhl}{yellow}. Note that Image 1 is generated by \modelname, and Image 2 is generated by SDXL.
}
\label{tab:gptv-example-2}
\centering
\begin{tcolorbox} 
    \centering
    \small
    \begin{tabular}{p{0.98\columnwidth}}
\textbf{Aspect:} Overal Design

\textbf{Expanded Prompt:} 3D render showcasing a tranquil indoor setting. Against a wall stands a large window, allowing a cascade of soft daylight to enter the room. This gentle illumination highlights a scene on a table beside the window: a perfect cube with a cone precisely positioned on its top. The cone's pointed tip contrasts with the cube's sharp edges, and the sunlight casts subtle shadows, emphasizing the geometric interplay. The background is minimalistic, ensuring the focus remains on the cone, cube, and the interplay of natural light.\\

\textbf{Image 1:}\\
\includegraphics[width=0.5\textwidth]{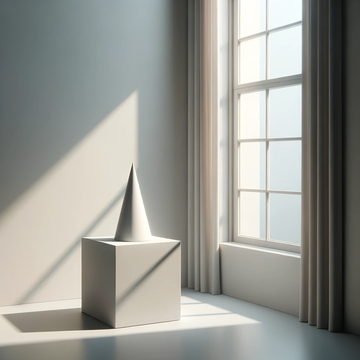}\\
\textbf{Image 2:}\\ 
\includegraphics[width=0.5\textwidth]{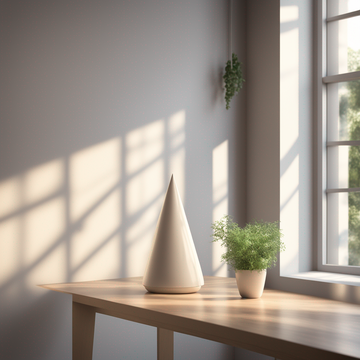}\\

--------------------------------------------------------------------------------------------------------------------

\textbf{GPT-4V:}\\
Image 1 beautifully highlights the geometric interplay and natural light emphasized in the caption. The minimalistic design keeps the focus on the cube, cone, and shadow interplay, offering a precise interpretation of the caption in a visually appealing manner. \colorbox{yellowhl}{Image 2, although serene, lacks the cube, deviating from the specified scene.} | \colorbox{yellowhl}{Image 1}
    \end{tabular}
\end{tcolorbox}
\end{table}

\begin{table}[t]
\caption{Example result from GPT-4V. Given the text prompt and two images, GPT-4V compares the two image, and makes a reasonable assessment. The key rationale is highlight in \colorbox{yellowhl}{yellow}. Note that Image 1 is generated by \modelname, and Image 2 is generated by Midjourney.
}
\label{tab:gptv-example-3}
\centering
\begin{tcolorbox} 
    \centering
    \small
    \begin{tabular}{p{0.98\columnwidth}}
\textbf{Aspect:} Overall Design

\textbf{Expanded Prompt:} A wide depiction of London in the year 1816. The cityscape showcases iconic landmarks like the Tower Bridge and the River Thames, although they appear more historical and less developed than in modern times. Cobblestone streets are bustling with pedestrians in period attire, horse-drawn carriages, and street vendors. Smoke rises from chimneys, and the skyline is dotted with spires, domes, and brick buildings. The overall ambiance is that of a city in the midst of the Industrial Revolution, with a mix of grandeur and grit.\\

\textbf{Image 1:}\\
\includegraphics[width=0.7\textwidth]{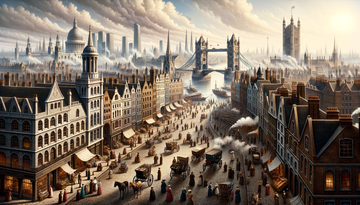}\\
\textbf{Image 2:}\\ 
\includegraphics[width=0.7\textwidth]{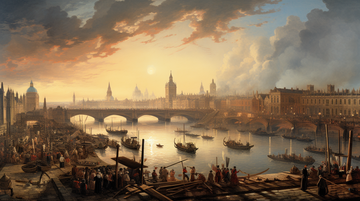}\\

--------------------------------------------------------------------------------------------------------------------

\textbf{GPT-4V:} Image 1 offers a detailed depiction aligning well with the Institute of Revolution ambiance; however, \colorbox{yellowhl}{Image 2's picturesque quality, combined with the attention to lighting and atmosphere}, provides a striking visual experience that \colorbox{yellowhl}{feels more immersive and aligns coherently} with the translational qualities of the caption. | \colorbox{yellowhl}{Image 2}\\

    \end{tabular}
\end{tcolorbox}
\end{table}

\begin{table}[t]
\caption{Example result from GPT-4V. Given the text prompt and two images, GPT-4V compares the two image, and makes a reasonable assessment. The key rationale is highlight in \colorbox{yellowhl}{yellow}. Note that Image 1 is generated by \modelname, and Image 2 is generated by Midjourney.
}
\label{tab:gptv-example-4}
\centering
\begin{tcolorbox} 
    \centering
    \small
    \begin{tabular}{p{0.98\columnwidth}}
\textbf{Aspect:} Overall Aesthetics

\textbf{Expanded Prompt:} Photo of an intense action movie scene inside a restaurant. In side view, a lady with blue eyes and a short salt and pepper buzz cut stands assertively, her attire consisting of a black tactical shirt, dark jeans, a utility belt, leather gloves, and combat boots. She tightly grips a matte black semi-automatic pistol, searching for another individual menacingly brandishing a knife, hiding in low position behind one of the tables. The blurred background reveals overturned tables and broken dishes scattered across the floor, symbolizing the fierce battle that has just occurred.

\\

\textbf{Image 1:}\\
\includegraphics[width=0.7\textwidth]{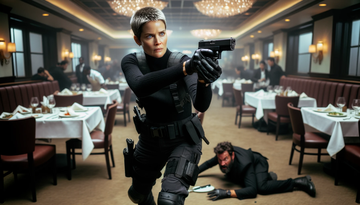}\\
\textbf{Image 2:}\\ 
\includegraphics[width=0.7\textwidth]{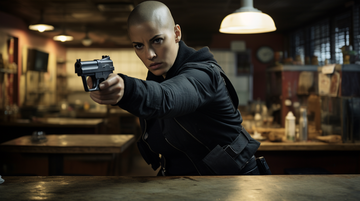}\\

--------------------------------------------------------------------------------------------------------------------

\textbf{GPT-4V:} Image 1 accurately captures the intense action scene in a restaurant. The assertive lady, the menacing individual, overturned tables, and scattered dishes are effectively depicted. However, the pistol isn't explicitly matte black. Image 2, though clear and intense, lacks the depicted chaos and the individual with a knife. However, it offers better visual aesthetics. \colorbox{yellowhl}{Despite not strictly adhering to the caption, Image 2 is chosen for its superior visual quality.}| \colorbox{yellowhl}{Image 2}\\

    \end{tabular}
\end{tcolorbox}
\end{table}

\begin{table}[t]
\caption{Failure case in GPT-4V evaluation. Incorrect rationale is highlight in \colorbox{redhl}{red}. Note that Image 1 is generated by Midjourney, and Image 2 is generated by SDXL.
}
\label{tab:gptv-fail-1}
\centering
\begin{tcolorbox} 
    \centering
    \small
    \begin{tabular}{p{0.98\columnwidth}}
\textbf{Aspect:} Image-Text Alignment

\textbf{Expanded Prompt:} Wide photo set within the University of Washington's library. Three teddy bears, distinct in color and appearance, appear as if they are deeply engrossed in their academic pursuits. They are surrounded by neat stacks of books with meticulous details on wooden desks that reflect the library's rich history. The room's ambiance is brought to life by the stained glass windows, allowing a soft light to seep in and highlight the texture and content of the books. A vintage globe stands prominently on one of the desks, adding to the academic environment. The medium shot focuses on the teddy bears' upper bodies, skillfully juxtaposing the charm of the teddies with the solemnity of the library setting.

\\

\textbf{Image 1:}\\
\includegraphics[width=0.7\textwidth]{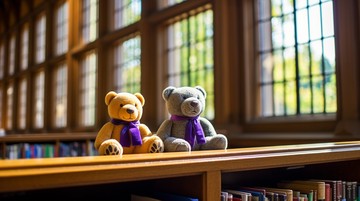}\\
\textbf{Image 2:}\\ 
\includegraphics[width=0.7\textwidth]{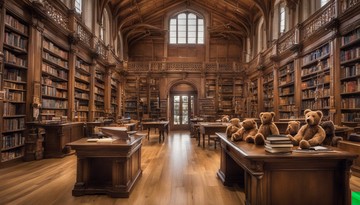}\\

--------------------------------------------------------------------------------------------------------------------

\textbf{GPT-4V:} Image 1 has two teddy bears and shows a closer focus on them with a clear view of the stained glass windows. However, \colorbox{redhl}{Image 2 captures three distinct teddy bears}, neat stacks of books, and the room’s rich ambiance, aligning more accurately with the caption’s details. No vintage globe is visible in either image. | Image 2\\

    \end{tabular}
\end{tcolorbox}
\end{table}

\begin{table}[t]
\caption{Failure case in GPT-4V evaluation. Incorrect rationale is highlight in \colorbox{redhl}{red}. Note that Image 1 is generated by \modelname, and Image 2 is generated by Midjourney.
}
\label{tab:gptv-fail-2}
\centering
\begin{tcolorbox} 
    \centering
    \small
    \begin{tabular}{p{0.98\columnwidth}}
\textbf{Aspect:} Text Rendering

\textbf{Expanded Prompt:} A photo of a white paper on a wooden table detailing sales metrics. The table has columns labeled 'Country', 'Sales Volume', 'Revenue', 'Profit', and 'Profit Margin'. Rows list metrics for USA, China, Australia, India, and South Korea. USA has the highest sales volume at 40,080 and revenue of \$15,971,880. Profit margins range from 17.5\% for India to 19.4\% for Australia. The bottom row aggregates the totals, showing a combined sales volume of 141,282 and total revenue of \$67,754,238. The average profit margin is 18.8\%.

\\

\textbf{Image 1:}\\
\includegraphics[width=0.7\textwidth]{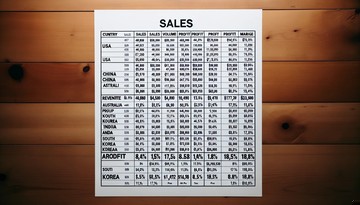}\\
\textbf{Image 2:}\\ 
\includegraphics[width=0.7\textwidth]{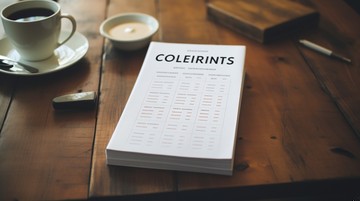}\\

--------------------------------------------------------------------------------------------------------------------

\textbf{GPT-4V:} Image 1's text is legible yet inconsistent with multiple errors and formatting issues making it not harmonious with the overall style or design. \colorbox{redhl}{Image 2, albeit not perfectly matching the caption, offers a cleaner, more structured text rendering,} \colorbox{redhl}{with legible fonts and well-aligned content.} It has an aesthetically pleasing and professional presentation, which is missing in Image 1. | Image 2\\

    \end{tabular}
\end{tcolorbox}
\end{table}

\clearpage
\subsection{Limitations of \modelname}
\label{sec:failure}
\begin{figure*}[h!]
\centering
\vspace{-20mm}
\centerline{\includegraphics[width=1.5\textwidth]{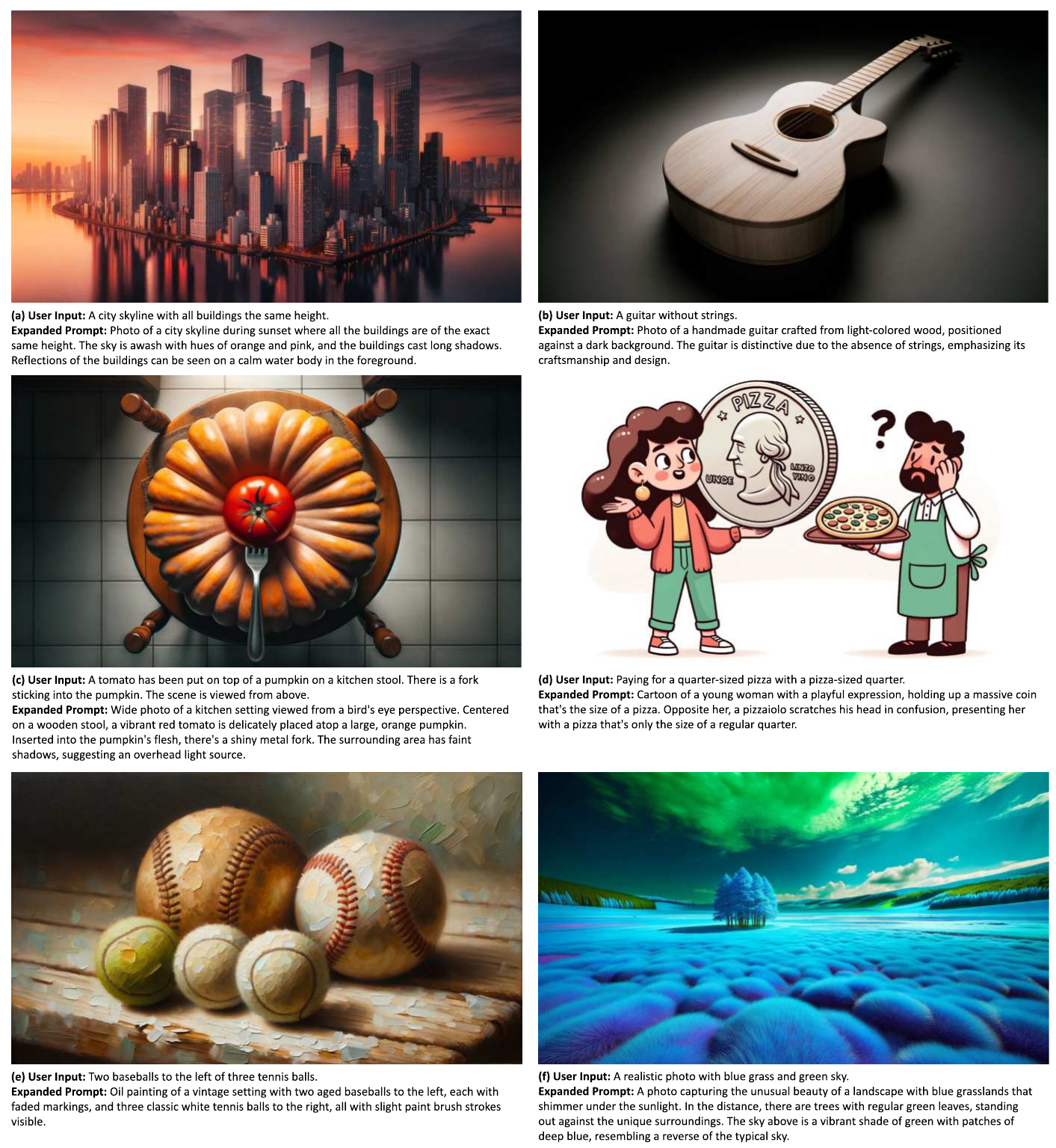}}
\caption[failure cases: uncommon scenes]{Failure cases in generating uncommon scenes. Check Section~\ref{sec:failure} for detailed discussions. 
}
\label{fig:fail_uncommon}
\end{figure*}

We next discuss the representative failure cases and model limitations. First, \modelname may still fail on certain challenging prompts that describe uncommon or complicated scenes. For example, ``all buildings of the same height'' in Figure~\ref{fig:fail_uncommon}(a), ``guitar without string'' in (b), ``fork in the pumpkin'' in (c), ``quarter-sized pizza'' in (d), ``to the left of'' in (e), and the green grass in the left- and right-most part of (f). 

\begin{figure*}[h!]
\centering
\vspace{-20mm}
\centerline{\includegraphics[width=1.5\textwidth]{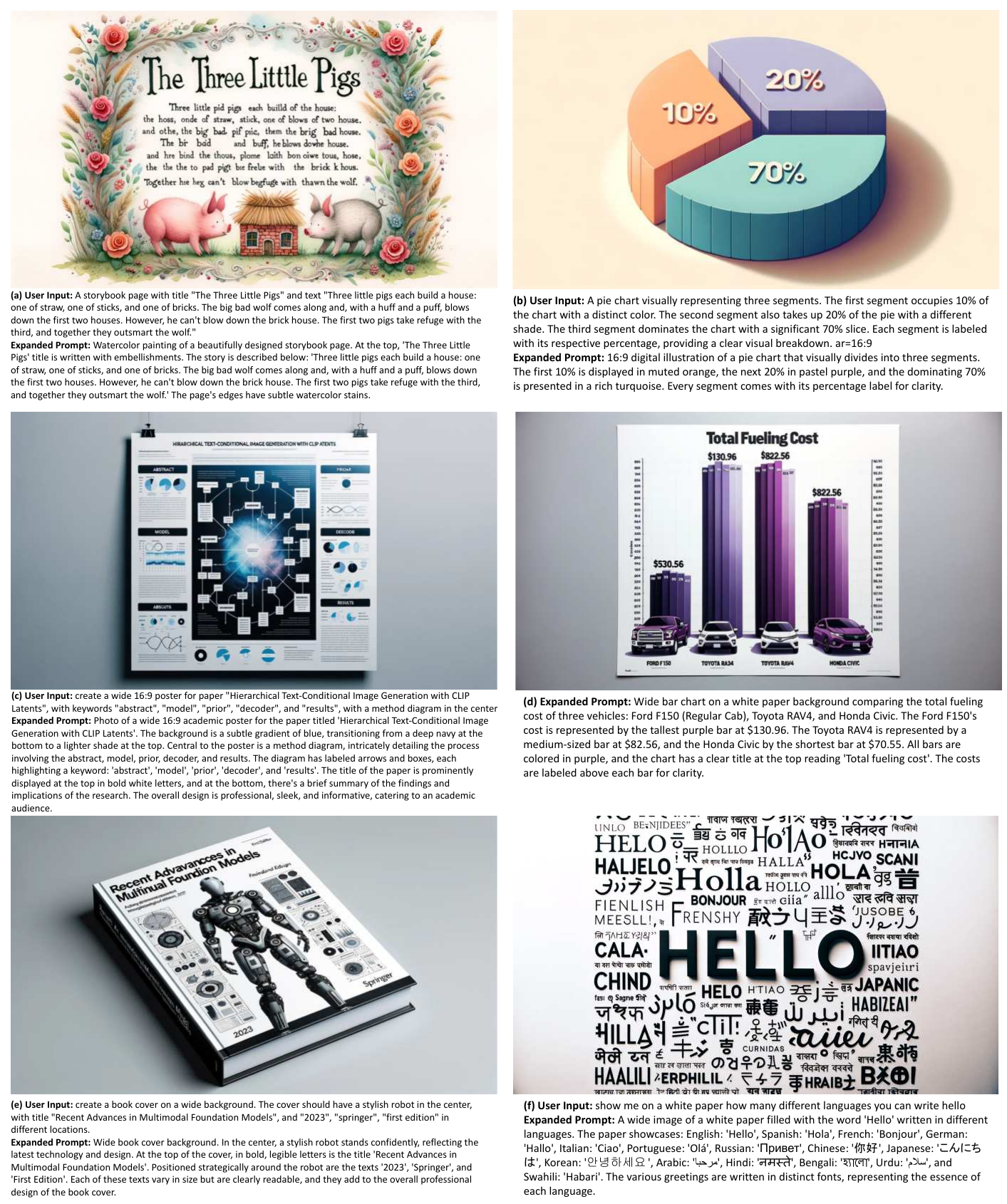}}
\vspace{-2mm}
\caption[failure cases: document design]{Failure cases in generating document designs. Check Section~\ref{sec:failure} for detailed discussions.
}
\label{fig:fail_doc}
\end{figure*}

\modelname has shown an impressive performance in text rendering and layout composition. However, document generation still remains a formidable challenge, hindering the achievement of flawless design outputs. Further enhancing the model's text rendering capabilities would significantly elevate the quality of visual design, as exemplified by the need for precise text generation in storybooks, posters, and book covers shown in Figure~\ref{fig:fail_doc}(a,c,e). In addition to generating accurate Latin characters, there is a need for the model to improve visual and scene text semantic alignments (\eg, the incorrect pie chart portion in (b)), incorporate support for customizable fonts (\eg, for the chart title in (d)), and extend its capabilities to include multiple languages as in (f).

\begin{figure*}[h!]
\centering
\vspace{-20mm}
\centerline{\includegraphics[width=1.5\textwidth]{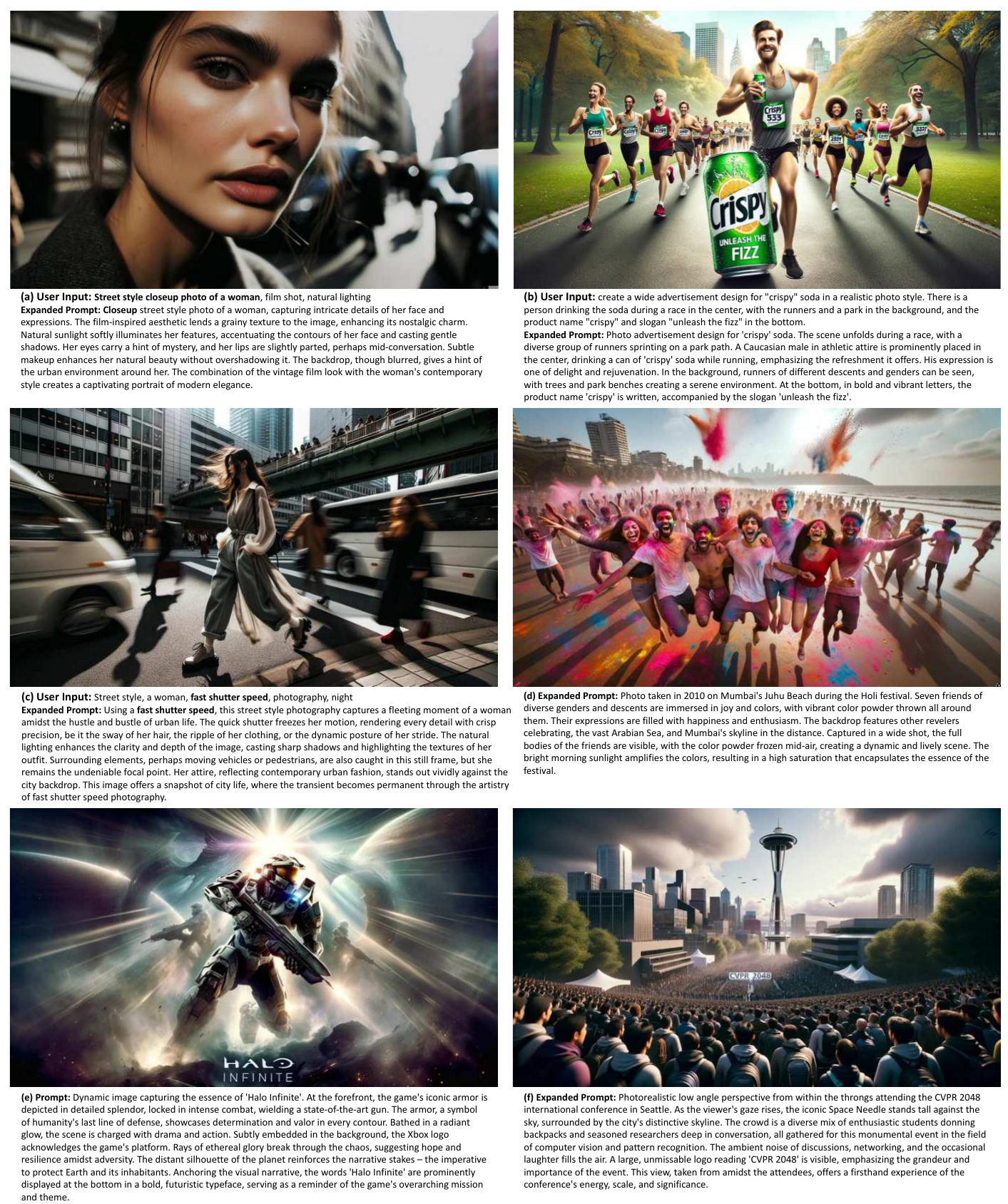}}
\caption[failure cases: image generation]{Other failure cases in image generation. Check Section~\ref{sec:failure} for detailed discussions.
}
\label{fig:failure_image}
\end{figure*}

We observe that the generation artifacts still exist in certain types of generated images. Notably, the skin texture in Figure~\ref{fig:failure_image}(a), and the human faces in the crowded scene (b), appear to be somewhat unnatural. Additionally, the model might also misunderstand certain generation settings, such as the camera setting ``fast shutter speed'' in (c), and the person counts in (d). 

Finally, \modelname currently has limited support for extended image generation functionalities~\cite{zhang2023adding,yang2023idea2img,li2023multimodal}, such as editing uploaded images~\cite{meng2021sdedit,hertz2022prompt,brooks2023instructpix2pix,kawar2023imagic}, concept customization~\cite{ruiz2023dreambooth,kumari2023multi,avrahami2023break,chen2023subject,shi2023instantbooth}, style transfer~\cite{gatys2015neural,Huang_2018_ECCV,NIPS2017_dc6a6489,choi2020stargan}, region control~\cite{yang2022reco,li2023gligen}, spatial condition~\cite{zhang2023adding,avrahami2023spatext}, \etc. Several of these extended functionalities may ease and enhance the visual design process. For example, the incorporation of image condition input could empower designers to refine and build upon existing designs, such as the ``halo armor'' in Figure~\ref{fig:failure_image}(e) or their prior designs, instead of starting from scratch. The region control~\cite{yang2022reco} may allow designers to more precisely place texts and other visual elements.


\clearpage
\section{Conclusions}
We have presented \benchname, a novel image generation benchmark constructed for visual design scenarios. This benchmark is systematically organized with samples categorized according to the design technical capability and application scenarios. We showcase \modelname's strong capability in assisting genuine visual design applications. 
Leveraging the comprehensive design category topology, curated evaluation samples, a visual gallery comprising state-of-the-art T2I models, and the easily replicable GPT-4V-powered evaluator, we aspire for \benchname to establish a solid foundation for design-centric generative models, thereby aiding designers more effectively in real-world tasks.

\section*{Acknowledgment}
We express our gratitude to all contributors from OpenAI for their technical efforts on the \modelname project~\cite{dalle3paper,dalle3system,dalle3blog}. Our sincere appreciation goes to Aditya Ramesh, Li Jing, Tim Brooks, and James Betker at OpenAI, who have provided thoughtful feedback on this work. We are profoundly thankful to Misha Bilenko for his invaluable guidance and support. We also extend heartfelt thanks to our Microsoft colleagues for their insights, with special acknowledgment to Jamie Huynh, Nguyen Bach, Ehsan Azarnasab, Faisal Ahmed, Lin Liang, Chung-Ching Lin, Ce Liu, and Zicheng Liu.

\clearpage
{
\bibliographystyle{plain}
\bibliography{egbib}
}

\appendix
\clearpage
\section{\benchname Gallery: \\ Comparisons among SDXL, Midjourney, Ideogram, Firefly2, and \modelname}
\label{sec:gallery}

\begin{figure}
\centering
\vspace{-20mm}
\begin{adjustbox}{center}
\small
\begin{tabular}{x{0.496\textwidth}@{}x{0.496\textwidth}@{}x{0.496\textwidth}}
SDXL & Midjourney & DALL-E 3 \\
\includegraphics[width=0.486\textwidth,height=0.2777\textwidth]{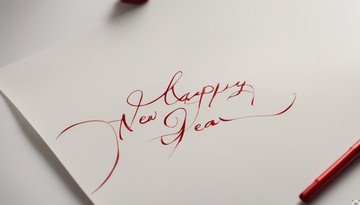} & \includegraphics[width=0.486\textwidth,height=0.2777\textwidth]{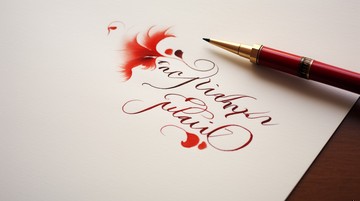} & \includegraphics[width=0.486\textwidth,height=0.2777\textwidth]{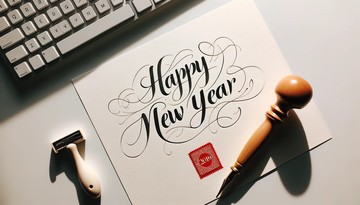} \\
\includegraphics[width=0.486\textwidth,height=0.2777\textwidth]{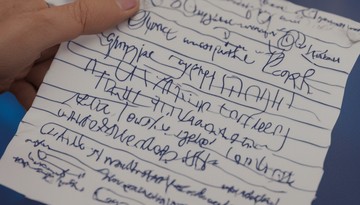} & \includegraphics[width=0.486\textwidth,height=0.2777\textwidth]{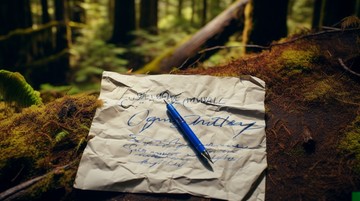} & \includegraphics[width=0.486\textwidth,height=0.2777\textwidth]{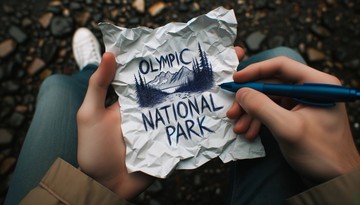} \\
\includegraphics[width=0.486\textwidth,height=0.2777\textwidth]{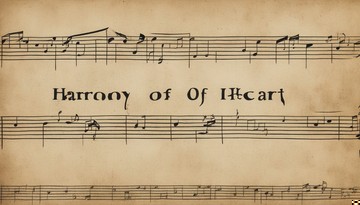} & \includegraphics[width=0.486\textwidth,height=0.2777\textwidth]{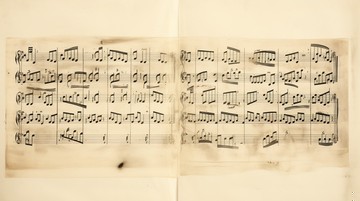} & \includegraphics[width=0.486\textwidth,height=0.2777\textwidth]{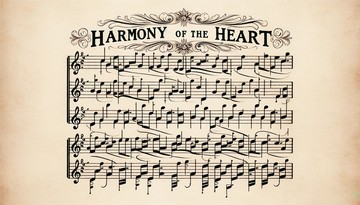} \\
\includegraphics[width=0.486\textwidth,height=0.2777\textwidth]{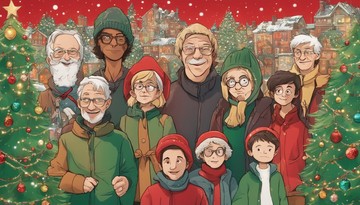} & \includegraphics[width=0.486\textwidth,height=0.2777\textwidth]{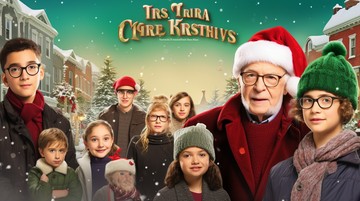} & \includegraphics[width=0.486\textwidth,height=0.2777\textwidth]{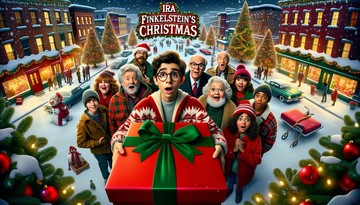} \\
\includegraphics[width=0.486\textwidth,height=0.2777\textwidth]{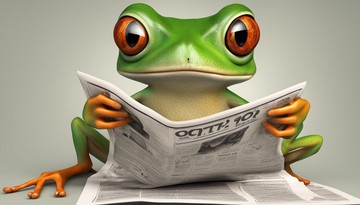} & \includegraphics[width=0.486\textwidth,height=0.2777\textwidth]{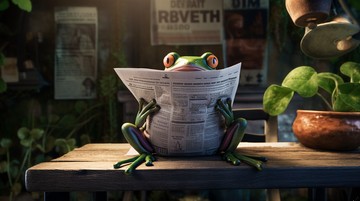} & \includegraphics[width=0.486\textwidth,height=0.2777\textwidth]{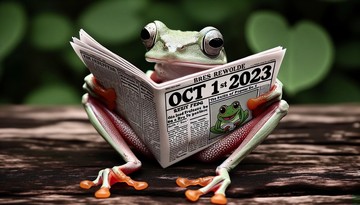} \\
\includegraphics[width=0.486\textwidth,height=0.2777\textwidth]{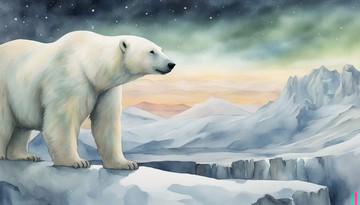} & \includegraphics[width=0.486\textwidth,height=0.2777\textwidth]{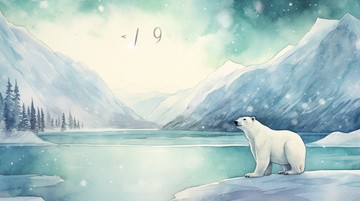} & \includegraphics[width=0.486\textwidth,height=0.2777\textwidth]{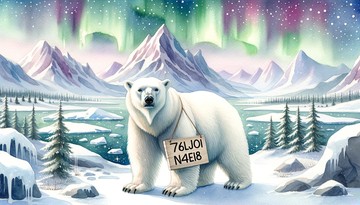}
\end{tabular}
\end{adjustbox}
\caption[text rendering comparisons]{Qualitative comparisons among SDXL, Midjourney v5.2, and \modelname. Check Figure~\ref{fig:text_style}, Figure~\ref{fig:text_lowfreq} for complete prompts.}
\label{fig:compare_text_5_6}
\end{figure}

\begin{figure}
\centering
\vspace{-20mm}
\begin{adjustbox}{center}
\small
\begin{tabular}{x{0.496\textwidth}@{}x{0.496\textwidth}@{}x{0.496\textwidth}}
Ideogram & Firefly 2 & DALL-E 3 \\
\includegraphics[width=0.486\textwidth,height=0.2777\textwidth]{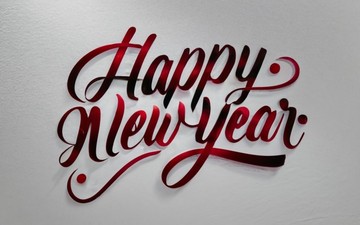} & \includegraphics[width=0.486\textwidth,height=0.2777\textwidth]{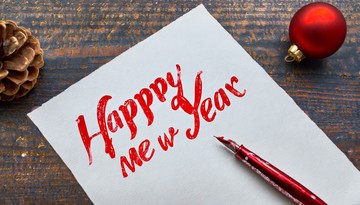} & \includegraphics[width=0.486\textwidth,height=0.2777\textwidth]{raw_images/DE3/text_1_0.jpg} \\
\includegraphics[width=0.486\textwidth,height=0.2777\textwidth]{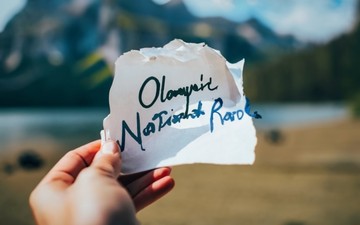} & \includegraphics[width=0.486\textwidth,height=0.2777\textwidth]{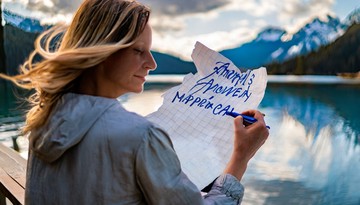} & \includegraphics[width=0.486\textwidth,height=0.2777\textwidth]{raw_images/DE3/text_2_0.jpg} \\
\includegraphics[width=0.486\textwidth,height=0.2777\textwidth]{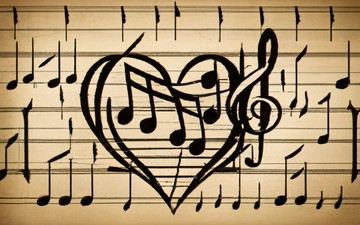} & \includegraphics[width=0.486\textwidth,height=0.2777\textwidth]{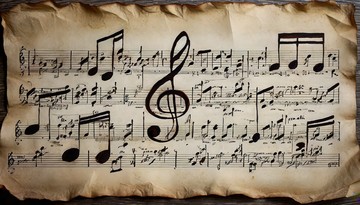} & \includegraphics[width=0.486\textwidth,height=0.2777\textwidth]{raw_images/DE3/text_5_0.jpg} \\
\includegraphics[width=0.486\textwidth,height=0.2777\textwidth]{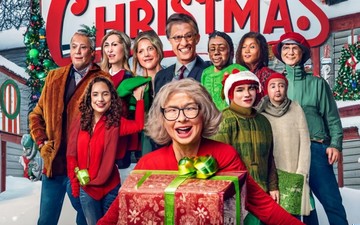} & \includegraphics[width=0.486\textwidth,height=0.2777\textwidth]{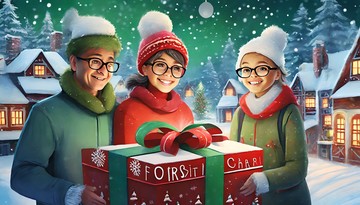} & \includegraphics[width=0.486\textwidth,height=0.2777\textwidth]{raw_images/DE3/text_6_0.jpg} \\
\includegraphics[width=0.486\textwidth,height=0.2777\textwidth]{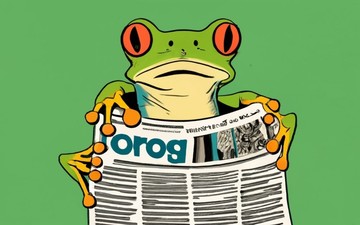} & \includegraphics[width=0.486\textwidth,height=0.2777\textwidth]{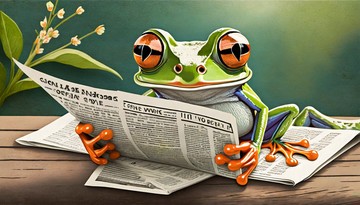} & \includegraphics[width=0.486\textwidth,height=0.2777\textwidth]{raw_images/DE3/text_7_0.jpg} \\
\includegraphics[width=0.486\textwidth,height=0.2777\textwidth]{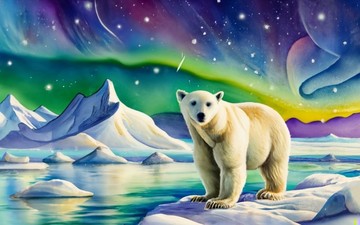} & \includegraphics[width=0.486\textwidth,height=0.2777\textwidth]{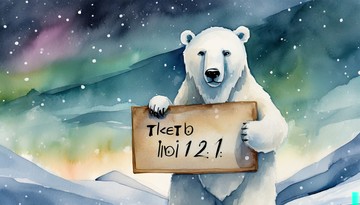} & \includegraphics[width=0.486\textwidth,height=0.2777\textwidth]{raw_images/DE3/text_10_0.jpg}
\end{tabular}
\end{adjustbox}
\caption[text rendering comparisons]{Qualitative comparisons among Ideogram, Firefly 2, and \modelname. Check Figure~\ref{fig:text_style}, Figure~\ref{fig:text_lowfreq} for complete prompts.}
\label{fig:firefly_text_5_6}
\end{figure}

\begin{figure}
\centering
\vspace{-20mm}
\begin{adjustbox}{center}
\small
\begin{tabular}{x{0.496\textwidth}@{}x{0.496\textwidth}@{}x{0.496\textwidth}}
SDXL & Midjourney & DALL-E 3 \\
\includegraphics[width=0.486\textwidth,height=0.2777\textwidth]{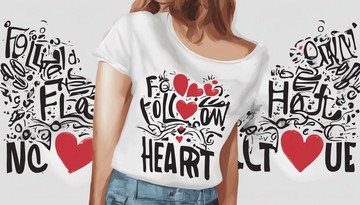} & \includegraphics[width=0.486\textwidth,height=0.2777\textwidth]{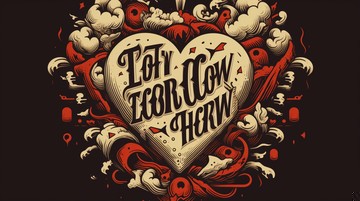} & \includegraphics[width=0.486\textwidth,height=0.2777\textwidth]{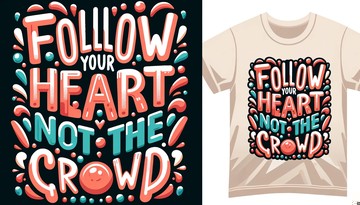} \\
\includegraphics[width=0.486\textwidth,height=0.2777\textwidth]{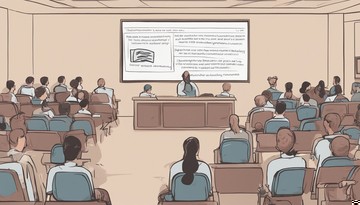} & \includegraphics[width=0.486\textwidth,height=0.2777\textwidth]{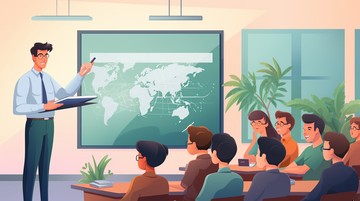} & \includegraphics[width=0.486\textwidth,height=0.2777\textwidth]{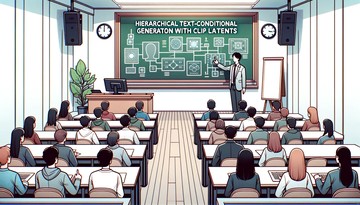} \\
\includegraphics[width=0.486\textwidth,height=0.2777\textwidth]{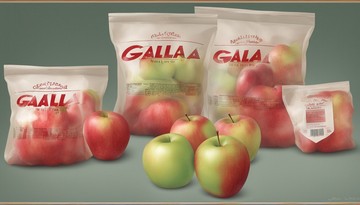} & \includegraphics[width=0.486\textwidth,height=0.2777\textwidth]{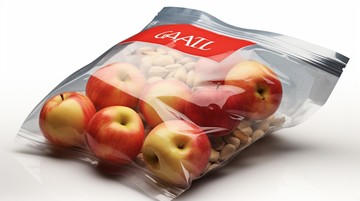} & \includegraphics[width=0.486\textwidth,height=0.2777\textwidth]{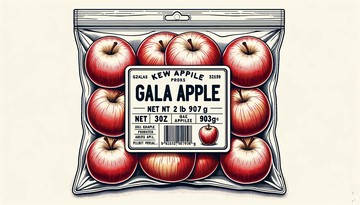} \\
\includegraphics[width=0.486\textwidth,height=0.2777\textwidth]{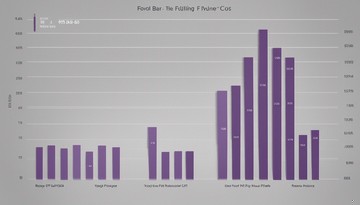} & \includegraphics[width=0.486\textwidth,height=0.2777\textwidth]{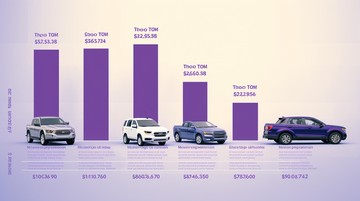} & \includegraphics[width=0.486\textwidth,height=0.2777\textwidth]{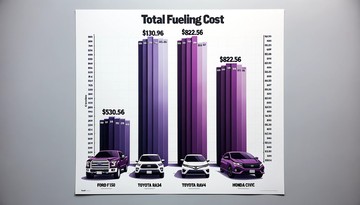} \\
\includegraphics[width=0.486\textwidth,height=0.2777\textwidth]{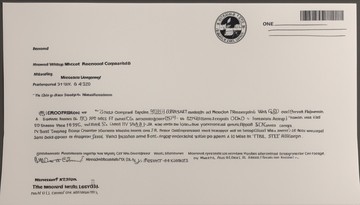} & \includegraphics[width=0.486\textwidth,height=0.2777\textwidth]{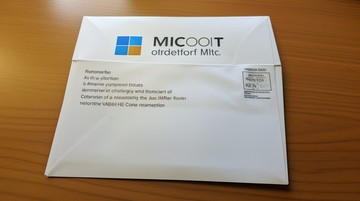} & \includegraphics[width=0.486\textwidth,height=0.2777\textwidth]{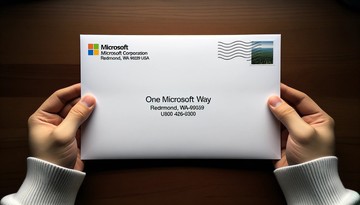} \\
\includegraphics[width=0.486\textwidth,height=0.2777\textwidth]{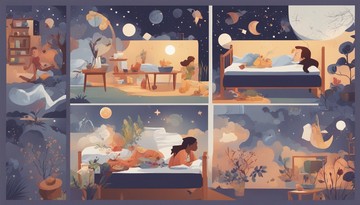} & \includegraphics[width=0.486\textwidth,height=0.2777\textwidth]{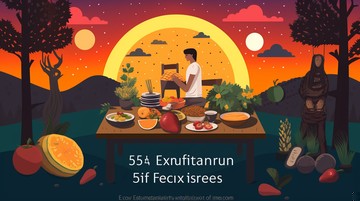} & \includegraphics[width=0.486\textwidth,height=0.2777\textwidth]{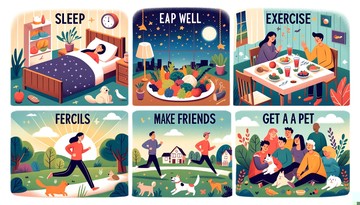}
\end{tabular}
\end{adjustbox}
\caption[layout and document comparisons]{Qualitative comparisons among SDXL, Midjourney v5.2, and \modelname. Check Figure~\ref{fig:text_long}, Figure~\ref{fig:capability_layout_composition_1}, Figure~\ref{fig:motion_graphics} for complete prompts.}
\label{fig:compare_text_7_8_9}
\end{figure}

\begin{figure}
\centering
\vspace{-20mm}
\begin{adjustbox}{center}
\small
\begin{tabular}{x{0.496\textwidth}@{}x{0.496\textwidth}@{}x{0.496\textwidth}}
Ideogram & Firefly 2 & DALL-E 3 \\
\includegraphics[width=0.486\textwidth,height=0.2777\textwidth]{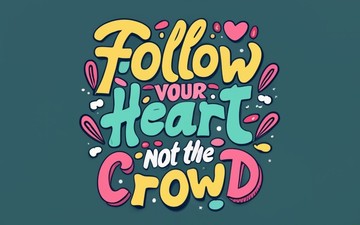} & \includegraphics[width=0.486\textwidth,height=0.2777\textwidth]{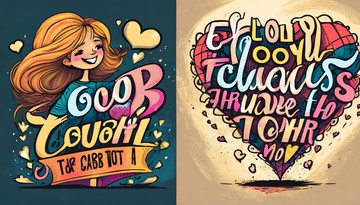} & \includegraphics[width=0.486\textwidth,height=0.2777\textwidth]{raw_images/DE3/text_13_0.jpg} \\
\includegraphics[width=0.486\textwidth,height=0.2777\textwidth]{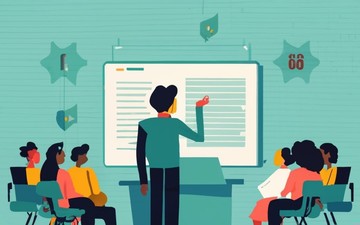} & \includegraphics[width=0.486\textwidth,height=0.2777\textwidth]{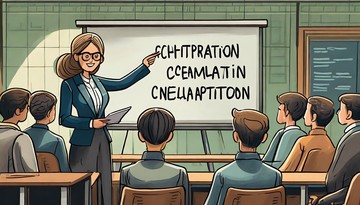} & \includegraphics[width=0.486\textwidth,height=0.2777\textwidth]{raw_images/DE3/text_15_0.jpg} \\
\includegraphics[width=0.486\textwidth,height=0.2777\textwidth]{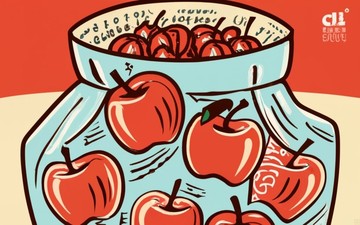} & \includegraphics[width=0.486\textwidth,height=0.2777\textwidth]{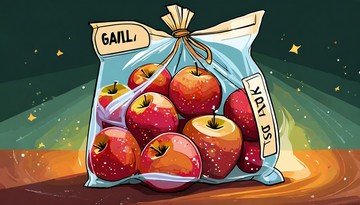} & \includegraphics[width=0.486\textwidth,height=0.2777\textwidth]{raw_images/DE3/text_17_0.jpg} \\
\includegraphics[width=0.486\textwidth,height=0.2777\textwidth]{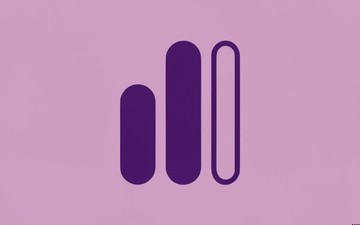} & \includegraphics[width=0.486\textwidth,height=0.2777\textwidth]{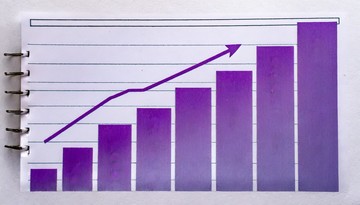} & \includegraphics[width=0.486\textwidth,height=0.2777\textwidth]{raw_images/DE3/text_21_0.jpg} \\
\includegraphics[width=0.486\textwidth,height=0.2777\textwidth]{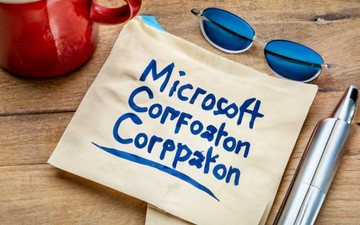} & \includegraphics[width=0.486\textwidth,height=0.2777\textwidth]{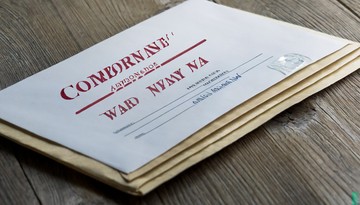} & \includegraphics[width=0.486\textwidth,height=0.2777\textwidth]{raw_images/DE3/text_27_0.jpg} \\
\includegraphics[width=0.486\textwidth,height=0.2777\textwidth]{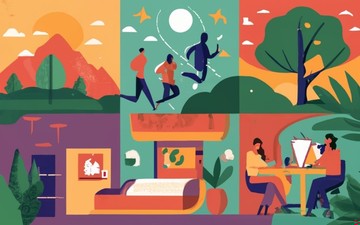} & \includegraphics[width=0.486\textwidth,height=0.2777\textwidth]{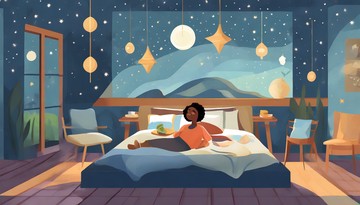} & \includegraphics[width=0.486\textwidth,height=0.2777\textwidth]{raw_images/DE3/motiongraphic_4_0.jpg}
\end{tabular}
\end{adjustbox}
\caption[layout and document comparisons]{Qualitative comparisons among Ideogram, Firefly 2, and \modelname. Check Figure~\ref{fig:text_long}, Figure~\ref{fig:capability_layout_composition_1}, Figure~\ref{fig:motion_graphics} for complete prompts.}
\label{fig:firefly_text_7_8_9}
\end{figure}

\begin{figure}
\centering
\vspace{-20mm}
\begin{adjustbox}{center}
\small
\begin{tabular}{x{0.496\textwidth}@{}x{0.496\textwidth}@{}x{0.496\textwidth}}
SDXL & Midjourney & DALL-E 3
\end{tabular}
\end{adjustbox}
\centerline{\includegraphics[width=1.5\textwidth]{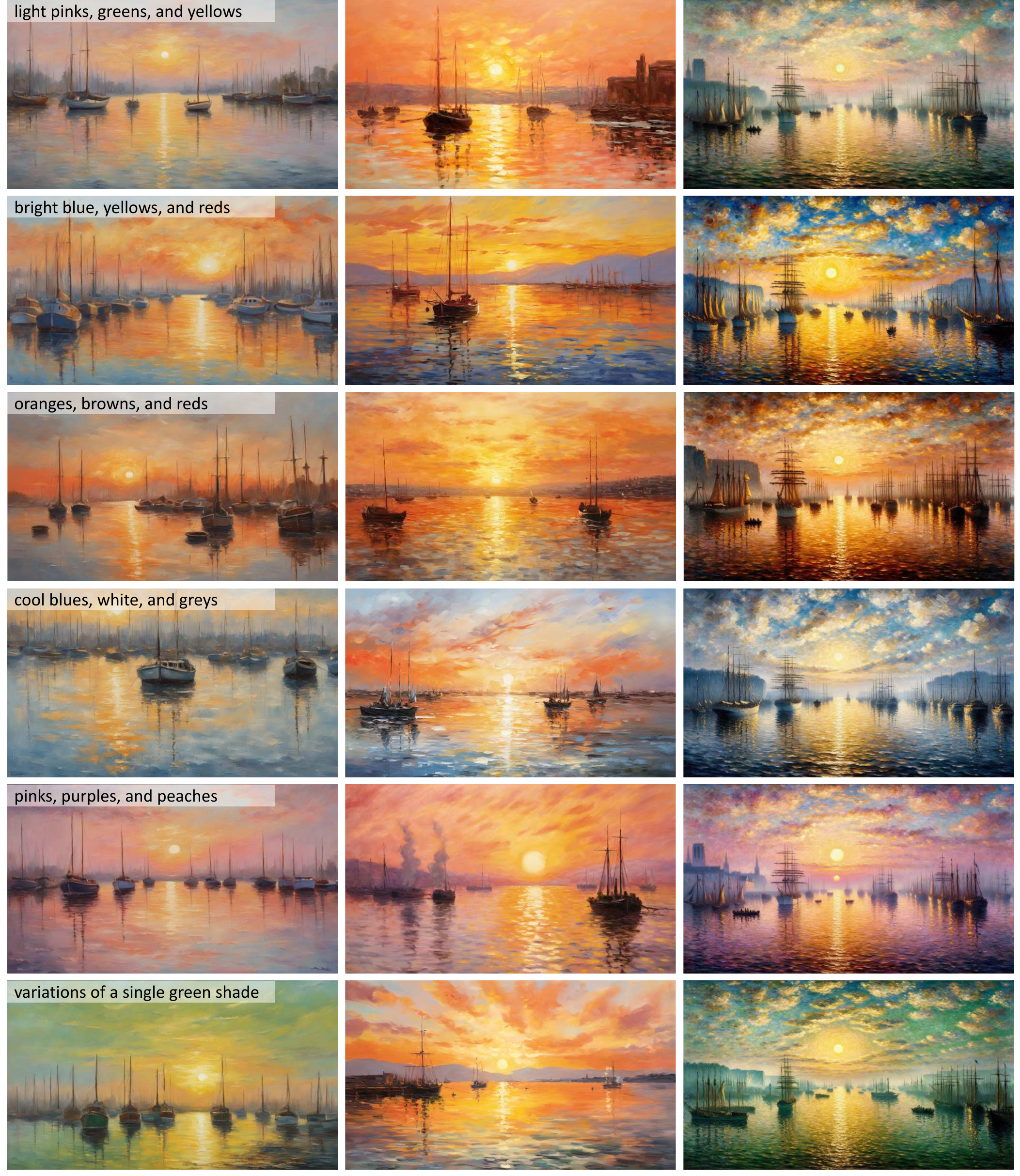}}
\caption[color comparisons]{Qualitative comparisons among SDXL, Midjourney v5.2, and \modelname. Check Figure~\ref{fig:color_variants_2} for complete prompts.}
\label{fig:compare_color_10}
\end{figure}

\begin{figure}
\centering
\vspace{-20mm}
\begin{adjustbox}{center}
\small
\begin{tabular}{x{0.496\textwidth}@{}x{0.496\textwidth}@{}x{0.496\textwidth}}
Ideogram & Firefly 2 & DALL-E 3
\end{tabular}
\end{adjustbox}
\centerline{\includegraphics[width=1.5\textwidth]{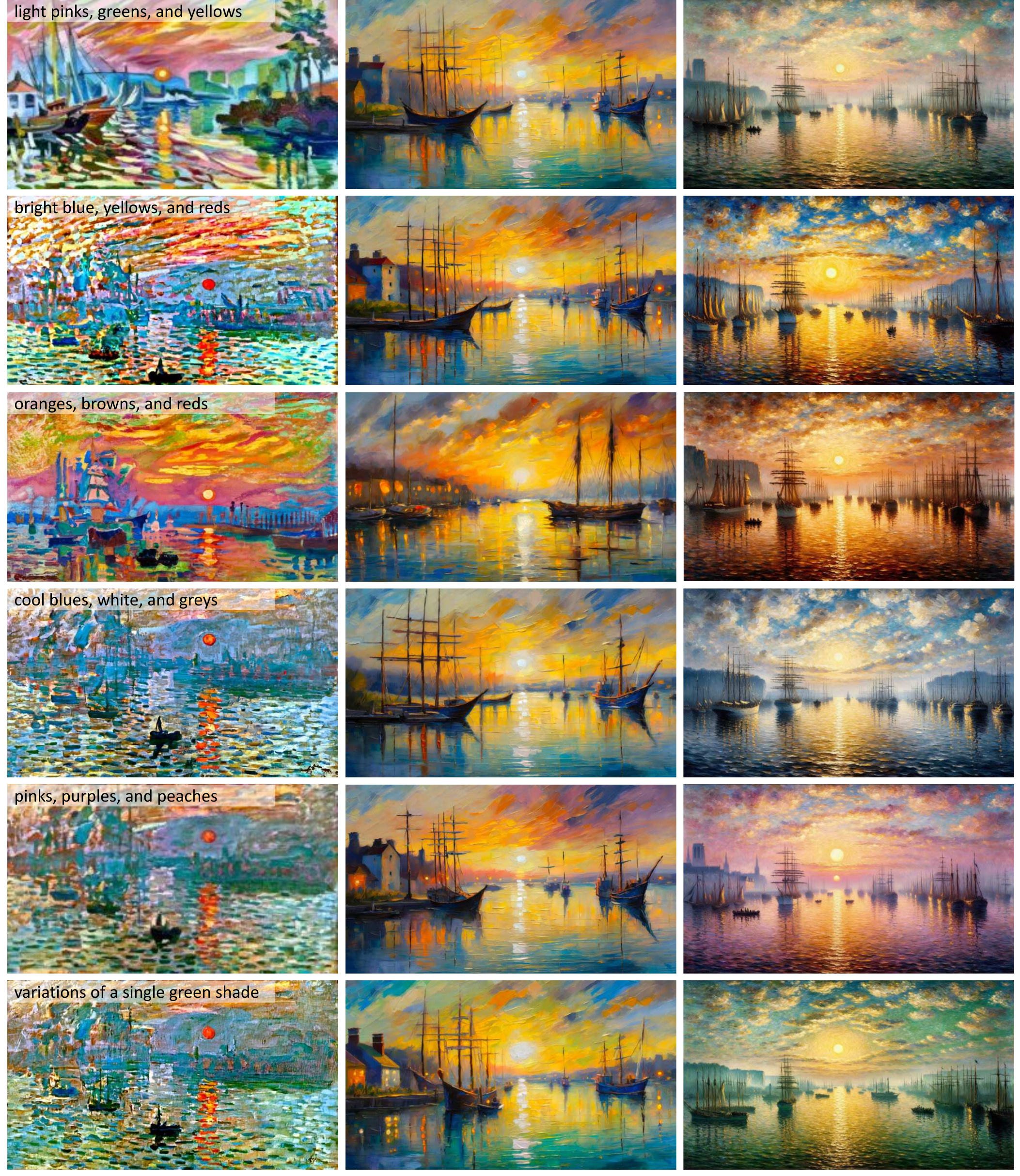}}
\caption[color comparisons]{Qualitative comparisons among Ideogram, Firefly 2, and \modelname. Check Figure~\ref{fig:color_variants_2} for complete prompts.}
\label{fig:firefly_color_10}
\end{figure}

\begin{figure*}[h!]
\centering
\vspace{-20mm}
\centerline{\includegraphics[width=1.5\textwidth]{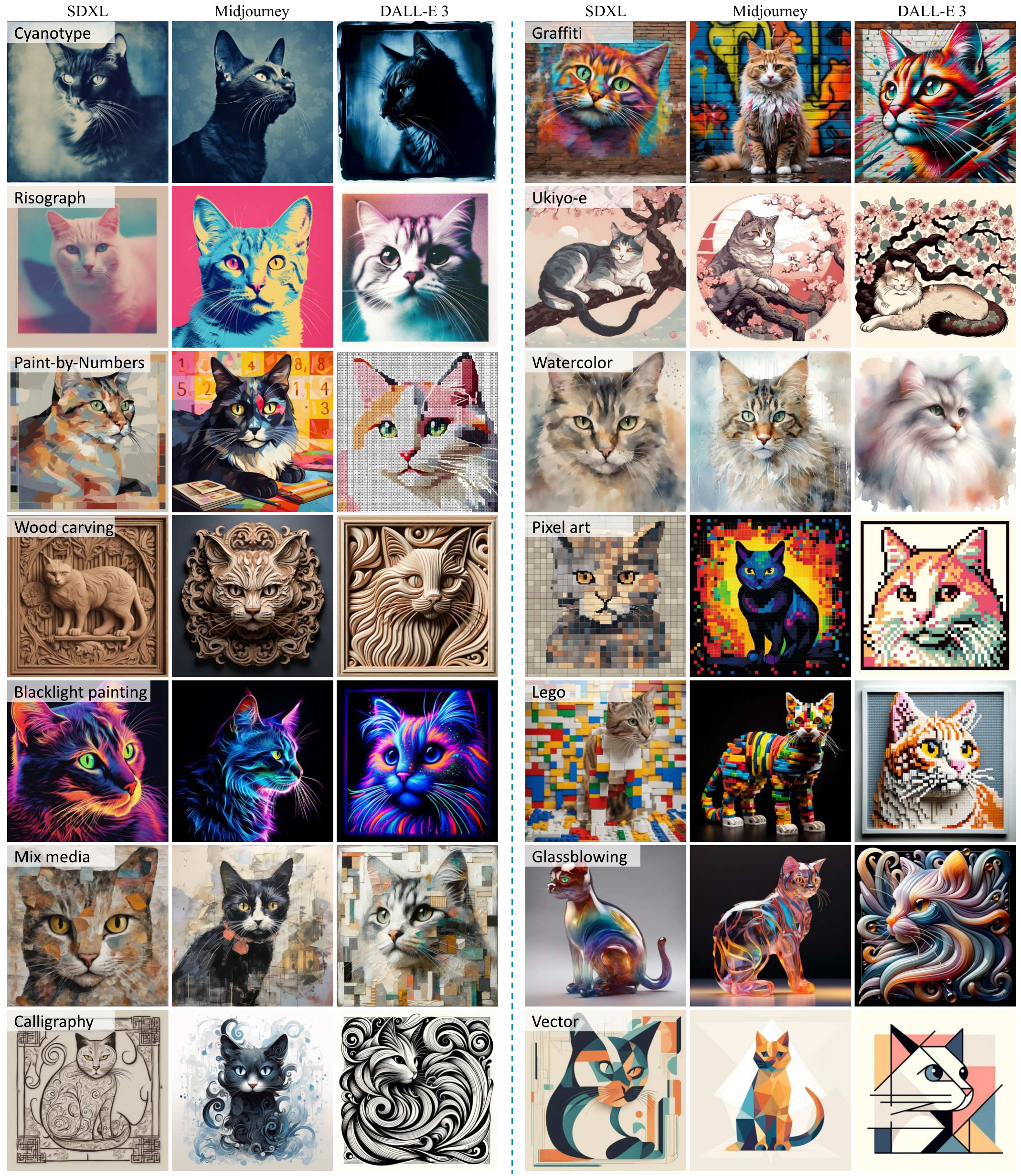}}
\caption[artistic medium comparisons]{Qualitative comparisons among SDXL, Midjourney v5.2, and \modelname. Check Figure~\ref{fig:material_style_design_1}, Figure~\ref{fig:material_style_design_2} for complete prompts.
}
\label{fig:compare_material_13_14}
\end{figure*}

\begin{figure*}[h!]
\centering
\vspace{-20mm}
\centerline{\includegraphics[width=1.5\textwidth]{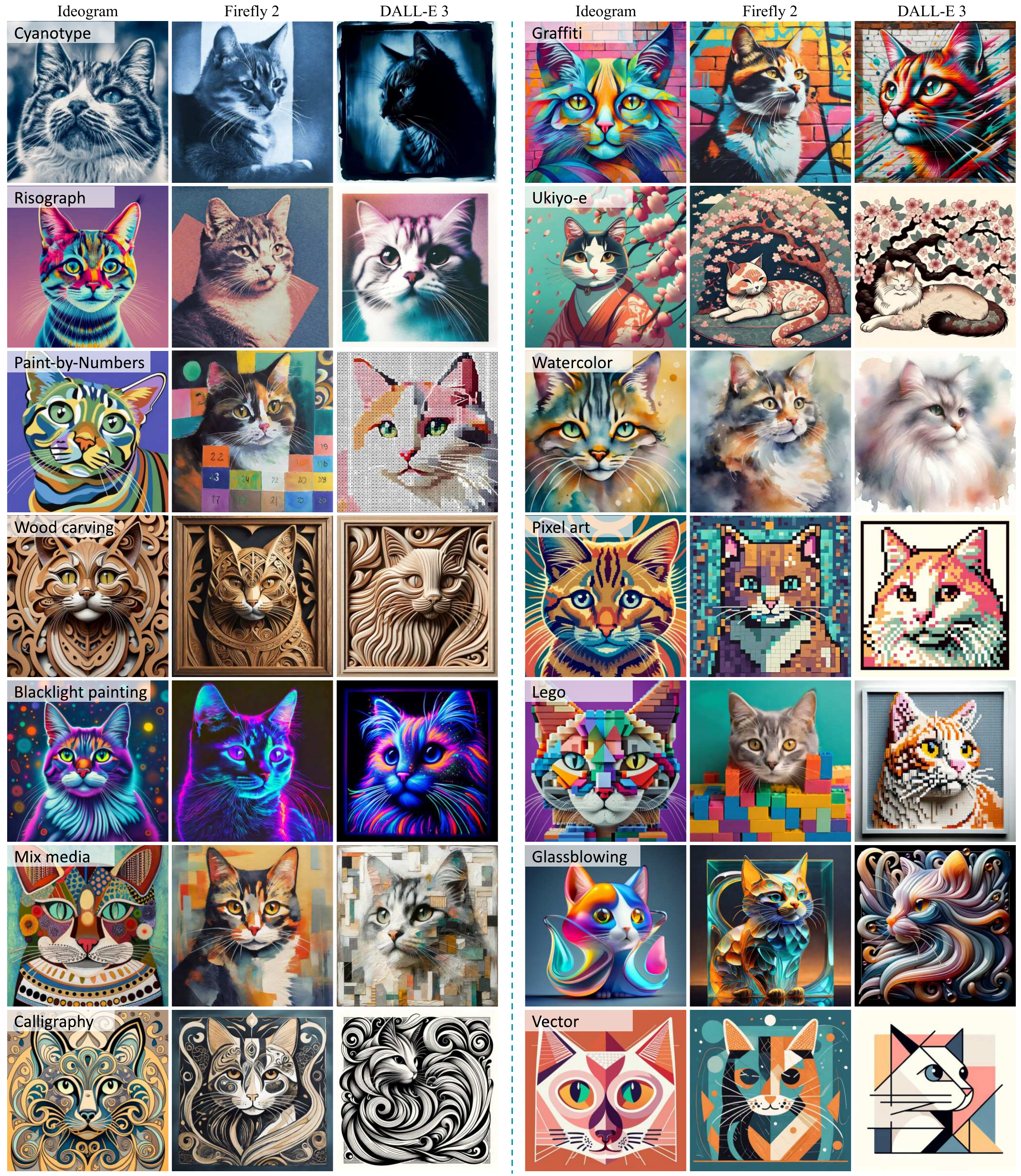}}
\caption[artistic medium comparisons]{Qualitative comparisons among Ideogram, Firefly 2, and \modelname. Check Figure~\ref{fig:material_style_design_1}, Figure~\ref{fig:material_style_design_2} for complete prompts.
}
\label{fig:firefly_material_13_14}
\end{figure*}

\begin{figure*}[h!]
\centering
\vspace{-20mm}
\centerline{\includegraphics[width=1.5\textwidth]{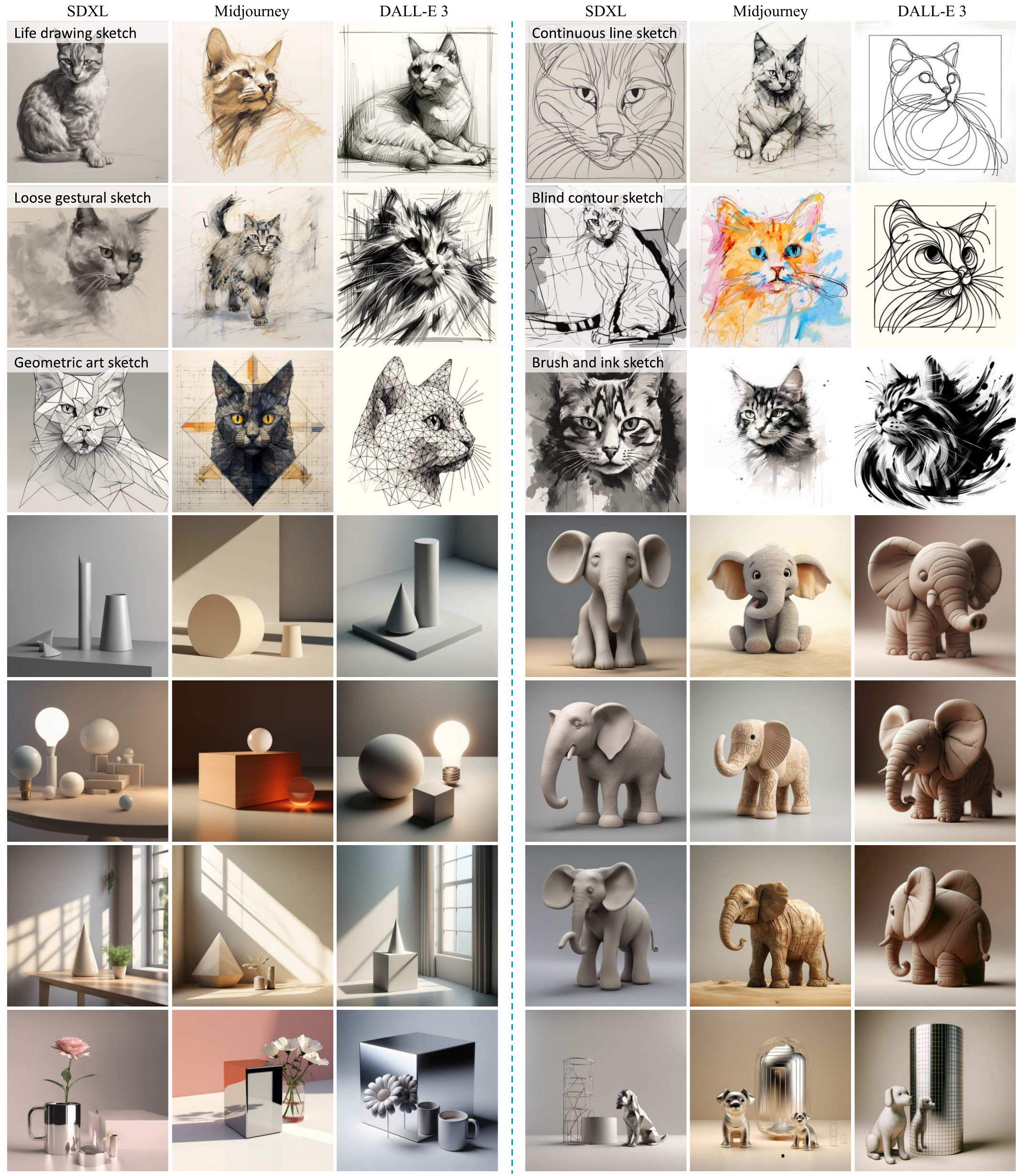}}
\caption[style and 3D comparisons]{Qualitative comparisons among SDXL, Midjourney v5.2, and \modelname. Check Figure~\ref{fig:material_style_design_3}, Figure~\ref{fig:capability_3d} for complete prompts.
}
\label{fig:compare_style3D_12_15}
\end{figure*}

\begin{figure*}[h!]
\centering
\vspace{-20mm}
\centerline{\includegraphics[width=1.5\textwidth]{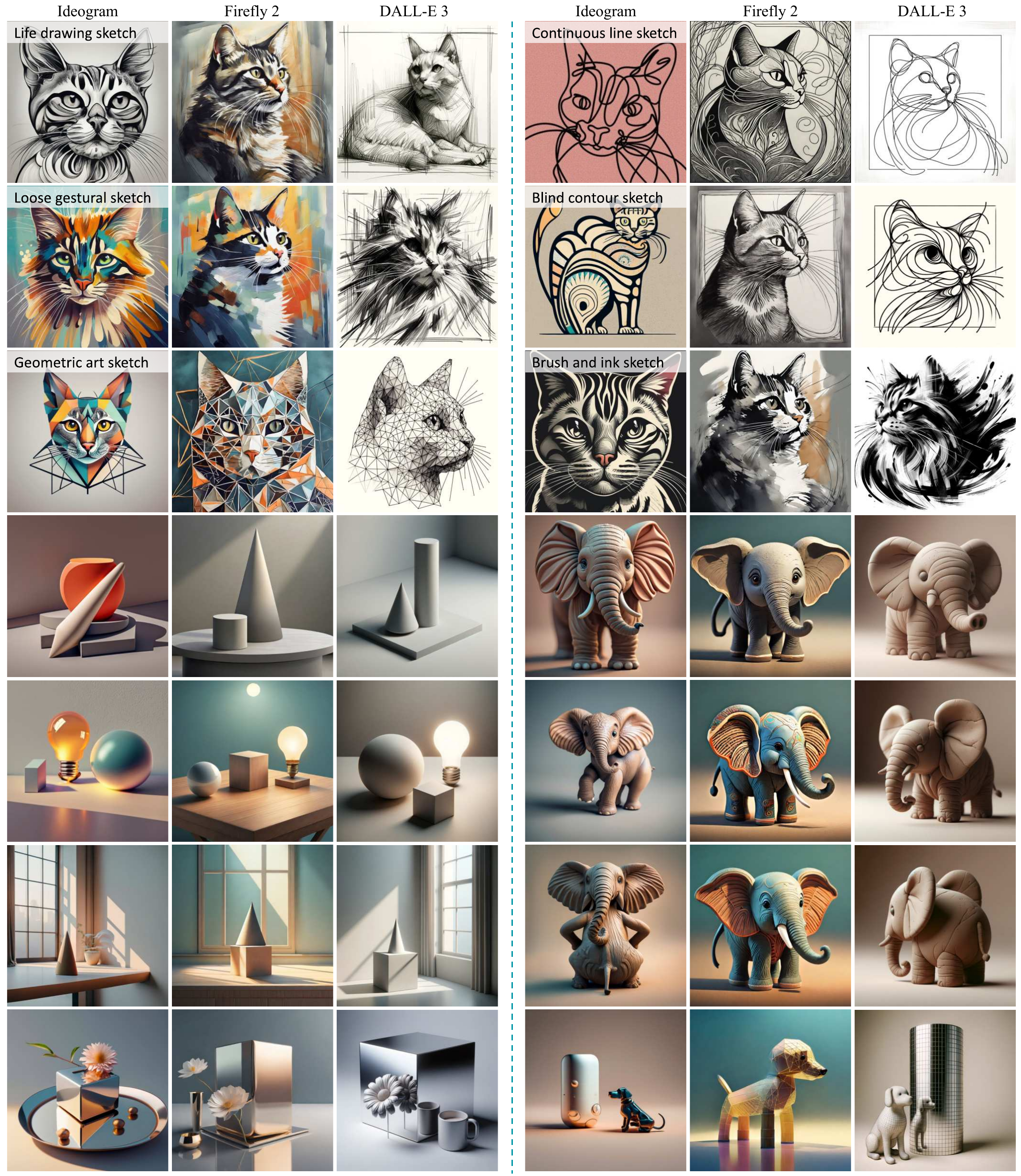}}
\caption[style and 3D comparisons]{Qualitative comparisons among Ideogram, Firefly 2, and \modelname. Check Figure~\ref{fig:material_style_design_3}, Figure~\ref{fig:capability_3d} for complete prompts.
}
\label{fig:firefly_style3D_12_15}
\end{figure*}

\begin{figure}
\centering
\vspace{-20mm}
\begin{adjustbox}{center}
\small
\begin{tabular}{x{0.496\textwidth}@{}x{0.496\textwidth}@{}x{0.496\textwidth}}
SDXL & Midjourney & DALL-E 3
\end{tabular}
\end{adjustbox}
\centerline{\includegraphics[width=1.5\textwidth]{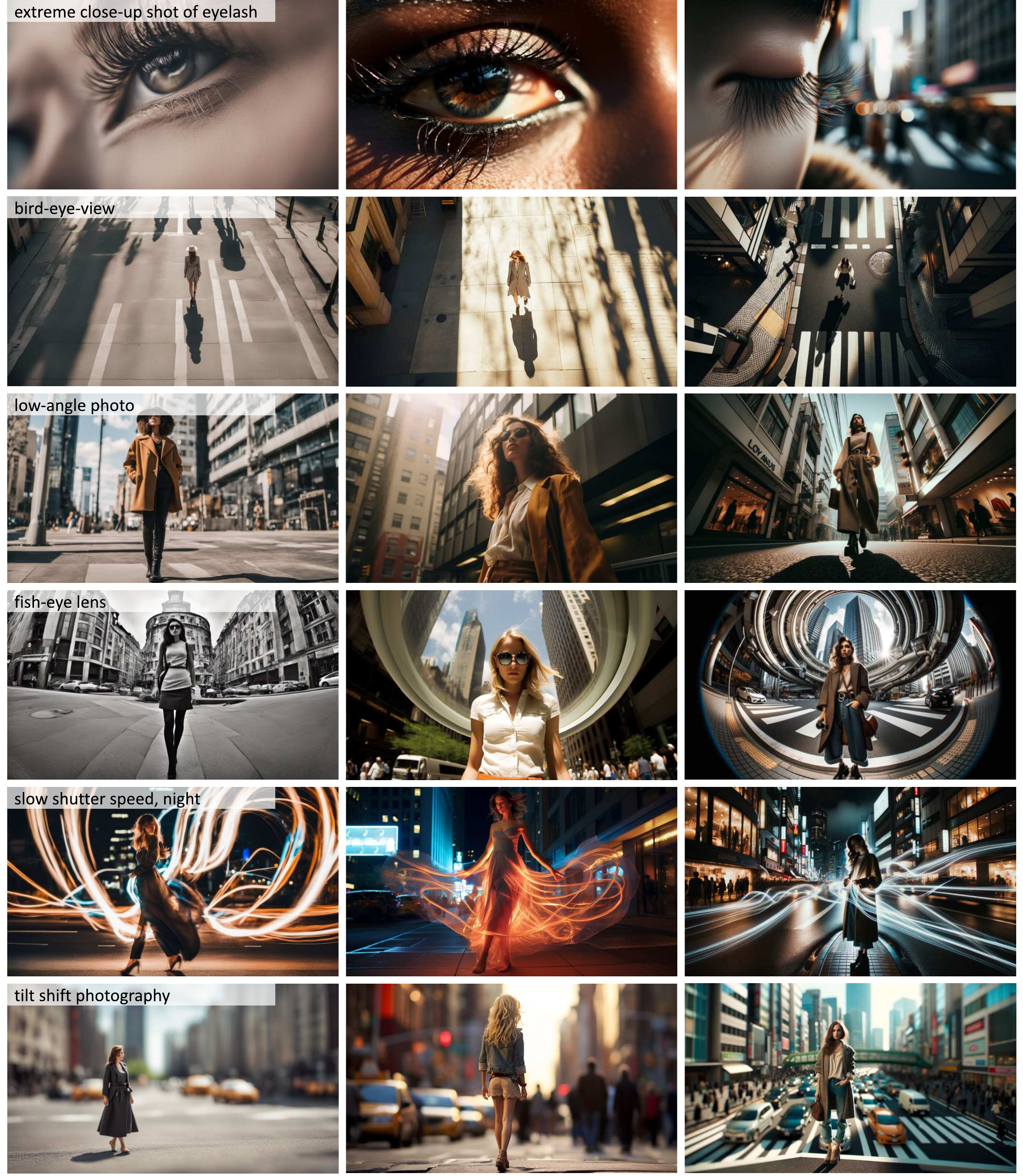}}
\caption[camera settings comparisons]{Qualitative comparisons among SDXL, Midjourney v5.2, and \modelname. Check Figure~\ref{fig:camera_view_1}, Figure~\ref{fig:camera_setting_1} for complete prompts.
}
\label{fig:compare_camera_16_17}
\end{figure}

\begin{figure}
\centering
\vspace{-20mm}
\begin{adjustbox}{center}
\small
\begin{tabular}{x{0.496\textwidth}@{}x{0.496\textwidth}@{}x{0.496\textwidth}}
Ideogram & Firefly 2 & DALL-E 3
\end{tabular}
\end{adjustbox}
\centerline{\includegraphics[width=1.5\textwidth]{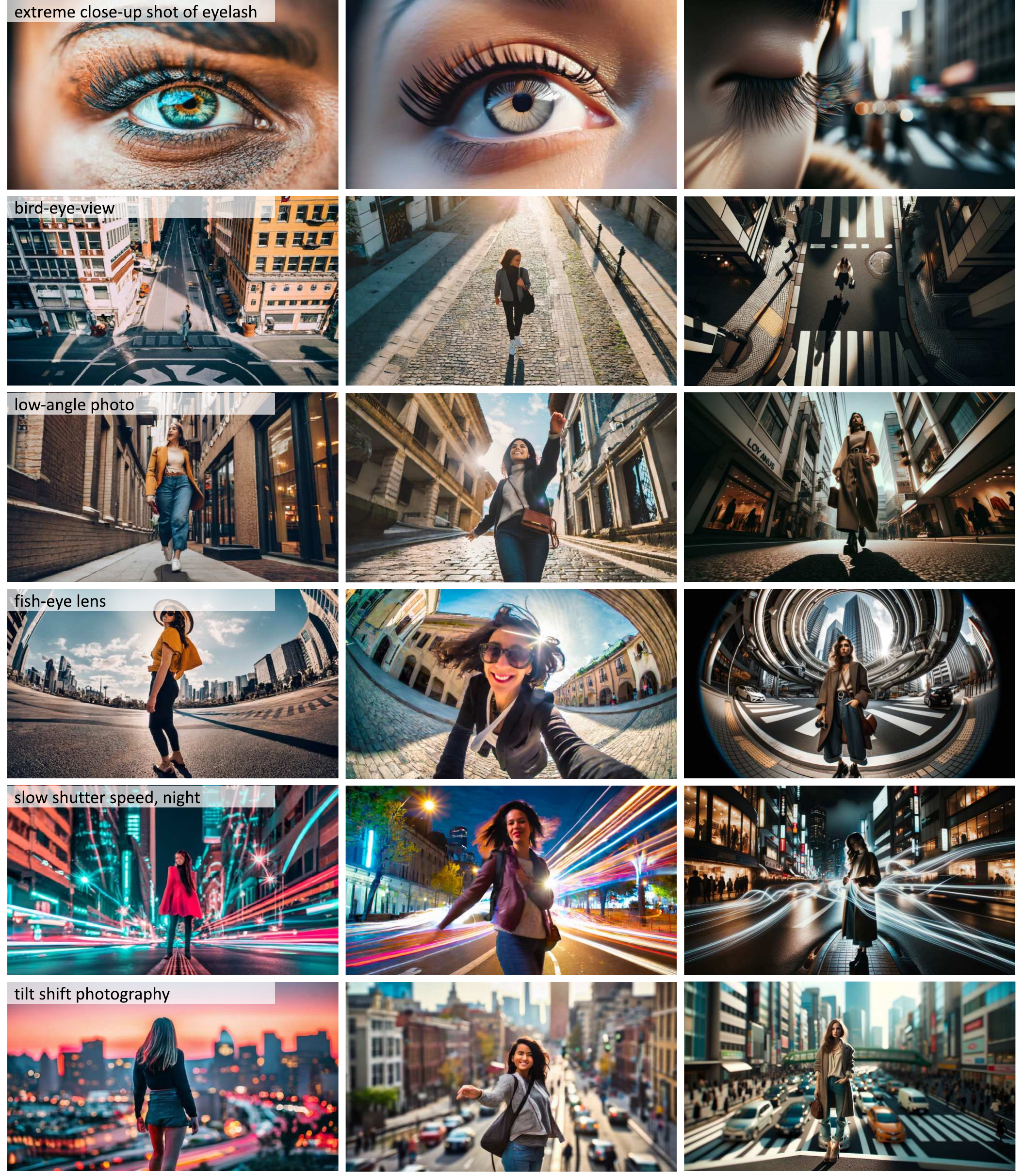}}
\caption[color comparisons]{Qualitative comparisons among Ideogram, Firefly 2, and \modelname. Check Figure~\ref{fig:camera_view_1}, Figure~\ref{fig:camera_setting_1} for complete prompts.
}
\label{fig:firefly_camera_16_17}
\end{figure}

\begin{figure}
\centering
\vspace{-20mm}
\begin{adjustbox}{center}
\small
\begin{tabular}{x{0.496\textwidth}@{}x{0.496\textwidth}@{}x{0.496\textwidth}}
SDXL & Midjourney & DALL-E 3 \\
\includegraphics[width=0.486\textwidth,height=0.2777\textwidth]{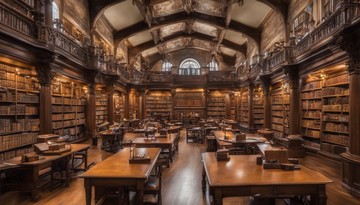} & \includegraphics[width=0.486\textwidth,height=0.2777\textwidth]{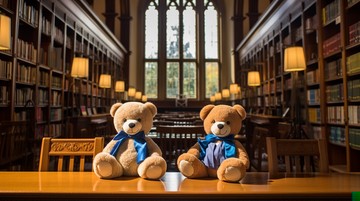} & \includegraphics[width=0.486\textwidth,height=0.2777\textwidth]{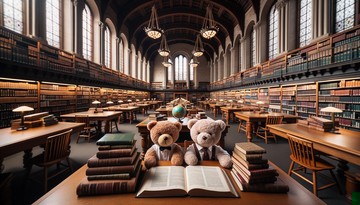} \\
\includegraphics[width=0.486\textwidth,height=0.2777\textwidth]{raw_images/SDXL/crowd_1_0.jpg} & \includegraphics[width=0.486\textwidth,height=0.2777\textwidth]{raw_images/MJ/crowd_1_0.jpg} & \includegraphics[width=0.486\textwidth,height=0.2777\textwidth]{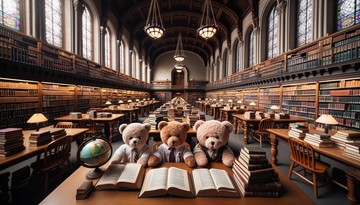} \\
\includegraphics[width=0.486\textwidth,height=0.2777\textwidth]{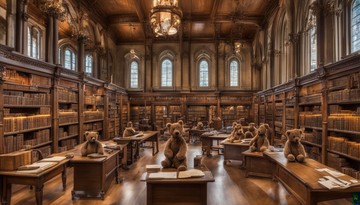} & \includegraphics[width=0.486\textwidth,height=0.2777\textwidth]{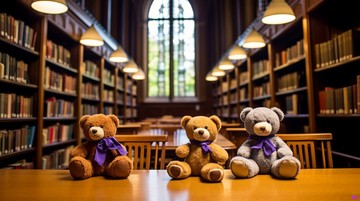} & \includegraphics[width=0.486\textwidth,height=0.2777\textwidth]{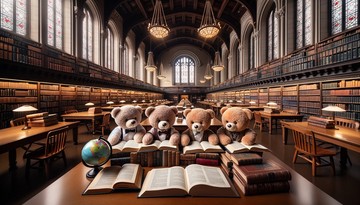} \\
\includegraphics[width=0.486\textwidth,height=0.2777\textwidth]{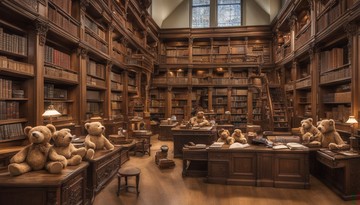} & \includegraphics[width=0.486\textwidth,height=0.2777\textwidth]{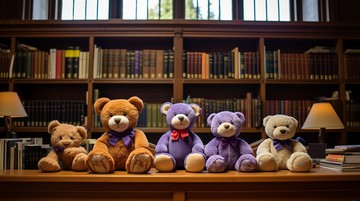} & \includegraphics[width=0.486\textwidth,height=0.2777\textwidth]{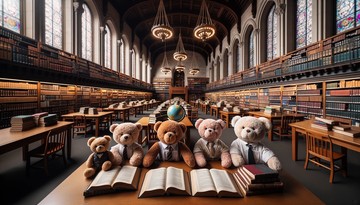} \\
\includegraphics[width=0.486\textwidth,height=0.2777\textwidth]{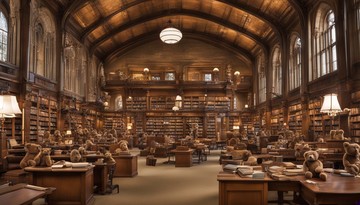} & \includegraphics[width=0.486\textwidth,height=0.2777\textwidth]{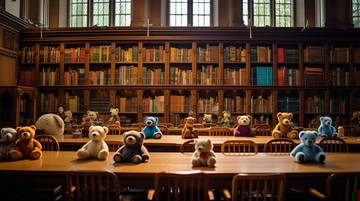} & \includegraphics[width=0.486\textwidth,height=0.2777\textwidth]{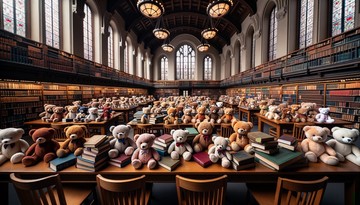} \\
\includegraphics[width=0.486\textwidth,height=0.2777\textwidth]{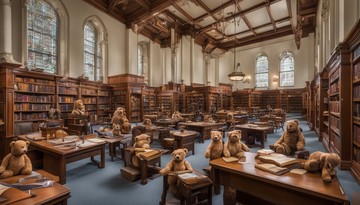} & \includegraphics[width=0.486\textwidth,height=0.2777\textwidth]{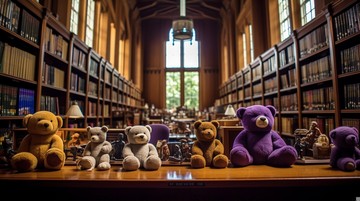} & \includegraphics[width=0.486\textwidth,height=0.2777\textwidth]{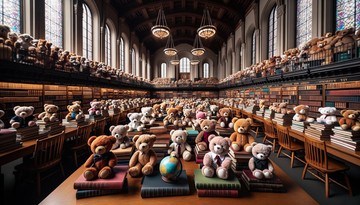}
\end{tabular}
\end{adjustbox}
\caption[crowded scene comparisons]{Qualitative comparisons among SDXL, Midjourney v5.2, and \modelname. Check Figure~\ref{fig:crowd1}, Figure~\ref{fig:crowd2} for complete prompts.}
\label{fig:compare_crowd_18_19}
\end{figure}

\begin{figure}
\centering
\vspace{-20mm}
\begin{adjustbox}{center}
\small
\begin{tabular}{x{0.496\textwidth}@{}x{0.496\textwidth}@{}x{0.496\textwidth}}
Ideogram & Firefly 2 & DALL-E 3 \\
\includegraphics[width=0.486\textwidth,height=0.2777\textwidth]{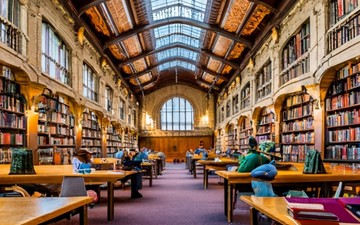} & \includegraphics[width=0.486\textwidth,height=0.2777\textwidth]{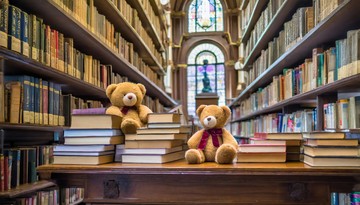} & \includegraphics[width=0.486\textwidth,height=0.2777\textwidth]{raw_images/DE3/crowd_0_0.jpg} \\
\includegraphics[width=0.486\textwidth,height=0.2777\textwidth]{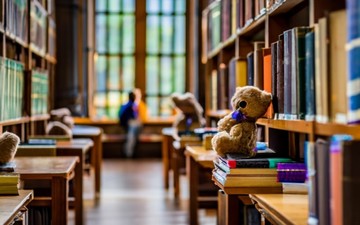} & \includegraphics[width=0.486\textwidth,height=0.2777\textwidth]{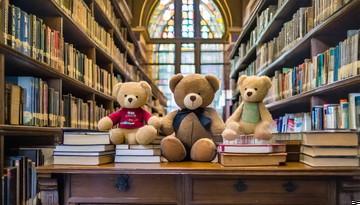} & \includegraphics[width=0.486\textwidth,height=0.2777\textwidth]{raw_images/DE3/crowd_1_0.jpg} \\
\includegraphics[width=0.486\textwidth,height=0.2777\textwidth]{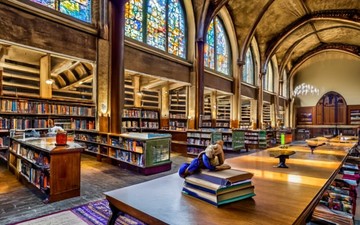} & \includegraphics[width=0.486\textwidth,height=0.2777\textwidth]{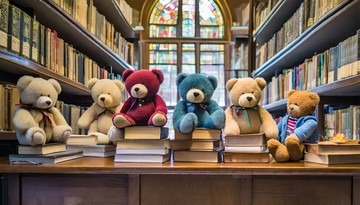} & \includegraphics[width=0.486\textwidth,height=0.2777\textwidth]{raw_images/DE3/crowd_2_0.jpg} \\
\includegraphics[width=0.486\textwidth,height=0.2777\textwidth]{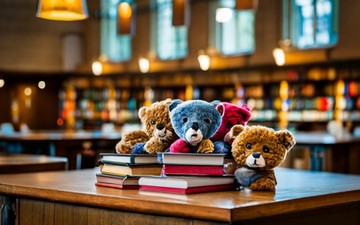} & \includegraphics[width=0.486\textwidth,height=0.2777\textwidth]{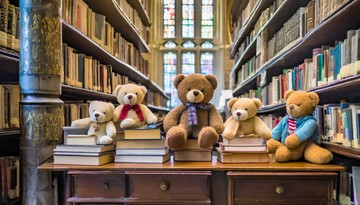} & \includegraphics[width=0.486\textwidth,height=0.2777\textwidth]{raw_images/DE3/crowd_3_0.jpg} \\
\includegraphics[width=0.486\textwidth,height=0.2777\textwidth]{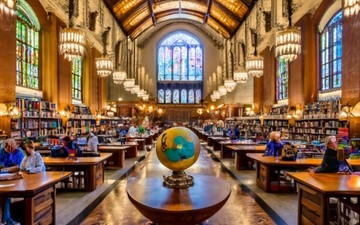} & \includegraphics[width=0.486\textwidth,height=0.2777\textwidth]{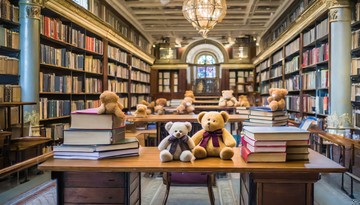} & \includegraphics[width=0.486\textwidth,height=0.2777\textwidth]{raw_images/DE3/crowd_4_0.jpg} \\
\includegraphics[width=0.486\textwidth,height=0.2777\textwidth]{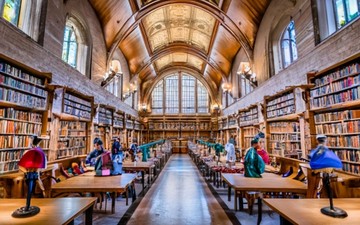} & \includegraphics[width=0.486\textwidth,height=0.2777\textwidth]{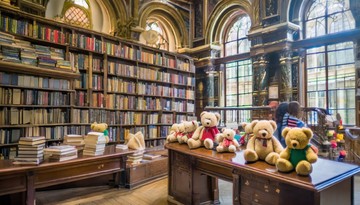} & \includegraphics[width=0.486\textwidth,height=0.2777\textwidth]{raw_images/DE3/crowd_5_0.jpg}
\end{tabular}
\end{adjustbox}
\caption[crowded scene comparisons]{Qualitative comparisons among Ideogram, Firefly 2, and \modelname. Check Figure~\ref{fig:crowd1}, Figure~\ref{fig:crowd2} for complete prompts.}
\label{fig:firefly_crowd_18_19}
\end{figure}

\begin{figure}
\centering
\vspace{-20mm}
\begin{adjustbox}{center}
\small
\begin{tabular}{x{0.496\textwidth}@{}x{0.496\textwidth}@{}x{0.496\textwidth}}
SDXL & Midjourney & DALL-E 3 \\
\includegraphics[width=0.486\textwidth,height=0.2777\textwidth]{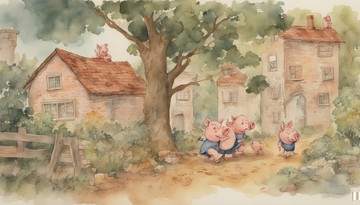} & \includegraphics[width=0.486\textwidth,height=0.2777\textwidth]{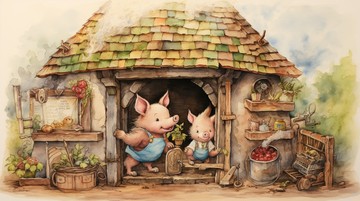} & \includegraphics[width=0.486\textwidth,height=0.2777\textwidth]{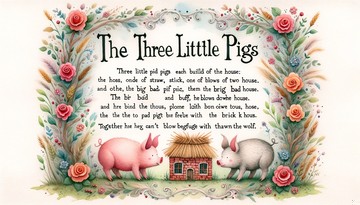} \\
\includegraphics[width=0.486\textwidth,height=0.2777\textwidth]{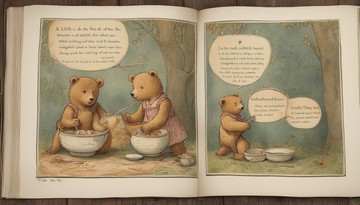} & \includegraphics[width=0.486\textwidth,height=0.2777\textwidth]{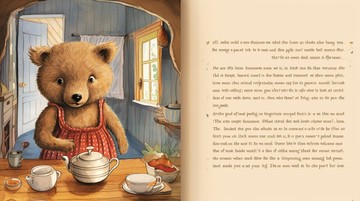} & \includegraphics[width=0.486\textwidth,height=0.2777\textwidth]{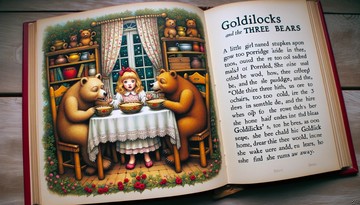} \\
\includegraphics[width=0.486\textwidth,height=0.2777\textwidth]{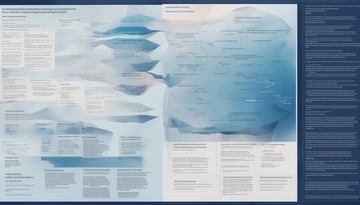} & \includegraphics[width=0.486\textwidth,height=0.2777\textwidth]{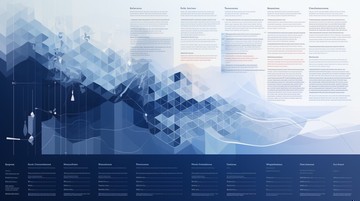} & \includegraphics[width=0.486\textwidth,height=0.2777\textwidth]{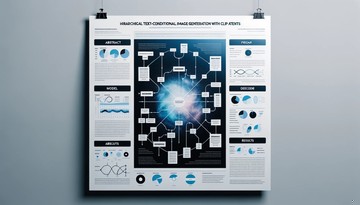} \\
\includegraphics[width=0.486\textwidth,height=0.2777\textwidth]{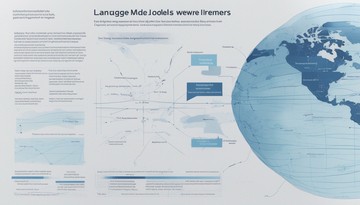} & \includegraphics[width=0.486\textwidth,height=0.2777\textwidth]{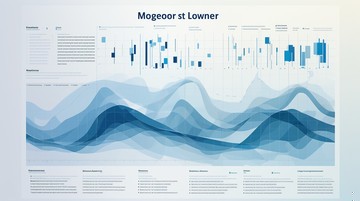} & \includegraphics[width=0.486\textwidth,height=0.2777\textwidth]{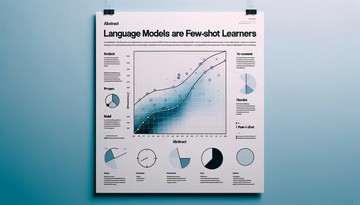} \\
\includegraphics[width=0.486\textwidth,height=0.2777\textwidth]{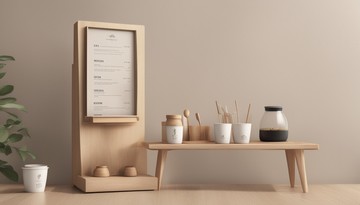} & \includegraphics[width=0.486\textwidth,height=0.2777\textwidth]{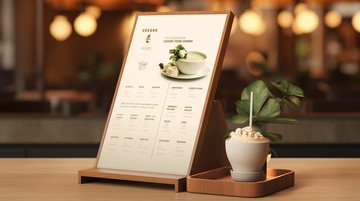} & \includegraphics[width=0.486\textwidth,height=0.2777\textwidth]{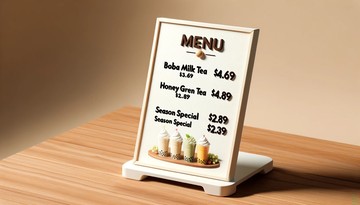} \\
\includegraphics[width=0.486\textwidth,height=0.2777\textwidth]{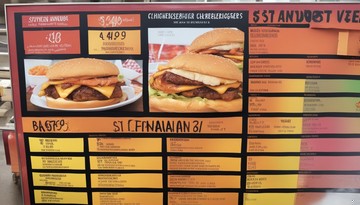} & \includegraphics[width=0.486\textwidth,height=0.2777\textwidth]{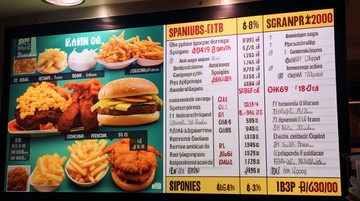} & \includegraphics[width=0.486\textwidth,height=0.2777\textwidth]{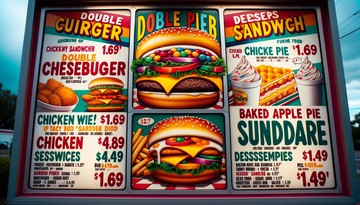}
\end{tabular}
\end{adjustbox}
\caption[storybooks, academic posters, and menus comparisons]{Qualitative comparisons among SDXL, Midjourney v5.2, and \modelname. Check Figure~\ref{fig:text_storybook} for complete prompts.
}
\label{fig:compare_storybook_20}
\end{figure}

\begin{figure}
\centering
\vspace{-20mm}
\begin{adjustbox}{center}
\small
\begin{tabular}{x{0.496\textwidth}@{}x{0.496\textwidth}@{}x{0.496\textwidth}}
Ideogram & Firefly 2 & DALL-E 3 \\
\includegraphics[width=0.486\textwidth,height=0.2777\textwidth]{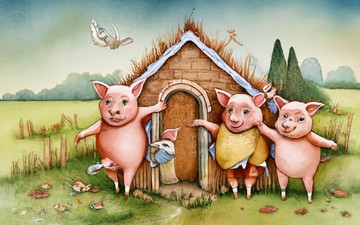} & \includegraphics[width=0.486\textwidth,height=0.2777\textwidth]{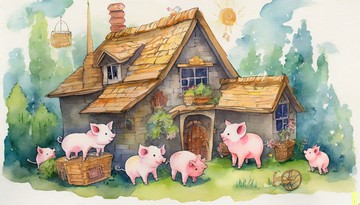} & \includegraphics[width=0.486\textwidth,height=0.2777\textwidth]{raw_images/DE3/storybook_0_0.jpg} \\
\includegraphics[width=0.486\textwidth,height=0.2777\textwidth]{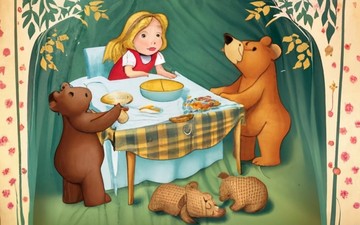} & \includegraphics[width=0.486\textwidth,height=0.2777\textwidth]{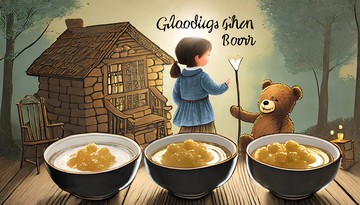} & \includegraphics[width=0.486\textwidth,height=0.2777\textwidth]{raw_images/DE3/storybook_1_0.jpg} \\
\includegraphics[width=0.486\textwidth,height=0.2777\textwidth]{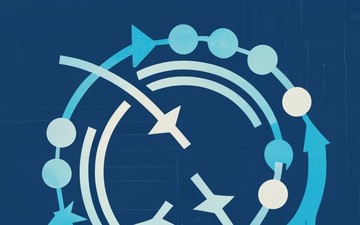} & \includegraphics[width=0.486\textwidth,height=0.2777\textwidth]{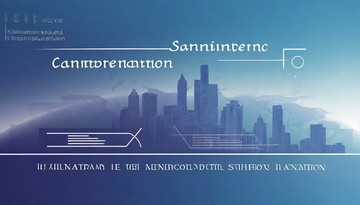} & \includegraphics[width=0.486\textwidth,height=0.2777\textwidth]{raw_images/DE3/storybook_2_0.jpg} \\
\includegraphics[width=0.486\textwidth,height=0.2777\textwidth]{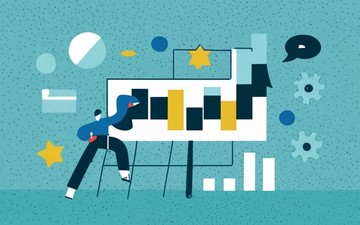} & \includegraphics[width=0.486\textwidth,height=0.2777\textwidth]{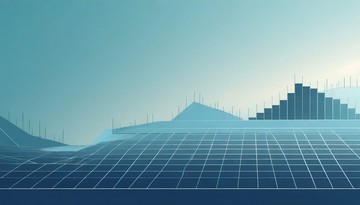} & \includegraphics[width=0.486\textwidth,height=0.2777\textwidth]{raw_images/DE3/storybook_3_0.jpg} \\
\includegraphics[width=0.486\textwidth,height=0.2777\textwidth]{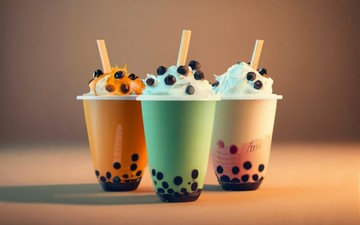} & \includegraphics[width=0.486\textwidth,height=0.2777\textwidth]{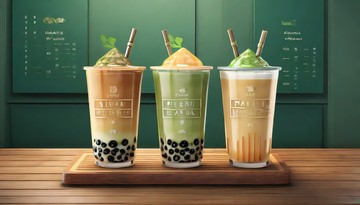} & \includegraphics[width=0.486\textwidth,height=0.2777\textwidth]{raw_images/DE3/storybook_4_0.jpg} \\
\includegraphics[width=0.486\textwidth,height=0.2777\textwidth]{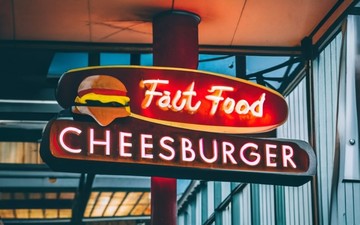} & \includegraphics[width=0.486\textwidth,height=0.2777\textwidth]{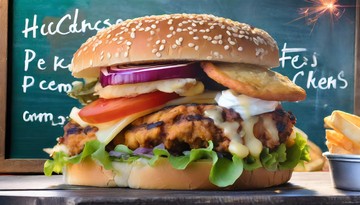} & \includegraphics[width=0.486\textwidth,height=0.2777\textwidth]{raw_images/DE3/storybook_5_0.jpg}
\end{tabular}
\end{adjustbox}
\caption[storybooks, academic posters, and menus comparisons]{Qualitative comparisons among Ideogram, Firefly 2, and \modelname. Check Figure~\ref{fig:text_storybook} for complete prompts.
}
\label{fig:flyfire_storybook_20}
\end{figure}

\begin{figure}
\centering
\vspace{-20mm}
\begin{adjustbox}{center}
\small
\begin{tabular}{x{0.496\textwidth}@{}x{0.496\textwidth}@{}x{0.496\textwidth}}
SDXL & Midjourney & DALL-E 3 \\
\includegraphics[width=0.486\textwidth,height=0.2777\textwidth]{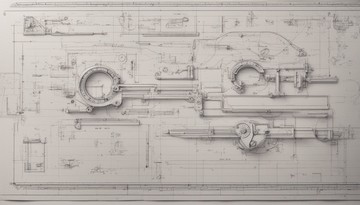} & \includegraphics[width=0.486\textwidth,height=0.2777\textwidth]{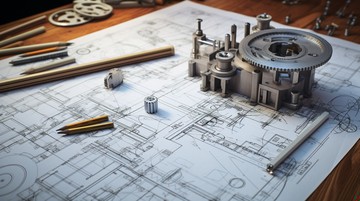} & \includegraphics[width=0.486\textwidth,height=0.2777\textwidth]{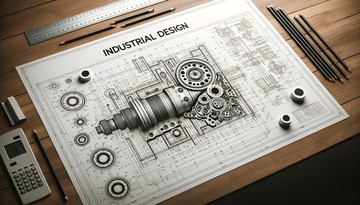} \\
\includegraphics[width=0.486\textwidth,height=0.2777\textwidth]{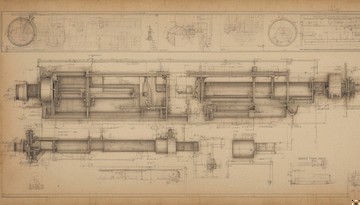} & \includegraphics[width=0.486\textwidth,height=0.2777\textwidth]{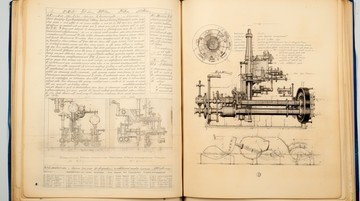} & \includegraphics[width=0.486\textwidth,height=0.2777\textwidth]{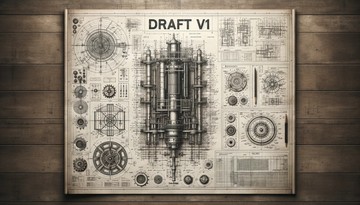} \\
\includegraphics[width=0.486\textwidth,height=0.2777\textwidth]{raw_images/SDXL/gui_2_0.jpg} & \includegraphics[width=0.486\textwidth,height=0.2777\textwidth]{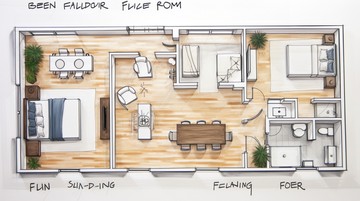} & \includegraphics[width=0.486\textwidth,height=0.2777\textwidth]{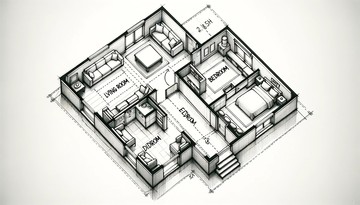} \\
\includegraphics[width=0.486\textwidth,height=0.2777\textwidth]{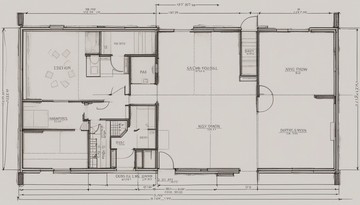} & \includegraphics[width=0.486\textwidth,height=0.2777\textwidth]{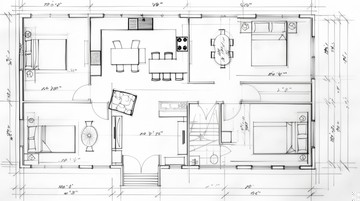} & \includegraphics[width=0.486\textwidth,height=0.2777\textwidth]{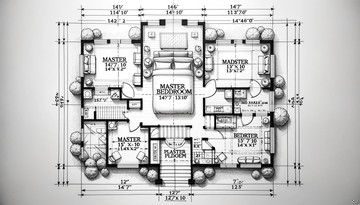} \\
\includegraphics[width=0.486\textwidth,height=0.2777\textwidth]{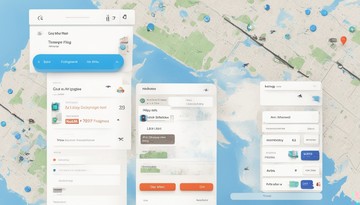} & \includegraphics[width=0.486\textwidth,height=0.2777\textwidth]{raw_images/MJ/gui_4_0.jpg} & \includegraphics[width=0.486\textwidth,height=0.2777\textwidth]{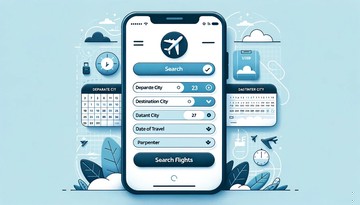} \\
\includegraphics[width=0.486\textwidth,height=0.2777\textwidth]{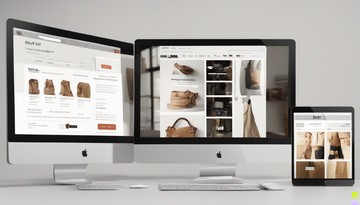} & \includegraphics[width=0.486\textwidth,height=0.2777\textwidth]{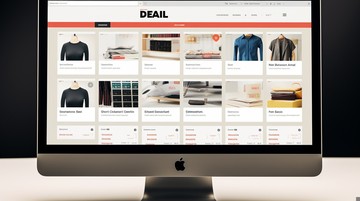} & \includegraphics[width=0.486\textwidth,height=0.2777\textwidth]{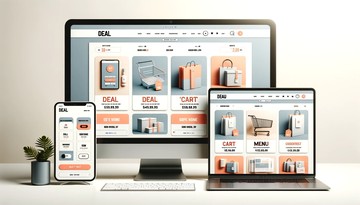}
\end{tabular}
\end{adjustbox}
\caption[industrial drafts, floorplans, and GUI comparisons]{Qualitative comparisons among SDXL, Midjourney v5.2, and \modelname. Check Figure~\ref{fig:text_GUI} for complete prompts.
}
\label{fig:compare_GUI_21}
\end{figure}

\begin{figure}
\centering
\vspace{-20mm}
\begin{adjustbox}{center}
\small
\begin{tabular}{x{0.496\textwidth}@{}x{0.496\textwidth}@{}x{0.496\textwidth}}
Ideogram & Firefly 2 & DALL-E 3 \\
\includegraphics[width=0.486\textwidth,height=0.2777\textwidth]{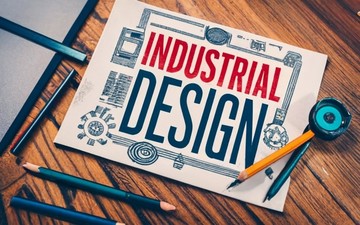} & \includegraphics[width=0.486\textwidth,height=0.2777\textwidth]{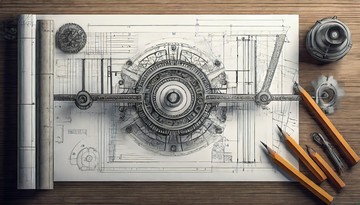} & \includegraphics[width=0.486\textwidth,height=0.2777\textwidth]{raw_images/DE3/gui_0_0.jpg} \\
\includegraphics[width=0.486\textwidth,height=0.2777\textwidth]{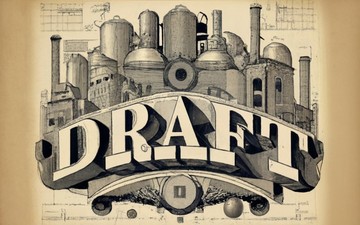} & \includegraphics[width=0.486\textwidth,height=0.2777\textwidth]{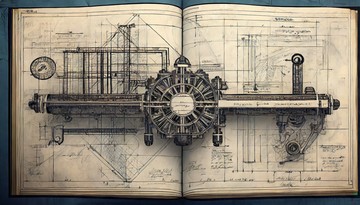} & \includegraphics[width=0.486\textwidth,height=0.2777\textwidth]{raw_images/DE3/gui_1_0.jpg} \\
\includegraphics[width=0.486\textwidth,height=0.2777\textwidth]{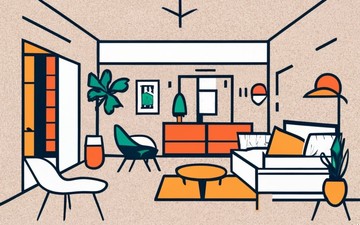} & \includegraphics[width=0.486\textwidth,height=0.2777\textwidth]{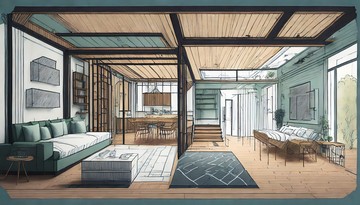} & \includegraphics[width=0.486\textwidth,height=0.2777\textwidth]{raw_images/DE3/gui_2_0.jpg} \\
\includegraphics[width=0.486\textwidth,height=0.2777\textwidth]{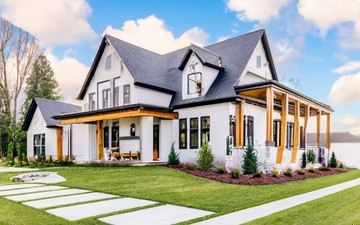} & \includegraphics[width=0.486\textwidth,height=0.2777\textwidth]{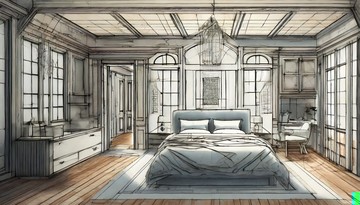} & \includegraphics[width=0.486\textwidth,height=0.2777\textwidth]{raw_images/DE3/gui_3_0.jpg} \\
\includegraphics[width=0.486\textwidth,height=0.2777\textwidth]{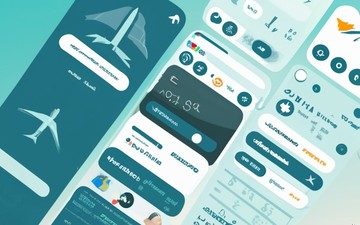} & \includegraphics[width=0.486\textwidth,height=0.2777\textwidth]{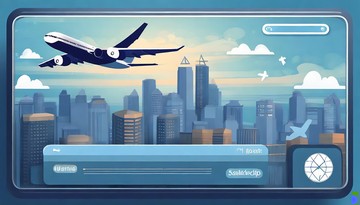} & \includegraphics[width=0.486\textwidth,height=0.2777\textwidth]{raw_images/DE3/gui_4_0.jpg} \\
\includegraphics[width=0.486\textwidth,height=0.2777\textwidth]{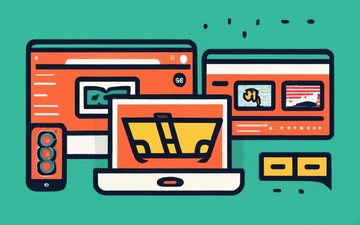} & \includegraphics[width=0.486\textwidth,height=0.2777\textwidth]{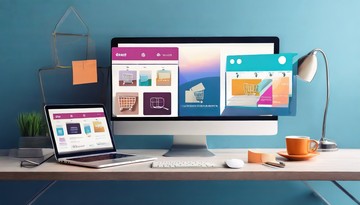} & \includegraphics[width=0.486\textwidth,height=0.2777\textwidth]{raw_images/DE3/gui_5_0.jpg}
\end{tabular}
\end{adjustbox}
\caption[industrial drafts, floorplans, and GUI comparisons]{Qualitative comparisons among Ideogram, Firefly 2, and \modelname. Check Figure~\ref{fig:text_GUI} for complete prompts.
}
\label{fig:flyfire_GUI_21}
\end{figure}

\begin{figure}
\centering
\vspace{-20mm}
\begin{adjustbox}{center}
\small
\begin{tabular}{x{0.496\textwidth}@{}x{0.496\textwidth}@{}x{0.496\textwidth}}
SDXL & Midjourney & DALL-E 3 \\
\includegraphics[width=0.486\textwidth,height=0.2777\textwidth]{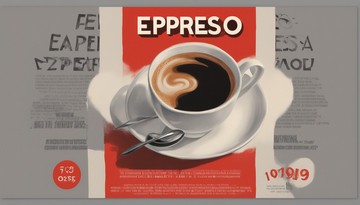} & \includegraphics[width=0.486\textwidth,height=0.2777\textwidth]{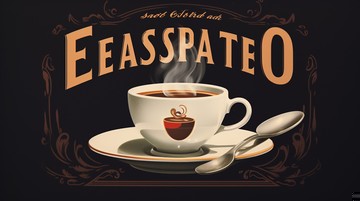} & \includegraphics[width=0.486\textwidth,height=0.2777\textwidth]{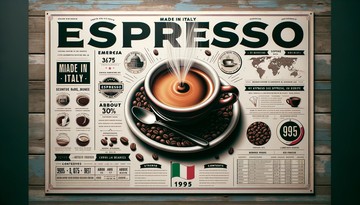} \\
\includegraphics[width=0.486\textwidth,height=0.2777\textwidth]{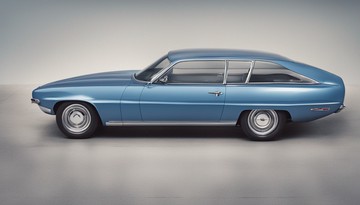} & \includegraphics[width=0.486\textwidth,height=0.2777\textwidth]{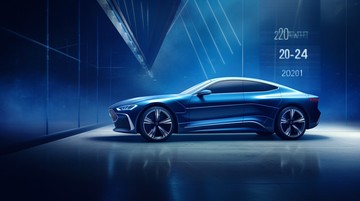} & \includegraphics[width=0.486\textwidth,height=0.2777\textwidth]{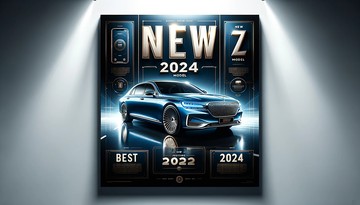} \\
\includegraphics[width=0.486\textwidth,height=0.2777\textwidth]{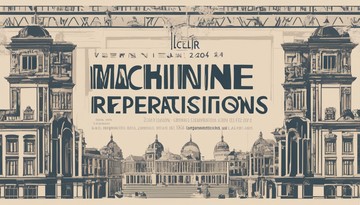} & \includegraphics[width=0.486\textwidth,height=0.2777\textwidth]{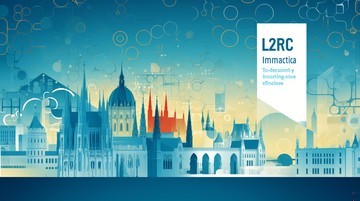} & \includegraphics[width=0.486\textwidth,height=0.2777\textwidth]{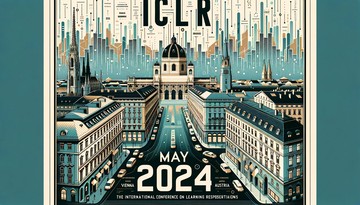} \\
\includegraphics[width=0.486\textwidth,height=0.2777\textwidth]{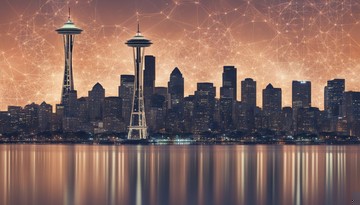} & \includegraphics[width=0.486\textwidth,height=0.2777\textwidth]{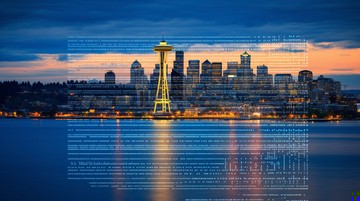} & \includegraphics[width=0.486\textwidth,height=0.2777\textwidth]{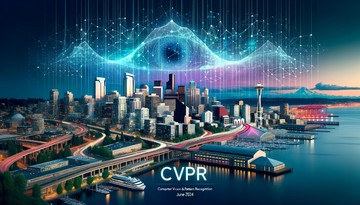} \\
\includegraphics[width=0.486\textwidth,height=0.2777\textwidth]{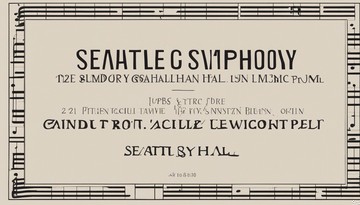} & \includegraphics[width=0.486\textwidth,height=0.2777\textwidth]{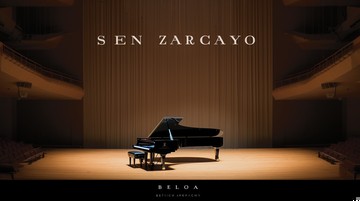} & \includegraphics[width=0.486\textwidth,height=0.2777\textwidth]{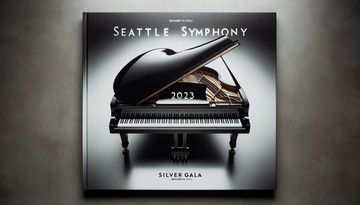} \\
\includegraphics[width=0.486\textwidth,height=0.2777\textwidth]{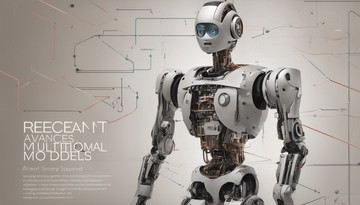} & \includegraphics[width=0.486\textwidth,height=0.2777\textwidth]{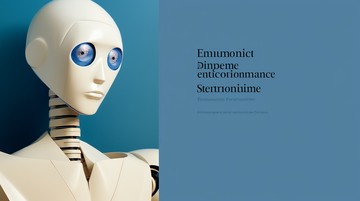} & \includegraphics[width=0.486\textwidth,height=0.2777\textwidth]{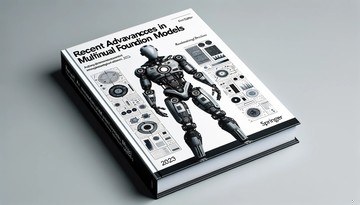}
\end{tabular}
\end{adjustbox}
\caption[ads, posters, and book cover comparisons]{Qualitative comparisons among SDXL, Midjourney v5.2, and \modelname. Check Figure~\ref{fig:text_poster} for complete prompts.
}
\label{fig:compare_poster_22}
\end{figure}

\begin{figure}
\centering
\vspace{-20mm}
\begin{adjustbox}{center}
\small
\begin{tabular}{x{0.496\textwidth}@{}x{0.496\textwidth}@{}x{0.496\textwidth}}
Ideogram & Firefly 2 & DALL-E 3 \\
\includegraphics[width=0.486\textwidth,height=0.2777\textwidth]{raw_images/Ideogram/poster_0_0.jpeg} & \includegraphics[width=0.486\textwidth,height=0.2777\textwidth]{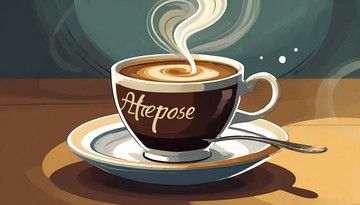} & \includegraphics[width=0.486\textwidth,height=0.2777\textwidth]{raw_images/DE3/poster_0_0.jpg} \\
\includegraphics[width=0.486\textwidth,height=0.2777\textwidth]{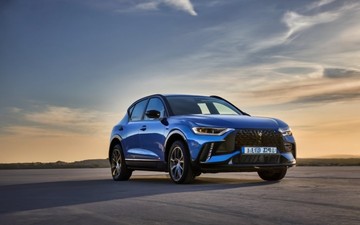} & \includegraphics[width=0.486\textwidth,height=0.2777\textwidth]{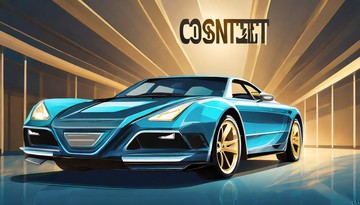} & \includegraphics[width=0.486\textwidth,height=0.2777\textwidth]{raw_images/DE3/poster_1_0.jpg} \\
\includegraphics[width=0.486\textwidth,height=0.2777\textwidth]{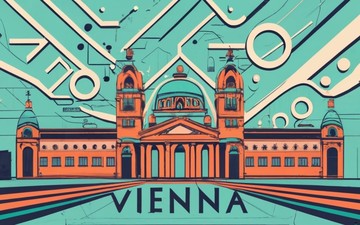} & \includegraphics[width=0.486\textwidth,height=0.2777\textwidth]{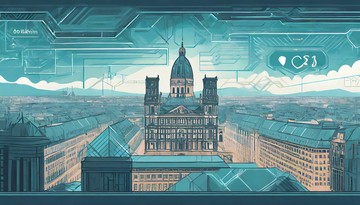} & \includegraphics[width=0.486\textwidth,height=0.2777\textwidth]{raw_images/DE3/poster_2_0.jpg} \\
\includegraphics[width=0.486\textwidth,height=0.2777\textwidth]{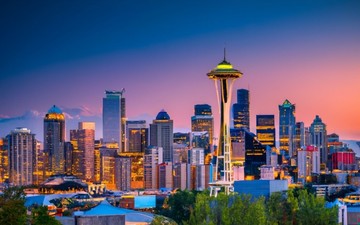} & \includegraphics[width=0.486\textwidth,height=0.2777\textwidth]{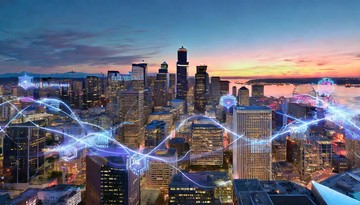} & \includegraphics[width=0.486\textwidth,height=0.2777\textwidth]{raw_images/DE3/poster_3_0.jpg} \\
\includegraphics[width=0.486\textwidth,height=0.2777\textwidth]{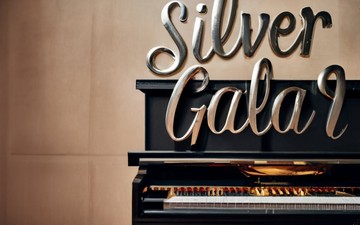} & \includegraphics[width=0.486\textwidth,height=0.2777\textwidth]{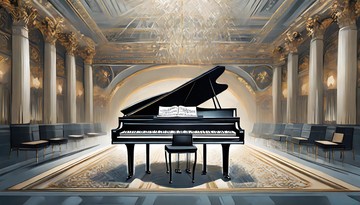} & \includegraphics[width=0.486\textwidth,height=0.2777\textwidth]{raw_images/DE3/poster_4_0.jpg} \\
\includegraphics[width=0.486\textwidth,height=0.2777\textwidth]{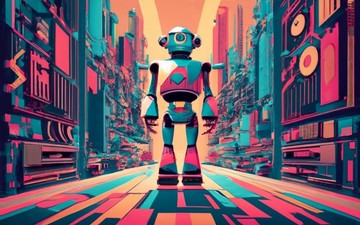} & \includegraphics[width=0.486\textwidth,height=0.2777\textwidth]{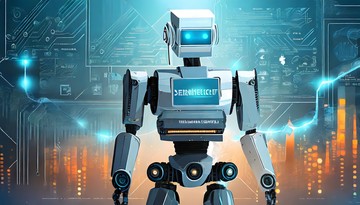} & \includegraphics[width=0.486\textwidth,height=0.2777\textwidth]{raw_images/DE3/poster_5_0.jpg}
\end{tabular}
\end{adjustbox}
\caption[ads, posters, and book cover comparisons]{Qualitative comparisons among Ideogram, Firefly 2, and \modelname. Check Figure~\ref{fig:text_poster} for complete prompts.
}
\label{fig:flyfire_poster_22}
\end{figure}

\begin{figure}
\centering
\vspace{-20mm}
\begin{adjustbox}{center}
\small
\begin{tabular}{x{0.496\textwidth}@{}x{0.496\textwidth}@{}x{0.496\textwidth}}
SDXL & Midjourney & DALL-E 3 \\
\includegraphics[width=0.486\textwidth,height=0.2777\textwidth]{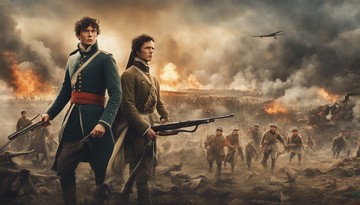} & \includegraphics[width=0.486\textwidth,height=0.2777\textwidth]{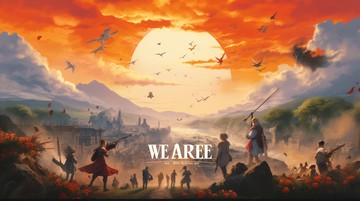} & \includegraphics[width=0.486\textwidth,height=0.2777\textwidth]{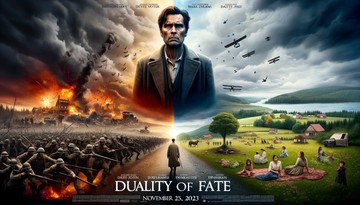} \\
\includegraphics[width=0.486\textwidth,height=0.2777\textwidth]{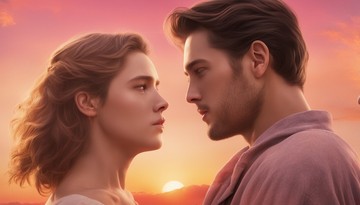} & \includegraphics[width=0.486\textwidth,height=0.2777\textwidth]{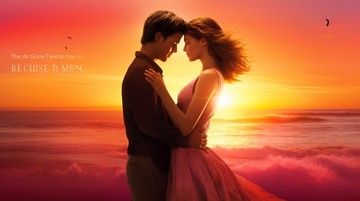} & \includegraphics[width=0.486\textwidth,height=0.2777\textwidth]{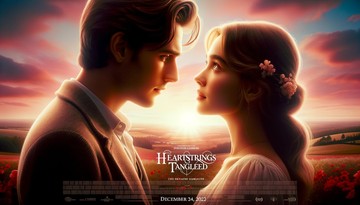} \\
\includegraphics[width=0.486\textwidth,height=0.2777\textwidth]{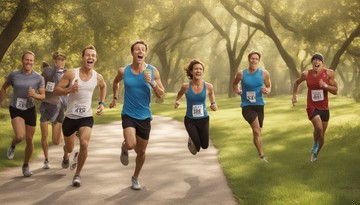} & \includegraphics[width=0.486\textwidth,height=0.2777\textwidth]{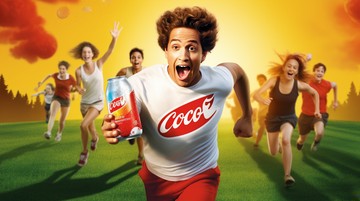} & \includegraphics[width=0.486\textwidth,height=0.2777\textwidth]{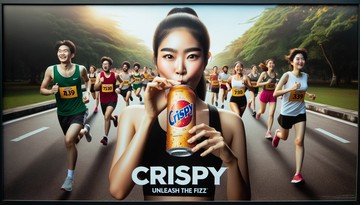} \\
\includegraphics[width=0.486\textwidth,height=0.2777\textwidth]{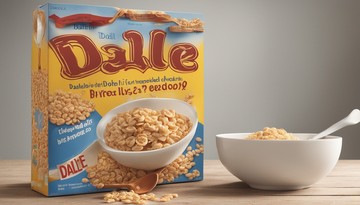} & \includegraphics[width=0.486\textwidth,height=0.2777\textwidth]{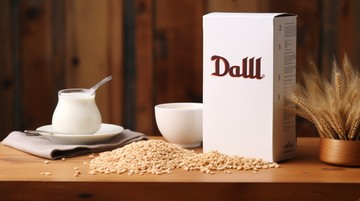} & \includegraphics[width=0.486\textwidth,height=0.2777\textwidth]{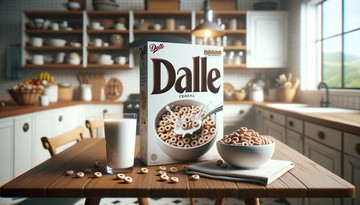} \\
\includegraphics[width=0.486\textwidth,height=0.2777\textwidth]{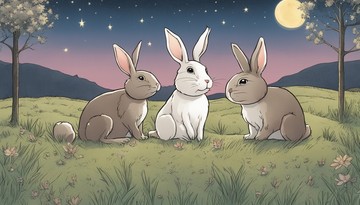} & \includegraphics[width=0.486\textwidth,height=0.2777\textwidth]{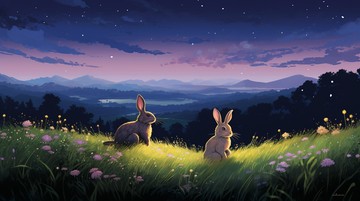} & \includegraphics[width=0.486\textwidth,height=0.2777\textwidth]{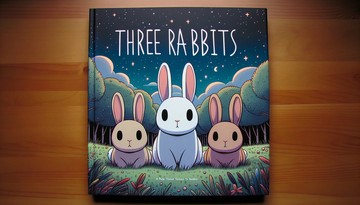} \\
\includegraphics[width=0.486\textwidth,height=0.2777\textwidth]{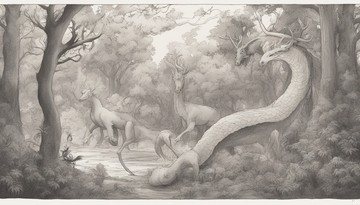} & \includegraphics[width=0.486\textwidth,height=0.2777\textwidth]{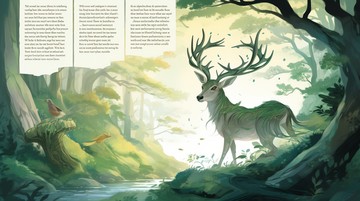} & \includegraphics[width=0.486\textwidth,height=0.2777\textwidth]{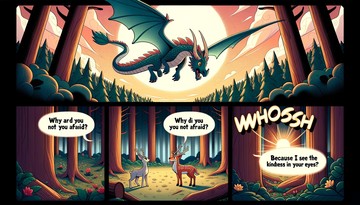}
\end{tabular}
\end{adjustbox}
\caption[movie posters and ads comparisons]{Qualitative comparisons among SDXL, Midjourney v5.2, and \modelname. Check Figure~\ref{fig:text_ads} for complete prompts.
}
\label{fig:compare_ads_23}
\end{figure}

\begin{figure}
\centering
\vspace{-20mm}
\begin{adjustbox}{center}
\small
\begin{tabular}{x{0.496\textwidth}@{}x{0.496\textwidth}@{}x{0.496\textwidth}}
Ideogram & Firefly 2 & DALL-E 3 \\
\includegraphics[width=0.486\textwidth,height=0.2777\textwidth]{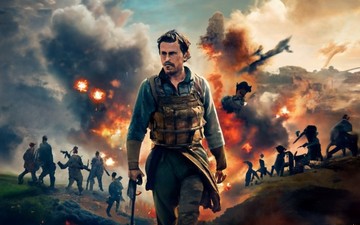} & \includegraphics[width=0.486\textwidth,height=0.2777\textwidth]{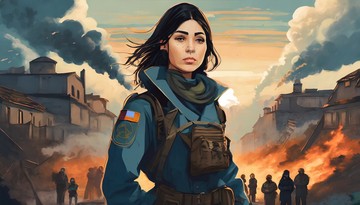} & \includegraphics[width=0.486\textwidth,height=0.2777\textwidth]{raw_images/DE3/ads_0_0.jpg} \\
\includegraphics[width=0.486\textwidth,height=0.2777\textwidth]{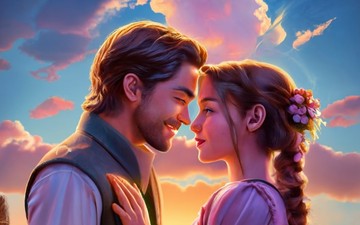} & \includegraphics[width=0.486\textwidth,height=0.2777\textwidth]{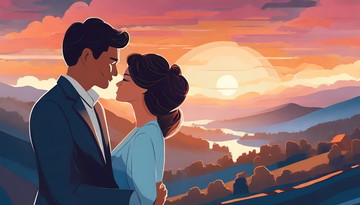} & \includegraphics[width=0.486\textwidth,height=0.2777\textwidth]{raw_images/DE3/ads_1_0.jpg} \\
\includegraphics[width=0.486\textwidth,height=0.2777\textwidth]{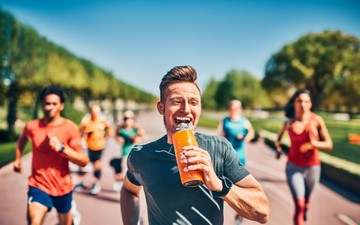} & \includegraphics[width=0.486\textwidth,height=0.2777\textwidth]{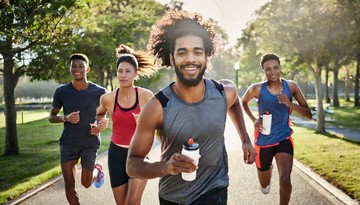} & \includegraphics[width=0.486\textwidth,height=0.2777\textwidth]{raw_images/DE3/ads_2_0.jpg} \\
\includegraphics[width=0.486\textwidth,height=0.2777\textwidth]{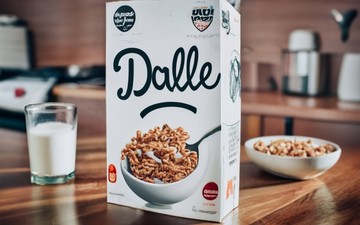} & \includegraphics[width=0.486\textwidth,height=0.2777\textwidth]{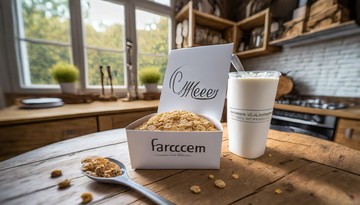} & \includegraphics[width=0.486\textwidth,height=0.2777\textwidth]{raw_images/DE3/ads_3_0.jpg} \\
\includegraphics[width=0.486\textwidth,height=0.2777\textwidth]{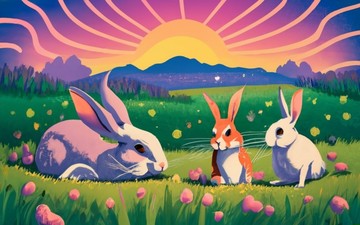} & \includegraphics[width=0.486\textwidth,height=0.2777\textwidth]{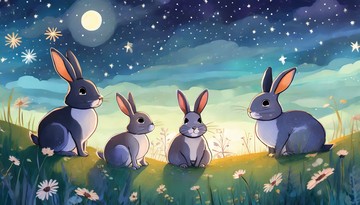} & \includegraphics[width=0.486\textwidth,height=0.2777\textwidth]{raw_images/DE3/ads_4_0.jpg} \\
\includegraphics[width=0.486\textwidth,height=0.2777\textwidth]{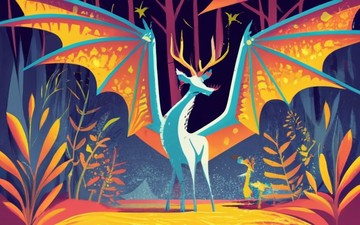} & \includegraphics[width=0.486\textwidth,height=0.2777\textwidth]{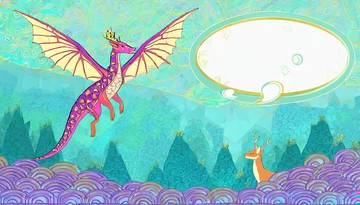} & \includegraphics[width=0.486\textwidth,height=0.2777\textwidth]{raw_images/DE3/ads_5_0.jpg}
\end{tabular}
\end{adjustbox}
\caption[movie posters and ads comparisons]{Qualitative comparisons among Ideogram, Firefly 2, and \modelname. Check Figure~\ref{fig:text_ads} for complete prompts.
}
\label{fig:flyfire_ads_23}
\end{figure}

\begin{figure}
\centering
\vspace{-20mm}
\begin{adjustbox}{center}
\small
\begin{tabular}{x{0.496\textwidth}@{}x{0.496\textwidth}@{}x{0.496\textwidth}}
SDXL & Midjourney & DALL-E 3 \\
\includegraphics[width=0.486\textwidth,height=0.2777\textwidth]{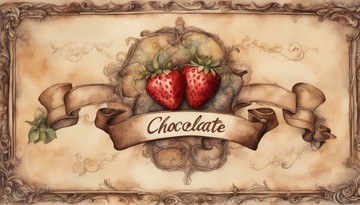} & \includegraphics[width=0.486\textwidth,height=0.2777\textwidth]{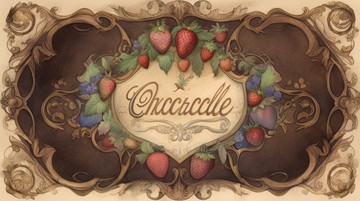} & \includegraphics[width=0.486\textwidth,height=0.2777\textwidth]{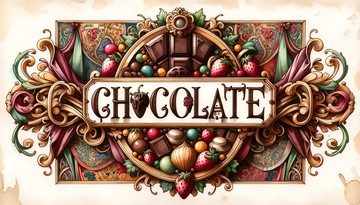} \\
\includegraphics[width=0.486\textwidth,height=0.2777\textwidth]{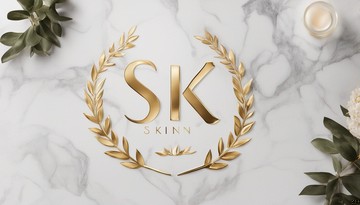} & \includegraphics[width=0.486\textwidth,height=0.2777\textwidth]{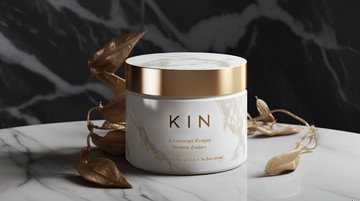} & \includegraphics[width=0.486\textwidth,height=0.2777\textwidth]{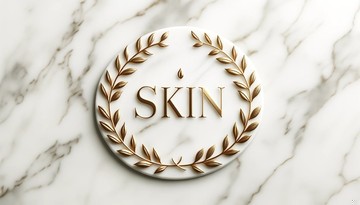} \\
\includegraphics[width=0.486\textwidth,height=0.2777\textwidth]{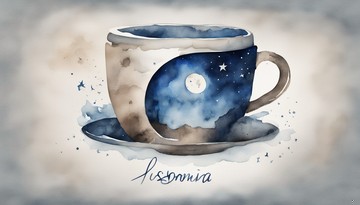} & \includegraphics[width=0.486\textwidth,height=0.2777\textwidth]{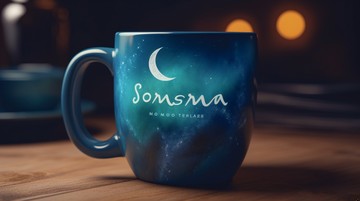} & \includegraphics[width=0.486\textwidth,height=0.2777\textwidth]{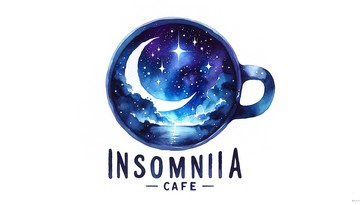} \\
\includegraphics[width=0.486\textwidth,height=0.2777\textwidth]{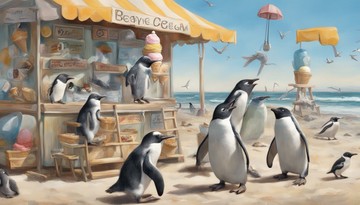} & \includegraphics[width=0.486\textwidth,height=0.2777\textwidth]{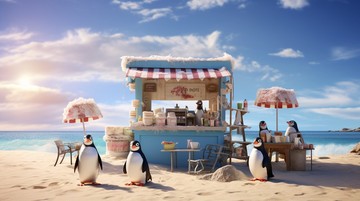} & \includegraphics[width=0.486\textwidth,height=0.2777\textwidth]{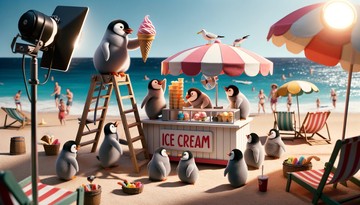} \\
\includegraphics[width=0.486\textwidth,height=0.2777\textwidth]{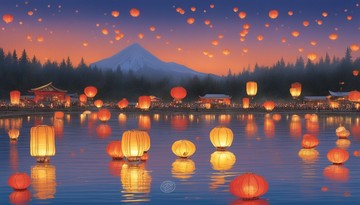} & \includegraphics[width=0.486\textwidth,height=0.2777\textwidth]{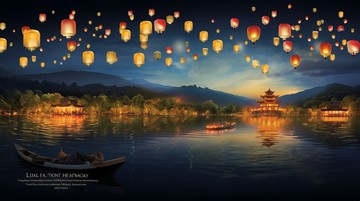} & \includegraphics[width=0.486\textwidth,height=0.2777\textwidth]{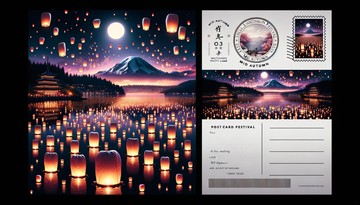} \\
\includegraphics[width=0.486\textwidth,height=0.2777\textwidth]{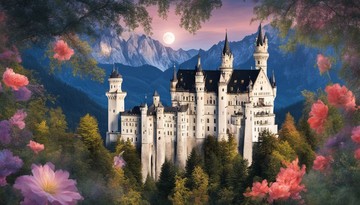} & \includegraphics[width=0.486\textwidth,height=0.2777\textwidth]{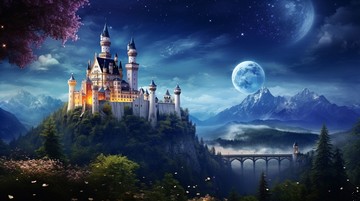} & \includegraphics[width=0.486\textwidth,height=0.2777\textwidth]{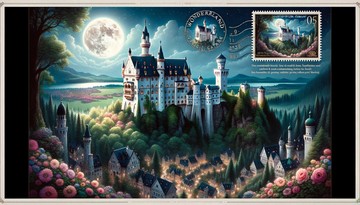}
\end{tabular}
\end{adjustbox}
\caption[infographics design comparisons]{Qualitative comparisons among SDXL, Midjourney v5.2, and \modelname. Check Figure~\ref{fig:scenario_logo_postcard_1} for complete prompts.
}
\label{fig:compare_logo_24}
\end{figure}

\begin{figure}
\centering
\vspace{-20mm}
\begin{adjustbox}{center}
\small
\begin{tabular}{x{0.496\textwidth}@{}x{0.496\textwidth}@{}x{0.496\textwidth}}
Ideogram & Firefly 2 & DALL-E 3 \\
\includegraphics[width=0.486\textwidth,height=0.2777\textwidth]{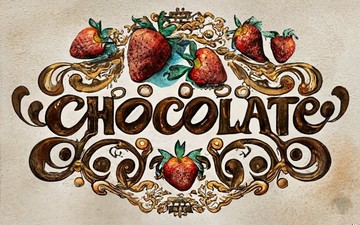} & \includegraphics[width=0.486\textwidth,height=0.2777\textwidth]{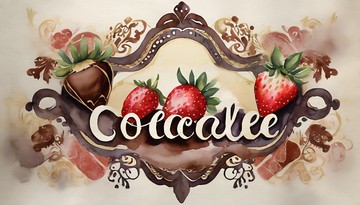} & \includegraphics[width=0.486\textwidth,height=0.2777\textwidth]{raw_images/DE3/logo_0_0.jpg} \\
\includegraphics[width=0.486\textwidth,height=0.2777\textwidth]{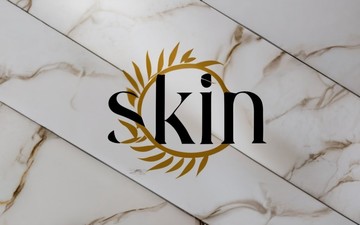} & \includegraphics[width=0.486\textwidth,height=0.2777\textwidth]{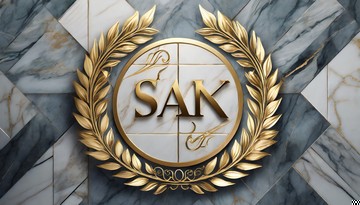} & \includegraphics[width=0.486\textwidth,height=0.2777\textwidth]{raw_images/DE3/logo_1_0.jpg} \\
\includegraphics[width=0.486\textwidth,height=0.2777\textwidth]{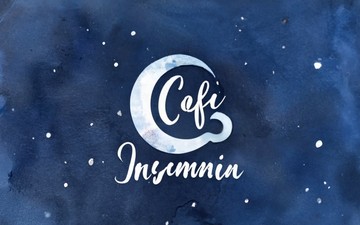} & \includegraphics[width=0.486\textwidth,height=0.2777\textwidth]{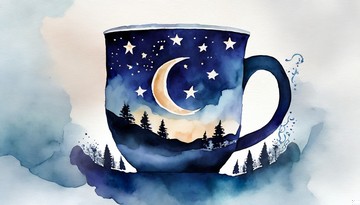} & \includegraphics[width=0.486\textwidth,height=0.2777\textwidth]{raw_images/DE3/logo_2_0.jpg} \\
\includegraphics[width=0.486\textwidth,height=0.2777\textwidth]{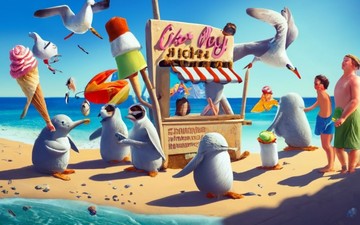} & \includegraphics[width=0.486\textwidth,height=0.2777\textwidth]{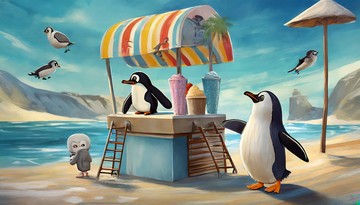} & \includegraphics[width=0.486\textwidth,height=0.2777\textwidth]{raw_images/DE3/logo_3_0.jpg} \\
\includegraphics[width=0.486\textwidth,height=0.2777\textwidth]{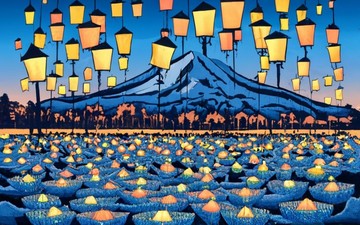} & \includegraphics[width=0.486\textwidth,height=0.2777\textwidth]{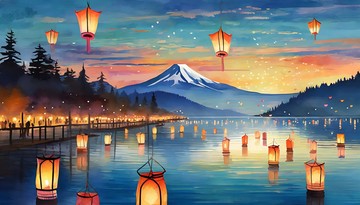} & \includegraphics[width=0.486\textwidth,height=0.2777\textwidth]{raw_images/DE3/postcard_0_0.jpg} \\
\includegraphics[width=0.486\textwidth,height=0.2777\textwidth]{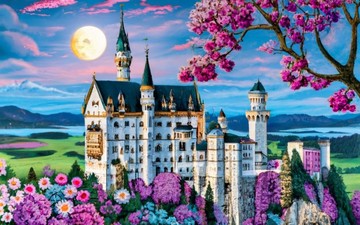} & \includegraphics[width=0.486\textwidth,height=0.2777\textwidth]{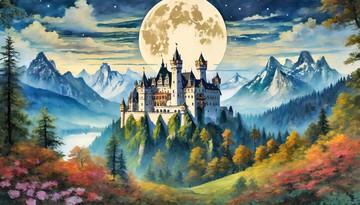} & \includegraphics[width=0.486\textwidth,height=0.2777\textwidth]{raw_images/DE3/postcard_1_0.jpg}
\end{tabular}
\end{adjustbox}
\caption[infographics design comparisons]{Qualitative comparisons among Ideogram, Firefly 2, and \modelname. Check Figure~\ref{fig:scenario_logo_postcard_1} for complete prompts.
}
\label{fig:flyfire_logo_24}
\end{figure}

\begin{figure}
\centering
\vspace{-20mm}
\begin{adjustbox}{center}
\small
\begin{tabular}{x{0.496\textwidth}@{}x{0.496\textwidth}@{}x{0.496\textwidth}}
SDXL & Midjourney & DALL-E 3 \\
\includegraphics[width=0.486\textwidth,height=0.2777\textwidth]{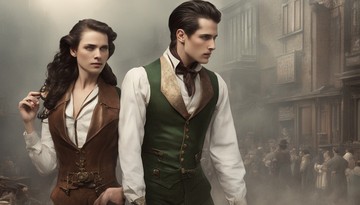} & \includegraphics[width=0.486\textwidth,height=0.2777\textwidth]{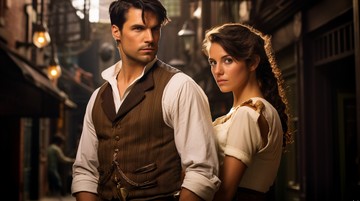} & \includegraphics[width=0.486\textwidth,height=0.2777\textwidth]{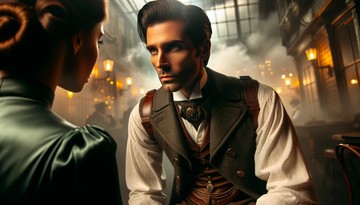} \\
\includegraphics[width=0.486\textwidth,height=0.2777\textwidth]{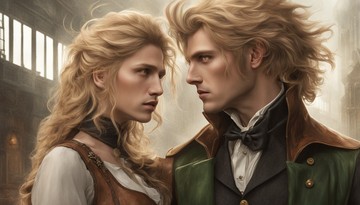} & \includegraphics[width=0.486\textwidth,height=0.2777\textwidth]{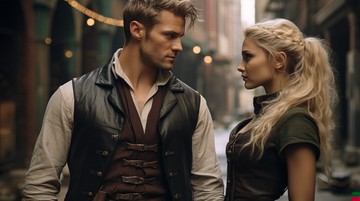} & \includegraphics[width=0.486\textwidth,height=0.2777\textwidth]{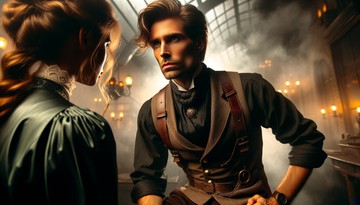} \\
\includegraphics[width=0.486\textwidth,height=0.2777\textwidth]{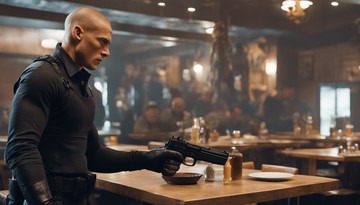} & \includegraphics[width=0.486\textwidth,height=0.2777\textwidth]{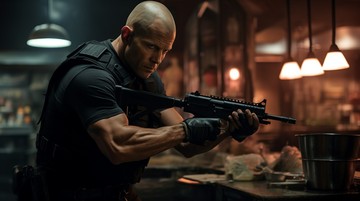} & \includegraphics[width=0.486\textwidth,height=0.2777\textwidth]{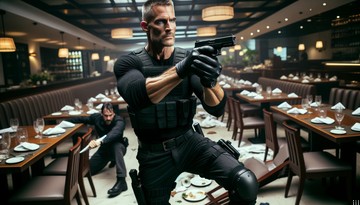} \\
\includegraphics[width=0.486\textwidth,height=0.2777\textwidth]{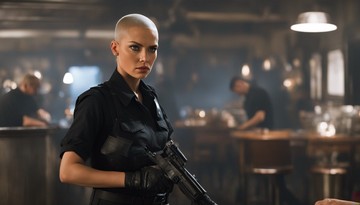} & \includegraphics[width=0.486\textwidth,height=0.2777\textwidth]{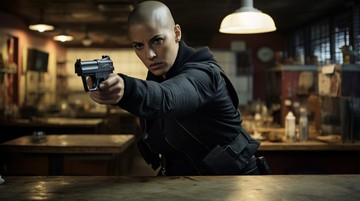} & \includegraphics[width=0.486\textwidth,height=0.2777\textwidth]{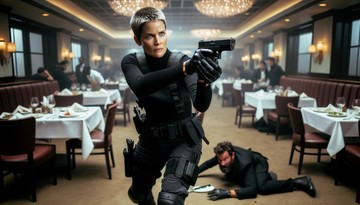} \\
\includegraphics[width=0.486\textwidth,height=0.2777\textwidth]{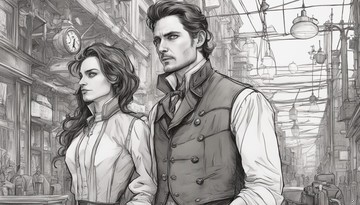} & \includegraphics[width=0.486\textwidth,height=0.2777\textwidth]{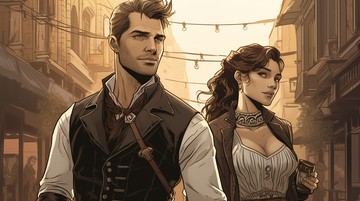} & \includegraphics[width=0.486\textwidth,height=0.2777\textwidth]{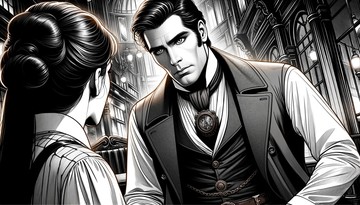} \\
\includegraphics[width=0.486\textwidth,height=0.2777\textwidth]{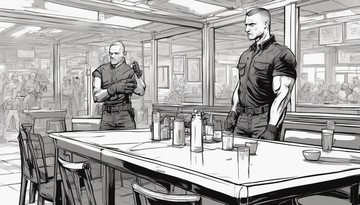} & \includegraphics[width=0.486\textwidth,height=0.2777\textwidth]{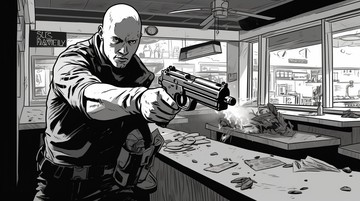} & \includegraphics[width=0.486\textwidth,height=0.2777\textwidth]{raw_images/DE3/cinematic_5_0.jpg}
\end{tabular}
\end{adjustbox}
\caption[cinematic scene comparisons]{Qualitative comparisons among SDXL, Midjourney v5.2, and \modelname. Check Figure~\ref{fig:text_cinematic} for complete prompts.
}
\label{fig:compare_cinematic_27}
\end{figure}

\begin{figure}
\centering
\vspace{-20mm}
\begin{adjustbox}{center}
\small
\begin{tabular}{x{0.496\textwidth}@{}x{0.496\textwidth}@{}x{0.496\textwidth}}
Ideogram & Firefly 2 & DALL-E 3 \\
\includegraphics[width=0.486\textwidth,height=0.2777\textwidth]{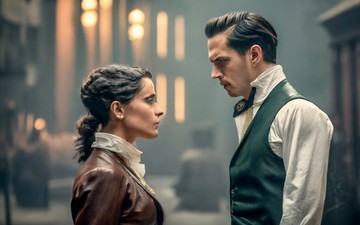} & \includegraphics[width=0.486\textwidth,height=0.2777\textwidth]{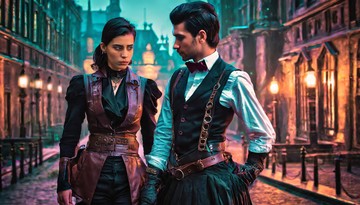} & \includegraphics[width=0.486\textwidth,height=0.2777\textwidth]{raw_images/DE3/cinematic_0_0.jpg} \\
\includegraphics[width=0.486\textwidth,height=0.2777\textwidth]{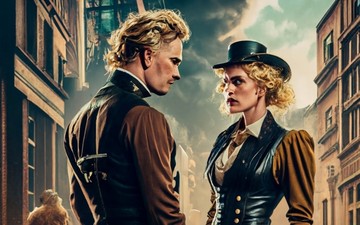} & \includegraphics[width=0.486\textwidth,height=0.2777\textwidth]{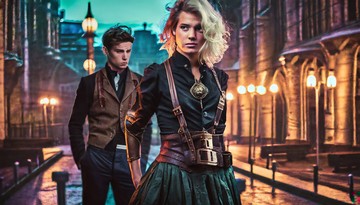} & \includegraphics[width=0.486\textwidth,height=0.2777\textwidth]{raw_images/DE3/cinematic_1_0.jpg} \\
\includegraphics[width=0.486\textwidth,height=0.2777\textwidth]{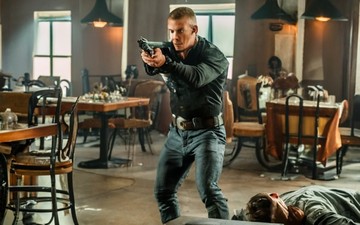} & \includegraphics[width=0.486\textwidth,height=0.2777\textwidth]{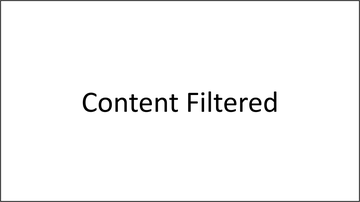} & \includegraphics[width=0.486\textwidth,height=0.2777\textwidth]{raw_images/DE3/cinematic_2_0.jpg} \\
\includegraphics[width=0.486\textwidth,height=0.2777\textwidth]{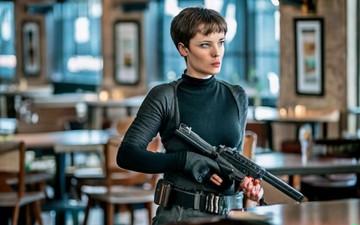} & \includegraphics[width=0.486\textwidth,height=0.2777\textwidth]{raw_images/emptyimage.png} & \includegraphics[width=0.486\textwidth,height=0.2777\textwidth]{raw_images/DE3/cinematic_3_0.jpg} \\
\includegraphics[width=0.486\textwidth,height=0.2777\textwidth]{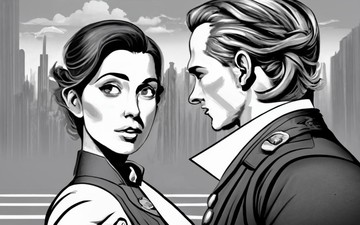} & \includegraphics[width=0.486\textwidth,height=0.2777\textwidth]{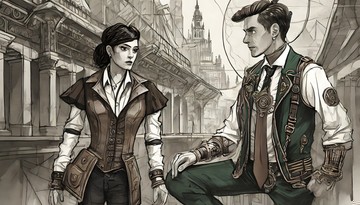} & \includegraphics[width=0.486\textwidth,height=0.2777\textwidth]{raw_images/DE3/cinematic_4_0.jpg} \\
\includegraphics[width=0.486\textwidth,height=0.2777\textwidth]{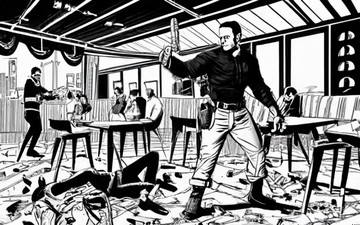} & \includegraphics[width=0.486\textwidth,height=0.2777\textwidth]{raw_images/emptyimage.png} & \includegraphics[width=0.486\textwidth,height=0.2777\textwidth]{raw_images/DE3/cinematic_5_0.jpg}
\end{tabular}
\end{adjustbox}
\caption[cinematic scene comparisons]{Qualitative comparisons among Ideogram, Firefly 2, and \modelname. Check Figure~\ref{fig:text_cinematic} for complete prompts. Firefly refuses to generate images for certain prompts, and we use a special ``Content Filtered'' as a placeholder image.
}
\label{fig:flyfire_cinematic_27}
\end{figure}

\begin{figure}
\centering
\vspace{-20mm}
\begin{adjustbox}{center}
\small
\begin{tabular}{x{0.496\textwidth}@{}x{0.496\textwidth}@{}x{0.496\textwidth}}
SDXL & Midjourney & DALL-E 3 \\
\includegraphics[width=0.486\textwidth,height=0.2777\textwidth]{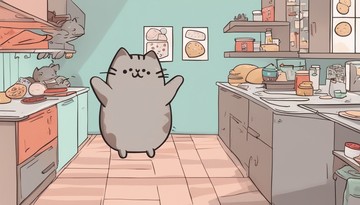} & \includegraphics[width=0.486\textwidth,height=0.2777\textwidth]{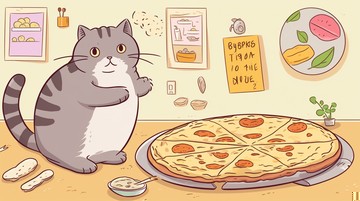} & \includegraphics[width=0.486\textwidth,height=0.2777\textwidth]{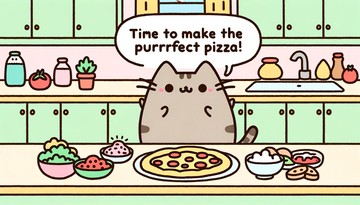} \\
\includegraphics[width=0.486\textwidth,height=0.2777\textwidth]{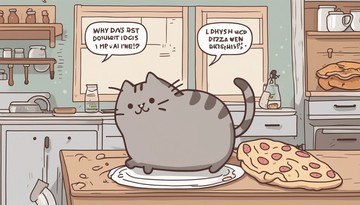} & \includegraphics[width=0.486\textwidth,height=0.2777\textwidth]{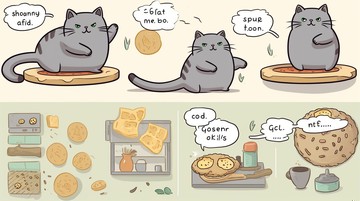} & \includegraphics[width=0.486\textwidth,height=0.2777\textwidth]{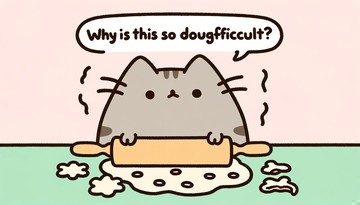} \\
\includegraphics[width=0.486\textwidth,height=0.2777\textwidth]{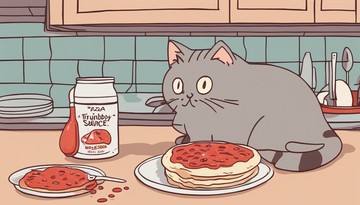} & \includegraphics[width=0.486\textwidth,height=0.2777\textwidth]{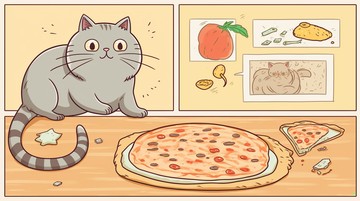} & \includegraphics[width=0.486\textwidth,height=0.2777\textwidth]{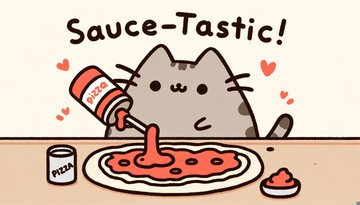} \\
\includegraphics[width=0.486\textwidth,height=0.2777\textwidth]{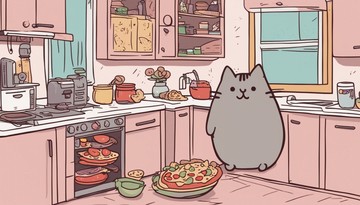} & \includegraphics[width=0.486\textwidth,height=0.2777\textwidth]{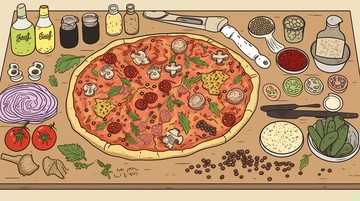} & \includegraphics[width=0.486\textwidth,height=0.2777\textwidth]{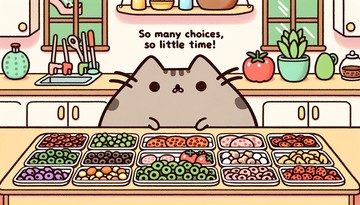} \\
\includegraphics[width=0.486\textwidth,height=0.2777\textwidth]{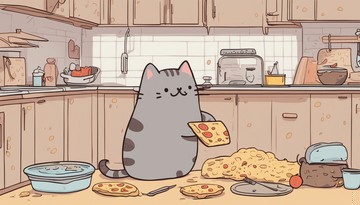} & \includegraphics[width=0.486\textwidth,height=0.2777\textwidth]{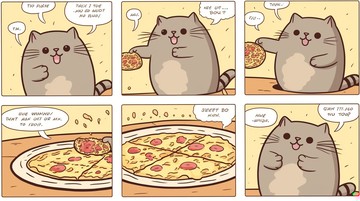} & \includegraphics[width=0.486\textwidth,height=0.2777\textwidth]{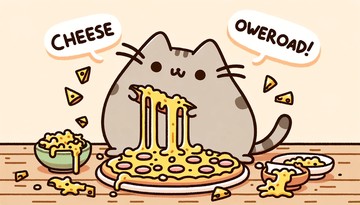} \\
\includegraphics[width=0.486\textwidth,height=0.2777\textwidth]{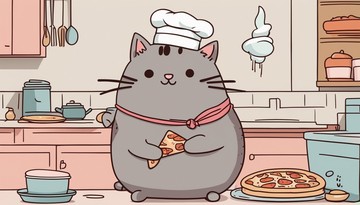} & \includegraphics[width=0.486\textwidth,height=0.2777\textwidth]{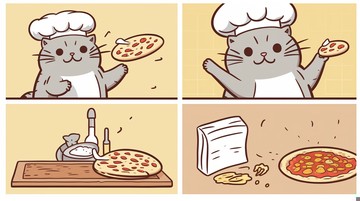} & \includegraphics[width=0.486\textwidth,height=0.2777\textwidth]{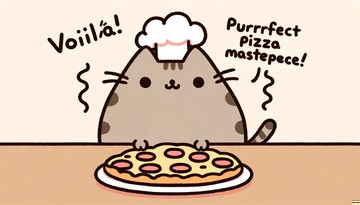}
\end{tabular}
\end{adjustbox}
\caption[comic strip comparisons]{Qualitative comparisons among SDXL, Midjourney v5.2, and \modelname. Check Figure~\ref{fig:text_comic_2} for complete prompts.
}
\label{fig:compare_comic_29}
\end{figure}

\begin{figure}
\centering
\vspace{-20mm}
\begin{adjustbox}{center}
\small
\begin{tabular}{x{0.496\textwidth}@{}x{0.496\textwidth}@{}x{0.496\textwidth}}
Ideogram & Firefly 2 & DALL-E 3 \\
\includegraphics[width=0.486\textwidth,height=0.2777\textwidth]{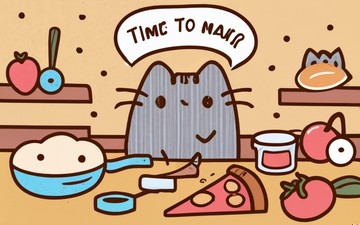} & \includegraphics[width=0.486\textwidth,height=0.2777\textwidth]{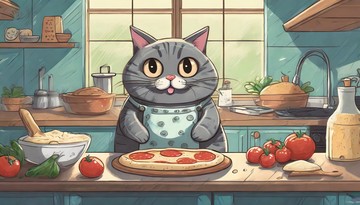} & \includegraphics[width=0.486\textwidth,height=0.2777\textwidth]{raw_images/DE3/comic_6_0.jpg} \\
\includegraphics[width=0.486\textwidth,height=0.2777\textwidth]{raw_images/Ideogram/comic_7_0.jpeg} & \includegraphics[width=0.486\textwidth,height=0.2777\textwidth]{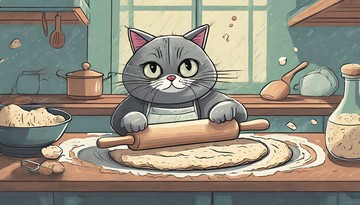} & \includegraphics[width=0.486\textwidth,height=0.2777\textwidth]{raw_images/DE3/comic_7_0.jpg} \\
\includegraphics[width=0.486\textwidth,height=0.2777\textwidth]{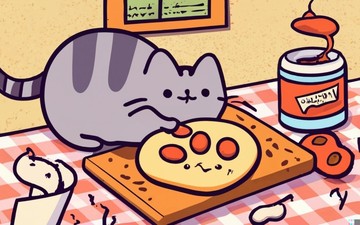} & \includegraphics[width=0.486\textwidth,height=0.2777\textwidth]{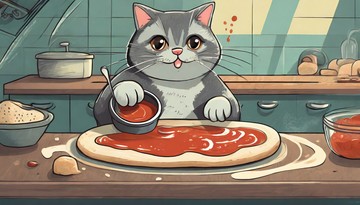} & \includegraphics[width=0.486\textwidth,height=0.2777\textwidth]{raw_images/DE3/comic_8_0.jpg} \\
\includegraphics[width=0.486\textwidth,height=0.2777\textwidth]{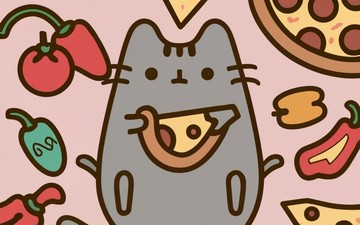} & \includegraphics[width=0.486\textwidth,height=0.2777\textwidth]{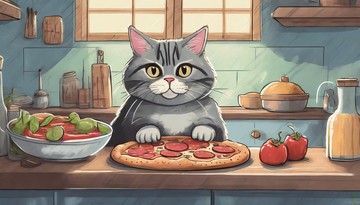} & \includegraphics[width=0.486\textwidth,height=0.2777\textwidth]{raw_images/DE3/comic_9_0.jpg} \\
\includegraphics[width=0.486\textwidth,height=0.2777\textwidth]{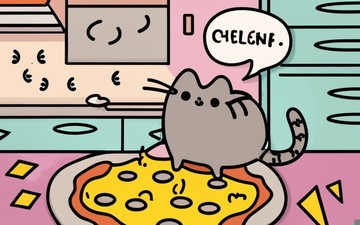} & \includegraphics[width=0.486\textwidth,height=0.2777\textwidth]{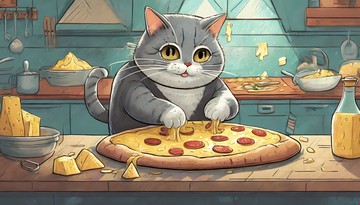} & \includegraphics[width=0.486\textwidth,height=0.2777\textwidth]{raw_images/DE3/comic_10_0.jpg} \\
\includegraphics[width=0.486\textwidth,height=0.2777\textwidth]{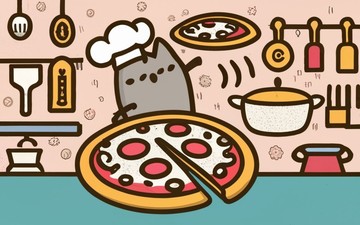} & \includegraphics[width=0.486\textwidth,height=0.2777\textwidth]{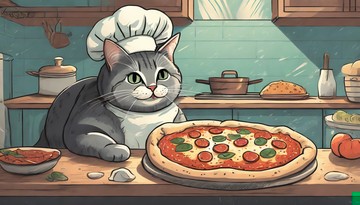} & \includegraphics[width=0.486\textwidth,height=0.2777\textwidth]{raw_images/DE3/comic_11_0.jpg}
\end{tabular}
\end{adjustbox}
\caption[comic strip comparisons]{Qualitative comparisons among Ideogram, Firefly 2, and \modelname. Check Figure~\ref{fig:text_comic_2} for complete prompts.
}
\label{fig:flyfire_comic_29}
\end{figure}

\begin{figure}
\centering
\vspace{-20mm}
\begin{adjustbox}{center}
\small
\begin{tabular}{x{0.496\textwidth}@{}x{0.496\textwidth}@{}x{0.496\textwidth}}
SDXL & Midjourney & DALL-E 3
\end{tabular}
\end{adjustbox}
\centerline{\includegraphics[width=1.5\textwidth]{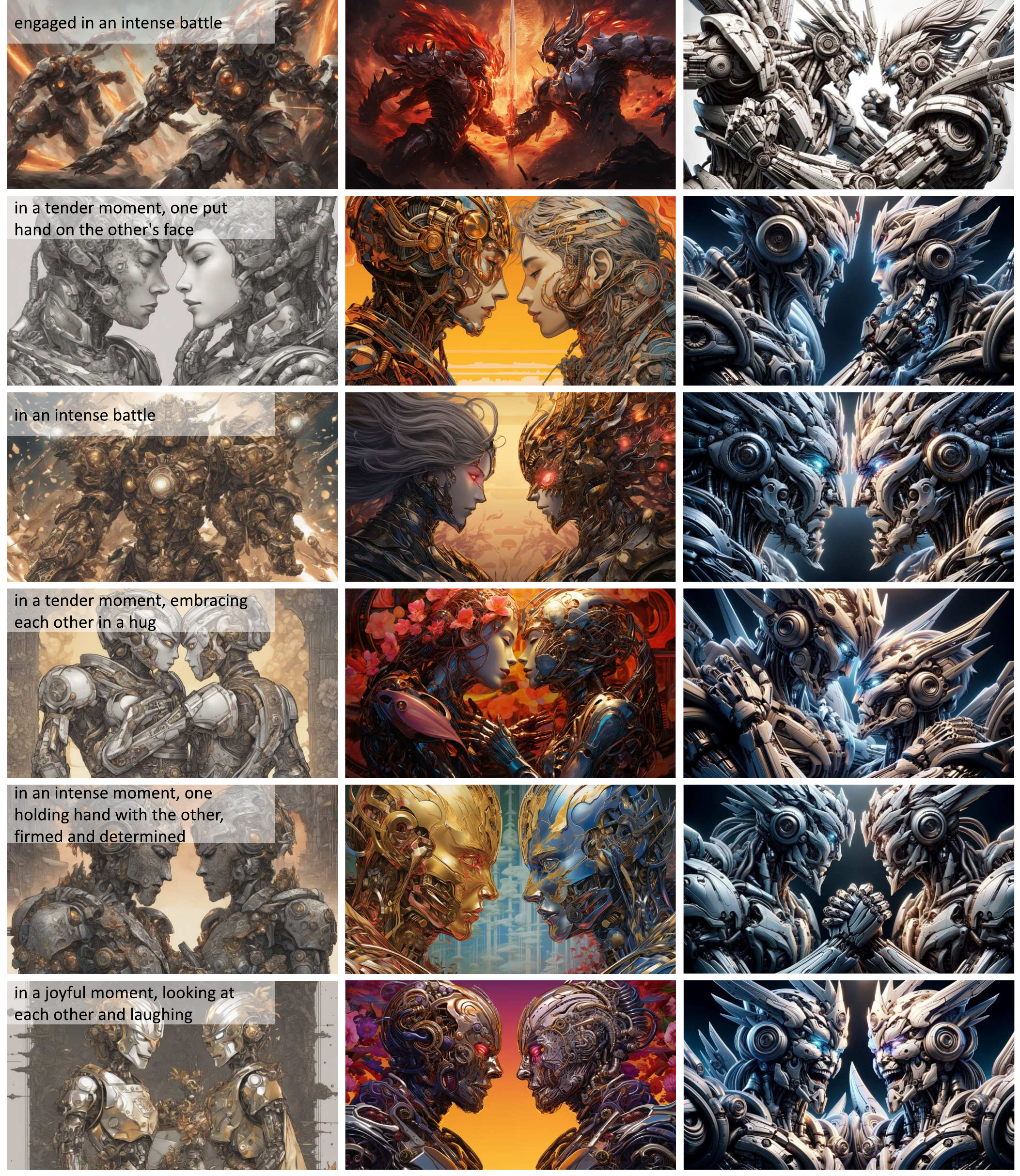}}
\caption[storyboard comparisons]{Qualitative comparisons among SDXL, Midjourney v5.2, and \modelname. Check Figure~\ref{fig:Scenario_Storyboard_Warriors} for complete prompts.
}
\label{fig:compare_storyboard_31}
\end{figure}

\begin{figure}
\centering
\vspace{-20mm}
\begin{adjustbox}{center}
\small
\begin{tabular}{x{0.496\textwidth}@{}x{0.496\textwidth}@{}x{0.496\textwidth}}
Ideogram & Firefly 2 & DALL-E 3
\end{tabular}
\end{adjustbox}
\centerline{\includegraphics[width=1.5\textwidth]{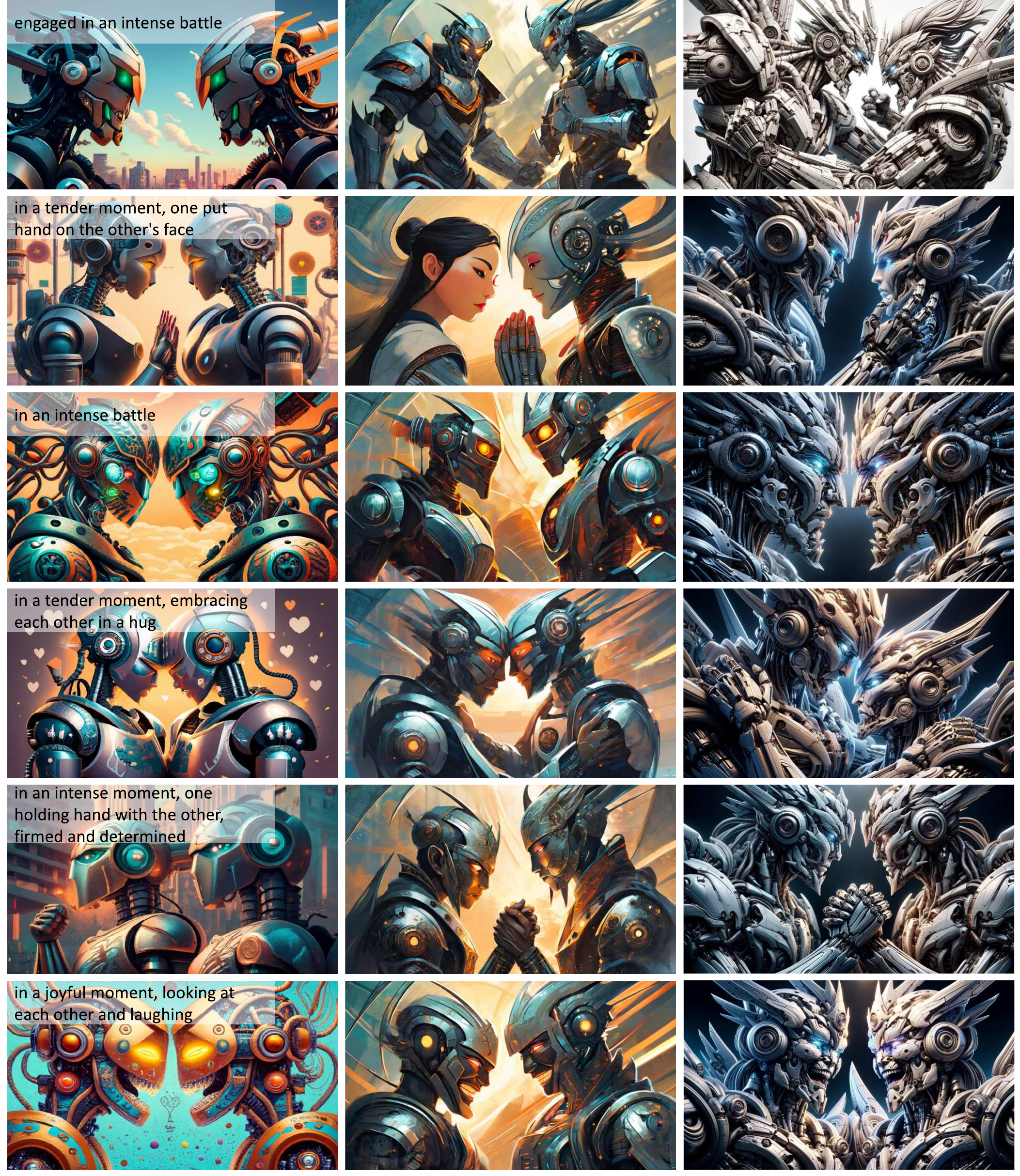}}
\caption[storyboard comparisons]{Qualitative comparisons among Ideogram, Firefly 2, and \modelname. Check Figure~\ref{fig:Scenario_Storyboard_Warriors} for complete prompts.
}
\label{fig:firefly_storyboard_31}
\end{figure}

\begin{figure}
\centering
\vspace{-20mm}
\begin{adjustbox}{center}
\small
\begin{tabular}{x{0.496\textwidth}@{}x{0.496\textwidth}@{}x{0.496\textwidth}}
SDXL & Midjourney & DALL-E 3 \\
\includegraphics[width=0.486\textwidth,height=0.2777\textwidth]{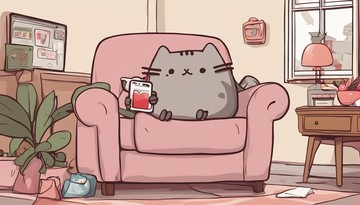} & \includegraphics[width=0.486\textwidth,height=0.2777\textwidth]{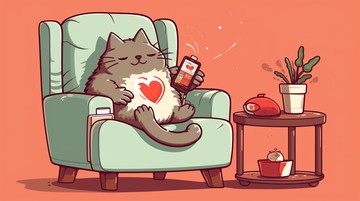} & \includegraphics[width=0.486\textwidth,height=0.2777\textwidth]{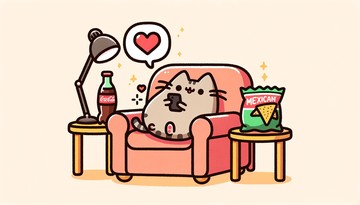} \\
\includegraphics[width=0.486\textwidth,height=0.2777\textwidth]{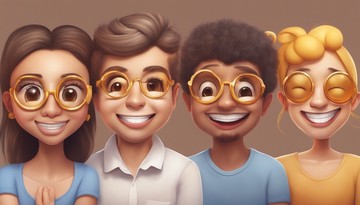} & \includegraphics[width=0.486\textwidth,height=0.2777\textwidth]{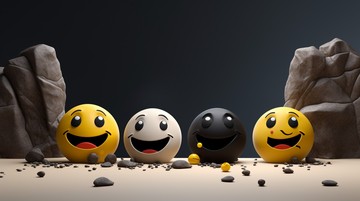} & \includegraphics[width=0.486\textwidth,height=0.2777\textwidth]{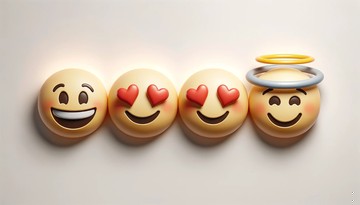} \\
\includegraphics[width=0.486\textwidth,height=0.2777\textwidth]{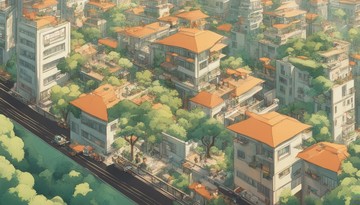} & \includegraphics[width=0.486\textwidth,height=0.2777\textwidth]{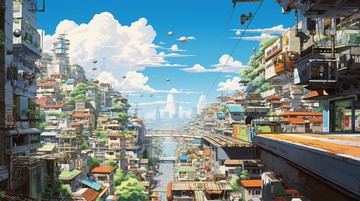} & \includegraphics[width=0.486\textwidth,height=0.2777\textwidth]{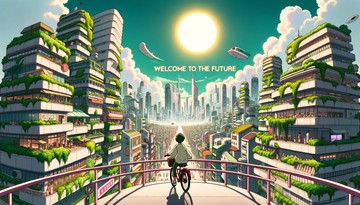} \\
\includegraphics[width=0.486\textwidth,height=0.2777\textwidth]{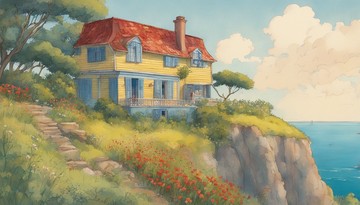} & \includegraphics[width=0.486\textwidth,height=0.2777\textwidth]{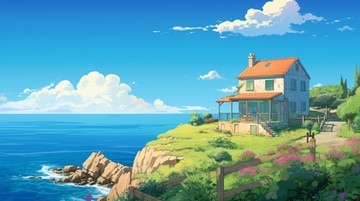} & \includegraphics[width=0.486\textwidth,height=0.2777\textwidth]{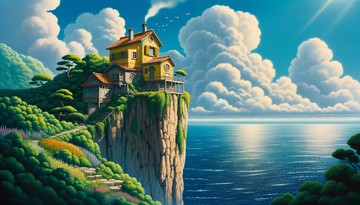} \\
\includegraphics[width=0.486\textwidth,height=0.2777\textwidth]{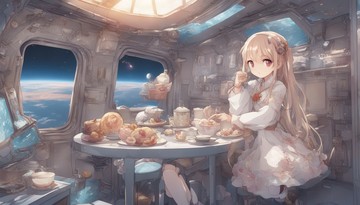} & \includegraphics[width=0.486\textwidth,height=0.2777\textwidth]{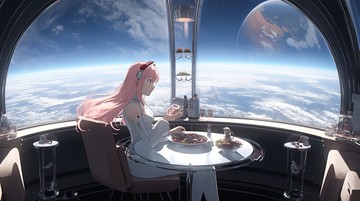} & \includegraphics[width=0.486\textwidth,height=0.2777\textwidth]{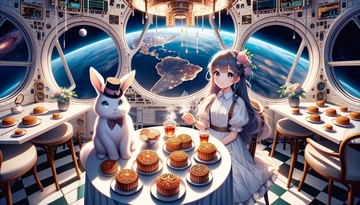} \\
\includegraphics[width=0.486\textwidth,height=0.2777\textwidth]{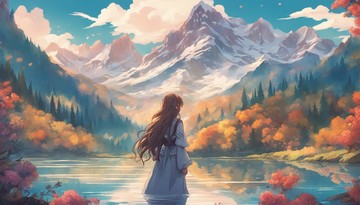} & \includegraphics[width=0.486\textwidth,height=0.2777\textwidth]{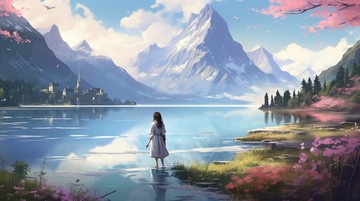} & \includegraphics[width=0.486\textwidth,height=0.2777\textwidth]{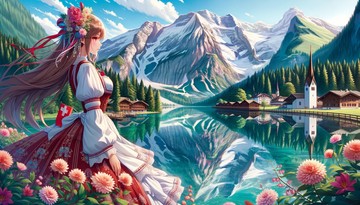}
\end{tabular}
\end{adjustbox}
\caption[cartoon comparisons]{Qualitative comparisons among SDXL, Midjourney v5.2, and \modelname. Check Figure~\ref{fig:scenario_cartoon_gaming_comic_emoji_2} for complete prompts.
}
\label{fig:compare_cartoon_32}
\end{figure}

\begin{figure}
\centering
\vspace{-20mm}
\begin{adjustbox}{center}
\small
\begin{tabular}{x{0.496\textwidth}@{}x{0.496\textwidth}@{}x{0.496\textwidth}}
Ideogram & Firefly 2 & DALL-E 3 \\
\includegraphics[width=0.486\textwidth,height=0.2777\textwidth]{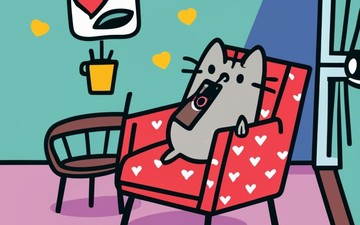} & \includegraphics[width=0.486\textwidth,height=0.2777\textwidth]{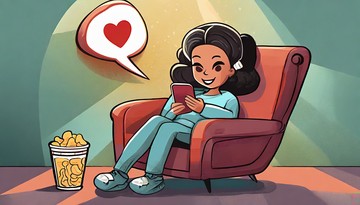} & \includegraphics[width=0.486\textwidth,height=0.2777\textwidth]{raw_images/DE3/cartoon_0_0.jpg} \\
\includegraphics[width=0.486\textwidth,height=0.2777\textwidth]{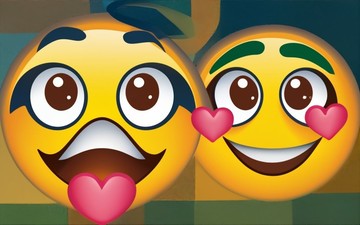} & \includegraphics[width=0.486\textwidth,height=0.2777\textwidth]{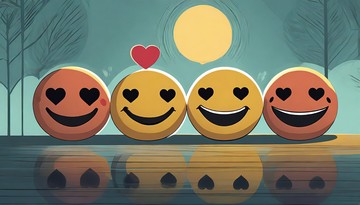} & \includegraphics[width=0.486\textwidth,height=0.2777\textwidth]{raw_images/DE3/cartoon_1_0.jpg} \\
\includegraphics[width=0.486\textwidth,height=0.2777\textwidth]{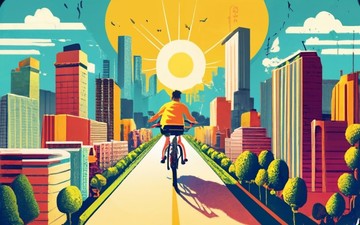} & \includegraphics[width=0.486\textwidth,height=0.2777\textwidth]{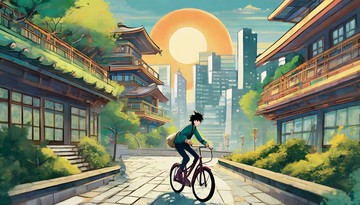} & \includegraphics[width=0.486\textwidth,height=0.2777\textwidth]{raw_images/DE3/cartoon_2_0.jpg} \\
\includegraphics[width=0.486\textwidth,height=0.2777\textwidth]{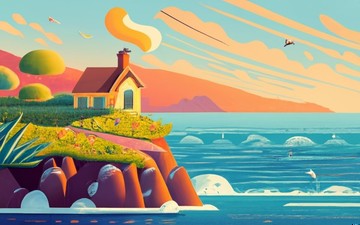} & \includegraphics[width=0.486\textwidth,height=0.2777\textwidth]{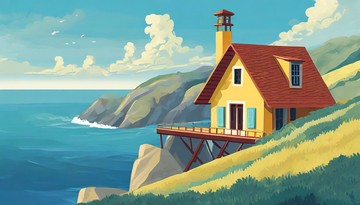} & \includegraphics[width=0.486\textwidth,height=0.2777\textwidth]{raw_images/DE3/cartoon_3_0.jpg} \\
\includegraphics[width=0.486\textwidth,height=0.2777\textwidth]{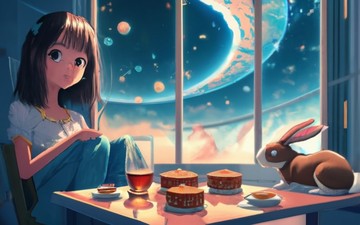} & \includegraphics[width=0.486\textwidth,height=0.2777\textwidth]{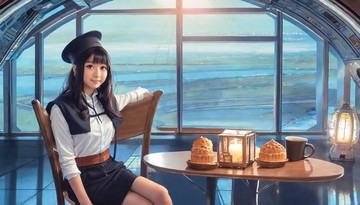} & \includegraphics[width=0.486\textwidth,height=0.2777\textwidth]{raw_images/DE3/cartoon_4_0.jpg} \\
\includegraphics[width=0.486\textwidth,height=0.2777\textwidth]{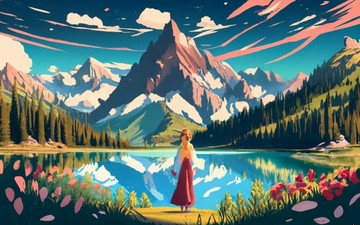} & \includegraphics[width=0.486\textwidth,height=0.2777\textwidth]{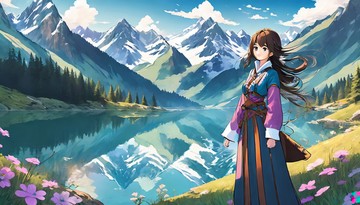} & \includegraphics[width=0.486\textwidth,height=0.2777\textwidth]{raw_images/DE3/cartoon_5_0.jpg}
\end{tabular}
\end{adjustbox}
\caption[cartoon comparisons]{Qualitative comparisons among Ideogram, Firefly 2, and \modelname. Check Figure~\ref{fig:scenario_cartoon_gaming_comic_emoji_2} for complete prompts.
}
\label{fig:flyfire_cartoon_32}
\end{figure}

\begin{figure}
\centering
\vspace{-20mm}
\begin{adjustbox}{center}
\small
\begin{tabular}{x{0.496\textwidth}@{}x{0.496\textwidth}@{}x{0.496\textwidth}}
SDXL & Midjourney & DALL-E 3 \\
\includegraphics[width=0.486\textwidth,height=0.2777\textwidth]{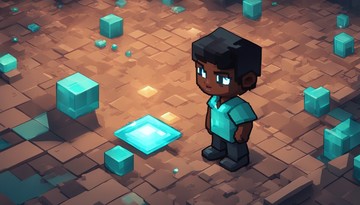} & \includegraphics[width=0.486\textwidth,height=0.2777\textwidth]{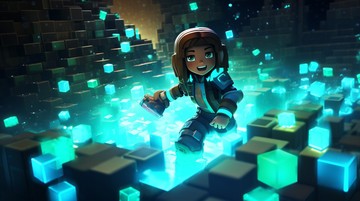} & \includegraphics[width=0.486\textwidth,height=0.2777\textwidth]{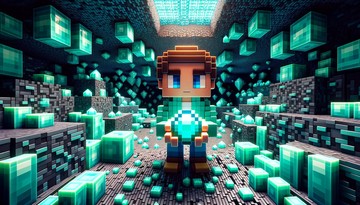} \\
\includegraphics[width=0.486\textwidth,height=0.2777\textwidth]{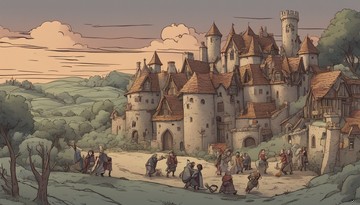} & \includegraphics[width=0.486\textwidth,height=0.2777\textwidth]{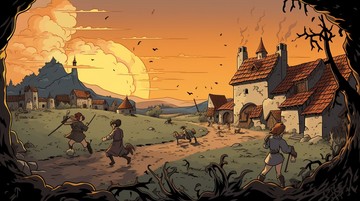} & \includegraphics[width=0.486\textwidth,height=0.2777\textwidth]{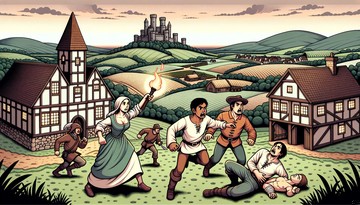} \\
\includegraphics[width=0.486\textwidth,height=0.2777\textwidth]{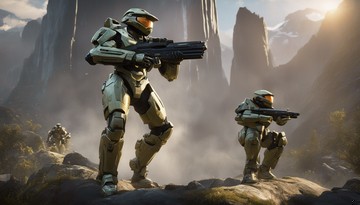} & \includegraphics[width=0.486\textwidth,height=0.2777\textwidth]{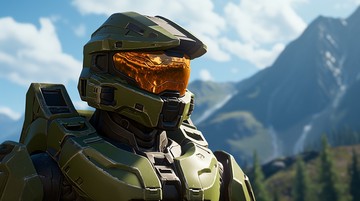} & \includegraphics[width=0.486\textwidth,height=0.2777\textwidth]{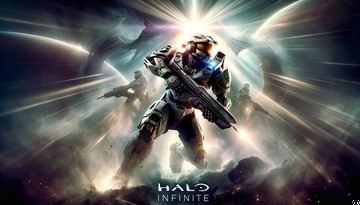} \\
\includegraphics[width=0.486\textwidth,height=0.2777\textwidth]{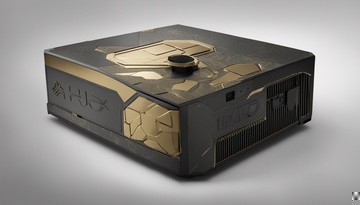} & \includegraphics[width=0.486\textwidth,height=0.2777\textwidth]{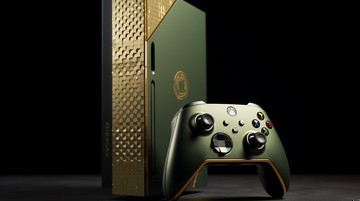} & \includegraphics[width=0.486\textwidth,height=0.2777\textwidth]{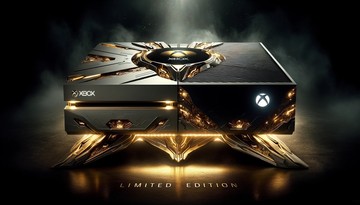} \\
\includegraphics[width=0.486\textwidth,height=0.2777\textwidth]{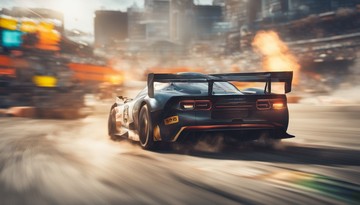} & \includegraphics[width=0.486\textwidth,height=0.2777\textwidth]{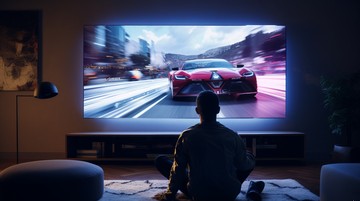} & \includegraphics[width=0.486\textwidth,height=0.2777\textwidth]{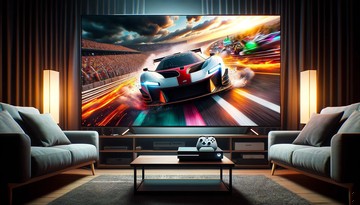} \\
\includegraphics[width=0.486\textwidth,height=0.2777\textwidth]{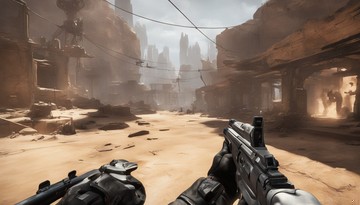} & \includegraphics[width=0.486\textwidth,height=0.2777\textwidth]{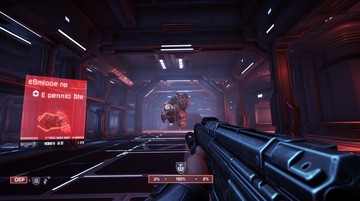} & \includegraphics[width=0.486\textwidth,height=0.2777\textwidth]{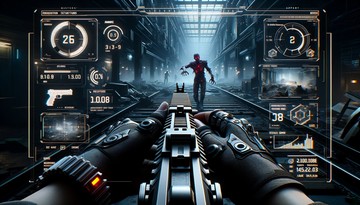}
\end{tabular}
\end{adjustbox}
\caption[game design comparisons]{Qualitative comparisons among SDXL, Midjourney v5.2, and \modelname. Check Figure~\ref{fig:scenario_cartoon_gaming_comic_emoji_1} for complete prompts.
}
\label{fig:compare_game_33}
\end{figure}

\begin{figure}
\centering
\vspace{-20mm}
\begin{adjustbox}{center}
\small
\begin{tabular}{x{0.496\textwidth}@{}x{0.496\textwidth}@{}x{0.496\textwidth}}
Ideogram & Firefly 2 & DALL-E 3 \\
\includegraphics[width=0.486\textwidth,height=0.2777\textwidth]{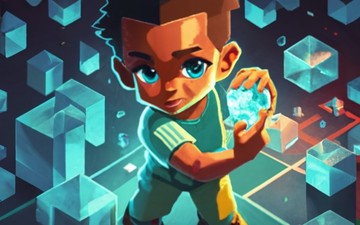} & \includegraphics[width=0.486\textwidth,height=0.2777\textwidth]{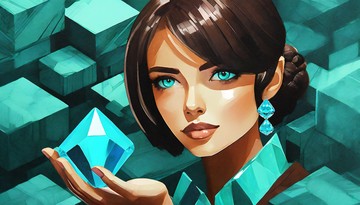} & \includegraphics[width=0.486\textwidth,height=0.2777\textwidth]{raw_images/DE3/game_0_0.jpg} \\
\includegraphics[width=0.486\textwidth,height=0.2777\textwidth]{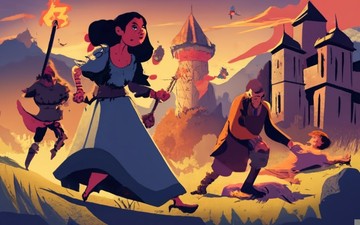} & \includegraphics[width=0.486\textwidth,height=0.2777\textwidth]{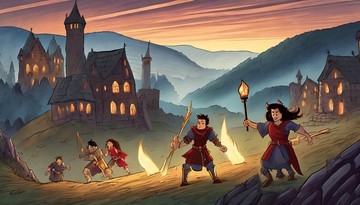} & \includegraphics[width=0.486\textwidth,height=0.2777\textwidth]{raw_images/DE3/game_1_0.jpg} \\
\includegraphics[width=0.486\textwidth,height=0.2777\textwidth]{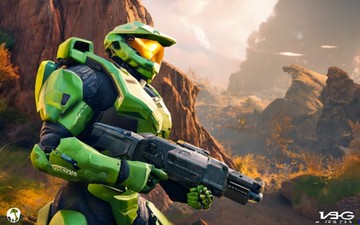} & \includegraphics[width=0.486\textwidth,height=0.2777\textwidth]{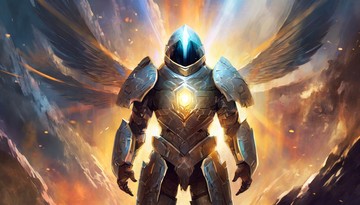} & \includegraphics[width=0.486\textwidth,height=0.2777\textwidth]{raw_images/DE3/game_2_0.jpg} \\
\includegraphics[width=0.486\textwidth,height=0.2777\textwidth]{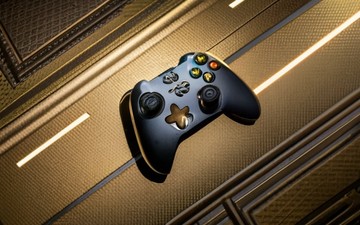} & \includegraphics[width=0.486\textwidth,height=0.2777\textwidth]{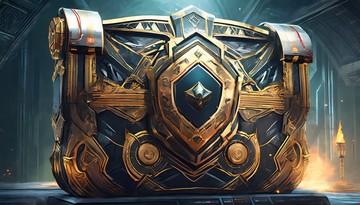} & \includegraphics[width=0.486\textwidth,height=0.2777\textwidth]{raw_images/DE3/game_3_0.jpg} \\
\includegraphics[width=0.486\textwidth,height=0.2777\textwidth]{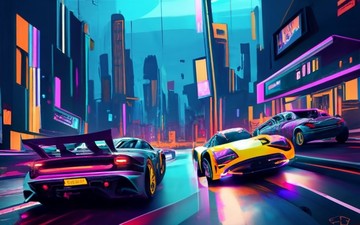} & \includegraphics[width=0.486\textwidth,height=0.2777\textwidth]{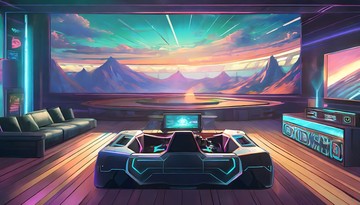} & \includegraphics[width=0.486\textwidth,height=0.2777\textwidth]{raw_images/DE3/game_4_0.jpg} \\
\includegraphics[width=0.486\textwidth,height=0.2777\textwidth]{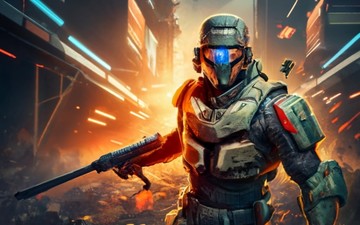} & \includegraphics[width=0.486\textwidth,height=0.2777\textwidth]{raw_images/emptyimage.png} & \includegraphics[width=0.486\textwidth,height=0.2777\textwidth]{raw_images/DE3/game_5_0.jpg}
\end{tabular}
\end{adjustbox}
\caption[game design comparisons]{Qualitative comparisons among Ideogram, Firefly 2, and \modelname. Check Figure~\ref{fig:scenario_cartoon_gaming_comic_emoji_1} for complete prompts. Firefly refuses to generate images for certain prompts, and we use a special ``Content Filtered'' as a placeholder image.
}
\label{fig:flyfire_game_33}
\end{figure}

\begin{figure}
\centering
\vspace{-20mm}
\begin{adjustbox}{center}
\small
\begin{tabular}{x{0.496\textwidth}@{}x{0.496\textwidth}@{}x{0.496\textwidth}}
SDXL & Midjourney & DALL-E 3 \\
\includegraphics[width=0.486\textwidth,height=0.2777\textwidth]{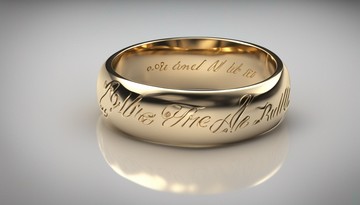} & \includegraphics[width=0.486\textwidth,height=0.2777\textwidth]{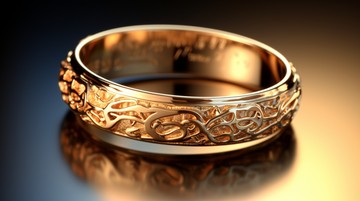} & \includegraphics[width=0.486\textwidth,height=0.2777\textwidth]{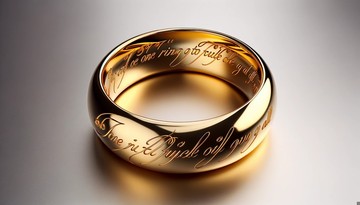} \\
\includegraphics[width=0.486\textwidth,height=0.2777\textwidth]{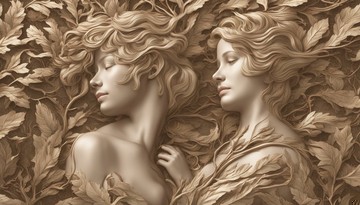} & \includegraphics[width=0.486\textwidth,height=0.2777\textwidth]{raw_images/MJ/product_1_0.jpg} & \includegraphics[width=0.486\textwidth,height=0.2777\textwidth]{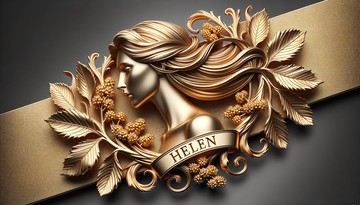} \\
\includegraphics[width=0.486\textwidth,height=0.2777\textwidth]{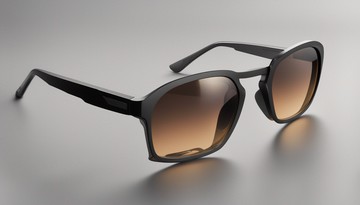} & \includegraphics[width=0.486\textwidth,height=0.2777\textwidth]{raw_images/MJ/product_2_0.jpg} & \includegraphics[width=0.486\textwidth,height=0.2777\textwidth]{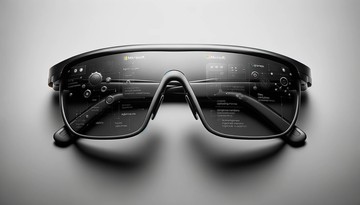} \\
\includegraphics[width=0.486\textwidth,height=0.2777\textwidth]{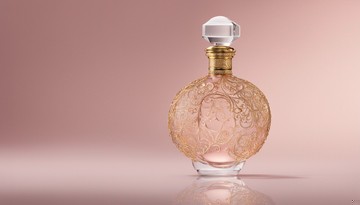} & \includegraphics[width=0.486\textwidth,height=0.2777\textwidth]{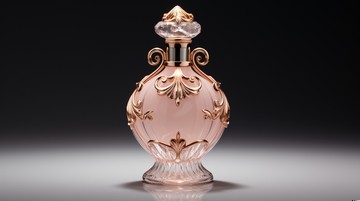} & \includegraphics[width=0.486\textwidth,height=0.2777\textwidth]{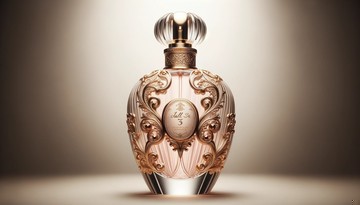} \\
\includegraphics[width=0.486\textwidth,height=0.2777\textwidth]{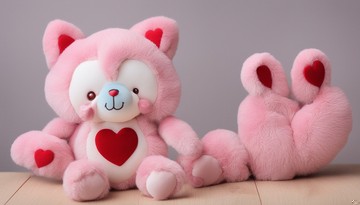} & \includegraphics[width=0.486\textwidth,height=0.2777\textwidth]{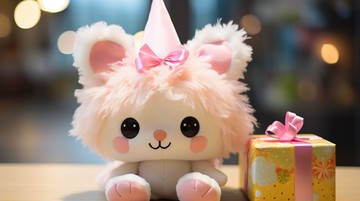} & \includegraphics[width=0.486\textwidth,height=0.2777\textwidth]{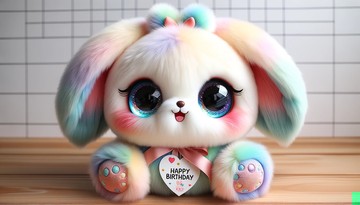} \\
\includegraphics[width=0.486\textwidth,height=0.2777\textwidth]{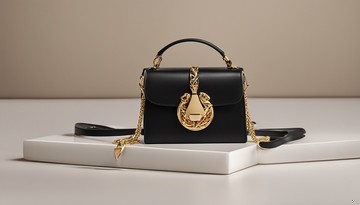} & \includegraphics[width=0.486\textwidth,height=0.2777\textwidth]{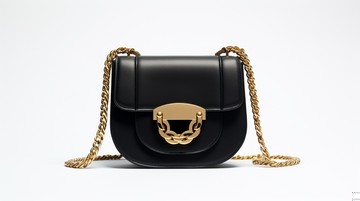} & \includegraphics[width=0.486\textwidth,height=0.2777\textwidth]{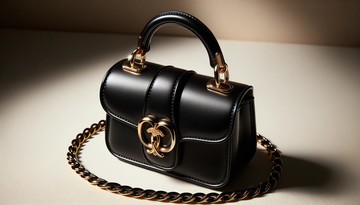}
\end{tabular}
\end{adjustbox}
\caption[product design comparisons]{Qualitative comparisons among SDXL, Midjourney v5.2, and \modelname. Check Figure~\ref{fig:scenario_product_1} for complete prompts.
}
\label{fig:compare_product1_34}
\end{figure}

\begin{figure}
\centering
\vspace{-20mm}
\begin{adjustbox}{center}
\small
\begin{tabular}{x{0.496\textwidth}@{}x{0.496\textwidth}@{}x{0.496\textwidth}}
Ideogram & Firefly 2 & DALL-E 3 \\
\includegraphics[width=0.486\textwidth,height=0.2777\textwidth]{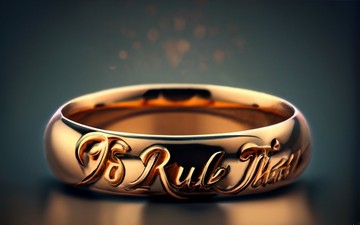} & \includegraphics[width=0.486\textwidth,height=0.2777\textwidth]{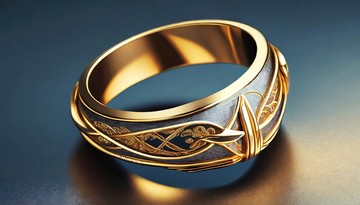} & \includegraphics[width=0.486\textwidth,height=0.2777\textwidth]{raw_images/DE3/product_0_0.jpg} \\
\includegraphics[width=0.486\textwidth,height=0.2777\textwidth]{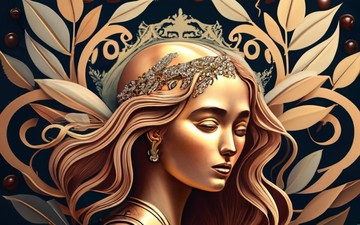} & \includegraphics[width=0.486\textwidth,height=0.2777\textwidth]{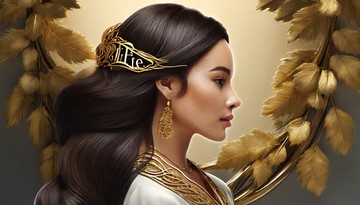} & \includegraphics[width=0.486\textwidth,height=0.2777\textwidth]{raw_images/DE3/product_1_0.jpg} \\
\includegraphics[width=0.486\textwidth,height=0.2777\textwidth]{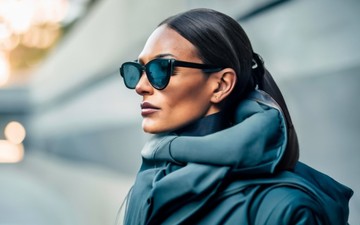} & \includegraphics[width=0.486\textwidth,height=0.2777\textwidth]{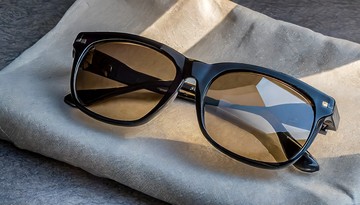} & \includegraphics[width=0.486\textwidth,height=0.2777\textwidth]{raw_images/DE3/product_2_0.jpg} \\
\includegraphics[width=0.486\textwidth,height=0.2777\textwidth]{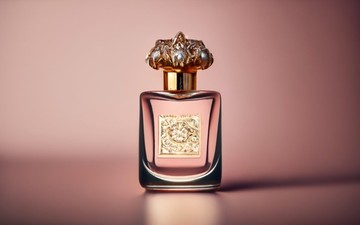} & \includegraphics[width=0.486\textwidth,height=0.2777\textwidth]{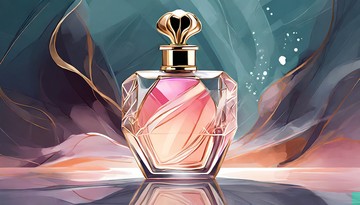} & \includegraphics[width=0.486\textwidth,height=0.2777\textwidth]{raw_images/DE3/product_3_0.jpg} \\
\includegraphics[width=0.486\textwidth,height=0.2777\textwidth]{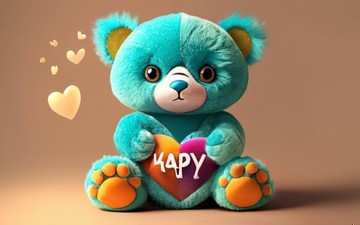} & \includegraphics[width=0.486\textwidth,height=0.2777\textwidth]{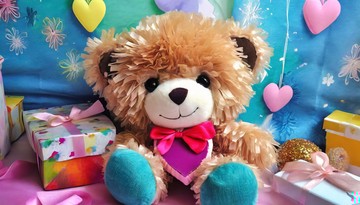} & \includegraphics[width=0.486\textwidth,height=0.2777\textwidth]{raw_images/DE3/product_4_0.jpg} \\
\includegraphics[width=0.486\textwidth,height=0.2777\textwidth]{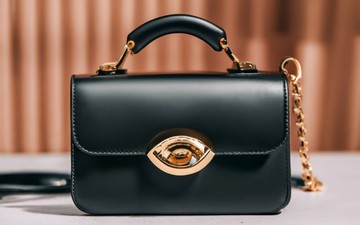} & \includegraphics[width=0.486\textwidth,height=0.2777\textwidth]{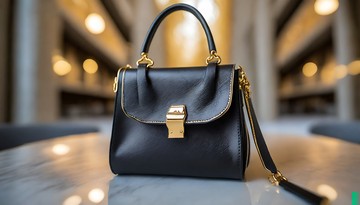} & \includegraphics[width=0.486\textwidth,height=0.2777\textwidth]{raw_images/DE3/product_5_0.jpg}
\end{tabular}
\end{adjustbox}
\caption[product design comparisons]{Qualitative comparisons among Ideogram, Firefly 2, and \modelname. Check Figure~\ref{fig:scenario_product_1} for complete prompts.
}
\label{fig:flyfire_product1_34}
\end{figure}

\begin{figure}
\centering
\vspace{-20mm}
\begin{adjustbox}{center}
\small
\begin{tabular}{x{0.496\textwidth}@{}x{0.496\textwidth}@{}x{0.496\textwidth}}
SDXL & Midjourney & DALL-E 3 \\
\includegraphics[width=0.486\textwidth,height=0.2777\textwidth]{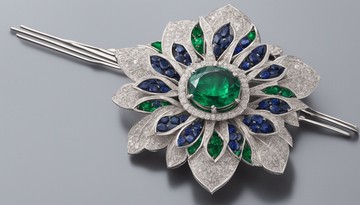} & \includegraphics[width=0.486\textwidth,height=0.2777\textwidth]{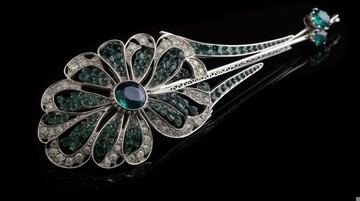} & \includegraphics[width=0.486\textwidth,height=0.2777\textwidth]{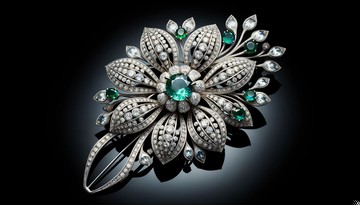} \\
\includegraphics[width=0.486\textwidth,height=0.2777\textwidth]{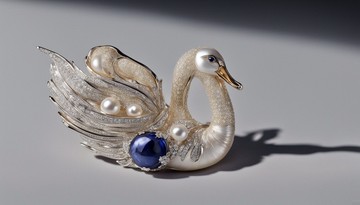} & \includegraphics[width=0.486\textwidth,height=0.2777\textwidth]{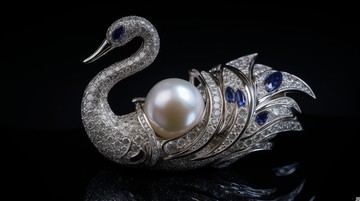} & \includegraphics[width=0.486\textwidth,height=0.2777\textwidth]{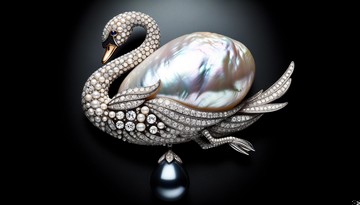} \\
\includegraphics[width=0.486\textwidth,height=0.2777\textwidth]{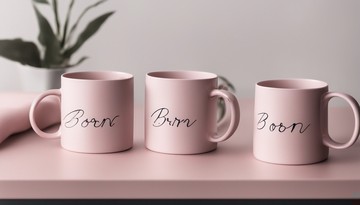} & \includegraphics[width=0.486\textwidth,height=0.2777\textwidth]{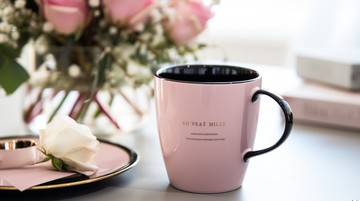} & \includegraphics[width=0.486\textwidth,height=0.2777\textwidth]{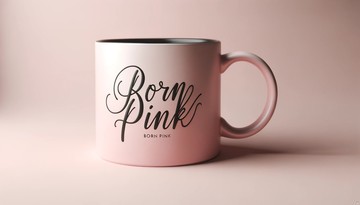} \\
\includegraphics[width=0.486\textwidth,height=0.2777\textwidth]{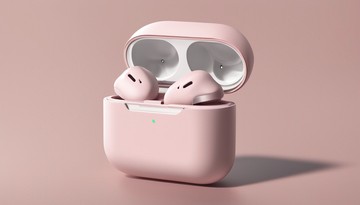} & \includegraphics[width=0.486\textwidth,height=0.2777\textwidth]{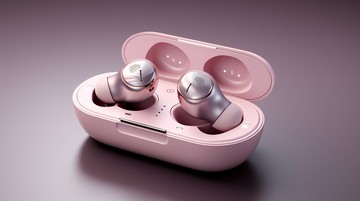} & \includegraphics[width=0.486\textwidth,height=0.2777\textwidth]{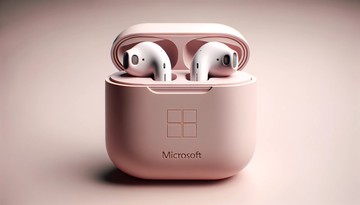} \\
\includegraphics[width=0.486\textwidth,height=0.2777\textwidth]{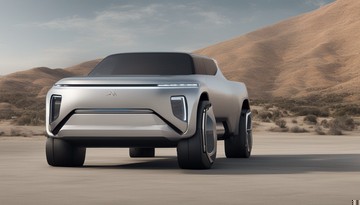} & \includegraphics[width=0.486\textwidth,height=0.2777\textwidth]{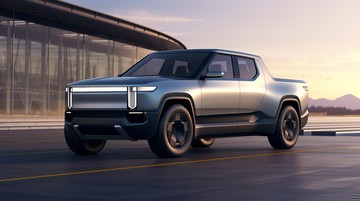} & \includegraphics[width=0.486\textwidth,height=0.2777\textwidth]{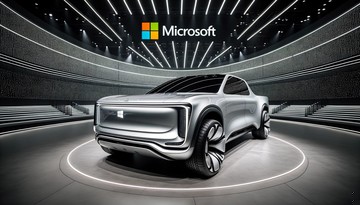} \\
\includegraphics[width=0.486\textwidth,height=0.2777\textwidth]{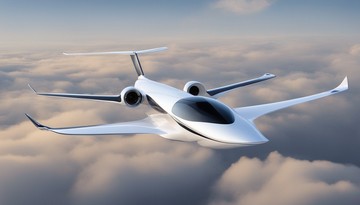} & \includegraphics[width=0.486\textwidth,height=0.2777\textwidth]{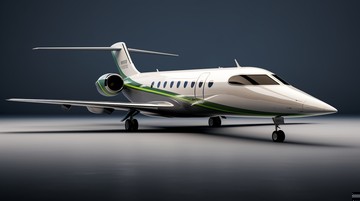} & \includegraphics[width=0.486\textwidth,height=0.2777\textwidth]{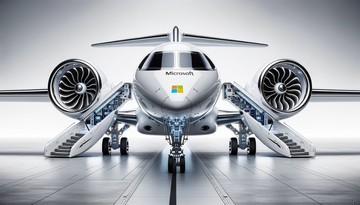}
\end{tabular}
\end{adjustbox}
\caption[product design comparisons]{Qualitative comparisons among SDXL, Midjourney v5.2, and \modelname. Check Figure~\ref{fig:scenario_product_2} for complete prompts.
}
\label{fig:compare_product1_35}
\end{figure}

\begin{figure}
\centering
\vspace{-20mm}
\begin{adjustbox}{center}
\small
\begin{tabular}{x{0.496\textwidth}@{}x{0.496\textwidth}@{}x{0.496\textwidth}}
Ideogram & Firefly 2 & DALL-E 3 \\
\includegraphics[width=0.486\textwidth,height=0.2777\textwidth]{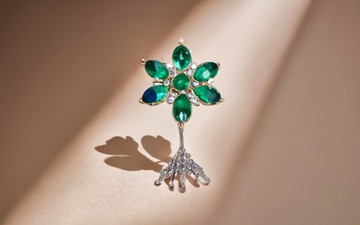} & \includegraphics[width=0.486\textwidth,height=0.2777\textwidth]{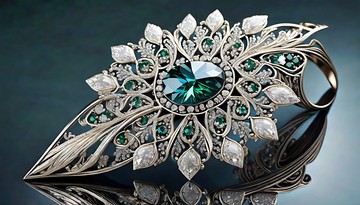} & \includegraphics[width=0.486\textwidth,height=0.2777\textwidth]{raw_images/DE3/product_6_0.jpg} \\
\includegraphics[width=0.486\textwidth,height=0.2777\textwidth]{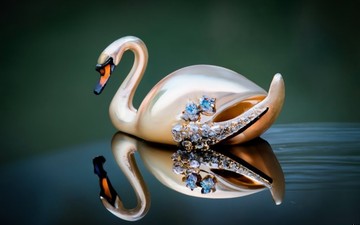} & \includegraphics[width=0.486\textwidth,height=0.2777\textwidth]{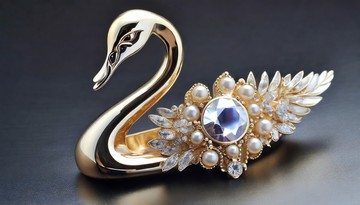} & \includegraphics[width=0.486\textwidth,height=0.2777\textwidth]{raw_images/DE3/product_7_0.jpg} \\
\includegraphics[width=0.486\textwidth,height=0.2777\textwidth]{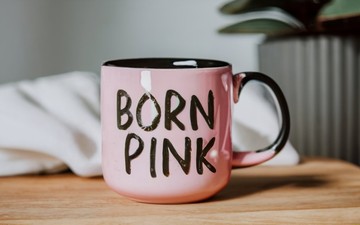} & \includegraphics[width=0.486\textwidth,height=0.2777\textwidth]{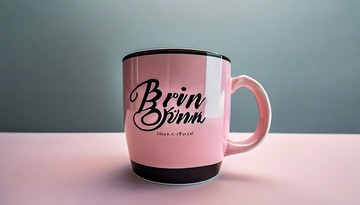} & \includegraphics[width=0.486\textwidth,height=0.2777\textwidth]{raw_images/DE3/product_8_0.jpg} \\
\includegraphics[width=0.486\textwidth,height=0.2777\textwidth]{raw_images/Ideogram/product_9_0.jpeg} & \includegraphics[width=0.486\textwidth,height=0.2777\textwidth]{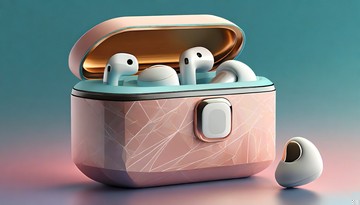} & \includegraphics[width=0.486\textwidth,height=0.2777\textwidth]{raw_images/DE3/product_9_0.jpg} \\
\includegraphics[width=0.486\textwidth,height=0.2777\textwidth]{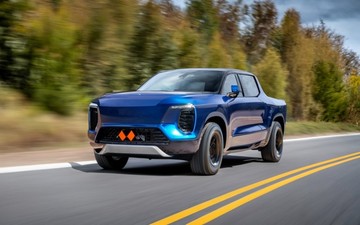} & \includegraphics[width=0.486\textwidth,height=0.2777\textwidth]{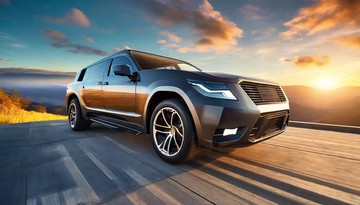} & \includegraphics[width=0.486\textwidth,height=0.2777\textwidth]{raw_images/DE3/product_10_0.jpg} \\
\includegraphics[width=0.486\textwidth,height=0.2777\textwidth]{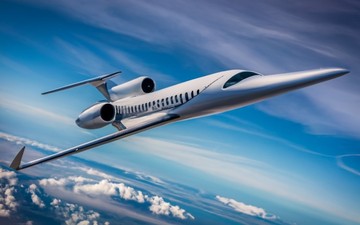} & \includegraphics[width=0.486\textwidth,height=0.2777\textwidth]{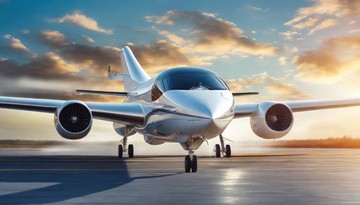} & \includegraphics[width=0.486\textwidth,height=0.2777\textwidth]{raw_images/DE3/product_11_0.jpg}
\end{tabular}
\end{adjustbox}
\caption[product design comparisons]{Qualitative comparisons among Ideogram, Firefly 2, and \modelname. Check Figure~\ref{fig:scenario_product_2} for complete prompts.
}
\label{fig:flyfire_product2_35}
\end{figure}

\begin{figure}
\centering
\vspace{-20mm}
\begin{adjustbox}{center}
\small
\begin{tabular}{x{0.496\textwidth}@{}x{0.496\textwidth}@{}x{0.496\textwidth}}
SDXL & Midjourney & DALL-E 3 \\
\includegraphics[width=0.486\textwidth,height=0.2777\textwidth]{raw_images/SDXL/fashion_0_0.jpg} & \includegraphics[width=0.486\textwidth,height=0.2777\textwidth]{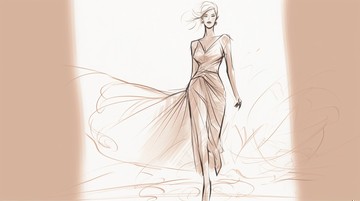} & \includegraphics[width=0.486\textwidth,height=0.2777\textwidth]{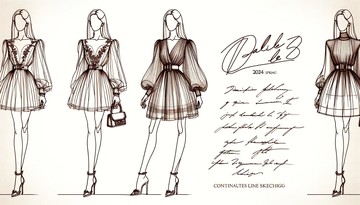} \\
\includegraphics[width=0.486\textwidth,height=0.2777\textwidth]{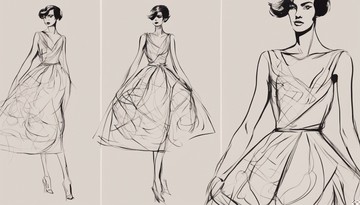} & \includegraphics[width=0.486\textwidth,height=0.2777\textwidth]{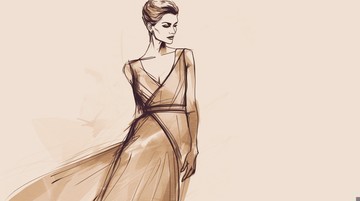} & \includegraphics[width=0.486\textwidth,height=0.2777\textwidth]{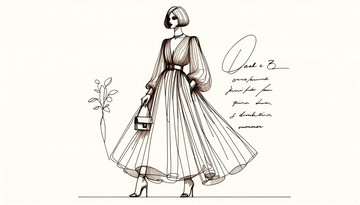} \\
\includegraphics[width=0.486\textwidth,height=0.2777\textwidth]{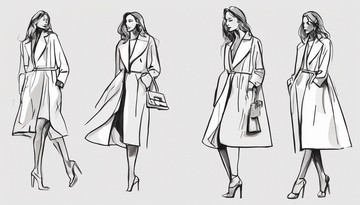} & \includegraphics[width=0.486\textwidth,height=0.2777\textwidth]{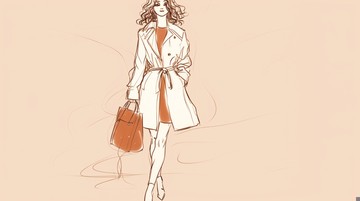} & \includegraphics[width=0.486\textwidth,height=0.2777\textwidth]{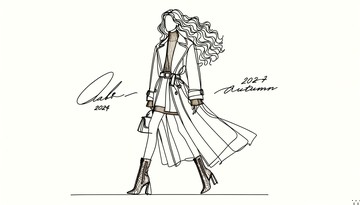} \\
\includegraphics[width=0.486\textwidth,height=0.2777\textwidth]{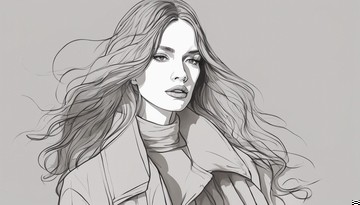} & \includegraphics[width=0.486\textwidth,height=0.2777\textwidth]{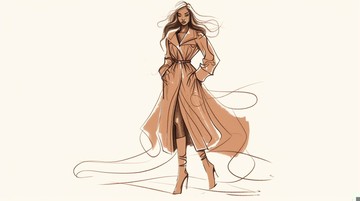} & \includegraphics[width=0.486\textwidth,height=0.2777\textwidth]{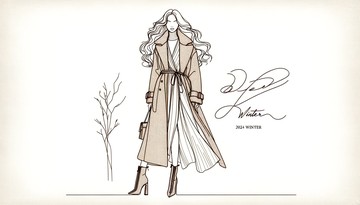} \\
\includegraphics[width=0.486\textwidth,height=0.2777\textwidth]{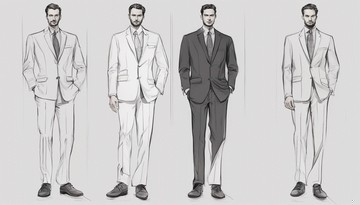} & \includegraphics[width=0.486\textwidth,height=0.2777\textwidth]{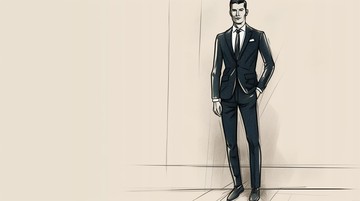} & \includegraphics[width=0.486\textwidth,height=0.2777\textwidth]{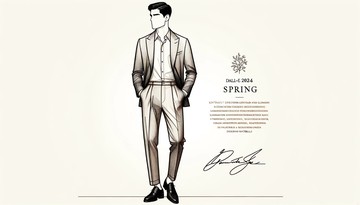} \\
\includegraphics[width=0.486\textwidth,height=0.2777\textwidth]{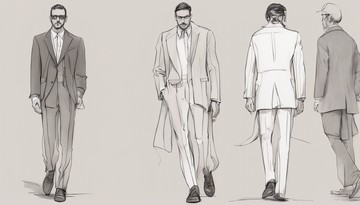} & \includegraphics[width=0.486\textwidth,height=0.2777\textwidth]{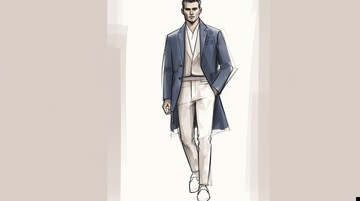} & \includegraphics[width=0.486\textwidth,height=0.2777\textwidth]{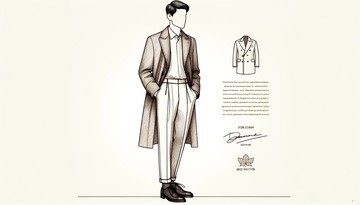}
\end{tabular}
\end{adjustbox}
\caption[fashion design comparisons]{Qualitative comparisons among SDXL, Midjourney v5.2, and \modelname. Check Figure~\ref{fig:Scenario_Fashion_Design_1} for complete prompts.
}
\label{fig:compare_fashion_36}
\end{figure}

\begin{figure}
\centering
\vspace{-20mm}
\begin{adjustbox}{center}
\small
\begin{tabular}{x{0.496\textwidth}@{}x{0.496\textwidth}@{}x{0.496\textwidth}}
Ideogram & Firefly 2 & DALL-E 3 \\
\includegraphics[width=0.486\textwidth,height=0.2777\textwidth]{raw_images/Ideogram/fashion_0_0.jpeg} & \includegraphics[width=0.486\textwidth,height=0.2777\textwidth]{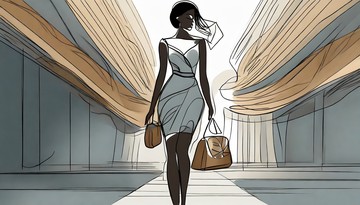} & \includegraphics[width=0.486\textwidth,height=0.2777\textwidth]{raw_images/DE3/fashion_0_0.jpg} \\
\includegraphics[width=0.486\textwidth,height=0.2777\textwidth]{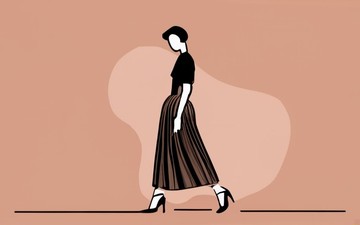} & \includegraphics[width=0.486\textwidth,height=0.2777\textwidth]{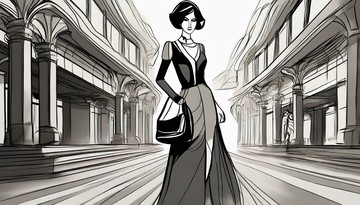} & \includegraphics[width=0.486\textwidth,height=0.2777\textwidth]{raw_images/DE3/fashion_1_0.jpg} \\
\includegraphics[width=0.486\textwidth,height=0.2777\textwidth]{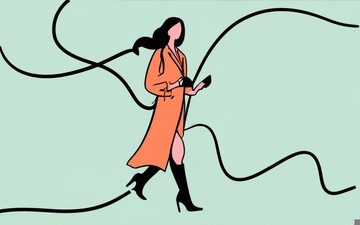} & \includegraphics[width=0.486\textwidth,height=0.2777\textwidth]{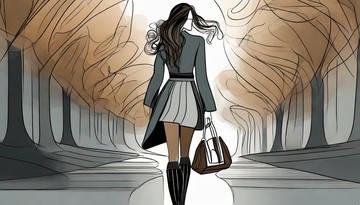} & \includegraphics[width=0.486\textwidth,height=0.2777\textwidth]{raw_images/DE3/fashion_2_0.jpg} \\
\includegraphics[width=0.486\textwidth,height=0.2777\textwidth]{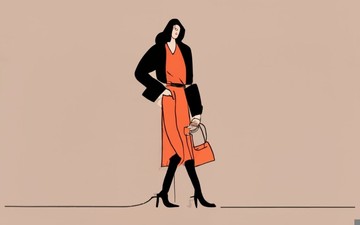} & \includegraphics[width=0.486\textwidth,height=0.2777\textwidth]{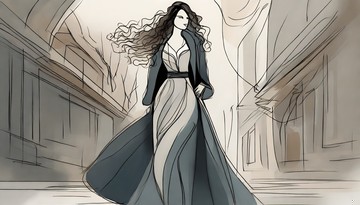} & \includegraphics[width=0.486\textwidth,height=0.2777\textwidth]{raw_images/DE3/fashion_3_0.jpg} \\
\includegraphics[width=0.486\textwidth,height=0.2777\textwidth]{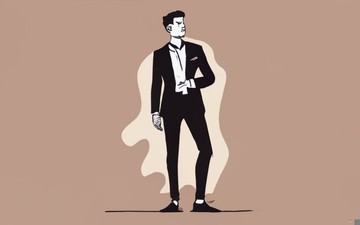} & \includegraphics[width=0.486\textwidth,height=0.2777\textwidth]{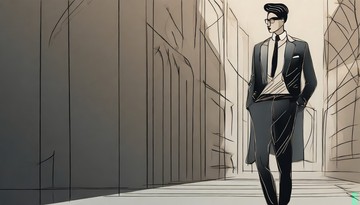} & \includegraphics[width=0.486\textwidth,height=0.2777\textwidth]{raw_images/DE3/fashion_4_0.jpg} \\
\includegraphics[width=0.486\textwidth,height=0.2777\textwidth]{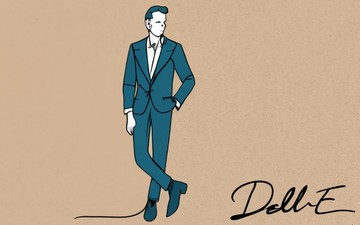} & \includegraphics[width=0.486\textwidth,height=0.2777\textwidth]{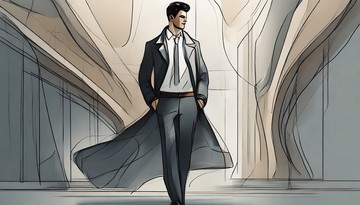} & \includegraphics[width=0.486\textwidth,height=0.2777\textwidth]{raw_images/DE3/fashion_5_0.jpg}
\end{tabular}
\end{adjustbox}
\caption[fashion design comparisons]{Qualitative comparisons among Ideogram, Firefly 2, and \modelname. Check Figure~\ref{fig:Scenario_Fashion_Design_1} for complete prompts.
}
\label{fig:flyfire_fashion_36}
\end{figure}

\begin{figure}
\centering
\vspace{-20mm}
\begin{adjustbox}{center}
\small
%
\end{adjustbox}
\centerline{\includegraphics[width=1.5\textwidth]{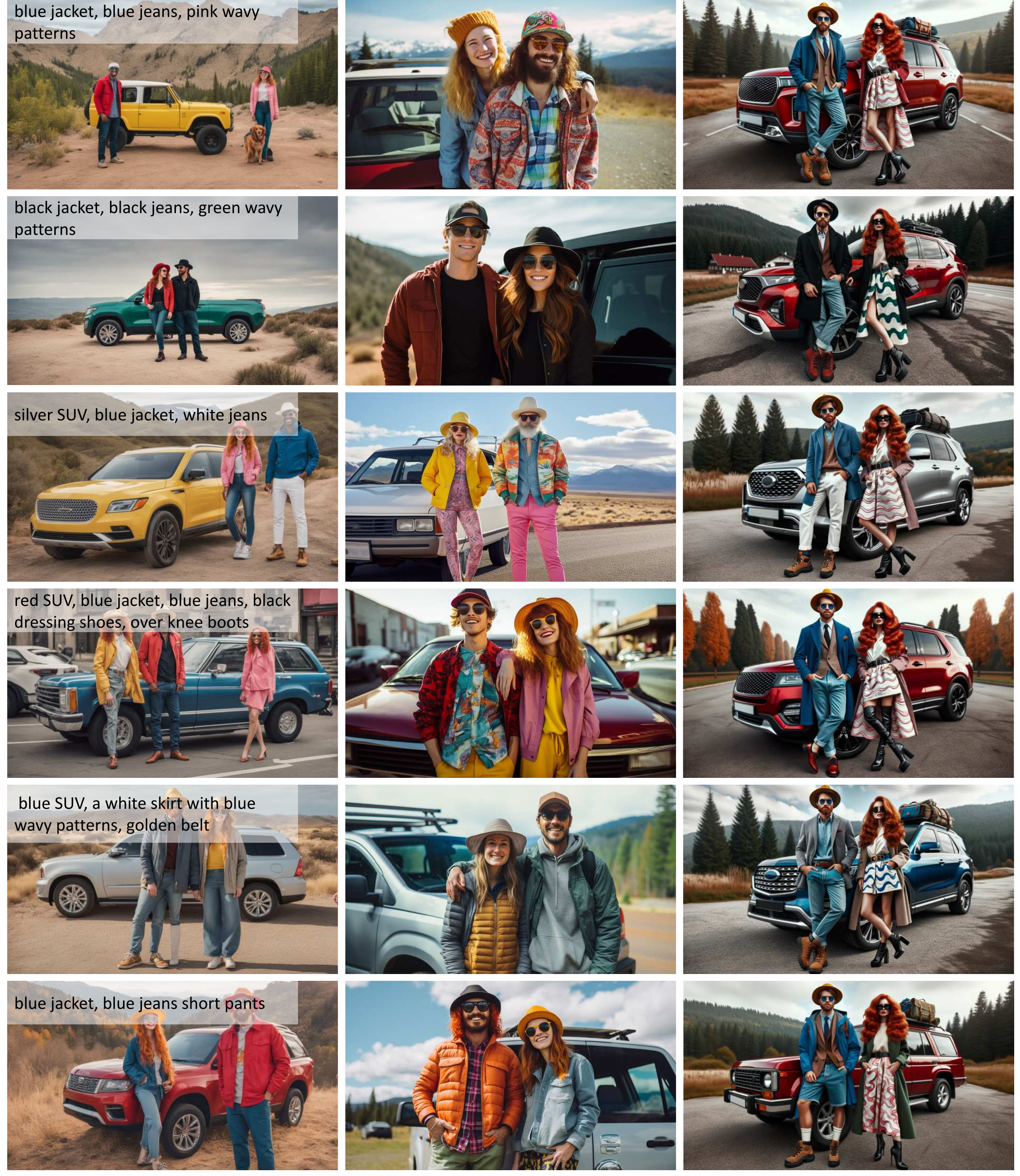}}
\caption[camera settings comparisons]{Qualitative comparisons among SDXL, Midjourney v5.2, and \modelname. Check Figure~\ref{fig:Capability_change_attribute_1}, Figure~\ref{fig:camera_setting_1} for complete prompts.
}
\label{fig:compare_attribute_37}
\end{figure}

\begin{figure}
\centering
\vspace{-20mm}
\begin{adjustbox}{center}
\small
%
\end{adjustbox}
\centerline{\includegraphics[width=1.5\textwidth]{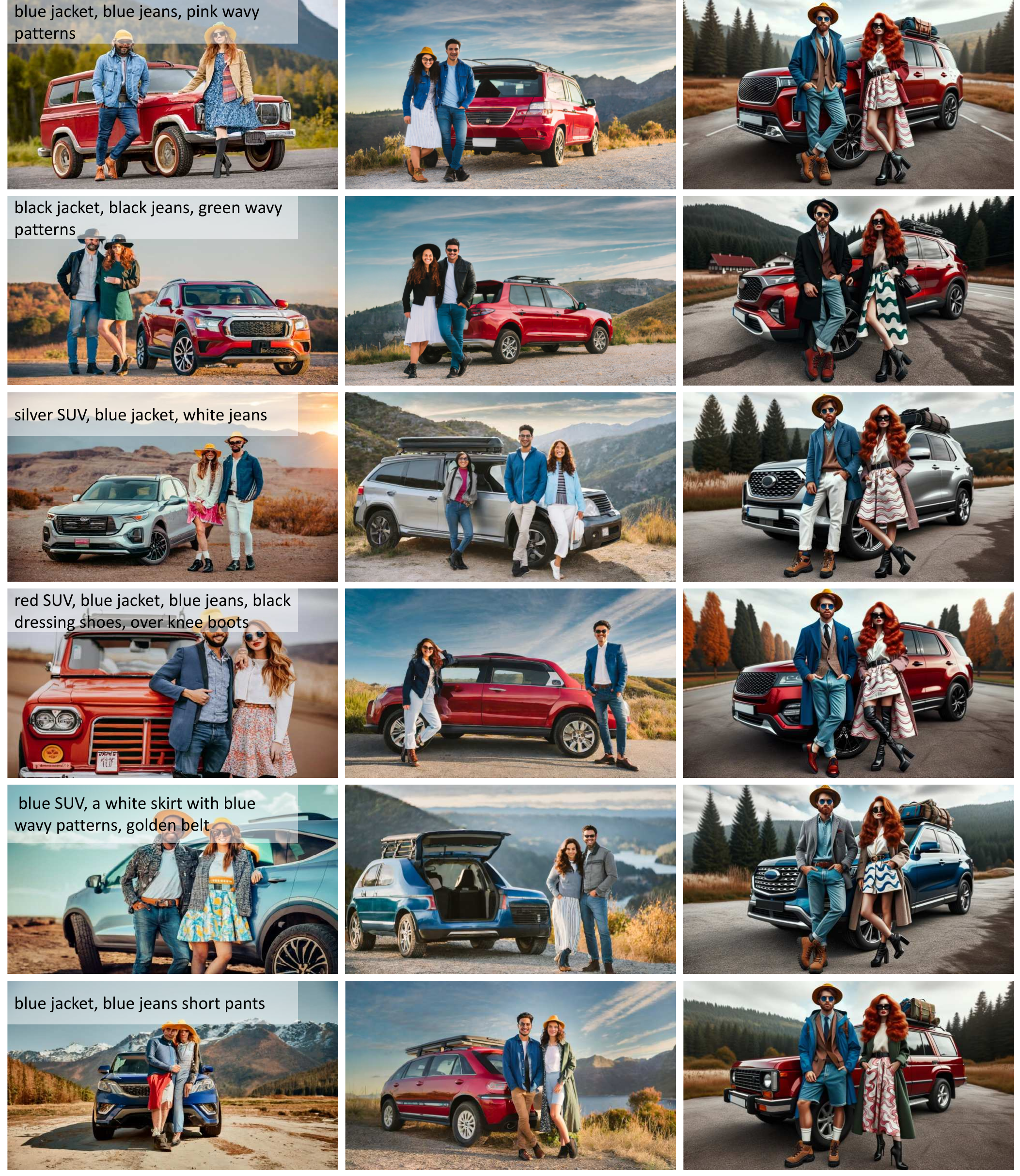}}
\caption[color comparisons]{Qualitative comparisons among Ideogram, Firefly 2, and \modelname. Check Figure~\ref{fig:Capability_change_attribute_1} for complete prompts.
}
\label{fig:firefly_attribute_37}
\end{figure}

Figures~\ref{fig:compare_text_5_6}-\ref{fig:firefly_knolling_40} visualize the images in the \benchname gallery, containing generation results from SDXL v1.0~\cite{podell2023sdxl}, Midjourney v5.2~\cite{Midjourney}, Ideogram~\cite{Ideogram}, Firefly 2~\cite{Firefly}, and \modelname~\cite{dalle3paper,dalle3system}. We use the Hugging Face Diffusers to run SDXL inference~\footnote{\url{https://huggingface.co/stabilityai/stable-diffusion-xl-base-1.0}}. We obtain generation results for the remaining models via their web interface~\footnote{\url{https://discord.com/invite/midjourney}}\footnote{\url{https://ideogram.ai/}}\footnote{\url{https://firefly.adobe.com/}}\footnote{\url{https://chat.openai.com/}}, respectively.










\begin{figure}
\centering
\vspace{-20mm}
\begin{adjustbox}{center}
\small
%
\end{adjustbox}
\caption[3d art comparisons]{Qualitative comparisons among SDXL, Midjourney v5.2, and \modelname. Check Figure~\ref{fig:scenario_3D_historical_art_1} for complete prompts.
}
\label{fig:compare_3dart_36}
\end{figure}
\begin{figure}
\centering
\vspace{-20mm}
\begin{adjustbox}{center}
\small
%
\end{adjustbox}
\caption[3d art comparisons]{Qualitative comparisons among Ideogram, Firefly 2, and \modelname. Check Figure~\ref{fig:scenario_3D_historical_art_1} for complete prompts.
}
\label{fig:flyfire_3dart_38}
\end{figure}

\begin{figure}
\centering
\vspace{-20mm}
\begin{adjustbox}{center}
\small
%
\end{adjustbox}
\caption[historical art comparisons]{Qualitative comparisons among SDXL, Midjourney v5.2, and \modelname. Check Figure~\ref{fig:scenario_historical_art_2} for complete prompts.
}
\label{fig:compare_historical_39}
\end{figure}
\begin{figure}
\centering
\vspace{-20mm}
\begin{adjustbox}{center}
\small
%
\end{adjustbox}
\caption[historical art comparisons]{Qualitative comparisons among Ideogram, Firefly 2, and \modelname. Check Figure~\ref{fig:scenario_historical_art_2} for complete prompts.
}
\label{fig:firefly_historical_39}
\end{figure}

\begin{figure}
\centering
\vspace{-20mm}
\begin{adjustbox}{center}
\small
%
\end{adjustbox}
\caption[knolling design comparisons]{Qualitative comparisons among SDXL, Midjourney v5.2, and \modelname. Check Figure~\ref{fig:Scenario_knolling} for complete prompts.
}
\label{fig:compare_knolling_40}
\end{figure}
\begin{figure}
\centering
\vspace{-20mm}
\begin{adjustbox}{center}
\small
%
\end{adjustbox}
\caption[knolling design comparisons]{Qualitative comparisons among Ideogram, Firefly 2, and \modelname. Check Figure~\ref{fig:Scenario_knolling} for complete prompts.
}
\label{fig:firefly_knolling_40}
\end{figure}

\begin{figure*}[h!]
\centering
\vspace{-25mm}
\centerline{\includegraphics[width=1.5\textwidth]{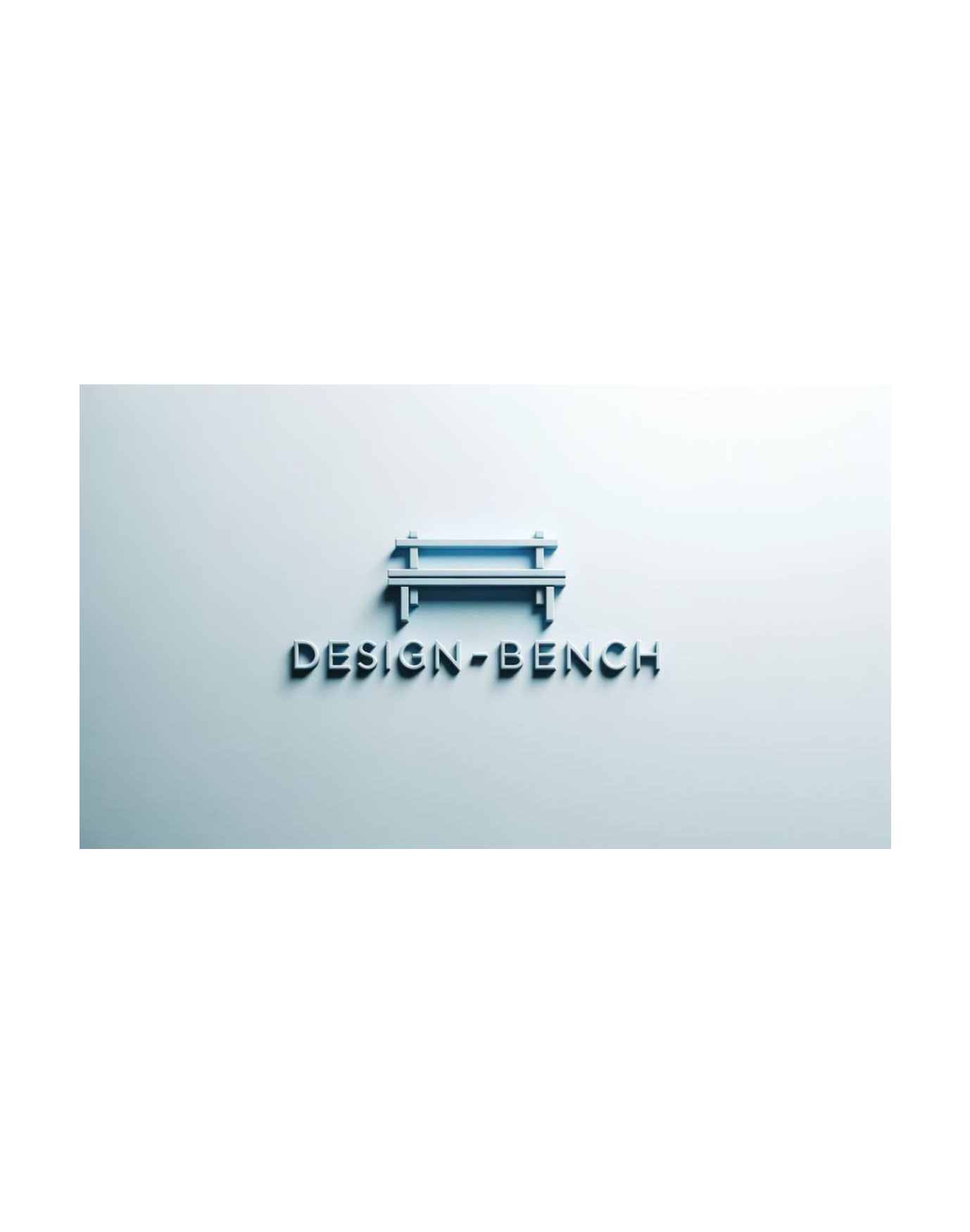}}
\vspace{-25mm}
\caption[\benchname logo design by \modelname.]{\benchname logo design by \modelname. Expanded prompt: 2D vector logo on a wide white background. The brand name 'Design-Bench' is written in a minimalist typeface, sleek with precision spacing. Above it is a simple bench emblem representing benchmarking and design. The color theme is a soft mute blue, reflecting professionalism and elegance.
}
\label{fig:logo_ending_page}
\end{figure*}

\end{document}